\documentclass[12pt]{thesis-umich}[thesis]
\usepackage{url}
\usepackage{graphicx}
\usepackage{subcaption}
\usepackage{amsmath}
\usepackage[inline]{enumitem}
\usepackage{booktabs}
\usepackage{multicol}
\usepackage{color}
\usepackage{booktabs}
\usepackage{multirow}
\usepackage{xspace}
\usepackage{xcolor}
\usepackage{natbib}
\usepackage{amsmath,bm,amsfonts,dsfont,bbm}
\usepackage{hyperref}
\usepackage{mdframed}
\usepackage{csquotes}
\usepackage{epigraph}
\usepackage{geometry}
\usepackage{pdfpages}
\usepackage{graphicx}
\usepackage{fullpage}
\usepackage{amsfonts}
\usepackage{amsmath}
\usepackage{amsthm}  
\usepackage{longtable}
\usepackage{xspace}
\usepackage[titletoc]{appendix}
\newtheorem{definition}{Definition}
\usepackage{algorithm}
\usepackage{algpseudocode}
\usepackage{subcaption}
\usepackage{amssymb}
\usepackage{bbm}
\usepackage{multirow}
\usepackage{pifont}%
\usepackage{latexsym}

\usepackage{tikz}
\usetikzlibrary{positioning}
\usetikzlibrary{calc}
\usetikzlibrary{arrows}
\usetikzlibrary{decorations.markings}
\usetikzlibrary{shapes.misc}
\usetikzlibrary{matrix,shapes,arrows,fit,tikzmark}
\usetikzlibrary{arrows.meta}
\usetikzlibrary{patterns}
\usetikzlibrary{patterns.meta}
\usetikzlibrary{decorations.pathreplacing,calc,positioning}
\usetikzlibrary{arrows,shapes}
\usetikzlibrary{arrows,chains,positioning,scopes,quotes}
\usetikzlibrary{calc,trees,positioning,arrows,chains,shapes.geometric,%
	decorations.pathreplacing,decorations.pathmorphing,shapes,%
	matrix,shapes.symbols}
\usetikzlibrary{shapes.callouts}
\usepackage[most]{tcolorbox}
\usepackage{array}
\usepackage{colortbl}
\usepackage{dcolumn}
\newcolumntype{d}[1]{D{.}{.}{#1}}  %

\usepackage{geometry}
\usepackage[activate={true,nocompatibility}, kerning=true,spacing=true, stretch=10,shrink=30]{microtype}
\microtypecontext{spacing=nonfrench}

\tikzset{
	>=stealth',
	punktchain/.style={
		rectangle, 
		rounded corners, 
		draw=black, %
		text width=5cm, 
		minimum height=2cm, 
		text centered, 
		inner sep=0,outer sep=0,
		},
	line/.style={draw, thick, <-},
	element/.style={
		tape,
		top color=white,
		bottom color=blue!50!black!60!,
		minimum width=8em,
		draw=blue!40!black!90, very thick,
		text width=10em, 
		minimum height=3.5em, 
		text centered, 
		on chain},
	every join/.style={->, thick,shorten >=1pt},
	decoration={brace},
	tuborg/.style={decorate},
	tubnode/.style={midway, right=2pt},
}

\usepackage{tikz}
\usepackage{geometry}
\usepackage{adjustbox}
\usetikzlibrary{positioning}

\department{Computer Science}

\definecolor{dkgreen}{RGB}{0,130,0}
\definecolor{aqua}{rgb}{0.0, 0.4, 1.0}
\title{\large{Understanding and evaluating computer vision models\\ through the lens of counterfactuals}}

\author{Pushkar Shukla}

\department{Computer Science}
\date{}
\email{pushkar92@gmail.com}

\frontpagestyle{7} %

\acknowledgments[7]{A PhD thesis is not just a collection of research aimed at advancing the frontiers of knowledge; it’s also a multi-year endurance test disguised as academia—filled with growth, setbacks, far too many late nights, and the occasional existential crisis. Along the way, you realize it’s not a solo adventure; many people, both academically and personally, play vital roles in helping you survive (and sometimes even enjoy) the journey. In my case, I have been fortunate to have many such people whose support, encouragement, and kindness helped me reach this point. To borrow a line from my advisor’s own PhD acknowledgments: “Barring an unexpected Academy Award, this may be my only opportunity to publicly thank you.” I would add: while I may not deserve the award for Best Actor, many of you certainly deserve recognition for Best Supporting Actor.

Thank you for walking alongside me during this journey.

First and foremost, I would like to thank my advisor, Prof. Matthew Turk. Matthew has been a constant source of support—not just as my PhD advisor at TTIC but also during my time as a master's student at UC Santa Barbara. He has always been calm and composed, qualities I have learned from during both good and challenging times. I am immensely grateful for his unwavering support and for always finding time for me, even while serving as the President of TTIC. He has pushed me to strive for excellence in the gentlest of ways, and his mentorship has been one of the most important influences on my academic and personal journey.\\
I would like to thank my committee members, Leonid Sigal and Matthew Walter—it was a pleasure to also serve as a TA for one of Matthew’s courses. Leon, your suggestions during our discussions were highly beneficial and contributed meaningfully to my work.

I would also like to thank all members of the TTIC family—TTIC is truly a wonderful place to pursue a PhD. I am especially grateful to Madhur Tulsiani and Julia Chuzhoy for their initial support, particularly in helping me survive my Algorithms coursework. I am also thankful to Greg Shakhnarovich for his guidance during my Machine Learning class.

The administrative team at TTIC—Amy Minick, Chrissy Coleman, Brandie Jones, Mary Marre, Adam Bohlander, Jessica Jacobson, Erica Cocom, Rose Bradford, Erica Cocom, Celste Ki, Randall Landsberg, Alicia McClarin and Deree Kobets—deserves my heartfelt thanks for their patience and support in helping me navigate countless forms and deadlines, despite my chronic tardiness. Your reminders and willingness to assist were invaluable.  A special thanks to Mary Marre for always keeping me informed about leftover food in the fridge waiting to be rescued — those timely updates truly brightened many of my days! I am also deeply grateful to Amy Minick for her endless patience and helpful suggestions, especially when it came to navigating the labyrinth of visa paperwork. I am grateful to Randy for giving me a chance to be a part of the TTIC outreach initiative and a taking the time out to give extremely valuable feedback on my defense presentation. Also, a note of thanks to Jerry and Ayesha for cheering me up with random conversations. 

I am grateful to all my research collaborators who made this journey possible: Gaurav Bhatt, Kartik Hosanagar, Kiri Salij, Alexander Tolbert, Dhruv Srikanth, Lee Cohen, Emily Diana, Vineeth Balasubramanian, and Sushil Bharati—thank you for your insightful discussions and collaborations. Gaurav, our friendship has been long-standing, and it was great to finally work together on a project. I owe special thanks to Aditya Chinchure (Adi) for extensive collaborations—honestly, a large part of this thesis would not have been possible without your support.I thoroughly enjoyed those late-night conversations about research — where every idea sounded brilliant at 2 a.m. but much less so after coffee the next morning. Your visit to TTIC as an intern, including our trip to Yosemite and the West Coast, will always be memorable.

This journey would not have been possible without the friendships I’ve been fortunate to form. Among my friends at TTIC, I am especially grateful to Shubham Toshniwal for his advice in my early years and for our exploratory trips to random restaurants in Chicago—even during COVID times, masked up. Shashank Srivastava deserves thanks for helping me through coursework and qualifying exams. Sudarshan Babu, living as roommates for two years was indeed an adventure — I’m just glad we both survived with our sanity (mostly) intact. David Yunis, thank you for those deep, thoughtful conversations over Diet Coke in Chicago bars—you are truly a gem of a person. I am also grateful to Max and Kavya for making me the honorary third wing of Dumbledore’s Army. I am thankful to Ankita, Mrinal, Takuma, Gene, Han, Kezziah, Teddy, Kshitij and Nirmit for some great interactions during and outside TTIC. 

My heartfelt thanks extend to my friends in Chicago beyond TTIC: Ansh, Neha, Peter,  Shehjar, and Apeksha. Shehjar and Apeksha, it has been very special to be the godfather to your daughter. I am also grateful to my running group—thank you for giving me something to look forward to every Saturday during long runs. 
I am thankful to Aditya Raj for his unwavering support and for introducing me to the philosophical perspective of Sir Christian Cole. In the past three years, you have been my go-to person for every kind of help. I am grateful to Eleven Dew Drops for the positive impact during a low phase of my PhD. Although our interactions were brief, the impact was deep and much needed.

To my college friends—Lokesh, Hitesh, Jeetu, Avinash, Shailesh, Kapil, Parashar, Meghna, and Sharma—thank you for four wonderful years and for your continued encouragement afterward. Even after eleven years, our friendship remains strong. Lokesh, you were an incredible roommate—thank you for tolerating all my eccentricities in undergrad.  To my childhood friends, Mishra, Anindya, and Archit, I am proud of how our bond has only grown stronger over time. Very few people have believed in me as much as you have. I am not someone who openly romanticizes friendship, but secretly, I cherish it the most.

Two practices that have stayed with me through the toughest phases of this journey are reading the Gita and practicing Vipassana. I would urge every PhD student to cultivate some form of spiritual practice. On days when reviews were harsh or experiments failed, it was reassuring to hear a quiet voice within say: “This too shall pass.” Unfortunately, that voice did not write rebuttals or fix broken code. I would like to sincerely thank Ajay Ji, Vinika Didi, Vivek Ji, and Aditya Ji for their consistent encouragement and for sharing thoughtful perspectives on life. Their guidance, often drawing from the teachings of the Gita—particularly the emphasis on focusing on one’s actions without attachment to outcomes—has been a valuable source of clarity and balance throughout this journey.

The last three years wouldn’t have been nearly as fulfilling or productive without my meditation buddies, Gaurav and Kartik. I still can’t believe I’ve been regular at something for this long—thank you for keeping me accountable. This practice has had a meaningful and lasting impact on both my personal and academic life. Beyond our the practice I really admire how the two of you push me to be a better person in different aspects of my life. I really look up to the two of you. 

I am also deeply grateful to the teachers who shaped me. Archana Ma’am supported me in my early years and laid the foundation for my academic journey. Bilal Sir played a pivotal role in instilling in me a love for programming, a passion that has stayed with me ever since. During my undergraduate years, Kamal Kapoor Sir provided constant encouragement and support, helping me navigate both academic and personal challenges. Varun Pratap Singh Sir was the first to suggest that I consider pursuing a PhD abroad—he saw potential in me that I had not yet recognized in myself. Among all my teachers, the one to whom I am most indebted is Ankush Mittal Sir. His guidance shaped me not just as a student but as an individual, and I will always remain grateful for his belief in my potential and for showing me the path that brought me here.

Coming to family: I am grateful to all my grandparents for their love and affection — thanks to them, I’ve developed an above-average capacity for unsolicited advice, overfeeding others, and expressing heartfelt concern, all clear signs of deep affection. I am especially thankful to my grandfathers for all those wonderful stories and for patiently playing cricket with me all day long, and to my grandmothers for their tireless efforts to overfeed me as a child — a mission they took very seriously and executed with great success. Kamala Awasthi and Ravi-Shankar Shukla, I still miss you from the deepest corners of my heart. To my uncle and aunt, Sudhanshu and Kanchan, thank you for showing me how to treat others with kindness and care. Your excitement about me pursuing my studies in the US was truly unmatched — even if, at the time, the rising price of bhaat(rice) in India might have suggested there were more pressing things to worry about. To Mausa and Mausi, thank you for treating me like your own child. A note of thanks to Kanishka for inspiring me to write that paper in 2017, which set me on this journey. To my cousins — Shruti, Pavani, Ashutosh, Saarika, Supriya, Saurabh, and Shashank — thank you for your steady support. A special thanks to Shashank for being my personal tech support, life coach, and emergency backup plan whenever things went sideways. Shruti and Pavani, thank you for the constant reminder that Winter is coming, and for being my Arya and Sansa in life — because, as you know, I know nothing.

I am deeply grateful to my parents, whose unwavering support has made this long and winding journey possible. Honestly, this PhD feels like a joint family project at this point. My mother, who once aspired to study mathematics but couldn’t, has been a quiet (and occasionally not-so-quiet) source of inspiration — I hope this PhD in an applied math field earns me at least some bragging rights on her behalf. My father’s unique philosophy on failure — that as a scientist, I should be well-trained by now to expect failure in 95$\%$ of the things I attempt — has kept me grounded and helped me maintain perspective (and my sense of humor) throughout this process. I suspect he’s just happy I’ve finally reached the 5$\%$ that worked.

Finally, I am grateful to God for providing me with the strength, clarity, and resilience to survive this journey. No amount of machine learning or scientific reasoning has calmed my anxiety as effectively as a quick prayer in moments of panic. Honestly, prayer has been my go-to debugging tool for life. Please continue to save me — and let’s just hope I pass this final hurdle, because at this point, divine intervention feels like my best bet.
\clearpage
\section*{Credit Assignment}
Good collaboration is often key to impactful research, and the work presented in this thesis has largely been the result of collaborative efforts. I would like to acknowledge and give due credit to my fellow collaborators. I also wish to emphasize that I personally value collaboration with researchers from diverse backgrounds, as it fosters deeper discussions and brings varied perspectives to the work.

The work in \textbf{Chapter 3} was a collaboration between myself, Sushil Bharati, and Matthew Turk. Sushil Bharati contributed by running experiments on the CelebA dataset, while Matthew Turk provided valuable guidance through insightful discussions.

The work in \textbf{Chapter 4} was a collaboration between myself, Dhruv Srikanth, Lee Cohen, and Matthew Turk. I led the conceptualization and prototyping of the approach, while Dhruv Srikanth played a key role in scaling and executing the experiments. Lee Cohen and Matthew Turk provided valuable feedback and were actively involved in project discussions.

The work in \textbf{Chapter 5} was a collaboration between myself, Aditya Chinchure, Gaurav Bhatt, Kiri Salij, Kartik Hosanagar, Leonid Sigal, and Matthew Turk. I led the conceptualization of the project, while Aditya Chinchure led the experimental efforts, though both of us contributed across these aspects. Gaurav Bhatt conducted comparisons between our approach and ITI-GEN, and Kiri Salij contributed code for using image captions to evaluate TTI models. Leonid Sigal, Matthew Turk, and Kartik Hosanagar provided valuable feedback throughout, with Kartik playing an instrumental role in designing the user study and robustness analyses for VQA.

The work in \textbf{Chapters 6 and 7} was a collaboration between myself, Aditya Chinchure, Emily Diana, Alexander Tolbert, Kartik Hosanagar, Leonid Sigal, Vineeth Balasubramanian, and Matthew Turk. I led the conceptualization and developed the initial working prototype for intersectional bias mitigation, while Aditya led the experimental efforts, with both of us contributing collaboratively across tasks. Discussions with Emily Diana and Alexander Tolbert primarily focused on intersectionality and causality, drawing on their expertise in these areas. Leonid Sigal, Matthew Turk, Kartik Hosanagar, and Vineeth Balasubramanian were involved in providing broader feedback and guidance throughout the project.
}
\committee{ %
	Professor Matthew Turk, Chair \\
	Professor Matthew Walter \\
	Professor Leonid Sigal \\
	
} 

\showlistoftables
\abstract{

Counterfactual reasoning—the practice of systematically exploring “what if” scenarios by making systematic changes to inputs, observing how a model’s behavior changes, and drawing inferences about the model—has become a central tool in interpretable and fair AI. This thesis introduces several frameworks that harness counterfactuals to explain, audit, and mitigate bias in both vision classifiers and generative models. By systematically varying semantically meaningful attributes while holding others fixed, these methods leverage counterfactuals as a lens to uncover spurious correlations, interrogate causal dependencies, and ultimately build models that are more robust and fair.

The thesis is organized into two major parts, based on the nature of the underlying models. The first part focuses on vision classification models and introduces two novel approaches. The first, CAVLI, combines attribution-based explanations (LIME) with concept-level representations (TCAV) to measure the extent to which a model’s decisions rely on human-interpretable visual concepts. By producing localized heatmaps and a Concept Dependency Score, CAVLI identifies instances when a model’s decisions rely on irrelevant cues, like background objects or spurious features. Building on this, the thesis proposes ASAC, a technique for generating adversarial counterfactuals that preserve visual semantics while perturbing protected attributes. Using a curriculum learning framework, this method fine-tunes biased classifiers on these targeted perturbations, leading to measurable improvements in both fairness metrics and accuracy. Unlike conventional generative counterfactuals, ASAC avoids introducing stereotype-laden artifacts, addressing both algorithmic and ethical pitfalls.

The second part of this thesis focuses on the use of counterfactuals to evaluate and mitigate biases in generative Text-to-Image (TTI) models. Through TIBET, a scalable pipeline is developed for dynamically evaluating prompt-sensitive biases by systematically varying identity-related terms in input prompts. This enables causal auditing of how attributes such as race, gender, and age influence image generation. Recognizing that social biases often interact, this thesis introduces BiasConnect, a causal graph-based diagnostic tool that measures intersectional effects between social attributes. Finally, InterMit is proposed as a modular, training-free algorithm that mitigates intersectional bias using causal sensitivity scores and user-defined fairness goals.

Together, these contributions demonstrate the practical utility of counterfactual reasoning in computer vision systems. By applying it across both discriminative and generative contexts—and by bridging interpretability, fairness, and causal analysis—this thesis lays the groundwork for a principled, scalable, and socially responsible approach to bias evaluation, mitigation and reasoning in modern computer vision systems.

}

\begin{document}

\chapter{Introduction}

\label{sec:intro}

\begin{figure}[ht]

    \centering
    \includegraphics[trim=300 80 300 200, clip,scale=0.3]
    {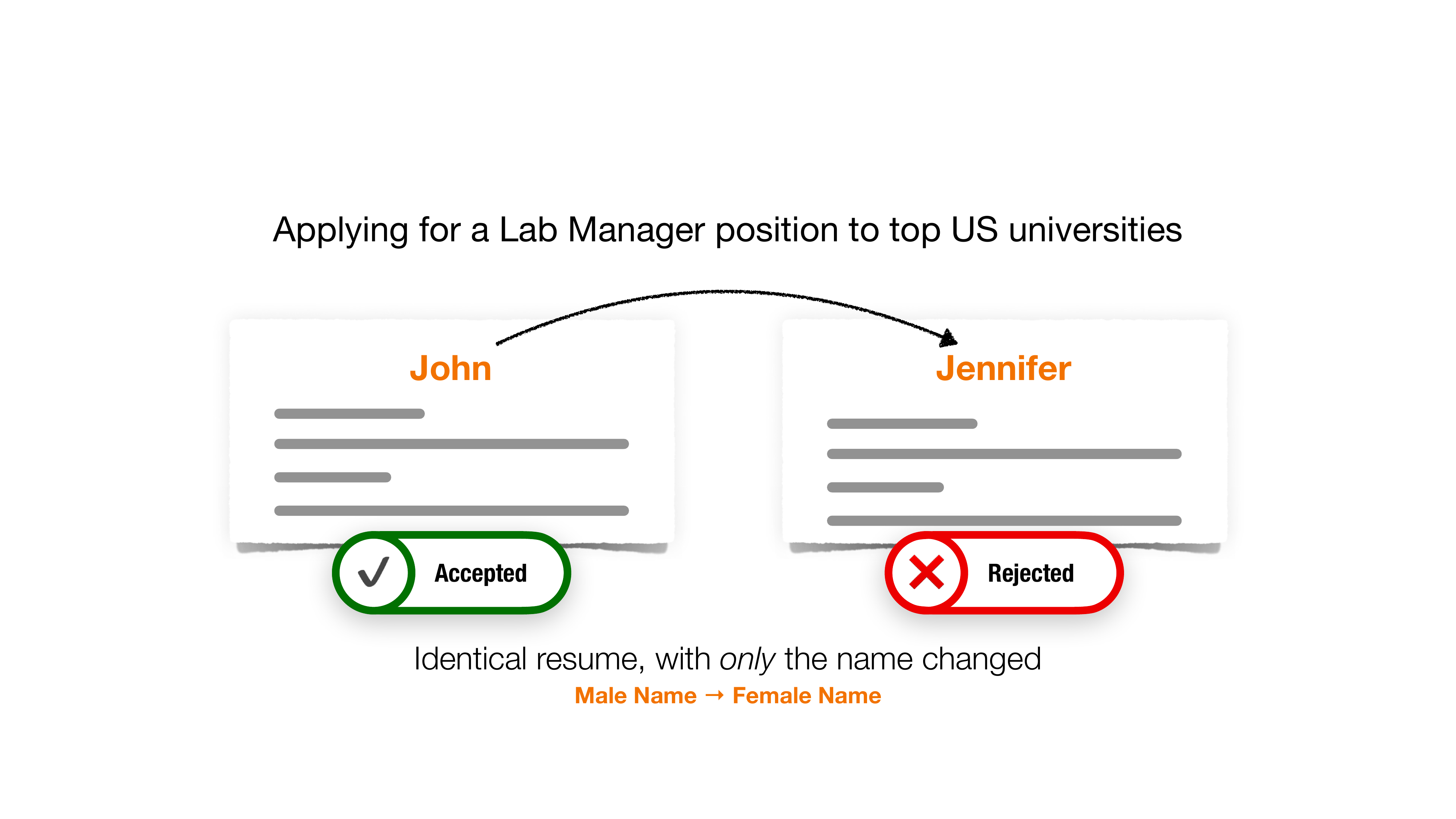}
    \caption{ Counterfactual reasoning reveals hiring bias. In a famous study conducted by Moss-Racusin et al. \cite{moss2012science}, identical resumes differing only in the applicant’s name (``John" vs. ``Jennifer") were submitted for a lab manager position. It was observed that faculty were significantly more likely to favor the male applicant. This style of reasoning—holding all factors constant while changing desired attributes—is referred to as counterfactual reasoning. This simple idea is at the heart of the methods and contributions in this thesis.}
\label{fig:c1_intro_teaser} 
\end{figure}
\clearpage

\section{Motivation}
As computer vision models become increasingly integrated into domains such as healthcare, finance, education, and creative industries, concerns around their fairness, transparency, and accountability have grown more prominent. Among these, bias in classification and generative models has emerged as particularly pernicious—manifesting in both overt and subtle ways that reflect and amplify societal stereotypes \cite{denton2019image,wang2020towards,zhang2020towards}. Models trained on large-scale web data often encode patterns that correlate sensitive attributes (e.g., gender, race) with undesired labels or aesthetics. For example, women may be overrepresented in submissive or decorative roles, while older individuals may be underrepresented altogether \cite{dash2022evaluating,balakrishnan2021towards}. Despite growing awareness of the risks associated with these models, systematic tools for understanding, evaluating, and improving their safety remain limited.. 

One powerful tool for understanding and improving machine learning models is  counterfactuals. As illustrated in the example of the hiring bias study (Figure \ref{fig:c1_intro_teaser}), counterfactuals can be thought of as variations in the input that are made in order to understand how the output changes. In this study, identical résumés were submitted with only one change—the applicant’s name, either ``John" or ``Jennifer." It was observed that faculty were more likely to favor ``John" over ``Jennifer." This simple change, while keeping everything else fixed, helped reveal how the outcome (hiring decisions) was influenced by gender. It is through such structured variations that we can draw inferences about the workings of a black-box model—referring to models like deep neural networks, where the mapping from input to output is not explicitly transparent or easily interpretable.

Counterfactual reasoning builds on this idea by formalizing it: it refers to the process of systematically altering parts of an input to an AI model and observing how the output changes. For instance, if we change one feature (e.g., the age or gender in a dataset) while holding all others constant and the model’s prediction changes, we can deduce that the altered feature had a causal influence on the outcome. This kind of reasoning allows researchers and practitioners to analyze model behavior, uncover biases, and make informed improvements to ensure fairness, robustness, and interpretability. At its core, counterfactual reasoning asks: What would the model do if a particular attribute were different, while everything else remained the same?
This concept of counterfactual reasoning can take slightly different forms depending on the domain and task.

In computer vision—particularly for classification tasks—it often involves altering specific visual concepts in an image, such as changing the apparent age or gender of a person, while holding other attributes constant. This allows us to examine how the model's predictions shift in response to targeted changes, revealing underlying dependencies and potential biases \cite{abid2022meaningfully,feder2021causalm,shukla2023cavli}.

The same principle applies to structured decision-making systems. For instance, in a bank loan approval model, one might ask: Would this applicant have been approved if their gender were different, assuming income, credit score, and employment status remained the same? Such counterfactual queries help diagnose whether sensitive attributes like gender or race are exerting unjustified influence over outcomes.

In multimodal generative systems like Text-to-Image (TTI) models, counterfactual reasoning typically involves modifying textual prompts—for example, changing “an old man at a church” to “an Asian old man at a church”—and observing how the generated images differ. These subtle changes can reveal hidden correlations between identity terms and visual outputs, surfacing stereotypes encoded in the model.

Although counterfactuals have gained prominence in recent years for diagnosing and improving computer vision models [12, 13, 14, 15], existing approaches remain limited in several important ways. Many methods generate counterfactuals that lack realism or causal grounding, making them less actionable for practitioners. Others focus narrowly on local, instance-specific explanations without offering broader insights into systemic issues such as dataset biases, fairness concerns, or the robustness of generative models. Moreover, evaluation criteria for counterfactuals in vision remain underdeveloped, often relying on ad hoc notions of plausibility rather than principled measures. This thesis addresses these gaps by developing methods that apply counterfactual reasoning not only to explain individual predictions, but also to evaluate and mitigate biases in both classification and generative models. In doing so, it moves counterfactual reasoning from being primarily a diagnostic tool to serving as a core approach for building fair, reliable, and interpretable vision systems.

This thesis explores several themes, all connected by the central idea of counterfactual reasoning. We begin with a basic question: How can we tell if a model’s decision—such as classifying an animal as a cow —depends on a human-defined concept like the presence of grasslands in the background? While such dependencies are easier to measure at the pixel level \cite{selvaraju2017grad}, we propose a method called \textbf{CAVLI} that uses counterfactual reasoning to quantify how much a decision relies on high-level, human-defined concepts.

Next, we question the reliability of the counterfactuals themselves. What if the model or engine used to generate them is biased? Relying on flawed counterfactuals for explanation or mitigation can reinforce existing biases and lead to misleading conclusions \cite{denton2019image}. To address this, we introduce in-situ counterfactual generation, where modifications are made directly within the model, avoiding external generative systems. Specifically, we propose a method based on adversarial examples—called \textbf{ASAC}—that produces targeted, model-aware counterfactuals. We demonstrate that these examples contribute not only to training fairer models but also to improving their overall performance.

%While counterfactual techniques have been widely explored in classification tasks \cite{mothilal2020explaining,sokol2019counterfactual,kasirzadeh2021use,christensen2023counterfactual}, their application to computer vision and generative models presents unique challenges. A significant limitation of current approaches is the reliance on external generative models (e.g., StyleGAN2 \cite{karras2020analyzing}) to produce image counterfactuals. These generators often encode their own biases—such as associating femininity with heavy makeup or masculinity with aggression—leading to spurious correlations that compromise the integrity of the counterfactuals themselves.  Using such flawed counterfactuals for explanation or mitigation can reinforce rather than resolve bias, resulting in misleading insights \cite{denton2019image}. To address this, our work proposes in-situ counterfactual generation, where image or prompt modifications are generated directly by the model without requiring any external distortion or manipulation. This approach is reflected in methods like \textbf{CAVLI} \cite{shukla2023cavli}, which uses attribution-based visual counterfactuals to analyze concept importance in vision models (e.g- "To what extent does the decision that a person is smiling depend on their gender?"), and \textbf{ASAC} \cite{shukla2024utilizing}, which generates adversarial counterfactuals for bias mitigation via curriculum learning. These techniques not only improve interpretability but also ensure that mitigation strategies are grounded in the behavior of the model being audited.

We also show that counterfactual reasoning can be used to evaluate biases in Text-to-Image models. We highlight that bias in generative models is not static. Rather, it is highly prompt-sensitive and dynamic—meaning that model outputs can vary widely based on how inputs are phrased or structured. This motivates the development of \textbf{TIBET} \cite{chinchure2023tibet}, a scalable framework for prompt-level counterfactual analysis in TTI models. TIBET uncovers how biases change across prompt variations, enabling both large-scale audits and deeper insights into prompt-induced bias shifts.

Another core motivation of this thesis is the recognition that different biases in vision models are rarely independent. Drawing from the framework of intersectionality \cite{crenshaw1989demarginalizing}, we argue that biases in generative models often emerge through complex interactions between social identities—such as gender, race, and age—and must be studied in combination rather than isolation. To this end, we develop tools like \textbf{BiasConnect} \cite{shukla2024utilizing}, which constructs pairwise causal graphs using counterfactual prompt interventions to measure how mitigating one type of bias (e.g., gender) affects another (e.g., age). This approach yields a structured diagnostic view of bias entanglement.

Finally, we extend this diagnosis into actionable mitigation through our proposed algorithm \textbf{InterMit} \cite{shukla2025mitigate}, which uses user-defined fairness priorities and intersectional sensitivity scores to guide efficient bias reduction in generative outputs. Unlike prior methods, InterMit does not require retraining and adapts mitigation steps based on both user goals and observed causal effects.

\begin{figure}[ht]
    \centering
    \includegraphics[width=0.75\textwidth]{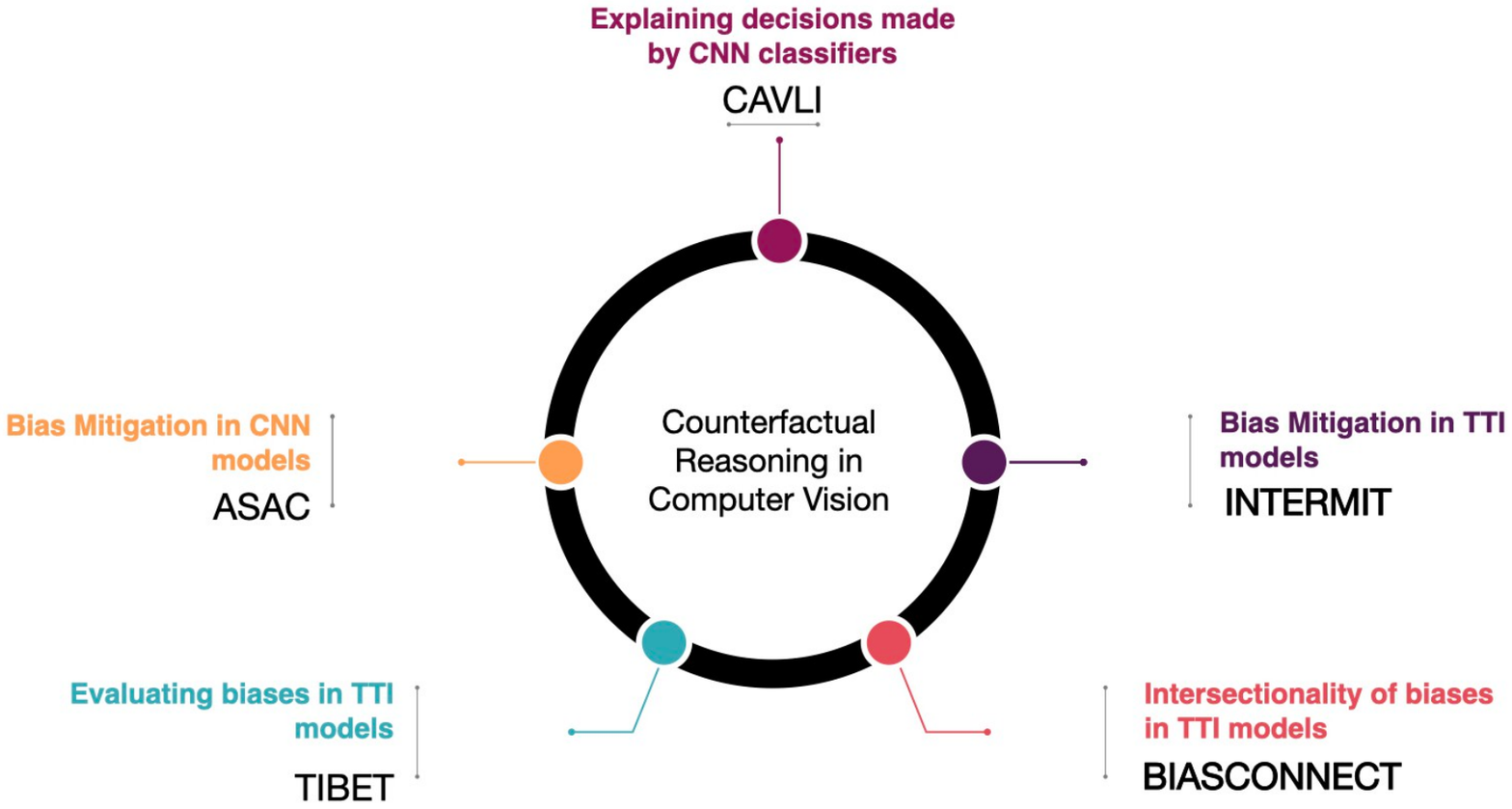}
    \caption{This figure illustrates how \textit{counterfactual reasoning} is employed throughout the thesis to probe and enhance the capabilities of computer vision models. In Chapter 3, we introduce \textbf{CAVLI}, a method for quantifying the influence of visual concepts on model decisions via counterfactuals. Chapter 4 leverages counterfactuals for bias mitigation using \textbf{ASAC}s, showing their efficacy in reducing unfair model behavior. In Chapter 5, we present \textbf{TIBET}, a framework for evaluating bias in text-to-image models through systematic counterfactual interventions. We further explore interactions between multiple bias dimensions in TTI models using our tool \textbf{BiasConnect} in Chapter 6, and  propose a strategy for intersectional bias mitigationin generative models  called \textbf{InterMit} in Chapter 7.}
    \label{fig:intro_overview}
\end{figure}

In summary, this thesis makes a case for using counterfactual reasoning not only as a diagnostic tool but as a core methodology to rethink how we evaluate and mitigate bias in modern AI systems. By applying counterfactual logic to both vision classifiers and generative models, we develop tools and algorithms that are conceptually unified, practically useful, and grounded in causal principles. A roadmap of these contributions is provided in Figure~\ref{fig:intro_overview}, and a chapter-wise breakdown follows in Section~\ref{sec:thesis_contributions}.

\section{What is a Counterfactual?}
\label{sec:counterfactual_definition}

\paragraph{Counterfactuals.} 
 We define a \textit{counterfactual} as an alternate version of an input to an AI model that differs from the original in a small set of semantically meaningful human defined concepts or data points, while holding all other attributes fixed. This concept-level definition generalizes across modalities. In computer vision, counterfactuals involve modifying high-level visual attributes—such as changing a person’s age, gender, or clothing in an image—while preserving the overall scene \cite{goyal2019counterfactual, abid2022meaningfully}. In natural language processing (NLP), counterfactuals can be constructed by altering specific words or phrases in a sentence that correspond to sensitive or influential concepts. For instance, changing “He is a nurse” to “She is a nurse” helps assess how language models associate gender with profession. Similarly, in structured decision-making tasks like credit scoring or loan approval, counterfactuals involve tweaking categorical inputs—e.g., changing an applicant’s gender from “male” to “female” while keeping income, credit score, and employment status fixed—to identify whether sensitive features unduly influence predictions \cite{wachter2017counterfactual}. Across all these domains, counterfactuals serve as targeted probes for uncovering hidden biases, testing model robustness, and generating interpretable explanations.

\paragraph{Counterfactual Reasoning.}
\textit{Counterfactual reasoning} refers to the systematic use of counterfactuals to isolate, test, and understand the causal impact of specific concepts on model behavior. It provides a powerful framework for evaluating fairness, generating explanations, and guiding mitigation strategies across both discriminative and generative tasks. By comparing model outputs before and after intervening on a target concept—while holding all else constant—researchers can attribute model decisions to specific concepts \cite{feder2021causalm, abid2022meaningfully} and uncover implicit biases embedded in data or learned representations \cite{chinchure2023tibet}. In this thesis, counterfactual reasoning underpins multiple contributions: from attributing decisions to visual concepts (e.g., via CAVLI) to evaluating bias in TTI models (e.g., using TIBET), and uncovering interactions between intersecting attributes (e.g., through BiasConnect). This reasoning framework not only helps expose undesirable behaviors but also lays the foundation for systematic bias mitigation.

\subsection{A Brief History of Counterfactuals in AI and Causal Inference}
\label{sec:counterfactual_history}

The concept of counterfactuals has longstanding origins in philosophy, social science, and causal inference. In philosophical logic, \textit{David Lewis} formalized counterfactual conditionals in his seminal work \textit{“Counterfactuals”} \cite{lewis1973counterfactuals}. He proposed a possible word semantics to evaluate statements like ``If A had happened, B would have followed,'' where the truth of such a counterfactual depends on its closeness to the actual world.

In statistics and the social sciences, the \textit{Potential Outcomes Framework} introduced by \textit{Donald Rubin} \cite{rubin1974estimating} laid a probabilistic foundation for counterfactual inference, imagining that each unit (e.g., an individual) has multiple potential outcomes under different treatment conditions. This framework underpins modern causal inference methods such as randomized control trials and treatment effect estimation.

The formal introduction of counterfactual reasoning into computer science was pioneered by \textit{Judea Pearl}, who developed \textit{Structural Causal Models (SCMs)} \cite{pearl2000causality}. SCMs provide a graphical and algebraic language for causal reasoning. Pearl also introduced the \textit{Ladder of Causality}, distinguishing three levels of reasoning: 
\begin{enumerate}
    \item \textbf{Association}: Observing statistical relationships (e.g., ``People who take aspirin tend to recover'').
    \item \textbf{Intervention}: Modeling outcomes of deliberate actions (e.g., ``What happens if we give aspirin?'').
    \item \textbf{Counterfactuals}: Hypothetical alternate outcomes (e.g., ``Would the patient have recovered if they had not taken aspirin?'').
\end{enumerate}
The third level—counterfactuals—is the most expressive and enables rich forms of explanation and introspection.

The integration of counterfactual reasoning into machine learning gained traction in the late 2010s, particularly in the domains of \textit{explainability} and \textit{algorithmic fairness}. Wachter et al. \cite{wachter2017counterfactual} and Mothilal et al. \cite{mothilal2020explaining} proposed counterfactual explanations as a method to interpret decisions of black-box models by identifying the smallest change in input that alters the output.

In the field of computer vision, early forms of counterfactuals appeared in the context of \textit{adversarial examples}, but the concept evolved into more semantically grounded image edits that aim to test model behavior in interpretable ways. Denton et al. \cite{denton2019image} introduced image counterfactuals using attribute-guided editing, while Wu et al. \cite{wu2021polyjuice} and Abid et al. \cite{abid2022meaningfully} extended counterfactual techniques to language and vision-language models.

This historical progression underscores how counterfactual reasoning has grown from a philosophical abstraction to a central tool in modern AI research, enabling deeper understanding, fairness auditing, and robust model design.

\subsection{The Utility of Counterfactuals in Computer Vision}
\label{sec:utility_counterfactuals}

Counterfactuals have emerged as a versatile and powerful tool in computer vision, enabling a deeper understanding of how vision models make decisions and how they behave under changes to their input. Their utility spans multiple dimensions:

\paragraph{(a) Interpretability.}
Counterfactuals are intuitive and human-understandable, often answering the question: ``What would need to change in the image for the model to make a different decision?'' This is particularly valuable in the context of black-box models, where internal decision mechanisms are inaccessible \cite{wachter2017counterfactual, feder2021causalm}. In vision tasks, counterfactuals can be generated by modifying image regions to identify semantically meaningful features responsible for a prediction. 

\paragraph{(b) Fairness and Bias Auditing.}
Counterfactuals have become central to fairness analysis in vision models by revealing disparities in model outputs across protected attributes such as race, gender, and age \cite{kusner2017counterfactual, denton2019image}. Vision-based counterfactuals can show whether a model would make a different decision if a person's apparent gender or ethnicity were changed. 

\paragraph{(c) Robustness and Generalization.}
Counterfactuals can be used to assess and improve model robustness by generating borderline or perturbed image examples. This helps evaluate whether a model's predictions are consistent under small semantic or visual changes \cite{christensen2023counterfactual}. Additionally, counterfactual images have been used to augment training data, improving generalization and reducing overfitting in computer vision models \cite{kaushik2019learning}.

\paragraph{(d) Causal Inference and Interventions.}
Counterfactual reasoning allows for causal analysis of vision model behavior by posing questions like: ``Would this image still be classified the same way if we intervened on a particular feature (e.g., background, object texture)?'' Such interventions help distinguish causal relationships from spurious correlations \cite{pearl2000causality}. This has applications in safety-critical systems, model debugging, and understanding unintended model behavior.

Overall, counterfactual reasoning provides a powerful lens for probing vision systems, not only by enabling better explanations, but also by guiding robust model design and ethical AI development.

\section{Thesis Contributions}
\label{sec:thesis_contributions}

We make the following key contributions in this thesis, unified by the use of counterfactual reasoning as a foundational lens to understand, evaluate, and mitigate bias in vision and generative models:

\begin{itemize}
\item \textit{ Concept Attribution in Vision Models (Chapter 3):}
We propose \textbf{CAVLI}, a method that leverages image counterfactuals to quantify how classifier decisions (e.g., labeling an image as a zebra) depend on specific visual concepts (e.g., stripes). By measuring the spatial overlap between pixels involved in decision-making and those tied to a concept, CAVLI provides interpretable insights into the role of semantic features in model predictions.

\item \textit{Counterfactual Generation for Bias Mitigation (Chapter 4):}  
We identify a key limitation of using generative models for counterfactual image synthesis in bias mitigation — namely, the introduction of spurious attribute correlations (e.g., associating gender with makeup). To address this, we propose a novel adversarial counterfactual strategy combined with a \textbf{curriculum-based fine-tuning approach} that mitigates bias while preserving semantic realism.
\item \textit{Dynamic Evaluation of Biases in TTI Models (Chapter 5):}
We recognize that biases in Text-to-Image (TTI) models are not static but dynamic and prompt-sensitive, with model behavior varying substantially across different inputs. To address this, we propose TIBET, a novel framework that uses prompt-level counterfactuals to dynamically evaluate and explain how bias manifests across diverse prompts. By generating controlled variations of input prompts (e.g., changing gender, ethnicity, or age terms) and analyzing resulting image attributes, TIBET allows for both quantitative bias measurement and causal reasoning about where and how different biases intersect. This enables large-scale, interpretable bias audits that account for the contextual fluidity of generative model behavior—going beyond static benchmarks to uncover more nuanced fairness failures.

\item \textit{Diagnosing Intersectional Bias in Generative Models (Chapter 6):}  
Recognizing \textbf{intersectionality} as a fundamental and understudied source of bias in generative AI, we introduce \textbf{BiasConnect}, a diagnostic framework that uses prompt-level counterfactual interventions to construct \textit{pairwise causal graphs} and compute \textit{Intersectional Sensitivity} scores. This enables structured, quantitative analysis of how mitigating bias along one axis (e.g., gender) may improve or worsen biases along others (e.g., age, ethnicity).

\item \textit{ Interventional Mitigation of Intersectional Bias (Chapter 7) :}  
Extending BiasConnect, we propose \textbf{InterMit}, a training-free, modular algorithm that uses causal sensitivity estimates to guide intersectional bias mitigation. InterMit incorporates user-defined priorities and target distributions to adaptively select mitigation directions, achieving lower residual bias and higher visual quality than prior approaches—while accounting for fairness trade-offs across multiple social dimensions.
\end{itemize}

\newpage
\chapter{Background}

\begin{figure}[ht]
    \centering
    \includegraphics[trim=220 180 250 180, clip,width=\textwidth]{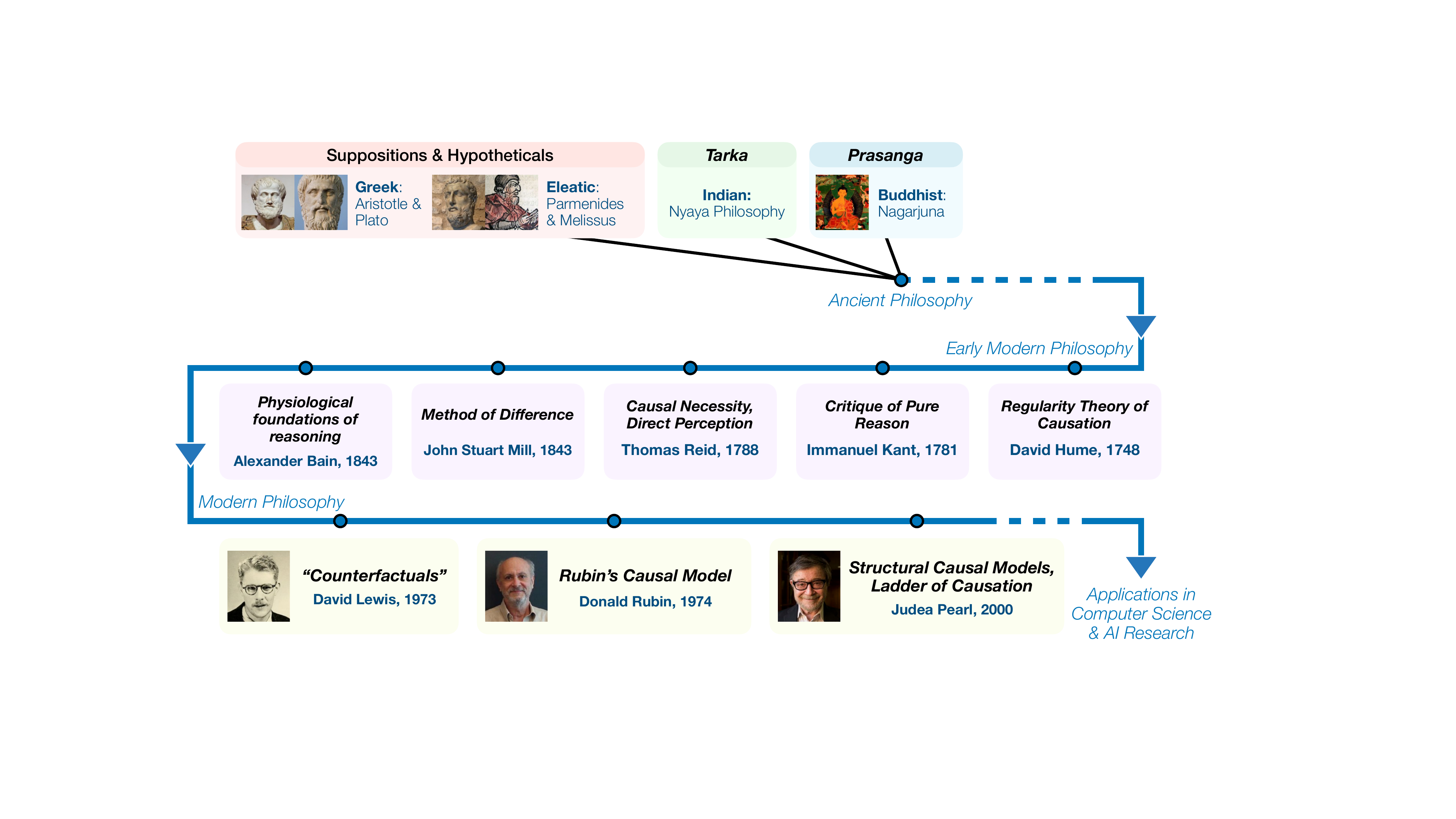}
   \caption{A brief timeline tracing the development of counterfactual reasoning. While the term ``counterfactual" was popularized by David Lewis in 1973 \cite{lewis1973counterfactuals}, the underlying idea of reasoning about alternative possibilities has roots in both ancient Western and Eastern philosophical traditions.}

    \label{fig:background_teaser}
\end{figure}
\clearpage
\section{Introduction}

This chapter provides a review of the existing literature on counterfactual reasoning, fairness, and explainability in machine learning, with a particular focus on computer vision and generative models. Since a large part of this thesis focuses on fairness and explainability, this review aims to give the necessary background to understand how these topics have developed and where current challenges remain.

The chapter begins by outlining the history of counterfactual thinking, tracing its roots from early philosophical ideas to its formal role in modern causal inference. It then discusses how counterfactuals are used in machine learning to improve understanding of model behavior, evaluate fairness, generate recourse, test robustness, and conduct causal analysis. The review highlights how these methods differ depending on the data type and how counterfactuals are constructed.

The chapter also surveys key definitions of fairness and examines how bias has been studied in both natural language processing and computer vision. Particular attention is given to work on intersectional biases, which considers how different social attributes like gender, race, and age combine in complex ways.

Finally, the chapter reviews explainability methods that aim to make model decisions more transparent, through techniques such as feature attribution, concept-based reasoning, and counterfactual examples. By reviewing this literature, the chapter sets the stage for the thesis’s contributions. In this thesis, counterfactual reasoning has been explored as a tool to address questions of fairness and explainability. This review identifies where further work is needed, especially in addressing biases in generative models.

\section{A brief history of counterfactuals}

Counterfactuals—and counterfactual thinking more broadly—have evolved over centuries across philosophy, logic, and psychology, long before being formally defined. Though the term itself is modern, the underlying mode of reasoning has deep roots in classical systems of thought. The early history of counterfactual thinking can be traced to ancient traditions such as Greek and Indian philosophy, where reasoning about alternatives played a central role in debates about knowledge, causality, and inference (see Figure~\ref{fig:background_teaser}).

In Greek philosophy, thinkers like Aristotle and Plato examined subjunctive suppositions—claims about what could have happened but did not—to explore the logic and metaphysics of possibility \cite{counterfactuals_stanford}. Their inquiries laid the groundwork for later developments in modal and conditional reasoning. Philosophers from the Eleatic school, including Parmenides and Melissus, also used hypothetical premises in their deductive arguments to reflect on necessity and the nature of being \cite{parmenides_melissus_counterfactual}. Similar forms of reasoning appear in Indian philosophy, where hypothetical and suppositional methods were central to classical logic \cite{ganeri2001philosophy,matilal1998character}. The Nyāya school developed \textit{tarka}, a structured technique for testing propositions by considering imagined alternatives and their consequences. In parallel, Buddhist thinkers such as Nāgārjuna employed \textit{prasaṅga} arguments—reductio-style reasoning that derived contradictions from assumed premises—to challenge views on reality and causation \cite{dreyfus1997recognizing}. These frameworks evolved independently from Greek thought, yet reflected a similarly rigorous engagement with counterfactual logic.

In early modern philosophy, David Hume offered a critical turning point. Although known for his regularity theory of causation, Hume’s claim that “if the cause had not been, the effect never had existed” \cite{hume1748enquiry} prefigures counterfactual accounts of dependence. While rejecting metaphysical necessity, Hume relied on imagined alternatives to explain causal inference. His critics and successors extended this tradition. Thomas Reid emphasized causal necessity and direct perception, challenging Hume’s skepticism \cite{reid1788essays}, while Immanuel Kant, in the \textit{Critique of Pure Reason}, proposed that causal structure is imposed by the mind, introducing a modal basis for reasoning about cause and effect \cite{kant1781critique}. In the 19th century, John Stuart Mill formalized a method resembling counterfactual testing—the \textit{Method of Difference}—arguing that if removing a condition eliminates an effect, a causal connection is likely \cite{mill1843logic}. Alexander Bain further examined the psychological foundations of such reasoning \cite{bain1859emotions}.

These philosophical developments were formalized in analytic philosophy by David Lewis, who is widely credited with defining the modern semantics of counterfactuals. In \textit{Counterfactuals} \cite{lewis1973counterfactuals}, Lewis introduced a possible-worlds framework in which a counterfactual is true if its consequent holds in the closest possible world(s) where the antecedent is true. This model gave counterfactuals a precise logical structure and made them central to debates in metaphysics, decision theory, and language. By embedding counterfactuals within modal logic, Lewis helped transition them from intuitive thought experiments to formal tools of analysis.

Building on this foundation, counterfactuals became integral to modern theories of causality in statistics and computer science. Donald Rubin’s potential outcomes framework—known as the Rubin Causal Model (RCM)—defines causal effects as differences between outcomes under different treatment assignments for the same unit \cite{rubin1974estimating}. Judea Pearl advanced causal inference by introducing \textit{Structural Causal Models} (SCMs), which unify counterfactual reasoning, graphical models, and algorithmic tools for causal analysis \cite{pearl2009causality}. SCMs use directed acyclic graphs (DAGs) to represent assumptions about causal relationships and are paired with structural equations that encode how variables influence one another. Within this framework, causal and counterfactual queries can be expressed formally and computed algorithmically, provided the causal structure is identified.

A key conceptual innovation in Pearl's framework is the \textit{Ladder of Causation}, which organizes causal reasoning into three hierarchical levels:

\begin{enumerate}
    \item \textbf{Association (Level 1):}  
    This level describes patterns in data that we observe. For example, we might look at the probability of a person \textit{recovering} (\( Y = y= recovered\)) given that they \textit{took a drug} (\( X=x=took\space drug \)). This is written as \( P(Y=y \mid X=x) \).  
    \begin{itemize}
        \item \( X \): the observed input or condition, like “took the drug”  
        \item \( Y \): the observed outcome, like “recovered”  
       
    \end{itemize}
    This tells us whether two things are related (i.e., correlated), but not whether one caused the other.

    \item \textbf{Intervention (Level 2):}  
    This level is about understanding \textit{causal effects}—what happens when we actively \textit{intervene} and change something. For instance, we might ask: “What’s the chance of recovery if we make someone take the drug?” This is written as \( P(Y \mid \texttt{do}(X)) \).  
    \begin{itemize}
        \item The \texttt{do}(X) notation means we are forcing the condition \( X \) to happen (e.g., giving the drug), rather than just observing it.  
    \end{itemize}

    \item \textbf{Counterfactuals (Level 3):}  
    This level goes one step further and asks: \textit{“What would have happened if things had been different?”}  
    For example, suppose a person took the drug (\( X = x \)) and recovered (\( Y = y \)). A counterfactual question would be:  
    \textit{“Would they still have recovered if they had \textbf{not} taken the drug?”}  
    This is written as:  
    \[
    P(Y| X = x')
    \]  
    \begin{itemize}
        \item \( Y \): the outcome we care about (e.g., recovery)  
        \item \( X \): the input condition (e.g., took the drug or not)  
        \item \( x' \): a different, hypothetical condition (e.g., did not take the drug)  
        \item \(P(Y| X = x') \): the alternate outcome that would have happened under the hypothetical condition
    \end{itemize}
    This level allows reasoning about individual-level alternate realities, and is the most powerful form of causal inference.
\end{enumerate}
\vspace{1em}

This hierarchy underscores the increasing expressive power of causal reasoning: while associations describe patterns in the data, interventions evaluate causal effects, and counterfactuals allow retrospective analysis of alternative outcomes. Pearl’s formalism not only generalizes earlier potential-outcome frameworks such as Rubin's but also supports practical algorithms for identification, estimation, and counterfactual generation across scientific domains.

Pearl's ladder clarifies the increasing levels of abstraction and information required for different types of causal inference. His work generalizes and extends the Rubin model by offering a formal calculus for computing not just treatment effects, but also complex, personalized counterfactuals in both experimental and observational settings. Together, SCMs and the Ladder of Causation form the theoretical core of contemporary causal inference and have influenced applications across fields such as epidemiology, economics, policy evaluation, and artificial intelligence.

In machine learning and AI, counterfactual reasoning is increasingly used for tasks such as explaining model decisions, fairness, and causal modeling. For instance, counterfactual explanations can reveal which minimal changes to inputs would alter a model's decision \cite{wachter2017counterfactual}, while fairness frameworks assess whether outcomes shift unjustifiably across protected attributes \cite{kusner2017counterfactual}. These applications often build on formal foundations from causal inference \cite{ pearl2009causality}, and their broader role will be discussed in detail in the next section.

\section{Counterfactuals: Definitions}

\subsection {Counterfactuals in the Rubin Causal Model}
In the Rubin Causal Model (RCM), counterfactuals are represented as \textit{potential outcomes} for each unit under different treatment conditions. Let \( Y_i(1) \) denote the potential outcome for unit \( i \) if it were assigned to the treatment condition, and \( Y_i(0) \) denote the potential outcome if assigned to the control condition.

For any given unit \( i \), only one of these outcomes can be observed in practice, depending on the treatment assignment:
\[
Y_i^{\text{obs}} = 
\begin{cases}
Y_i(1), & \text{if } T_i = 1 \\
Y_i(0), & \text{if } T_i = 0
\end{cases}
\]
The unobserved outcome is referred to as the \textit{counterfactual}.

The \textbf{individual causal effect} for unit \( i \) is defined as the difference between the potential outcomes under treatment and control:
\[
\tau_i = Y_i(1) - Y_i(0)
\]
However, because we cannot observe both \( Y_i(1) \) and \( Y_i(0) \) for the same unit, individual causal effects are generally not identifiable. Instead, the focus is often on estimating the \textbf{average treatment effect} (ATE):
\[
\text{ATE} = \mathbb{E}[Y(1)] - \mathbb{E}[Y(0)]
\]

In this framework, a counterfactual is formally the potential outcome under a treatment condition that did not occur. While the Rubin model does not frame counterfactuals as logical conditionals, the underlying idea corresponds to evaluating what \textit{would have happened} had a unit received a different treatment than it actually did.

\subsubsection{Counterfactuals in Pearl's Structural Causal Models}

In Pearl’s framework, a counterfactual expresses what the value of an outcome variable \( Y \) would have been, had a treatment variable \( X \) been set to a particular value \( x \), possibly contrary to fact, within a specified structural causal model. This is denoted formally as \( Y_{x} \), and it enables reasoning about alternative scenarios conditioned on observed data. Counterfactuals differ from interventional queries such as \( P(Y \mid \texttt{do}(X = x)) \) in that they refer to unit-specific, hypothetical outcomes, rather than population-level causal effects. Interventional queries are defined at the distributional level, 
describing how outcomes change under intervention across populations or subpopulations.
Counterfactuals, in contrast, are unit-specific: they ask what would have happened to 
an individual data point under a different treatment. Thus, interventions capture average causal effects, while counterfactuals capture hypothetical 
alternatives for specific units.

A \textbf{Structural Causal Model (SCM)} is defined as a tuple:
\[
M = (\mathbf{U}, \mathbf{V}, \mathbf{F}, P(\mathbf{U}))
\]
where:
\begin{itemize}
    \item \( \mathbf{U} \) are exogenous (background) variables,
    \item \( \mathbf{V} \) are endogenous (observed) variables,
    \item \( \mathbf{F} \) is a set of structural equations \( \{V_i = f_i(\text{pa}_i, U_i)\} \), where \( \text{pa}_i \subseteq \mathbf{V} \cup \mathbf{U} \) are the parents of \( V_i \),
    \item \( P(\mathbf{U}) \) is a probability distribution over \( \mathbf{U} \).
\end{itemize}

To evaluate a counterfactual expression such as ``\( Y \) would be \( y \) if \( X \) were \( x \), given that we observed \( X = x' \) and \( Y = y' \),'' Pearl proposes a \textbf{three-step procedure}:

\begin{enumerate}
    \item \textbf{Abduction}: Given observations \( X = x', Y = y' \), update the model by inferring the posterior distribution \( P(\mathbf{U} \mid X = x', Y = y') \) over the exogenous variables.
    
    \item \textbf{Action}: Modify the model by performing an intervention \( \texttt{do}(X = x) \), replacing the structural equation for \( X \) with the constant \( X = x \).
    
    \item \textbf{Prediction}: Use the modified model and the inferred \( \mathbf{U} \) to compute the resulting distribution over \( Y \). The counterfactual query is evaluated as:
    \[
    P(Y_{x} = y \mid X = x', Y = y') = \sum_{u} P(Y_{x} = y \mid u) P(u \mid X = x', Y = y')
    \]
\end{enumerate}

This formalization allows counterfactuals to be treated as computable objects under well-defined assumptions about the data-generating process, providing a foundation for retrospective causal analysis, policy evaluation, and algorithmic accountability.

\section{Adavantages and Challenges of Counterfactual Reasoning}

Counterfactual reasoning provides a rigorous way of asking ``what if'' questions, offering both conceptual advantages and practical limitations. This section outlines its major benefits and the challenges that arise when applying it to real-world domains.

\subsection{Advantages}  
A central strength of counterfactual reasoning is its ability to move beyond statistical association and explicitly model causation. Whereas correlation describes patterns of co-occurrence, causation asks whether manipulating one variable would bring about a change in another. This distinction is particularly valuable in domains dominated by spurious associations, such as high-dimensional observational datasets.

Counterfactuals also enable reasoning at the level of individuals rather than populations. While average causal effects summarize treatment impacts across groups, counterfactuals ask what would have happened to a specific unit under an alternative condition. This unit-level perspective is crucial for generating recourse in machine learning \cite{wachter2017counterfactual}, auditing fairness for individuals and subgroups \cite{kusner2017counterfactual}, and designing personalized interventions in medicine and policy.  

Finally, counterfactual reasoning is inherently action-oriented. By linking interventions to outcomes, it provides a framework for accountability and decision support. In applied machine learning, this translates into actionable recommendations, fairness interventions, and interpretability tools that highlight which features are causally relevant to a model’s predictions.

\subsection{Challenges}  
Despite these benefits, several fundamental challenges limit the practical use of counterfactual frameworks. Estimating direct and indirect (mediated) effects requires strong assumptions about the causal graph and the absence of hidden confounding . In real-world, high-dimensional domains, specifying all relevant mediators is rarely feasible, making such estimates fragile and highly sensitive to model misspecification.

Another limitation lies in the treatment of exogenous variables. Both potential outcomes and structural causal models assume that background factors can be enumerated or probabilistically modeled. In practice, however, many relevant influences remain unobserved. These hidden confounders bias both interventional and counterfactual estimates, constraining the reliability of inferences. Even when sensitivity analyses or partial identification techniques are employed, conclusions remain bounded by untestable assumptions.  

Finally, counterfactual reasoning is highly model-dependent. Small errors in specifying structural equations or independence assumptions can lead to drastically different conclusions, raising concerns about robustness, reproducibility, and interpretability.  

\medskip
In summary, counterfactual reasoning combines unique strengths—clarifying causation, enabling individual-level reasoning, and supporting actionability—with significant challenges, particularly in complex, high-dimensional domains where hidden confounding and model misspecification remain persistent obstacles.

\section{A Taxonomy of Counterfactuals in Machine Learning}
This section presents a taxonomy of counterfactuals in machine learning based on three orthogonal dimensions. The first dimension, \emph{use-case}, categorizes counterfactuals according to their functional role—such as explanation, fairness auditing, recourse generation, or causal inference. The second dimension, \emph{modality}, distinguishes counterfactuals by the type of input data they operate on, including text, images, tabular data, or multimodal combinations. The third dimension, \emph{type of construction}, captures how counterfactuals are generated—ranging from input-level perturbations to concept-level modifications, and from adversarial to causally grounded interventions. This structured classification clarifies the diverse forms counterfactuals can take and provides a foundation for analyzing their design and application across tasks and domains.

\subsection{By Use-Case}

Counterfactual methods in machine learning have been deployed for various use-cases ranging from interpretability and fairness to causal inference and robustness. This section categorizes the literature by primary application domains and highlights key works in each.

\subsubsection{Interpretability }

Counterfactual explanations are a widely used approach for understanding and interpreting the decisions of complex machine learning models. They aim to answer the question: “What minimal change to the input features would change the model's prediction?”. By systematically changing features (whether input features or human defined concepts), counterfactual reasoning provides a framework for understanding the dependece of model output on these features. This use-case emphasizes transparency and user trust, especially in high-stakes domains like finance, healthcare, and law.

One of the earliest applications of counterfactuals for interpreting machine learning models was by Wachter et al. \cite{wachter2017counterfactual}, who proposed an optimization-based method for generating counterfactual explanations without requiring access to model internals. Their work emphasized the legal and philosophical relevance of such explanations, particularly in the context of GDPR (European Union's data privacy law that gives individuals control over their personal data and enforces strict rules on how organizations collect, use, and protect it)  and individual rights. Building on this foundation, subsequent works have extended counterfactual reasoning in various directions—seeking not only faithful explanations, but also those that are diverse \cite{mothilal2020explaining}, logically grounded \cite{russell2019efficient}, multi-objective \cite{dandl2020multi}, actionable \cite{poyiadzi2020face}, and causally consistent \cite{karimi2020model}. Others have focused on enhancing human interpretability \cite{lucic2022focus}, addressing hidden confounding \cite{barocas2020hidden}, offering diagnostic black-box tools \cite{sharma2020certifai}, or grounding explanation quality in cognitive principles \cite{keane2020good}. Together, these efforts reflect a growing recognition of counterfactuals as a versatile and strong lens for understanding machine learning models.

\subsubsection{Fairness and Bias Auditing}

Counterfactuals are a powerful tool for assessing algorithmic fairness. The core idea is to evaluate whether a model's prediction would remain consistent had a sensitive attribute (e.g., race, gender) been different, while holding all else equal. This supports definitions of fairness grounded in causal invariance or conditional independence.

A core central theme to this idea is the definition of counterfactual fairness, introduced by Kusner et al. \cite{kusner2017counterfactual}. Counterfactual fairness states that a decision is fair towards an individual if the decision would have remained the same in a counterfactual world where the individual belonged to a different demographic group (e.g., race or gender), while keeping all other non-protected attributes constant. Since then, this idea has evolved through a variety of extensions. Scholars have introduced refinements that isolate unfair influence along specific causal pathways \cite{chiappa2019path}, integrated counterfactual reasoning into the training of fair representations through adversarial learning \cite{madras2019fairness}, and applied the framework to domain-specific settings like text classification and NLP auditing \cite{garg2019counterfactual,hall2020conditional}. Other efforts have explored semantic perturbation techniques to probe visual model fairness \cite{black2021fliptest}, explicitly constructed image-based counterfactuals to reveal bias in generative models \cite{balagopalan2022constructing}, and used simulation-based evaluations to analyze multimodal systems \cite{zhao2022evaluating}. Alongside this, surveys and methodological innovations have broadened the use of counterfactuals for fairness auditing and data augmentation during model training \cite{mehrabi2021survey,wu2021counterfactual}. Together, these contributions demonstrate how counterfactual reasoning has become a unifying thread across modalities and fairness paradigms, enabling both diagnostic and corrective interventions in machine learning systems.

%Kusner et al. \cite{kusner2017counterfactual} introduced the foundational notion of \textit{counterfactual fairness}, ensuring predictions are invariant across counterfactual worlds where only sensitive attributes change. Chiappa \cite{chiappa2019path} refined this idea with \textit{path-specific counterfactual fairness}, allowing one to isolate unfair causal pathways. Madras et al. \cite{madras2019fairness} incorporated causal reasoning into fairness-aware representations using adversarial training. Garg et al. \cite{garg2019counterfactual} studied counterfactual fairness in text classifiers, demonstrating robustness under perturbations of identity words. Hall et al. \cite{hall2020conditional} proposed \textit{conditional counterfactuals} to audit NLP models using targeted substitutions. Black et al. \cite{black2021fliptest} introduced FlipTest, a semantic perturbation framework to probe unfairness in vision systems. Balagopalan et al. \cite{balagopalan2022constructing} emphasized fairness in image generation by constructing explicitly controlled counterfactuals. Zhao et al. \cite{zhao2022evaluating} proposed counterfactual simulations to evaluate biases in large vision-language models. Mehrabi et al. \cite{mehrabi2021survey} reviewed multiple counterfactual fairness paradigms across modalities. Wu et al. \cite{wu2021counterfactual} proposed counterfactual data augmentation techniques for improving fairness during model training.

%These works span causal modeling, data augmentation, and adversarial training, all unified by a counterfactual lens to fairness.

\subsubsection{Algorithmic Recourse}
Building on the foundation of counterfactual reasoning, recourse-oriented counterfactuals aim to provide individuals with actionable guidance for changing unfavorable model outcomes. Research in this area has explored several key directions. Some works focus on generating sparse and feasible interventions using optimization-based methods \cite{ustun2019actionable}, while others emphasize causal consistency to ensure recommended changes are not only effective but also realistic within the constraints of the real world \cite{karimi2020model}. A separate line of work frames recourse as a problem of identifying plausible transitions along data manifolds, capturing feasibility in high-dimensional settings \cite{poyiadzi2020face}. Together, these efforts treat counterfactuals not just as explanations, but as practical tools that can help users understand and potentially improve their outcomes.

%Recourse-oriented counterfactuals provide actionable recommendations for individuals to change an unfavorable model decision. Ustun et al. \cite{ustun2019actionable} introduced a linear programming framework for generating sparse, feasible recourse. Karimi et al. \cite{karimi2020model} developed model-agnostic, causal recourse techniques. Poyiadzi et al. \cite{poyiadzi2020face} framed recourse as a geodesic problem in data manifolds. Wachter et al. \cite{wachter2017counterfactual} also motivated recourse by proposing human-readable changes to inputs. Raghavan et al. \cite{raghavan2022multi} extended counterfactual generation to multi-class and multi-outcome settings. 

\subsubsection{Robustness and Adversarial Testing}

In robustness testing, counterfactuals are widely used to probe a model’s sensitivity to small or structured input changes. These perturbations can expose vulnerabilities in how models generalize, even when the changes appear imperceptible to humans \cite{goodfellow2015explaining}. Over time, this idea has been extended to test robustness under broader distributional shifts, using stress tests and systematic evaluations across a range of inputs \cite{hendrycks2019benchmarking}. In vision and language, counterfactual examples have been used to highlight misclassifications and reveal brittle behavior in models \cite{morris2021reassessing, zmigrod2019counterfactual}. Increasingly, counterfactuals are also integrated with interpretability frameworks to offer more robust and transparent model explanations \cite{ribeiro2020beyond}. These trends reflect a shift from using counterfactuals solely for adversarial attacks to employing them as diagnostic tools for improving model reliability.

\subsubsection{Causal Inference and Treatment Effect Estimation}

Counterfactuals are fundamental to causal inference, where they enable reasoning about hypothetical interventions and treatment effects. The potential outcomes framework defines causal effects as differences between actual and counterfactual outcomes \cite{rubin1974estimating}, while structural causal models formalize this reasoning through graphical models and do-calculus \cite{pearl2009causality}. Recent advances leverage machine learning to improve counterfactual estimation under challenges like covariate shift. Approaches include representation learning for balanced comparisons \cite{shalit2017estimating, johansson2016learning}, latent variable models for handling hidden confounders \cite{louizos2017causal}, and generative models for estimating individual treatment effects \cite{yoon2018ganite}. These developments reflect a shift toward scalable, model-based causal inference rooted in counterfactual logic.

\subsection{By Input Modality}

The form and challenges of counterfactual reasoning vary significantly across input modalities. In this section, we categorize counterfactual methods by the type of data they operate on: text, vision, and tabular. Each modality imposes different constraints on how counterfactuals can be generated, evaluated, and interpreted.

\subsubsection{Text}

In language tasks, counterfactual generation is constrained by grammar, coherence, and semantic consitency. Even minor edits can drastically alter meaning, making fluency-preserving perturbations non-trivial. Methods have focused on identity-sensitive edits \cite{garg2019counterfactual}, morphological control \cite{zmigrod2019counterfactual}, and structured interventions for fairness and auditing \cite{hall2020conditional, wu2021counterfactual, lu2022learn}. More recent approaches aim to improve interpretability in sequence models \cite{rieger2020interpretations}, with surveys reviewing generation and evaluation frameworks tailored to text \cite{zhang2023survey}.

\subsubsection{Vision}

In vision, counterfactuals require semantically meaningful changes that remain photorealistic and causally informative. This often involves disentangling latent features like gender or emotion from pixel-level noise. Approaches range from attribute editing in face images \cite{singla2019explanation, balagopalan2022constructing} to counterfactual scene composition for VQA \cite{goyal2019counterfactual} and concept activation maps for interpretability \cite{chang2019explaining}. Additional methods probe bias in multimodal models \cite{zhao2022evaluating}, robustness to visual perturbations \cite{morris2021reassessing}, and interventions in generative latent spaces \cite{bau2019gandissect}.

\subsubsection{Tabular}
Tabular data offers the most direct setting for counterfactuals due to its structured and often low-dimensional nature. Optimization-based methods dominate this space, focusing on feasibility, diversity, and actionable recourse \cite{wachter2017counterfactual, ustun2019actionable, mothilal2020explaining}. Others integrate causal constraints \cite{karimi2020model}, manifold-aware interpolation \cite{poyiadzi2020face}, or human-aligned filtering mechanisms \cite{lucic2022focus, dandl2020multi}.

\subsection{By Type of Construction}

Counterfactuals differ not only in what they are used for or what data they operate on, but also in how they are constructed. This section distinguishes counterfactuals based on their construction strategy—ranging from low-level input perturbations to high-level semantic edits, and from causally grounded methods to adversarial variants.

\subsubsection{Input-Level Counterfactuals}

Input-level counterfactuals involve direct changes to the input features, often constrained to minimal perturbations. These are among the earliest and most practical forms, particularly in tabular and text domains.

Wachter et al. \cite{wachter2017counterfactual} proposed a gradient-based method to find minimally different inputs that flip model predictions. Dandl et al. \cite{dandl2020multi} formulated counterfactual generation as a multi-objective optimization problem over proximity and plausibility. Mothilal et al. \cite{mothilal2020explaining} emphasized diversity in generated counterfactuals for tabular classifiers. Russell \cite{russell2019efficient} used constraint solvers to generate logical and sparse counterfactuals. Poyiadzi et al. \cite{poyiadzi2020face} introduced geodesic counterfactuals that traverse realistic paths in the data manifold. Lucic et al. \cite{lucic2022focus} filtered input-level counterfactuals using human-aligned metrics. Dhamdhere et al. \cite{dhamdhere2019explaining} applied input perturbations for attribution analysis in saliency maps.

\subsubsection{Concept-Level Counterfactuals}

Concept-level counterfactuals manipulate high-level semantic factors (e.g., gender, pose, sentiment) that underlie the input. This often requires disentangled representations or causal structure.

Goyal et al. \cite{goyal2019counterfactual} generated concept-level visual counterfactuals in VQA using scene editing. Karimi et al. \cite{karimi2020model} imposed causal constraints to ensure edits preserve structural dependencies. Upadhyay et al. \cite{upadhyay2021robustness} applied interventions to disentangled text representations. Balagopalan et al. \cite{balagopalan2022constructing} developed image counterfactuals that modify specific concepts (e.g., skin tone). Mahajan et al. \cite{mahajan2019preserving} created counterfactuals that preserve task-specific concepts in vision tasks. Gauthier et al. \cite{gauthier2021counterfactual} applied concept-level edits for training robust language classifiers. Kim et al. \cite{kim2022counterfactual} proposed counterfactuals based on shifting latent attributes in multimodal inputs.

\subsubsection{Adversarial vs. Causal Counterfactuals}

Adversarial counterfactuals aim to test model robustness by finding near-indistinguishable inputs that cause misclassification. In contrast, causal counterfactuals focus on hypothetical scenarios rooted in structural causal models.

Goodfellow et al. \cite{goodfellow2015explaining} introduced adversarial examples via gradient-based attacks. Ribeiro et al. \cite{ribeiro2018anchors} proposed perturbation-based local explanations that test model stability. Hendricks et al. \cite{hendricks2018grounding} evaluated adversarial counterfactuals in image captioning. Pearl \cite{pearl2009causality} formalized causal counterfactuals using structural equations. Kusner et al. \cite{kusner2017counterfactual} defined counterfactual fairness via causal graphs. Chiappa \cite{chiappa2019path} developed path-specific counterfactuals. Karimi et al. \cite{karimi2020algorithmic} compared adversarial and causal counterfactuals for recourse. Mahajan et al. \cite{mahajan2023generating} applied structural interventions to generate interpretable counterfactuals across modalities.

\section{Fairness in Machine Learning}

\subsection{What Is Fairness?}

Fairness in machine learning broadly refers to the principle that algorithmic systems should not produce systematically discriminatory outcomes against individuals or groups based on socially salient attributes such as gender, race, caste, or socioeconomic status. The formalization of fairness has led to a proliferation of definitions, each capturing a different normative notion of equity. Three primary families of definitions dominate the literature:

\begin{itemize}
    \item \textbf{Group Fairness} focuses on ensuring parity of outcomes across different groups defined by protected attributes. Common criteria include:
    \begin{itemize}
        \item Statistical Parity (Demographic Parity): The predicted outcome should be independent of the sensitive attribute \cite{feldman2015certifying}.
        \item Equalized Odds: The model's true and false positive rates should be equal across groups \cite{hardt2016equality}.
        \item Equal Opportunity: A relaxation of equalized odds, requiring only equal true positive rates \cite{hardt2016equality}.
    \end{itemize}
    
    \item \textbf{Individual Fairness} posits that similar individuals should receive similar outcomes, operationalized through a task-specific similarity metric \cite{dwork2012fairness}.

    \item \textbf{Counterfactual Fairness} requires that a model's decision remain unchanged in a counterfactual world where the individual’s protected attribute had been different \cite{kusner2017counterfactual}.
\end{itemize}

These definitions are often mutually exclusive \cite{kleinberg2016inherent}, necessitating careful trade-offs in deployment contexts. Recent work explores relaxations and hybrid notions to better reflect practical constraints \cite{garg2020fairness, roessler2022trading}.
\subsection{Fairness in Natural Language Processing}

Natural Language Processing (NLP) models are known to inherit—and often exacerbate—social biases embedded in the large-scale corpora they are trained on. Multiple forms of such bias have been documented. For instance, coreference resolution systems tend to resolve gender-neutral roles like “the doctor” disproportionately as male \cite{zhao2018gender}. Word embeddings such as Word2Vec encode stereotypical analogies (e.g., “man:computer as woman:homemaker”), which can propagate harmful associations in downstream tasks \cite{bolukbasi2016man}. Language generation models have been shown to produce toxic or biased continuations when prompted with identity-related terms \cite{sheng2019woman}. Sentiment analysis systems, too, can demonstrate disparate performance across linguistic varieties like African-American Vernacular English (AAVE), reflecting socio-linguistic biases in training data \cite{sap2019risk}.

To address these challenges, a range of intervention strategies have been proposed. One early line of work focused on debiasing word embeddings by removing gender-associated subspaces \cite{bolukbasi2016man}. Other methods employ counterfactual data augmentation (CDA), generating paired sentences by swapping identity terms to ensure balanced model behavior across demographics \cite{zmigrod2019counterfactual, prabhu2023counterfactual}. More recently, causal probing techniques have been developed to measure and suppress the influence of sensitive attributes within model internals, either by adversarially minimizing their predictive power or by identifying and erasing specific latent features \cite{elazar2018adversarial, vig2020investigating}. These strategies reflect a growing effort to build fairer and more socially aware NLP systems.

Recent trends in large language model fairness focus on aligning generation with human values, auditing RLHF training pipelines \cite{ganguli2022red}, and building explainable fairness probes \cite{liang2022holistic}.

\subsection{Fairness in Computer Vision}

Fairness in computer vision has become an increasingly important topic due to the deployment of vision systems in socially sensitive contexts such as surveillance, hiring, medical imaging, and content moderation. Unlike structured tabular or textual data, visual data is high-dimensional, opaque, and difficult to semantically disentangle, making bias detection and mitigation particularly challenging. Moreover, societal power structures are often reflected and amplified in visual data, especially when datasets and benchmarks are scraped from the internet without auditing or curation \cite{birhane2021multimodal, denton2019image}.

\vspace{1em}
\noindent \textbf{Facial Recognition Biases.}
The most cited work in this space is the ``Gender Shades" study by Buolamwini and Gebru, which demonstrated stark disparities in commercial gender classification systems. Accuracy was as high as 99\% for lighter-skinned males but dropped below 65\% for darker-skinned females \cite{buolamwini2018gender}. This exposed the inadequacies of training data and systemic blind spots in commercial model pipelines. Subsequent work by Raji et al. \cite{raji2019actionable} further showed how public audits of model disparities led to measurable improvements, emphasizing the importance of transparency and third-party accountability.

\vspace{1em}
\noindent \textbf{Object Detection and Classification Biases.}
Wang et al. \cite{wang2020towards} demonstrated that object recognition systems trained on datasets like  COCO \cite {lin2014microsoft} and Open Images \cite{kuznetsova2020open} tend to entrench gender stereotypes—for instance, associating women with kitchen scenes and men with sports or machinery. Even when gender is not explicitly labeled, these associations emerge implicitly through correlated visual context. Such biases propagate downstream in captioning, summarization, and decision-support tasks.

\vspace{1em}
\noindent \textbf{Biases in Generative Models.}
Fairness challenges are even more pronounced in generative systems. Denton et al. \cite{denton2019image} and Birhane and Prabhu \cite{birhane2021multimodal} highlighted how training data scraped from uncurated internet sources reproduces misogynistic, racist, and class-based stereotypes in image generation. For example, prompts like “CEO” might yield mostly white male images, while prompts like “nurse” might yield primarily women of color. Zhao et al. \cite{zhao2021ethical} analyzed such systems to show how generative models reflect societal stereotypes in a feedback loop, raising complex ethical and epistemic questions about representation. Existing evaluation frameworks such as T2IAT \cite{wang2023t2iat}, DALL-Eval \cite{cho2023dall}, and other studies \cite{ghosh2023person, bianchi2023easily, friedrich2023fair} primarily assess biases along predefined axes, such as gender \cite{wang2023t2iat, cho2023dall, esposito2023mitigating, bianchi2023easily}, skin tone \cite{wang2023t2iat, cho2023dall, ghosh2023person, esposito2023mitigating, bianchi2023easily}, culture \cite{esposito2023mitigating, wang2023t2iat}, and geographical location \cite{esposito2023mitigating}. 
While these works offer key insights into single-axis bias detection and mitigation, they lack a systematic examination of how biases on one axis influence another—a core aspect of intersectionality. 

\vspace{1em}
\noindent \textbf{Intervention Strategies.}
To improve fairness and transparency in model predictions, various intervention strategies have emerged that modify datasets, model architectures, or interpretability tools to better account for sensitive attributes and human-aligned reasoning.

One class of approaches focuses on improving data quality and transparency. Structured documentation practices aim to make dataset collection and labeling processes more transparent and auditable \cite{gebru2018datasheets}, while data resampling and relabeling techniques have been used to reduce biased correlations, such as gender misclassification in image captions \cite{hendricks2018women}.

Another strategy involves generating controlled counterfactuals by manipulating specific input attributes. These interventions alter sensitive features like skin tone or hairstyle while keeping the overall content unchanged, enabling targeted fairness audits and behavior testing under attribute shifts \cite{bashir2021counterfactual}.

A third direction probes models at the conceptual level. Concept activation techniques allow researchers to quantify how much certain human-defined concepts influence model predictions \cite{kim2018interpretability}. Building on this, more recent approaches introduce concept bottlenecks that constrain the model to reason explicitly through interpretable features, leading to more faithful and fairness-preserving predictions \cite{yuksekgonul2022post}.

\vspace{1em}

\noindent \textbf{Ongoing Challenges and Directions.}
Despite this progress, vision fairness remains a moving target. There is no universal definition of fairness that suits all visual tasks. Moreover, techniques like concept bottlenecks and dataset balancing require human input, which reintroduces subjectivity. Another emerging area is fairness-aware diffusion models, where causal interventions are embedded directly into the generative process. Community-wide initiatives are needed to design benchmarks, metrics, and co-designed datasets that reflect global and intersectional notions of fairness.

Vision fairness remains an evolving area, with calls for more robust evaluation metrics, audit toolkits, and participatory approaches to dataset governance \cite{scheuerman2020datasets}.
\subsection{Intersectionality of Biases}
Intersectionality, introduced by Crenshaw \cite{crenshaw1989demarginalizing}, describes how multiple forms of oppression—such as racism, sexism, and classism—intersect to shape unique experiences of discrimination. Two key models define this concept: the additive model, where oppression accumulates across marginalized identities, and the interactive model, where these identities interact synergistically, creating effects beyond simple accumulation \cite{curry2018killing}. In the context of AI, most existing work \cite{diana2023correcting,kavouras2023fairness,kearns2018preventing, pmlr-v80-hebert-johnson18a} aligns more closely with the additive model, focusing on quantifying and mitigating biases in intersectional subgroups. This perspective has influenced fairness metrics \cite{diana2021minimax,foulds2020intersectional,ghosh2021characterizing} designed to assess subgroup-level performance, extending across various domains, including natural language processing (NLP) \cite{lalor2022benchmarking,lassen2023detecting,guo2021detecting,tan2019assessing} and recent large language models \cite{kirk_bias_2021,ma2023intersectional,devinney2024we,bai2025explicitly}, multimodal research \cite{howard2024socialcounterfactuals,hoepfinger2023racial}, and computer vision \cite{wang2020towards, steed2021image}. These approaches typically measure disparities across predefined demographic intersections and propose mitigation strategies accordingly. Our work aligns with the interactive model of intersectionality, using counterfactual-driven causal analysis in TTI models. Beyond subgroup analysis, we intervene on a single bias axis to assess its ripple effects on others, revealing independences and interactions.

\section{Explainability in Machine Learning}

\subsection{Motivations and Challenges}

As machine learning models become increasingly complex and opaque—especially deep neural networks—explainability has emerged as a central requirement for trustworthy AI systems. Explainability enables stakeholders to understand, trust, and audit model decisions. It is crucial not only for scientific insight and debugging but also for meeting regulatory and ethical standards, such as the ``right to explanation'' in the GDPR. However, creating faithful, interpretable, and user-aligned explanations remains a major technical and philosophical challenge.

\subsection{Approaches to Explainability}

Explainability techniques in machine learning can be broadly categorized into four major families: post-hoc attribution methods, concept-based explanations, visual counterfactual explanations, and natural language justifications. Each of these families offers a different lens into the model's decision-making process and is suited to different audiences and use-cases.

\textbf{Post-hoc attribution methods} attempt to explain a model's decision by assigning importance scores to individual input features. One of the earliest and most widely used methods is LIME \cite{ribeiro2016should}, which fits an interpretable linear model locally around the input to approximate the decision boundary. SHAP \cite{lundberg2017unified} builds on game-theoretic principles and uses Shapley values to ensure additive and consistent feature attributions. Integrated Gradients \cite{sundararajan2017axiomatic} offers an axiomatic approach by computing the integral of gradients along a straight-line path from a baseline to the input. DeepLIFT \cite{shrikumar2017learning} compares activations relative to a reference point to attribute differences in output scores. Layer-wise Relevance Propagation (LRP) \cite{bach2015pixel} decomposes the prediction by propagating relevance scores backward through the network layers. SmoothGrad \cite{smilkov2017smoothgrad} further improves the robustness of gradient-based saliency maps by averaging gradients over multiple noisy samples. While these methods are model-agnostic or gradient-based, they often lack stability and may fail to align with human intuitions.

\textbf{Concept-based explanation methods} move beyond raw features to attribute decisions to human-interpretable concepts. Testing with Concept Activation Vectors (TCAV) \cite{kim2018interpretability} computes directional derivatives of network predictions with respect to high-level concept vectors defined by curated example sets.  Automated Concept Extraction (ACE) \cite{ghorbani2019towards} extends this by clustering activations to discover coherent concepts in a completely unsupervised manner. Concept Whitening \cite{chang2020concept} reorients latent spaces to make concept directions orthogonal, enabling direct control and interpretation. Self-Explaining Neural Networks (SENN)  \cite{alvarez2018towards} builds interpretable models that decompose predictions into combinations of learned concepts with associated relevance scores. ProtoPNet  \cite{chen2019looks} learns prototypical patches from training data and uses similarity to these prototypes for classification, making the reasoning process visually intuitive. Another important direction is concept bottleneck models \cite{koh2020concept}, which require models to make predictions through a set of intermediate, user-defined concepts, thereby enforcing semantic transparency.

\textbf{Visual counterfactual explanations} aim to identify features that causally influence model decisions by generating alternative inputs that lead to different predictions. This often involves modifying images in a way that preserves semantic meaning while altering the outcome. Early methods achieved this by editing specific regions of an image to induce label changes \cite{goyal2019counterfactual}, while more advanced approaches learn disentangled latent representations to produce controlled interventional edits in videos \cite{wu2021dive} or high-dimensional image spaces \cite{pawelczyk2020learning}. Generative models such as GANs have also been employed to synthesize realistic counterfactuals for fairness assessment \cite{nemirovsky2020countergan}, and prompt-based interventions have been used to audit text-to-image systems for bias \cite{bashir2021counterfactual}. More recent work takes a perturbation-based approach, directly optimizing over image regions to uncover minimal deletions that cause prediction shifts \cite{chang2022fido}.

\textbf{Natural language explanations} provide accessible, human-readable insights into model behavior, especially in NLP and multimodal tasks. Several efforts have paired model predictions with commonsense-driven justifications to improve transparency and trust, often via datasets and models designed to output textual rationales \cite{rajani2019explain}. In visual settings, explanation frameworks combine language with visual grounding, linking answers with natural justifications and attention maps \cite{huk2018vqa, park2018multimodal}. Recent developments leverage large language models to generate chain-of-thought ``self-talk" explanations that enhance both interpretability and generalization \cite{lampinen2022can}. Despite their accessibility, concerns remain about the faithfulness of these explanations; empirical studies suggest that many textual justifications do not accurately reflect the model’s internal decision process \cite{wiegreffe2021measuring}.

Together, these techniques form the backbone of modern explainable AI research. However, each method has limitations in terms of stability, completeness, or faithfulness, and ongoing research seeks to unify causal reasoning, human alignment, and transparency within a coherent framework.

\chapter{CAVLI: Using counterfactuals to quantify concept influence on classifier decisions}
\epigraph{

		``We do not see things as they are, we see things as we are."
}{\textit{Anaïs Nin}}

\begin{figure*}[ht]
    \centering
     \includegraphics[trim=40 80 150 130, clip,width=0.9\textwidth]{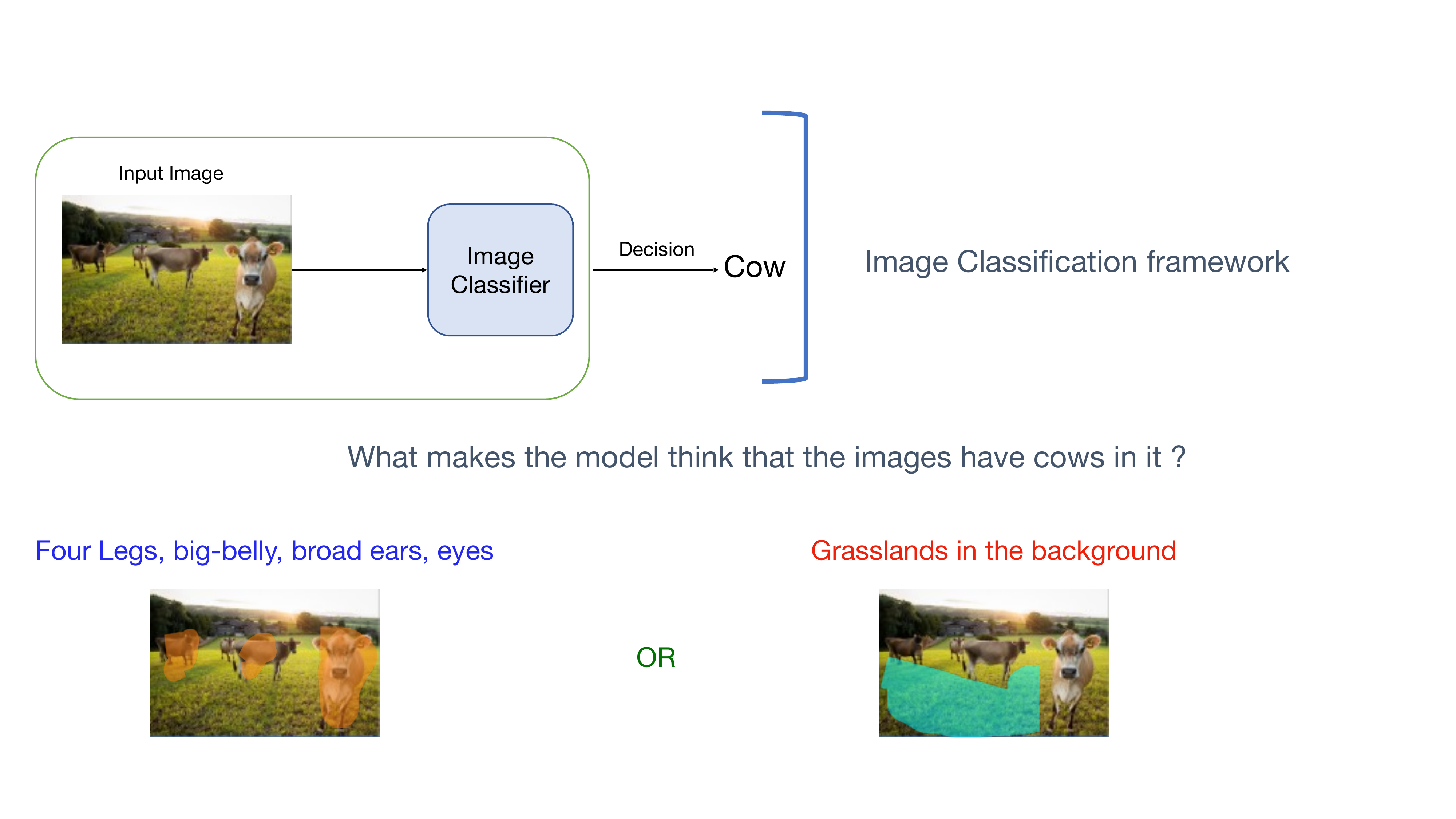}
     \captionof{figure}{This chapter begins with a central question: When a black-box model predicts a label (e.g., “Cow”), how can we determine which meaningful, human-interpretable concepts influenced that decision? For instance, did the model recognize the image as a cow because of meaningful features like its distinctive body structure, or was the decision influenced by spurious cues such as the presence of grasslands in the background? To address this, we introduce a method that quantifies the extent to which a model relies on human-defined concepts—offering a more interpretable alternative to conventional pixel-level explanations.}
     
     \label{fig:CAVLI_teaser}
\end{figure*}
\clearpage 

\section{Introduction}
\label{sec:CAVLI_intro}
Despite their widespread use, deep neural networks remain difficult to interpret. These models often behave as black boxes, making it hard to explain how they arrive at specific outputs. This lack of interpretability poses significant risks, especially when models are deployed in high-stakes areas such as healthcare, law enforcement, or financial decision-making. In such cases, the cost of a wrong prediction can be substantial, making transparency and accountability essential. Clear explanations not only help identify errors but also support human oversight and allow timely intervention when needed.

As discussed in earlier chapters, counterfactual reasoning provides a practical lens to examine model decisions. A counterfactual asks what would have happened if certain parts of the input were changed. Systematically changing the input and observing the changes in the output allows us to see what how removal or addition of certain features or concepts impact the model’s outcome. Counterfactuals help bridge the gap between the model's internal logic and human intuition. They shift the focus from static attributions to dynamic, intervention-based insights. This chapter builds on that foundation to explore a specific application of counterfactuals—measuring the role of human concepts in the decisions made by vision models.

In computer vision, this task is particularly challenging because the inputs are pixel arrays that lack an obvious mapping to high-level human concepts such as age, gender, or clothing type. Consider, for example, a binary classifier that detects whether a person is smiling. The input is an image, and the output is a yes-or-no answer. While the decision may appear simple, the factors driving it might include complex and implicit cues such as facial structure, hairstyle, or even background context. We may want to ask: Is the classifier making the decision based on genuine smile features, or is it influenced by confounding cues like gender or lighting?

To address such questions, we introduce a method called CAVLI that combines two popular approaches called T\textbf{CAV} and \textbf{LI}ME. This framework aims to quantify how much a neural network’s decision relies on a given human-defined concept. The key idea is to measure the overlap between image regions that contribute to the model’s decision and regions that are representative of the concept. For instance, we may want to test if a model identifying a cow relies heavily on the presence of grass in the background. If the model often uses grass as a signal to detect buffalo, it reveals a potential shortcut or bias in decision-making.

\begin{figure}
    \centering
  \includegraphics[scale=0.55]{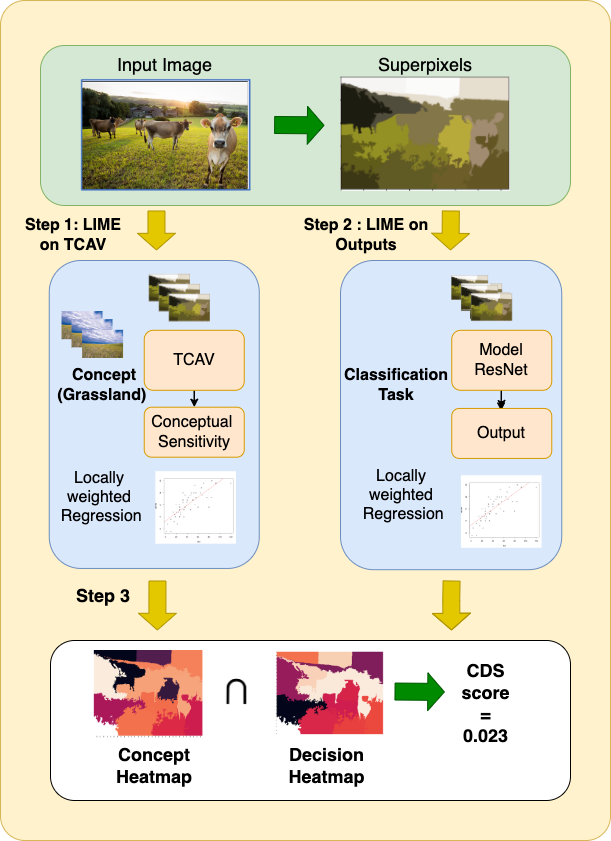}
  \caption{Overview of our proposed approach, CAVLI, to estimate the dependence of a concept (e.g., “grassland”) on a decision (e.g., “cow detection”). After decomposing the input image into superpixels, in Step 1 we find the regions of the image that have the highest association with the concept, defined by a set of images. In Step 2 we identify image regions with the highest involvement in the classification decision. Finally, we measure the overlap between the two in order to quantify the dependence.}
  \label{fig:block}
\end{figure}

As depicted in Figure \ref{fig:block} the high-level architecture of our approach consists of three main steps. In Step 1, we identify the regions of an image that have the highest association with a given human concept. We use a heatmap to highlight the regions that are most relevant to the concept of interest. In Step 2, we identify the image regions that have the highest involvement in the decision-making process of the neural network for a given input image. To achieve this, we use attribution methods to compute the contribution of each image region to the final decision. Finally, in Step 3, we measure the overlap between the image regions identified in Steps 1 and 2. This overlap gives us a measure of the degree to which the decision of the neural network is dependent on the human concept of interest. 

Our framework is based on the insight that if a model's decision depends on a concept, then there should be a high overlap in regions that are used by the model for decision-making and regions of the image that is used for concept modelling. For example (Figure \ref{fig:block}), if parts of the image that are used for classifying whether an animal is a cow or not are also associated with the concept of grassland, then the decision \texttt{cow} depends on the concept \texttt{grassland}. Our approach introduces the Concept Dependency Score (CDS), which quantifies the reliance of decisions on visual concepts. Apart from CDS we also present visual heatmaps that illustrate the overlap between decisions and concept regions. These heatmaps provide a visual understanding, enabling humans to see the extent to which a decision depends on a particular concept. The proposed approach, which combines Conditional Dependency Sampling (CDS) with heatmaps, offers multiple advantages. Firstly, the heatmap enables a computer vision practitioner to understand the overlap between regions used for decision-making and the regions associated with a given concept. This overlap is quantified using CDS. For example, a wildlife classifier should detect cows based on shape and texture, not just the background. If the CDS shows a high dependence on the background, it may indicate a dataset bias or shortcut learning. In other domains like medical imaging, such methods can reveal whether a diagnosis is based on relevant tissue features or artifacts in the scan.

Once a practitioner uses CAVLI, they can inspect a model’s decision alongside the Concept Dependency Score (CDS) and a visual heatmap to assess whether the decision aligns with the intended concept. The CDS quantifies reliance on the concept, while the heatmap shows where the model "looked" in the image—together offering a clear, interpretable summary of the model’s reasoning. Together, heatmaps and CDS can inform a practitioner whether a given decision made by the classifier is dependent on a visual concept and, if so, what regions in the image used for decision-making were associated with the concept. This visualization using heatmaps, coupled with the quantification using CDS, can serve as a yardstick for practitioners who want to ensure that decisions made by a computer vision model are based on the right concept. Higher CDS scores with desired or task-relevant concepts indicate that the model is relying on meaningful information to make its decision—such as focusing on facial features to detect a smile or using anatomical regions in a medical scan for diagnosis. This alignment suggests that the model has learned the intended patterns and is reasoning in a human-aligned way. In contrast, high CDS scores with irrelevant or sensitive concepts suggest potential problems. If the concept is social (e.g., gender, race), it may indicate bias, raising fairness concerns. If the concept is contextually irrelevant (e.g., background scenery in wildlife classification), it may signal that the model has picked up on spurious correlations—shortcuts that could fail under distribution shifts or real-world deployment. A classic example is ensuring that a model does not detect a buffalo simply because there is grass in the background. 

%Another utility of this approach lies in high-stakes scenarios, such as a court of law or medical diagnosis, where decisions made by AI models need to be well justified. Since our approach measures the overlap between image regions used for decision-making and image regions associated with visual concepts, and quantifies the dependence of a decision on a visual concept, it enables humans to inspect the dependence of a decision on different visual concepts and justify whether the decision is dependent on the right set of visual concepts or not.

Our method aligns with the broader goals of this thesis—leveraging counterfactual reasoning to improve the transparency and accountability of computer vision models. As part of our approach, we systematically blur different regions of an image to generate a diverse set of modified inputs. By observing how the model’s output changes when specific parts of the image are masked, we can quantify the decision’s dependence on those regions. This intervention-based process mirrors the counterfactual reasoning framework introduced in Chapter 1, where inputs are deliberately altered to assess their causal influence on the model’s prediction. By connecting decisions to concepts using counterfactual reasoning, CAVLI allows for systematic inspection and correction of model behavior. It supports both debugging and compliance in scenarios where accountability matters. This approach reflects the spirit of counterfactual reasoning: to ask not just what the model predicts, but why it predicts it—and how those reasons relate to human-understandable concepts.

\section{Methodology}
\label{alg:CAVLI_method}
\subsection{Notation}

Consider a trained neural network $F:X \rightarrow \{1,....,K\}$, on a dataset $X=\{\mathbf{x_1,x_2,..,.x_t}\}$ and associated labels $Y=\{y_1,y_2,...,y_t\}$, where $y_i$ $\in$ $\{0,1\}^{K}$ with $K$ classes. $F_k(\mathbf{x_i}):=h_l(f_l(\mathbf{x_i}))$, where $f_l\mathbf{(x_i)}$ are the output logits  of the $l^{\textrm{th}}$ layer and $h_l$ is the activation function of the $l^{\textrm{th}}$ layer.

\subsection{TCAV and LIME}
\label{alg:CAVLI_TCAV_and_LIME}
%Testing with Concept Activation Vectors}
TCAV \cite{kim2018interpretability} uses human-defined \textit{concepts} (e.g.,  ``gender'' or ``stripes'') instead of input features to provide explanations for a machine learning model.  To express a concept it finds a Concept Activation Vector (CAV) $ a\in \mathbb{R}^d$ (a layer with dimension $d$)  in the network's activation space \cite{schrouff2021best} that points in the direction of the concepts. This is achieved by training a classifier that distinguishes concept activations (``striped'' or ``dotted'') from activations of negative samples and taking a unit norm vector $\mathbf{v_c}$ orthogonal to its decision boundary. The inner product in Equation 1. denotes the similarity of the activstion to the required concept and $\mathbf{v_c}$ denotes the direction of the concept vector. This is defined as the Conceptual Sensitivity $\textrm{CS}$ of a given layer $l$ for the network's output class $k$ and the concept $C$:
\begin{equation}
    \textrm{CS}_{C,l}^k (F,\mathbf{x_i})=\nabla h_l(f_l(\mathbf{x_i}))^{T}\mathbf{v_c}
\end{equation}
The TCAV score is given as the ratio of the number of inputs with positive conceptual sensitivity to the number of inputs for a class.

%\subsection{LIME}
LIME \cite{ribeiro2016model} is a black box method for understanding local explanations of a machine learning model. In order to explain the prediction of a model $F$ on an image $\mathbf{x_i}$ it:

\begin{enumerate}[noitemsep]
\item Decomposes $\mathbf{x_i}$ in $r$ homogeneous image patches or superpixels.
\item Creates a set of new images $\mathbf{x_{i_1},.....,x_{i_n}}$ by selecting $n$ subsets of the superpixels
\item Queries the model for each of these images $y_{i_j}=F(x_{i_j})$ $\forall j \in \{1,2...,n\}$
\item Builds a local weighted surrogate model $\hat{\beta}_i$ fitting the $y_{i_j}$ to the presence or absence of superpixels.
\end {enumerate}

Each coefficient of $\hat{\beta}_i$ is associated with a superpixel of the original image $\mathbf{x_i}$. Intuitively, the more positive the value of the coefficient, the more important the superpixel is for the prediction of the model. Generally, the user visualizes $\hat{\beta_i}$ by highlighting the superpixels associated with the top positive coefficients.
\subsection{Superpixels}
\label{sec:superpixels}

Superpixels are perceptually coherent regions of an image obtained by grouping pixels with similar
color, texture, or spatial proximity \cite{achanta2012slic}. Instead of analyzing individual pixels,
which are both high-dimensional and visually unintuitive, superpixels provide a mid-level
representation that aligns more closely with human perception. By reducing the number of
primitives from thousands of pixels to a few hundred superpixels, they enable more efficient
processing while preserving important object boundaries.

Several algorithms have been proposed for generating superpixels, among which the Simple Linear
Iterative Clustering (SLIC) method is widely used due to its efficiency and effectiveness
\cite{achanta2012slic}. SLIC adapts the k-means clustering algorithm in a five-dimensional space
(Lab color values and pixel coordinates), ensuring that generated superpixels are spatially compact
and adhere well to object boundaries. The main steps of SLIC involve:

\begin{enumerate}[noitemsep]
    \item Initialize $k$ cluster centers uniformly across the image grid.
    \item Assign each pixel to the nearest cluster center in a joint color–spatial distance metric.
    \item Update cluster centers as the mean of assigned pixels.
    \item Iterate until convergence, producing approximately equally sized, homogeneous regions.
\end{enumerate}

For our purposes, superpixels serve two key roles. First, they act as the interpretable building
blocks over which perturbations are applied in LIME, creating counterfactual-like variants of the
original image. Second, they provide localized regions where concept-level sensitivities can be
measured using TCAV. Together, these properties make superpixels a natural choice for bridging
concept attribution and decision attribution in our framework.

\subsection{Proposed Approach}
\label{alg:CAVLI_method}

\begin{algorithm}[H]
\label{alg:CAVLI}

                \caption{ {CAVLI}}
                \begin{algorithmic}[1]
                     % \begin{algorithmic}[1]
    \State {Train a TCAV model for a given concept $C$, a model $F$, and a layer $l$, resulting in the CAV vector $\mathbf{v_c}$.}
    \State {Decompose the input image $x_i \in X$ into $r$ homogeneous superpixels $\{S\}$.}
    \State {Create new images $\{x_{i_1}, \ldots, x_{i_n}\}$ by masking $x_i$ over $n$ random subsets of $\{S\}$.}
    \State {Calculate conceptual sensitivities $z_{i_j} = \textrm{CS}_{C,l}^k(F,x_{i_j})$ for each $x_{i_j}$.}
    \State \parbox[t]{\dimexpr\linewidth-1em}{Fit a local surrogate $\hat{\alpha}_i$ using $z_{i_j}$ values and superpixel presence.}
    \State {Query the model for each image patch: $y_{i_j} = F(x_{i_j})$.}
    \State {Fit a second surrogate $\hat{\beta}_i$ using $y_{i_j}$ values and superpixel presence.}
    \State {Compute Pearson correlation $\gamma_i$ between coefficients of $\hat{\alpha}_i$ and $\hat{\beta}_i$.}
    \State {Calculate Concept Dependency Score: $\textrm{CDS}_i = \gamma_i \cdot \textrm{CS}_{C,l}^k(F,x_i)$.}

\end{algorithmic}
                % \end{algorithmic}
                \end{algorithm} 
We propose a hybrid TCAV-LIME-based approach as a solution to the problem.
%(Steps 1 to 5 in Algorithm 1).
Algorithm 1 describes our proposed framework. First, we try to understand how well the model captures a concept for an individual decision. We frame the question by trying to understand \textbf{what parts of the image the model associates the most with a specific attribute}. Techniques like TCAV cannot be used directly because of their global nature. We start by dividing the image into $r$ homogeneous superpixels using the SLIC \cite{achanta2012slic} superpixel algorithm. 
%We propose a hybrid TCAV-LIME-based approach as a solution to the problem (Steps 1 to 5 in Algorithm 1). 
The second part of our pipeline investigates the question, \textbf{for a given decision, what parts of the image were the most influential in making the decision?} We make use of LIME to generate these regions in the image (Steps 1, 2, 6, and 7). Finally, we measure the overlap between the two parts. The intuition behind the approach is that if there is a clear overlap between image patches that have a high dependency on a concept and image patches that have the highest weight in decision-making, then the decision made by the network is heavily dependent on the concept (Steps 8 and 9).

The coefficients of $\hat{\alpha_i}$ indicate the level of association between different superpixels and a particular concept. A higher coefficient value suggests that the model considers the concept to be more closely related to that region, and vice versa. Similarly,  coefficients of $\hat{\beta_i}$ corresponds to a superpixel in the original image $\mathbf{x_i}$. The higher the weight of the superpixel, the more significant its contribution to the model's decision-making process. We are interested in measuring whether the superpixels associated with the the given concepts are also associated with decisions made by the algorithm. We calculate the Pearson correlation $\gamma_i$  correlation of $\hat{\beta_i}$ and $\hat{\alpha_i}$ to measure the overlap between the two decisions. A larger value of $\gamma_i$ indicates that there is a high overlap between the regions of the image that the model associates with the concept and those it uses for the decision. The Concept Dependency Score $CDS_i$, is calculated as the product of $\gamma_i$ and $CS_{C,l}^k(F,x_{i})$, ensuring that relevant concepts are given higher values. For a qualitative understanding, the coefficients of $\hat{\alpha_i}$ associated with the superpixels can be represented as a concept heatmap. This heatmap gives us a visualization of what parts of the image are more likely to be associated with a concept.

Randomly turning off subsets of superpixels during the analysis creates a counterfactual-like effect by simulating alternative versions of the input image in which certain visual information is masked. Each such perturbation can be interpreted as a counterfactual example: “What would the model predict if this region were absent?” This allows us to estimate the influence of specific regions on the model’s decision without retraining or accessing model internals. By analyzing how the model’s prediction changes in response to the removal of particular superpixels, we can identify which parts of the image are most critical for its output. This perturbation-based approach is central to LIME and complements TCAV’s concept attribution, enabling a localized and interpretable view of the concept-decision relationship.

%\section{Justification for different components }

\section{Evaluation}
%\begin{figure}
  %\includegraphics[scale=0.4]{cvpr2023-author_kit-v1_1-1/latex/graphics/heat4.png}
  %\caption{We use concept and decision heatmaps to analyze a classifier's decisions and their dependence on a specific concept, such as identifying whether an image contains a buffalo and the useful parts of the image for decision-making. In this example, we focus on the concept of grasslands, and the concept heatmap displays the areas of the image that the model associates most strongly with this concept.}
  %\label{fig:heatmap}
%\end{figure}
\begin{table*}
\label{table:zebra}
\centering
 \caption{A comparison of mean CDS values and TCAV values of different concepts for the class Zebra in the ImageNet dataset.}

  \begin{tabular}{|l|l|l|l|l|l|l|l|l|}
    \hline
   \textbf{Model} &
      \multicolumn{2}{c|}{Stripes} &
      \multicolumn{2}{c|}{Grassland} &
      \multicolumn{2}{c|}{Indoor} &
      \multicolumn{2}{c|}{Horse}\\
    \hline
    & CDS & TCAV & CDS & TCAV & CDS & TCAV&CDS&TCAV \\
    \hline
    GoogleNet & 0.17 & 0.78 & 0.26 & 0.62 & 0.13& 0.12 &0.02&0.41\\
    \hline
    ResNet & 0.23& 0.87 & 0.26 & 0.81 & 0.11 & 0.48 & 0.11 & 0.51 \\
    \hline
    InceptionNet & 0.45 & 0.84 & 0.21& 0.71 & -0.13 & 0.43 & 0.16 & 0.35\\
    \hline
  \end{tabular}
  \end{table*}
 \begin{table*}
\label{table:basketball}
\centering
\caption{A comparison of mean CDS values and TCAV values of different concepts for the class basketball in the ImageNet dataset.}
  \begin{tabular}{|l|l|l|l|l|l|l|l|l|}
    \hline
    \textbf{Model} &
      \multicolumn{2}{c|}{Ball} &
      \multicolumn{2}{c|}{Jersey} &
      \multicolumn{2}{c|}{Female} &
      \multicolumn{2}{c|}{Race}\\
    \hline  
    \textbf{}& CDS & TCAV & CDS & TCAV & CDS & TCAV&CDS&TCAV \\
    \hline
    GoogleNet & 0.29 & 0.56 & 0.38 & 0.93 & -0.03& 0.26 & 0.24 & 0.46\\
    \hline
    ResNet & 0.27& 0.68 & 0.21 & 0.46 & -0.20  & 0.45 & 0.22 & 0.73 \\
    \hline
    InceptionNet & 0.41 & 0.87 & 0.05 & 0.31 & 0.09 &  0.31 & 0.18 & 0.57\\
    \hline
  \end{tabular}
  
\end{table*}
\subsection{ImageNet Dataset}
  In our initial experiments, we assess the effectiveness of CDS in explaining model decisions in terms of concepts. To establish a baseline, we compare the performance of TCAV with the average CDS scores across different samples.
  We propose a hypothesis that if there exists a correlation between the mean CDS scores and global concept methods like TCAV, it indicates that our metric is capable of accurately capturing the dependence between the model decisions and underlying concepts.We conducted experiments on the ImageNet dataset, using similar settings to Kim et al. \cite{kim2019learning} and Schrouff et al. \cite{schrouff2021best} to validate our model. Specifically, we focused on the Zebra and Basketball classes, using three different models (GoogleNet, ResNet-50, and InceptionNet) for each class. Our goal was to measure the average statistics for each class using 100 images per set and calculating the mean correlation across all CDS scores. The experiments were conducted on the penultimate layer of all models.

\textbf{Zebra.} We ran experiments similar to Schrouff et al. \cite{schrouff2021best} that focus on four different concepts: ``stripes,'' ``zigzagged,'' ``dotted,'' ``horse,'' and ``grasslands.'' The results of the experiments are presented in Table 3.1, which shows that ResNet and GoogleNet both exhibited the highest mean CDS score for the concept stripes and the lowest mean CDS score for the concept indoor within the Zebra class. For InceptionNet, the concept grassland was more strongly associated with the Zebra class. The TCAV scores, which serve as global indicators of concept dependency, followed a similar trend. This pattern suggests that, on average, the CDS scores resemble the TCAV score. It is worth noting that higher CDS values indicate greater dependency on a concept, while lower values indicate lower dependency on the concept.

\textbf{Basketball.} We examined four human concepts (``ball,'' ``jersey,'' ``gender,'' and ``race'') in a manner similar to the Zebra class. The results in Table 3.2 show higher mean CDS scores for ``jersey'' and ``ball,'' and a lower score for ``female.''  We trained a race concept classifier with positive class images of African American faces. Our results further confirm the previous findings of a correlation between decision made on the basketball class and  concept race.  \cite{schrouff2021best}.

\subsection{CelebA dataset}
We are interested in exploring whether our approach can detect biases in model decisions caused by unbalanced data. Through our experiments we are interested in measuring whether these confounds can be detected by our metric.
The CelebA dataset \cite{liu2015faceattributes} is known to have naturally occurring confounds.   We train a smile classifier in a biased setting, where the training set is subsampled to create a higher positive correlation between the female-smiling and male-non-smiling attributes. We analyze the average CDS scores for different subgroups on the test data, as shown in Table 3.3. We observe that the highest average CDS scores were for the ``female smiling'' group, while the lowest were for the ``male smiling'' group. 
%Our 
These experimental findings align with the existing biases present in the dataset.

\begin{table}

\centering
\caption{Average CDS scores for different subgroups in the CelebA dataset.}
\scalebox{1.0}{
\begin{tabular}{ |c|c|c| }
 \hline
 & Male & Female \\ 
 \hline
 Smile & 0.004   &  0.013  \\  
 \hline
 Non-Smile & 0.005 & 0.007 \\  
  \hline
\end{tabular}}

\end{table}

\subsection{Qualitative Analysis}
\begin{figure}
    \centering
  \includegraphics[scale=0.85]{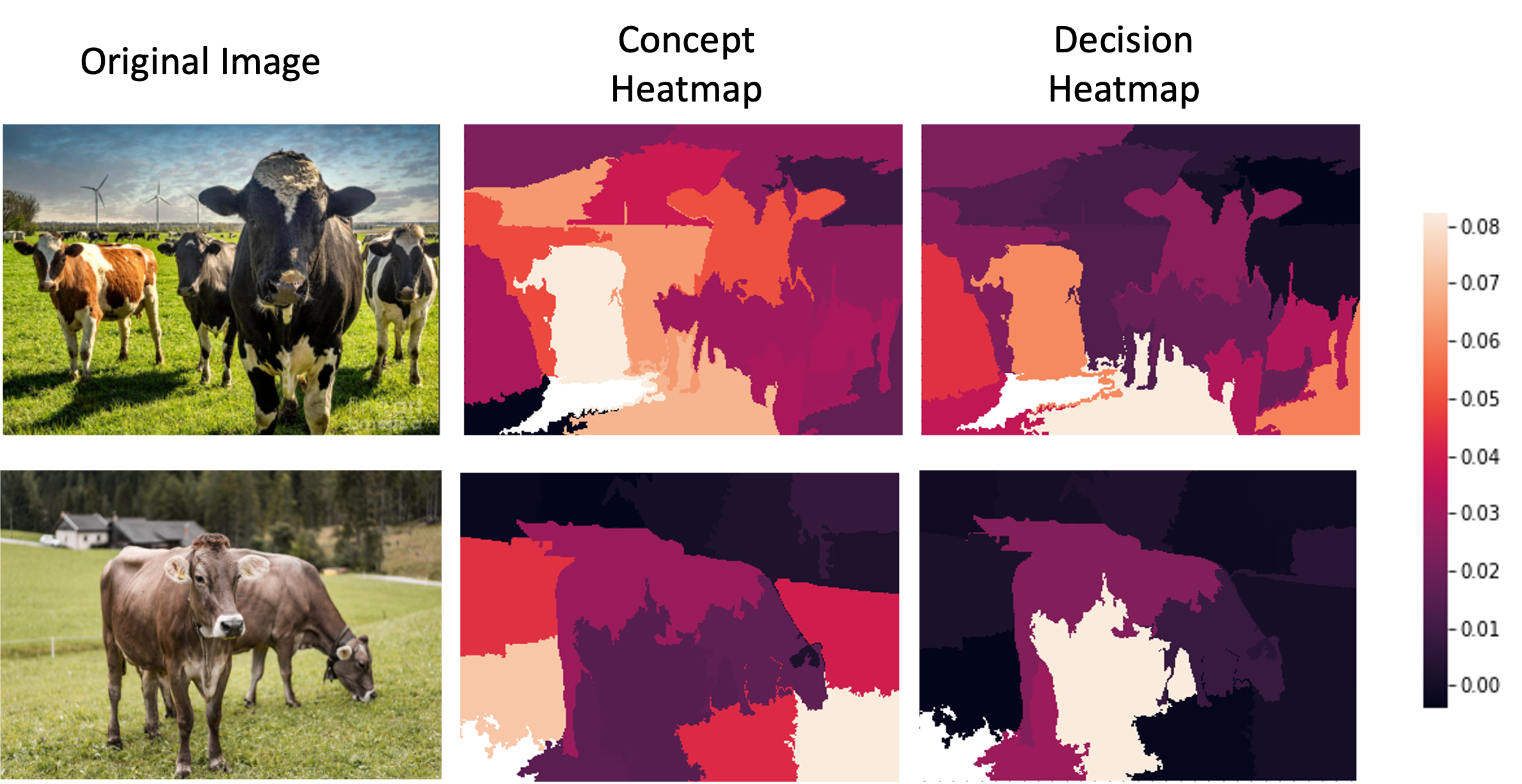}
  \caption{We use concept and decision heatmaps to analyze a classifier's decisions and their dependence on a specific concept, such as identifying whether an image contains a cow and what parts are the most useful in decision-making. Here we focus on grasslands, and the concept heatmap displays the areas of the image that the model associates most strongly with this concept.}
  \label{fig:heatmap}
\end{figure}

Our method generates concept heatmaps that illustrate the image regions and their dependence on human-defined concepts. These heatmaps aid in visually interpreting a model's image dependency, as shown in Figure \ref{fig:heatmap} for the cow-grasslands class-concept pair. The qualitative analysis reveals not only the parts of the image used for decision-making, but also whether the model links these parts to a concept. Additionally, this visual representation can identify spurious correlations where a classification decision (cow) is based on a concept (grassland) that is not directly related to the class.
\section{Conclusion}
This work introduces a novel framework for generating local, concept-based explanations in computer vision models by combining TCAV and LIME in a hybrid approach. At its core, the method uses counterfactual reasoning to understand the relationship between human-understandable concepts and model predictions. By selectively turning off superpixels to simulate counterfactual inputs, and measuring their alignment with concept representations, we create localized explanations that reflect how changes in visual concepts affect model behavior. This approach culminates in the Concept Dependency Score (CDS), which quantifies the overlap between concept relevance and decision salience.

The method aligns closely with the broader goals of this thesis, which leverages counterfactual reasoning to make black-box models more transparent and trustworthy. By operationalizing counterfactuals through perturbations, our method not only enhances interpretability but also provides actionable insights for auditing model decisions in sensitive applications. While the framework has limitations—including reliance on segmentation quality and the subjective nature of concept definitions—it provides a principled way to connect input-level interventions with higher-level semantic reasoning. Future work could extend this approach through richer concept libraries, human-in-the-loop validation, and deployment in real-world auditing settings. Overall, this chapter demonstrates how counterfactual reasoning can serve as a powerful foundation for concept-based interpretability in vision models.

\subsection{Limitations}
While CAVLI provides a general framework to assess the dependence of model decisions on human-defined concepts, it relies on the assumption that a greater spatial overlap between regions important for decision-making and regions associated with a concept implies stronger dependence on that concept. This reasoning aligns only with the first rung of Pearl’s ladder of causation—association—and does not account for deeper causal relationships. For example, if the regions associated with both skin color and smiling overlap significantly in an image, the resulting high CDS score may misleadingly suggest a strong dependence on both, even though only one may be causally relevant. Association, however, does not imply causation.

Moreover, CAVLI requires pre-defined concepts and lacks mechanisms for discovering or prioritizing concepts automatically. It does not determine which concepts are most relevant for a given image or task. While we address this limitation in the following chapter by introducing a method to automatically identify relevant concepts, the dependence on manually specified concepts remains a key constraint of the current approach.
\chapter{ASACs: Mitigating bias in computer vision models through adversarial counterfactuals}
\epigraph{“The wicked are not always villains, nor the good always heroes. We carry both in ourselves.”}{\textit{Rabindranath Tagore}}
\begin{figure*}[ht]
    \centering
     \includegraphics[width=0.70\textwidth]{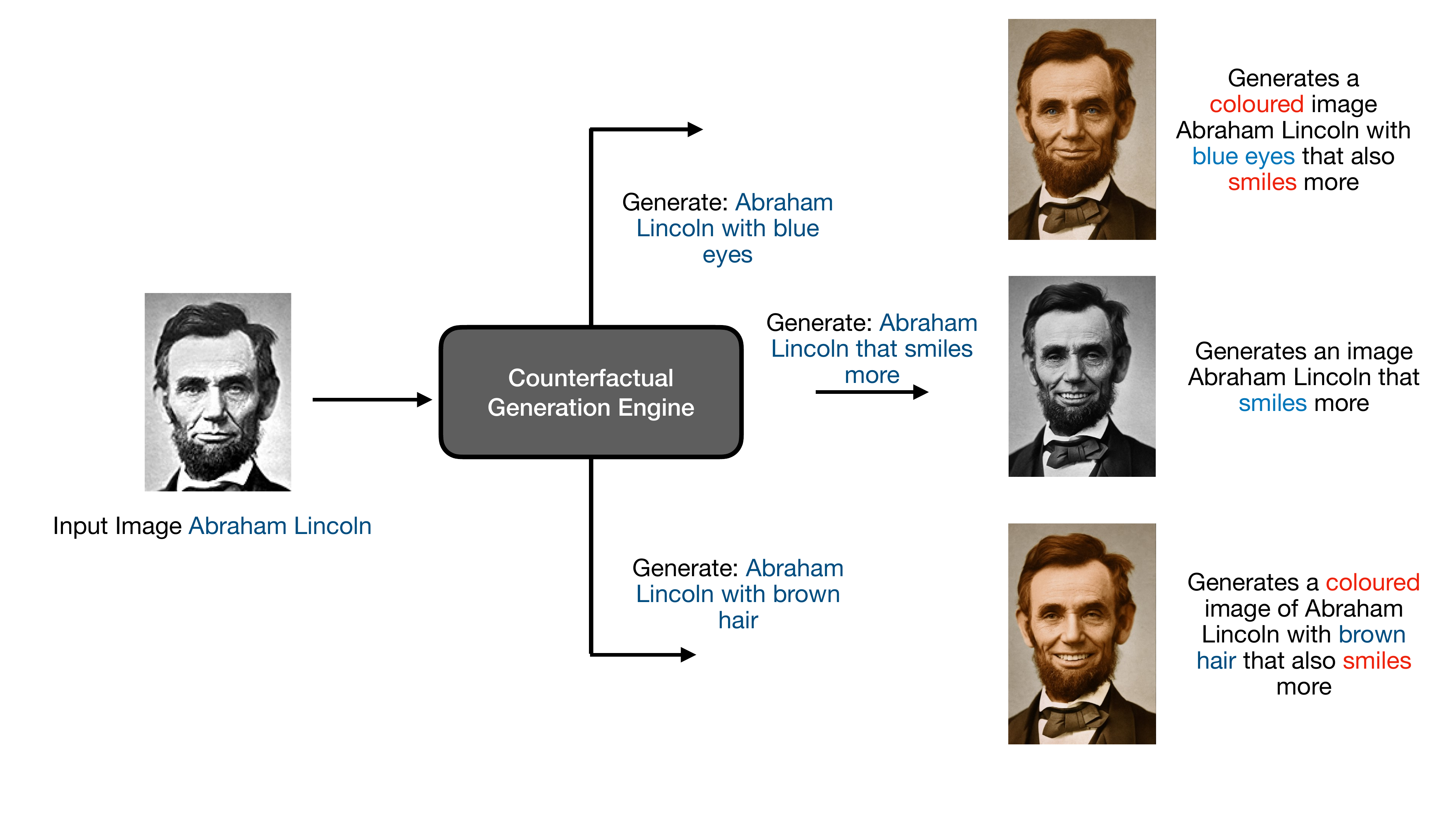}
     \captionof{figure}{What happens when we ask a counterfactual generation engine to modify an image along specific attributes? When asked to alter attributes like smile, eye color, or hair, standard counterfactual engines often introduce unintended changes—like clothing or background—revealing entangled attributes and stereotypes. This chapter attempts to address these problems by proposing a novel counterfactual generation scheme, called Attribute Specific Adversarial Counterfactuals (ASACs), and demonstrates that ASACs can be utilized for bias mitigation in computer vision classifiers.
 }
     \label{fig:ASAC_teaser}
\end{figure*}
\clearpage
\section{Introduction}
\label{sec:intro}
Computer vision systems trained on large amounts of data have been at the heart of recent technological innovation and growth. Such systems continue to find increasing use-cases in different areas such as healthcare, security, autonomous driving, remote sensing and education. However, despite the tremendous promise of large vision models, several studies have demonstrated the presence of unwanted societal biases in these models \cite{bolukbasi2016man,buolamwini2018gender,hendricks2018women}. Therefore, understanding and mitigating these biases is crucial toward deploying such systems in real-world applications.

Several approaches have been proposed to mitigate \cite{buolamwini2018gender,seyyed2021underdiagnosis,hendricks2018women,meister2023gender,wang2022revise,liu2019fair,joshi2022fair,wang2020towards,wang2023overwriting}, measure \cite{denton2019image,balakrishnan2021towards}, and explain \cite{abid2022meaningfully,feder2021causalm,wu2021polyjuice} biases in computer vision models. Among these approaches, the use of counterfactuals has emerged as a promising and prominent line of research. This chapter takes a step further in that direction by proposing a new approach to mitigate biases in image classifiers.

This chapter builds on that line of work by proposing a new counterfactual-based strategy to mitigate bias in image classifiers. However, before embracing counterfactuals as a tool for fairness, we pause to ask a fundamental question: \textbf{Are the counterfactuals we currently use valid and trustworthy?} Could they, instead of helping, introduce new artifacts, reduce model robustness, or even amplify the very biases they are meant to mitigate?

% \pushkar{ Does this definition look good} \textbf{A counterfactual for an image is defined as a transformed image produced by systematically replacing regions }\lee{perhaps parts is a better word? when I think of regions I think of specific patches} of the original image, such that a system’s decision on the transformed image is altered 

%Counterfactuals generated across dimensions in which biases may exist (e.g.,\ race, gender) can be used not only to quantify the degree of bias present in a model but also to mitigate such bias toward building more fair models. 
\begin{figure*}[tb]
  \centering
 
   \includegraphics[trim=0 190 0 0, clip,width=\textwidth]{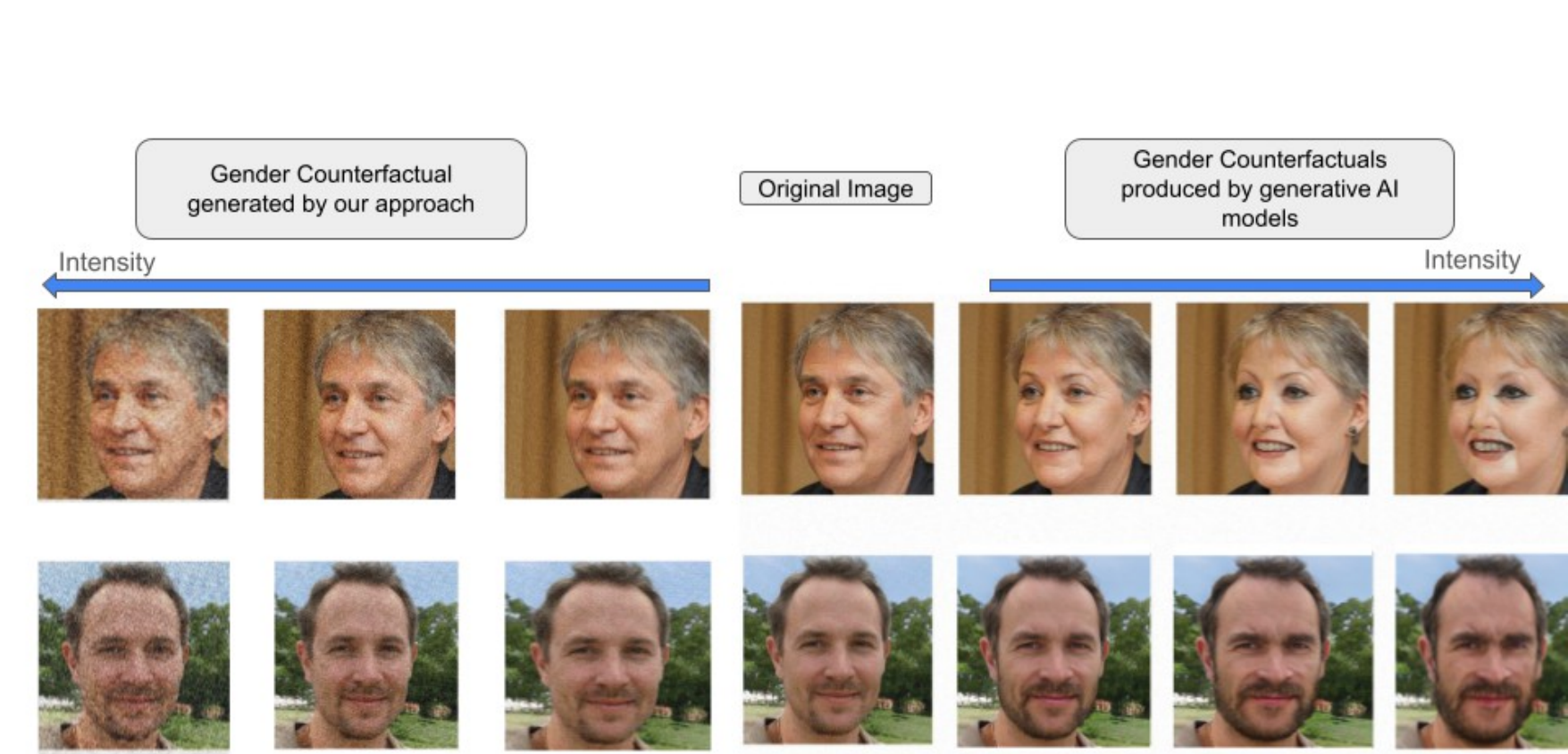}
   \caption{An example of gender-based counterfactuals generated by StyleGAN2 and our method for a smile classification task. In the first row, we attempt to generate more feminine-looking versions of a given face, and in the second row, more masculine-looking versions. The StyleGAN2 outputs (right) exhibit biased correlations—femininity is often expressed through darker lipstick and exaggerated smiles, while masculinity is associated with older or wrinkled facial features. In contrast, our method (left) produces Adversarial Semantic Attribute Counterfactuals (ASACs) that preserve the original face's overall visual identity while varying gender-relevant features in a more controlled and semantically faithful manner.}
\label{fig:style_gan2}
\end{figure*}

Indeed, existing counterfactual generation methods suffer from several critical limitations. Most rely on generative models to produce image variants along protected attributes such as race or gender. These generators, often trained on biased datasets, may produce counterfactuals that embed spurious correlations or stereotypical visual features. For example, Figure~\ref{fig:style_gan2} shows biased outputs from StyleGAN2 \cite{karras2020analyzing} for a smile classification task: when asked to generate more feminine faces, the model exaggerates smiles and adds heavy makeup; when asked to masculinize a face, it adds wrinkles or harsh facial structure. Such distortions can mislead both human and algorithmic decision-making, especially if used to train or fine-tune a supposedly fairer classifier.

A similar issue arises in more targeted edits (see Figure \ref{fig:ASAC_teaser}). What happens when we ask a counterfactual generation engine to modify an image along specific attributes such as smile, eye color, or hair? In practice, these edits often come bundled with unintended changes—alterations in clothing, background, or lighting—revealing deep entanglements between features and entrenched visual stereotypes. These artifacts undermine the reliability of counterfactuals and raise ethical concerns about representational harm, particularly when dealing with sensitive attributes.

Therefore, to ensure fair and responsible use of counterfactuals, we advocate for approaches that preserve semantic integrity, minimize spurious correlations, and explicitly guard against potential misuse. This chapter introduces such a method designed to generate precise, ethically sound counterfactuals that challenge biased model decisions without reinforcing harmful stereotypes.
Instead of relying on generative models, we propose using adversarial images as an alternative for producing counterfactuals in the context of bias mitigation. While adversarial examples are traditionally used to expose model vulnerabilities, we repurpose them to construct minimal, targeted perturbations that reveal and reduce unfair model dependencies and also improve the overall performance of the model. Our approach demonstrates that adversarial counterfactuals can not only improve fairness metrics but also enhance overall model accuracy.

%Our goal is twofold: to address ethical concerns and reduce spurious artifacts in generating image counterfactuals. First, generating counterfactuals along sensitive attributes like gender or age can raise privacy and fairness issues, as such transformations (e.g., creating a female version of a male subject) are both unnecessary and potentially harmful. Second, we aim to minimize spurious artifacts introduced during training, especially those stemming from third-party image generators. We hypothesize that these specifically designed adversarial images will reduce spurious biases and enhance model robustness and performance. 

Our goal is to improve the performance of a classifier as well as debias it against protected attributes without using images that may introduce or exacerbate  such biases in the model. We attempt to do so in a way that also address the ethical concerns around training on images created based on societal stereotypes. We use existing adversarial techniques to create counterfactuals that challenge vision models based on protected attributes. We refer to these samples as ``Attribute-Specific Adversarial Counterfactuals'' or ASACs. Our method ensures that ASACs retain the visual appearance of the original image, effectively addressing ethical concerns regarding the quality of image counterfactuals. By keeping the essence of the image unaltered, the likelihood of introducing spurious correlations (propagated by the generative model during the image generation process) into the fairness mitigation pipeline is markedly reduced.%\lee{do we have any results that are based on race noise?} 
%\pushkar{NO}

% \lee{unless this is different in applicative papers (compared to theory), we should mention the papers that are doing things similar to our paper, and what we do differently here (not just in the related work).}\pushkar{See Literature Review}

%Our findings suggest that the proposed training approach, not only improves fairness metrics but also maintains or increases the overall performance of the model. Overall, our contributions can be summarized as follows:
In addition, our work introduces a novel approach to improve both fairness and accuracy in these models by utilizing ASACs in a curriculum learning-based framework. Our approach utilizes ASACs generated at different noise magnitudes and assigns a learning curriculum to these samples based on their ability to deceive the model. We then fine-tune the model using the assigned curriculum. We validate our approach both qualitatively and quantitatively using experiments on various classifiers trained on multiple datasets and several target attributes. We show that our method generalizes to biased models of varying scales (between 8M and 24M parameters). Our contributions can be summarized as follows: (1) We introduce a bias-averse method for generating image counterfactuals using adversarial images to create Attribute-Specific Adversarial Counterfactuals (ASACs) for protected attributes. (2) We also present a novel curriculum learning-based fine-tuning approach that leverages ASACs to mitigate pre-existing biases in vision models. Finally, (3) we provide a comprehensive evaluation across various datasets, architectures and metrics, demonstrating that our method enhances fairness metrics without compromising the model's overall performance and in many cases, improving the models performance significantly.
%\begin{itemize}
    %\item We present a bias-averse approach to generating image counterfactuals, utilizing adversarial images to create ASACs specific to sensitive attributes.
    %\item We introduce a novel curriculum learning-based fine-tuning methodology that leverages ASACs to mitigate pre-existing biases in vision models.
    %\item We present a comprehensive evaluation across different datasets and evaluation metrics, demonstrating that our method improves fairness metrics without compromising the overall performance of the model.
%\end{itemize}

\section{Related Works}
\label{sec:related}

% \label{sub:adv_examples}

Adversarial examples have emerged as a notable challenge in the realm of computer vision \cite{maudslay2019s,wang2021adversarial,szegedy2013intriguing}. These examples are crafted with the specific goal of deceiving machine learning models by making subtle alterations to input, resulting in model misclassification. Adversarial attacks are specialized algorithms designed to generate such examples and have been the subject of extensive research. These attacks are categorized into black-box \cite{rahmati2020geoda,jiang2019black,narodytska2017simple} and white-box attacks \cite{szegedy2013intriguing,maudslay2019s}, depending on the attacker's access to model parameters. Our approach uses two white-box methods, FGSM \cite{goodfellow2014explaining} and PGD \cite{madry2017towards}.

\paragraph{\textbf{The Relationship Between Counterfactual and Adversarial Examples}}
While the connection between counterfactuals and adversarial images has been extensively explored in theoretical machine learning \cite{pawelczyk2022exploring,ustun2019actionable,van2021interpretable, karimi2020model,beutel2017data}, limited attention has been given to this relationship in computer vision \cite{zhang2020towards,qiu2020semanticadv,lim2023biasadv,wang2020score}. Among these approaches, work by Wang et al. \cite{wang2020towards} and Zhang et al. \cite{zhang2020towards} is particularly noteworthy because of its proximity to our approach. Wang et al. proposed a post-processing approach to mitigate biases using adversarial examples, uniquely incorporating a GAN-based loss function. Conversely, Zhang et al. employ an architecture similar to ours, generating adversarial perturbations to counteract biases. Our work, however, is distinct in proposing a novel training setup that introduces a curriculum learning-based strategy with the use of ASACs. Unlike previous methods that often rely on a single adversarial example, our approach systematically assesses the impact of each adversarial example on the original classifier, providing a more directed training regime. Compared to previous works our paper highlights how adversarial counterfactuals can not only prevent ethical issues compared to image generator-based counterfactuals but also prevent spurious correlations from being added in the de-biasing process. In addition, we conduct a more rigorous empirical analysis to show that the approach can not only improve on fairness metrics but also overall performance. 

\paragraph{\textbf{Curriculum Learning}} Curriculum learning is an effective training paradigm that gradually exposes models to progressively more complex examples, aiding in better generalization \cite{bengio2009curriculum,jiang2015self,matiisen2019teacher,kong2021adaptive,graves2017automated}. This approach strategically organizes training data, facilitating smoother convergence and improved performance. Several approaches have since been proposed to organize data that include sorting the data \cite{bengio2009curriculum}, adaptive organization \cite{matiisen2019teacher}, self-paced curriculum assignment \cite{matiisen2019teacher}, and teacher-based curriculum learning \cite{matiisen2019teacher}. To the best of our knowledge, we believe that we are the first to use a curriculum based framework with adversarial images to improve both fairness as well as performance of biased classifiers in computer vision.
\section{Notations}
\label{sec:ASAC_notation}
We define the data distribution as $D$, from which a set of images $\mathbf{X}$ are sampled. Along with the images, we sample the corresponding ground truth labels for a target attribute and a protected attribute defined by $\mathbf{Y}_{\textrm{target}}$ and $\mathbf{Y}_{\textrm{protected}}$ respectively. A single sample drawn from the set is defined as $(x,y_{target},y_{protected})$ where $x$ represents the input image and $(y_{\textrm{target}}, y_{\textrm{protected}})$ represent the ground truth label of the target and protected attributes respectively.

%(\mathbf-label pairs $(x,y)$ are sampled \textit{i.i.d.}, both for train and test data. Protected attribute pairs are denoted as $(x_a, y_a)$ and are sampled from the same data distribution as $(x,y)$ with adversarial perturbations relative to an attribute-specific image $x_a$ represented by $\delta_{x}$. As an example consider a smile classifier with gender as the protected attribute. Here $(x,y)$ is the input and true label of the smile classifier whereas $x_a$ denotes the image when the input image is $x$  is perturbed.\textbf{Give an example here }

Let $M_{(\theta, \rho)}: \textbf{X}\rightarrow \mathbf{\hat{Y}}_{\textrm{target}}$ be a target attribute classifier that maps the input data to the target class. After training $M_{(\theta, \rho)}$, we train a protected attribute classifier $C_{(\theta, \phi)}: \textbf{X}\rightarrow \mathbf{\hat{Y}}_{\textrm{protected}}$ that maps the same input data to a protected attribute class. Here $\theta$ denotes the parameters of the common backbone network shared by the two classifiers, whereas $\phi$ and $\rho$ are the linear layers for $C$ and $M$ respectively. This means that for a given an input image $x$, $\hat{y}_{\textrm{target}}= M_{(\theta, \rho)} (x) $ and $\hat{y}_{\textrm{protected}}=C_{(\theta, \phi)}(x)$.

\subsection{Evaluation Metrics}
\label{sec:ASAC_sub_eval_metrics}
To evaluate the fairness of our model, we employ three key metrics: the Difference in Demographic Parity (DDP), the Difference in Equalized Odds (DEO), and the Difference in Equalized Opportunity (DEOp) \cite{hardt2016equality}. Alongside these, we also measure the model's accuracy (ACC). Lower values for fairness metrics indicate a less biased model whereas a high value in accuracy indicates a more performant model. The metrics are based on definitions of fairness and were chosen to compare with previous works \cite{zhang2020towards, ramaswamy2021fair}.

The DEO focuses on the difference in the rates of false negatives and false positives between genders. A substantial DEO indicates a bias in which one group is either less likely to be incorrectly dismissed or more likely to be inaccurately favored. %The difference in Equalized opportunity is given as  DEO=$ |P(\hat{Y}=1|Y=y,A=a)-P(\hat{Y}=1|Y=y,A=a')|$,where  $y \in \{0,1\}$

\begin{definition}[Equalized Odds]
A predictor $\hat{Y}$ satisfies equalized odds with respect to protected attribute $A \in \{a,a'\}$ and outcome $Y$, if $\hat{Y}$ and $A$ are independent conditional on $Y$.
\begin{equation}
P(\hat{Y}=1|Y=y,A=a)=P(\hat{Y}=1|Y=y,A=a')  \quad .
\end{equation}
\text{where } $y \in \{0,1\}$. 
To estimate how far a predictor $\hat Y$ from having equalized odds, we  estimate the Difference in Equalized Odds (DEO), i.e., we estimate
\[
\sum_{y} |P(\hat{Y}=1|Y=y,A=a)-P(\hat{Y}=1|Y=y,A=a')
\]

\end{definition}

The second metric that we use to measure fairness is the Difference in Demographic Parity or DDP. DDP quantifies the absolute gap in approval rates (e.g., smile prediction) across different protected attributes.  A large DDP value suggests a tendency for individuals in a specific group to receive more favorable outcomes than their counterparts in the other group. 
\begin{definition}[Demographic Parity]
A predictor $\hat{Y}$ satisfies demographic parity with respect to protected attribute $A \in \{a, a'\}$ and outcome $Y$, if the following condition holds:
\begin{equation}
P(\hat{Y} = 1 | A = a) = P(\hat{Y} = 1 | A = a').
\end{equation}
To estimate how far  a predictor $\hat Y$ from having demographic parity, we  estimate the Difference in Demographic Parity (DDP), i.e., we estimate
\[
|P(\hat{Y}=1|A=a)-P(\hat{Y}=1|A=a')|.
\]
\end{definition}

The Difference in Equalized Opportunity (DEOp) measures the disparity in True Positive Rates between different demographic groups in a model, indicating bias in favorable outcomes. A DEOp of zero signifies equal accuracy across groups, representing a fairness ideal in model performance. The difference in Equalized Opportunity is given as follows. 
\begin{definition}[Equalized Opportunity]
A predictor $\hat{Y}$ satisfies equalized opportunity with respect to protected attribute $A \in \{a, a'\}$ and outcome $Y$, if the following condition holds for a specific outcome value $y$:
\[
P(\hat{Y}=1 | Y=1, A=a) = P(\hat{Y}=1 | Y=1, A=a').
\]
To measure how far a predictor $\hat Y$ from having demographic parity, we  estimate the Difference in Equalized Opportunity (DEOp), i.e., we estimate
\[
|P(\hat{Y}=1 | Y=1, A=a) - P(\hat{Y}=1 | Y=1, A=a')|.
\]
\end{definition}
% \subsection {Evaluation  Metrics} 
% DEOp=$|P(\hat{Y} = 1 | A = a) - P(\hat{Y} = 1 | A = a')|$

\begin{figure*}[tb]
  \centering
   \includegraphics[width=\linewidth]{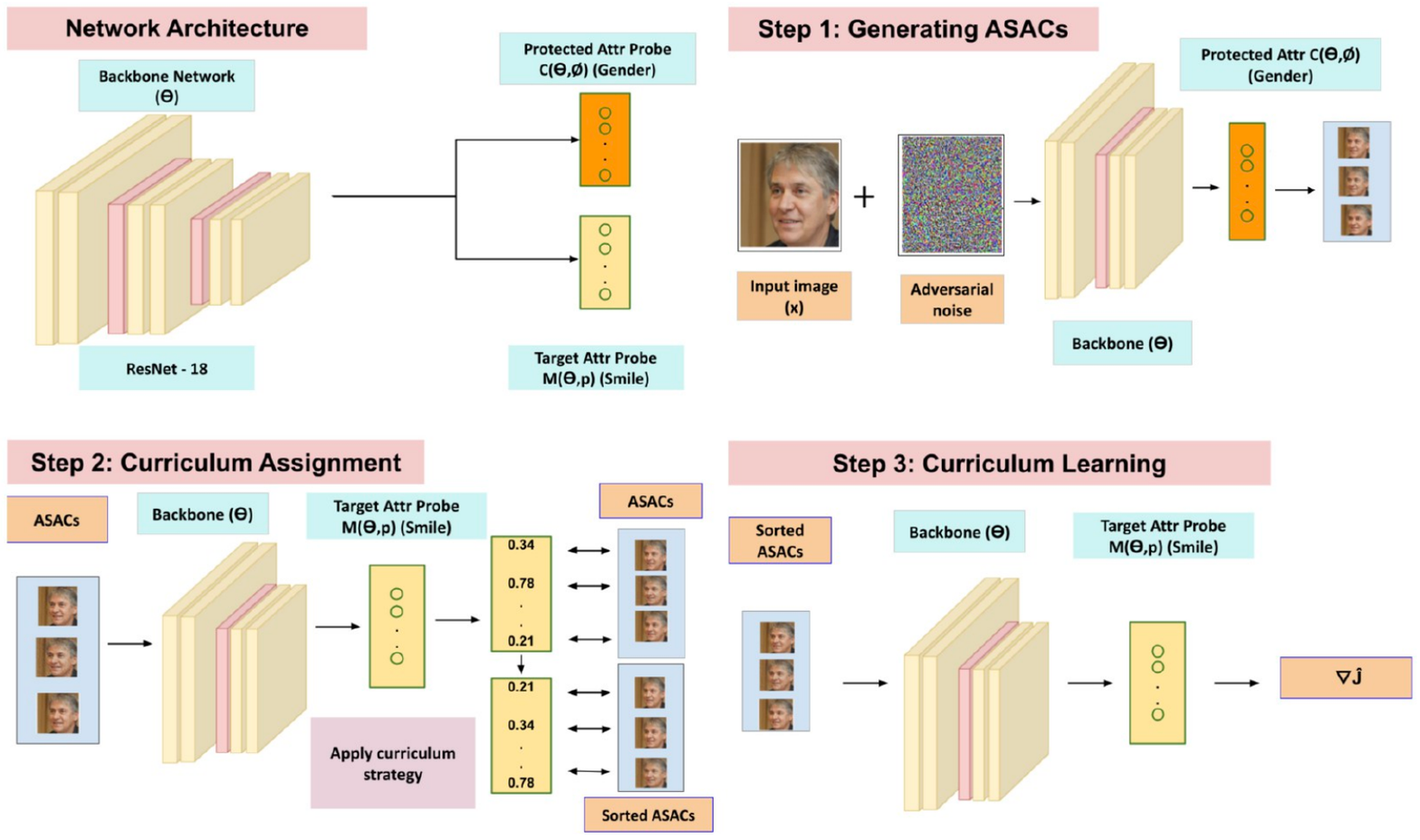}
   \caption{\textbf{Bias Mitigation Strategy}: Our proposed solution for mitigating biases in a model (e.g., smile classifier) $M_{(\theta,\rho)}$ involves training sensitive attribute classifier  $C_{(\theta,\phi)}$ (shown in the network architecture). We then follow a three-stage pipeline. (1) We generate ASACs that are capable of deceiving $C_{(\theta,\phi)}$. (2) We define a curriculum assignment strategy that organizes these ASACs based on the degree to which they deceive the original model $M_{(\theta,\rho)}$. (3) We fine-tune the original model $M_{(\theta,\rho)}$ using the organized ASACs.
}
\label{fig:three_step_pipeline}
\end{figure*}

\section{Method}
\label{sec:ASAC_method}

Our method for creating attribute-specific adversarial counterfactuals (ASACs) and utilizing them to fine-tune a biased model can be broken down into three stages. For every batch during the fine-tuning process, we first generate ASACs for the batch $X= \{x_1, x_2, \ldots,x_k\}$, aiming to mislead the classification model $M_{(\theta,\rho)}$ regarding a protected attribute $a$ (detailed in Section \ref{sub:gen_ASAC}). For example, when training a smile classifier, we produce adversarial images that prompt misclassification based on the protected attribute of gender. The next step (Section \ref{sub:curriculum}) evaluates the effectiveness of the generated ASACs. Continuing with our example, we measure the capacity of the ASACs to incorrectly influence the model's smile classification, despite the ASACs being generated to confuse the model with respect to the gender class. We assign a training curriculum by partitioning the batch of ASACs into mini-batches based on the ASAC's success in deceiving the model. This procedure can be seen in Algorithm \ref{alg:ASAC}. Finally, the concluding stage utilizes the curriculum and ASACs produced by taking gradient step for every mini-batch to fine-tune the original model $M_{(\theta,\rho)}$. This aims to reduce biases within the model and enhance its discriminative performance.

\begin{algorithm}[H]
\caption{Construct Minibatches for Training Based on Curriculum. Note that a gradient step is taken over each minibatch returned.}
\label{alg:ASAC}
\begin{algorithmic}[1]
    \State \textbf{Input:} $M_{(\theta, \rho)}$ (target classifier), $\{x_1, \dots, x_k\}$ (input batch), $\{\epsilon_1, \dots, \epsilon_l\}$ (noise magnitudes), $f_{\text{attack}}$ (adversarial attack), \textit{order} (sorting order)
    \State \textbf{Output:} $\text{minibatches}$: Partitioned minibatches sorted by difficulty score
    \vspace{0.5em}
    \For{$i = 1$ \textbf{to} $k$}
        \For{$j = 1$ \textbf{to} $l$}
            \State $x_{a_{ij}} \gets f_{\text{attack}}(x_i, \epsilon_j)$
            \State $\text{score} \gets DS(x_{a_{ij}}; y_{\text{target}}, M_{(\theta, \rho)})$
            \State $\text{ASACs} \gets \text{ASACs} \cup \{(x_{a_{ij}}, \text{score})\}$
        \EndFor
    \EndFor
    \State{ $\text{ASACs} \gets \text{Sort}(\text{ASACs}, \text{by score in \textit{order}})$}
   \State{ $\text{minibatches} \gets \text{Partition}(\text{ASACs}, l)$}
   \State{ \textbf{return} $\text{minibatches}$}
\end{algorithmic}
\end{algorithm}

\subsection{Generating attribute-specific adversarial examples}
\label{sub:gen_ASAC}
This section details our approach to constructing ASACs-adversarial images designed to mislead a model concerning a specific protected attribute, denoted as $a$. We start by employing a target classifier $M_{(\theta,\rho)}$ and develop an attribute classifier, $C_{(\theta, \phi)}$, to predict the protected attribute $a$ within the dataset $D$. Both models share a similar structure, with only the last layer differing, as depicted in Figure \ref{fig:three_step_pipeline}, thereby utilizing nearly identical representations for classification. They are trained on identical data distributions $D$. To construct each ASAC $x_a$, we augment the original image $x$ with noise $\delta_{x}$ using common methods to generate adversarial images as shown below.

\begin{equation}
\label{eq:adv_examples}
    x_a= x + \delta_{x}
\end{equation}

In our work, we experiment with two commonly used adversarial methods, Fast Gradient Sign Method (FGSM) \cite{goodfellow2014explaining} and Projected Gradient Descent (PGD) \cite{madry2017towards}, to generate images capable of deceiving the attribute classifier. FGSM is a single-step attack approach that perturbs input images based on the gradient sign of the loss function, as shown in Equation \ref{eq:fgsm}, with $\epsilon\in[0,1]$ denoting a scaling factor which we refer to as the noise magnitude. PGD iteratively perturbs input data similar to FGSM to maximize the loss function, aiming to find the smallest perturbation that causes misclassification.

\begin{equation}
\label{eq:fgsm}
 \delta_{x} = \varepsilon \times \text{sign} \left( \nabla_x J(\theta, \phi, x, y) \right)
\end{equation}
Here $J$ is any differentiable loss function on which the model is trained. It should be noted that ASAC's differ from a normal attacks because the images generated by ASAC's intend to fool the protected attribute classifier $C_{(\theta, \phi)}$. Hence, ASAC's are designed such that $C_{(\theta, \phi)}(x) \neq C_{(\theta, \phi)}(x_a) $. Since both the target classifier and the protected attribute classifier have the same backbone $\theta$ , the generated ASAC's can influence the target classsifer $M_{(\theta,\rho)}$'s understanding of the protected attribute. If $\hat{y}_{\textrm{target}}= M_{(\theta, \rho)} (x) $ is dependent on the protected attribute i.e., the target classifier is biased, then the generated ASAC may also be able to fool the target classifier and $ M_{(\theta, \rho)} (x) \neq  M_{(\theta, \rho)} (x_a)$ even though the ASACs were generated to fool the protected attribute classifier $C_{(\theta, \phi)}$.
 %Though this is a more computationally expensive attack, it often yields more potent adversarial examples than its single-step counterpart. We perform ablations (see Table \ref{tab:noise_mags}) with various noise magnitudes for each sample image to generate a range of adversarial images, enabling us to evaluate the methods' effectiveness under different conditions. We show that grouping ASACs from different noise magnitudes for each batch results in improved overall performance of the predictor $M(\theta, \rho)$ (see Table \ref{tab:noise_mags}). In addition, grouping ASACs across different noise magnitudes enables us to define an in-batch curriculum improving both fairness and performance of the predictor $M(\theta, \rho)$. \lee{I'm missing the explanation of what curriculum learning is...}
 
%\begin{figure}[tb]
  %\centering
   %\includegraphics[width=\linewidth]{fig/difficulty_score.png}
   %\caption{ This figure illustrates the impact of various noise magnitudes on the discriminative power of both the target and protected attribute classifiers for smile and gender respectively. Our notion of difficulty scores as shown in Equation \ref{eq:ds_2} quantifies the impact that each ASAC has on the original classifier $M(\theta,\phi)$ as shown by the orange data points.}
%\label{fig:confidence}
%\end{figure}

\subsection{Curriculum Learning} 
\label{sub:curriculum}
We propose a curriculum learning-based strategy for training our model. A curriculum learning approach trains the model from easy to hard examples or vice versa, thereby making the  learning process efficient and faster. Our approach comprises two main stages. First, we evaluate the influence of each ASAC on the classification model $M_{(\theta,\rho)}$ by assigning each ASAC a difficulty score. Subsequently, the model implements a curriculum for training, guided by the difficulty scores over the ASACs. %As shown in Algorithm \ref{alg:asacs_curriculum}, we sort and then partition the ASACs in a batch into mini-batches based on the difficulty scores and then take a gradient step through each mini-batch. %A  curriculum based  training strategy on ASACs aids in improve convergence speed accuracy, and fairness metrics.%By defining a curriculum based on ASACs from different noise magnitudes, we guide the adversarial fine-tuning process, enhancing convergence speed, accuracy, and fairness metrics.

\subsubsection{Computing Difficulty Scores}
\label{sub:difficulty}
The difficulty score is crucial in curriculum learning, as it measures the utility of each example and determines the order in which ASACs are presented to the model during adversarial training \cite{kong2021adaptive}. However, obtaining the optimal difficulty score a priori is not feasible in our setting. To address this, we define the difficulty score for any ASAC image target label pair $(x_a, y_{\textrm{target}})$ as the complement of the softmax value that the model $M_{(\theta,\rho)}$ returns for $x_a$ corresponding to the ground truth label $y_{\textrm{target}}$.
%The difficulty score plays a key role in curriculum learning as it measures the utility of each example and determines the order in which ASACs are presented to the model during adversarial training \cite{kong2021adaptive}. However, in our problem setting, it is not feasible to obtain the optimal difficulty score a priori. Therefore we circumvent this issue by , for any attribute-specific image $x_a$, we define the difficulty score as the complement of the softmax value the model $M$ returns on $x_a$:

\begin{equation}
\label{eq:ds}
DS(x_a; y_{\text{target}}, M_{\theta, \rho}) = 1 - \text{Softmax}(M_{\theta,\rho}(x_a))_{y_{\text{target}}}
\end{equation}

% \begin{equation}
% \label{eq:ds}
% DS(x_a, y_{\text{target}}; M_{\theta, \rho}) = 1 - \text{Softmax}(M_{\theta,\rho}(x_a))_{y_{\text{target}}}
% \end{equation}

% \begin{eq/uation}
%\label{eq:ds_1}
   % DS(x_a) = 1 - \text{Softmax}(x_a; M_{\theta,\rho}) 
    %\\ = 1 - \frac{\exp(M_{\theta,\rho}(x_a))}{\sum_{i=1}^{N}\exp(M_{\theta,\rho}(x_a(i)))}
%\end{equation}

%\noindent where $N$ is the number of classes for the target outcome. Note that here we use ASACs for adversarial curriculum learning, therefore, we  rewrite Equation (\ref{eq:ds_1}) as shown below:
%\begin{equation}
%\label{eq:ds_2}
    %DS(x'_a) = 1 - \text{Softmax}(x'_a; M_{\theta,\rho})
%\end{equation}

%A subtle but key point to note here is that the model used to determine the difficulty score and ultimately the curriculum is the target classifier $M(\theta, \rho)$, however, the examples being scored are the ASACs generated, i.e.,\ $x'_a$ which are generated as a function of the protected attributed classifier $C(\theta, \phi)$. %Figure \ref{fig:confidence} shows the effect of various noise magnitudes on the confidence of the target classifier $M(\theta, \rho)$ in it's prediction. This is analogous to $1 - DS(x'_a)$.

\subsubsection{Curriculum Assignment}
\label{sub:cl_fair_adv_learning}
%We propose a curriculum learning framework to adversarially fine-tune our target classifier $M(\theta, \rho)$ using ASACs. Each noise magnitude produces ASACs for a particular level of difficulty, to which the model responds differently (see Table \ref{tab:noise_mags} for an ablation on the effect of different noise magnitudes). As demonstrated in Figure \ref{fig:confidence}, the confidence of the model $M(\theta, \rho)$ in the misclassification of ASACs varies with the degree of noise magnitude used in producing the ASACs. Defining a curriculum on ASACs produced from various noise magnitudes allows us to guide the adversarial fine-tuning of $M(\theta, \rho)$, thereby leading to faster convergence, higher accuracy and improved fairness metrics. 
%We propose a curriculum learning framework to fine-tune our target classifier $M(\theta, \rho)$ using ASACs. ASACs, generated with different noise magnitudes, represent varying difficulty levels, affecting the model's confidence in misclassifications. By defining a curriculum based on ASACs from different noise magnitudes, we guide the adversarial fine-tuning process, enhancing convergence speed, accuracy, and fairness metrics.

For each batch of $k$ images $\{x_1, \dots , x_k\}$, we compute the ASACs as the Cartesian product of $k$ images with the set of $l$ noise magnitudes we use in the curriculum (including $\epsilon = 0$ i.e., the original images). For each of the samples in the resulting batch of size $k \times l$, we compute a difficulty score, $\textrm{DS}(x_a)$, detailed in Equation \ref{eq:ds}, that scores an ASAC based on the difficulty of the example for the target classifier to classify correctly. We then sort this batch based on the difficulty of each sample and the curriculum strategy chosen. Finally, we partition the sorted batch into $l$ mini-batches allowing us to take a gradient step through each mini-batch, thereby utilizing the ASACs in a curriculum. See Table \ref{tab:cl_order} for an ablation on the effect of a randomized, increasing, and decreasing order (\textit{w.r.t.} difficulty score) for samples in the curriculum. This procedure is illustrated in detail in Algorithm \ref{alg:ASAC} and enables what defines each in-batch curriculum in the fine-tuning process of the target classifier $M_{(\theta, \rho)}$.

\section {Experiment Setup}
\label{sec:exp_setup}
This section describes our experimental setup. We outline the datasets used, followed by the evaluation metrics that are employed to evaluate the model and finally describe the training details.

\subsection{Datasets}
\label{sub:dataset}
Similar to Wang et al.'s approach \cite{wang2022assessing}, we use the CelebA \cite{liu2015faceattributes} and  the UTK \cite{zhifei2017cvpr} dataset. CelebA contains 202,599 images annotated with 40 attributes and we focus on Smiling, Big Nose, and Wavy Hair, with Gender as the protected attribute. Similarly, The UTK dataset  comprises of over 20,000 images in the wild with only a single image per face.

\subsection{Training Details}
\label{sub:train_details}
%We utilize ResNet-18 \cite{he2016deep} as the general backbone architecture unless specified otherwise, pretrained for 50 epochs and fine-tuned with our proposed approach for 10 epochs, utilizing the same data distribution $D$. A batch size of 128 is used with Adam optimizer (learning rate $10^{-4}$). We experiment with $\epsilon \in [0,1]$ for varying adversarial noise levels in images, with smaller $\epsilon$ implying less noise and values closer to 1 indicating higher corruption.

While we experiment with various architectures to assess the generalizability of our approach, ResNet-18 \cite{he2016deep} serves as the default backbone architecture unless specified otherwise. The base model is pretrained for 50 epochs and subsequently fine-tuned using our proposed method for an additional 10 epochs, leveraging the same data distribution $\mathcal{D}$. Training is conducted with a batch size of 128 using the Adam optimizer \cite{kingma2014adam}, with hyperparameters $\beta_1 = 0.9$, $\beta_2 = 0.999$, and $\epsilon = 1 \times 10^{-8}$. Gradient clipping is applied with a maximum norm of 1.0.

To evaluate the effect of adversarial perturbations, we vary the noise magnitude $\epsilon$ within the range $[0,1]$, where smaller values correspond to subtler perturbations and values closer to 1 indicate higher image corruption. Unless otherwise noted, we use noise magnitude values of ${0.01, 0.001}$ and set the curriculum learning weight parameter to $\alpha = 0.5$.

Experiments are run on multiple hardware configurations. Most training and fine-tuning experiments are performed on a compute cluster equipped with a Nvidia RTX A6000 GPU (48 GB VRAM), 384 GB RAM, and 20 CPU cores, with each run taking approximately 60 to 120 minutes. Additional runs are executed on an Apple M1 Pro system (8-core CPU, 14-core GPU, 16-core Neural Engine, and 16 GB RAM), and a Nvidia A100 GPU (40 GB VRAM) with 25 GB RAM for large-scale comparisons. Unless stated otherwise, all models are trained and fine-tuned using ResNet-18 with the aforementioned default hyperparameters.
%-$18$ \dhruv{missing citation}\cite{he2016deep} as the backbone architecture. The models are pre-trained for $50$ epochs and fine-tuned using the method for adversarial counter tuning for $15$ epochs. We use a batch size of $128$ and train the model using the Adam optimizer \cite{kingma2014adam} with a learning rate of $10^{-4}$. We experiment with different values of $\epsilon \in [0,1]$ \lee{what are the exact values?} to inject varying amounts of adversarial noise in the images. A smaller value of $\epsilon$ signifies less adversarial noise added to images while values closer to $1$ indicate that the images have been largely corrupted.

\section{Experiments}
\label{sec:ASAC_results}

In this section, we present and discuss various experiments to evaluate and interpret our proposed approach. We evaluate our approach through quantitative performance metrics and fairness evaluations, followed by qualitative analyses using visualization methods for model understanding.

%In addition, we conduct ablations on different curriculum policies, noise magnitudes ($\epsilon$) and network architectures. We use two popular datasets, CelebA \cite{liu2015faceattributes} and Labeled Faces in the Wild (LFW) \cite{LFWTech}. To evaluate the fairness of our model, we employ three key metrics: the Difference in Demographic Parity (DDP), the Difference in Equalized Odds (DEO), and the Difference in Equalized Opportunity (DEOp) \cite{hardt2016equality}. Alongside these, we also measure the model's accuracy (ACC). Lower values for fairness metrics indicate a less biased model whereas a high value in accuracy indicates a more performant model. For more details about the metrics and their definition please refer to the supplementary material (Section \ref{sup:Deifintion}% and \ref{sup:metrics}

%Our results can broadly be divided into three parts. In the first, we quantitatively evaluate our models in terms of their performance as well on fairness metrics outlined in Section \ref{sub:eval_metrics}. The second part qualitatively evaluates the models trained through our approach; providing visualizations using interpretability methods for visualizing the disentanglement of latent representations and the regions of the image most involved in the discriminative abilities of the models. Finally, we perform ablations over curriculum policies employed, the effect of various noise magnitudes ($\epsilon$), and the impact of different weights ($\alpha$) in the adversarial loss proposed (see Equation \ref{eq:adv_loss}). 

\subsection{Quantitative Results}
\label{sub:quant_results}

\begin{table*}[t]
\caption{A comparison of our approach with other approaches \cite{zhang2020towards,ramaswamy2021fair} on the CelebA dataset.}
\label{tab:celeba_results}
%\vspace*{-0.15in}
\centering

\begin{tabular*}{\textwidth}{@{\extracolsep{\fill}} llrrrr}
\toprule
Label & Method  & DEO & DEOp &DDP& ACC\\
\midrule
% \multirow{}{}{} & \multirow{}{}{\centering Smile } & \multirow{}{}{} \\
\midrule
Smile & Base Classifier  &0.088 &0.066&	0.15&	84.29  \\
&Adversarial Training \cite{zhang2020towards}	&0.077&	0.051&	0.17	& 91.74 \\
&Counterfactual GAN based-training \cite{ramaswamy2021fair}  & 0.065 &0.050  &0.17  &  86.50 \\
&ASAC (with FGSM)&\textbf{0.050} &\textbf{0.043}	&0.15	&\textbf{91.91}\\
&ASAC (with PGD )&0.058 &0.045 &\textbf{0.14}	&91.20\\

\midrule
% \multirow{}{}{} & \multirow{}{}{\centering Big Nose} & \multirow{}{}{} \\
\midrule
Big Nose & Base Classifier  & \textbf{0.332} & \textbf{0.257} & \textbf{0.28} & 80.77  \\
&Adversarial Training \cite{zhang2020towards} & 0.396 & 0.311 & 0.35 & 81.53 \\
&Counterfactual GAN based-training \cite{ramaswamy2021fair}  &0.392  &0.301  &0.36 &  78.77 \\
&ASAC (with FGSM) & 0.354 & 0.267 & 0.30 & \textbf{82.05}\\
&ASAC (with PGD) & 0.374 & 0.281 & 0.31 & 81.03\\

\midrule
% \multirow{}{}{} & \multirow{}{}{\centering Wavy  Hair} & \multirow{}{}{} \\
\midrule
Wavy Hair & Base Classifier & 0.347 & 0.237 & 0.35 & 81.15\\
&Adversarial Training \cite{zhang2020towards}    & 0.359 & 0.264 & 0.40 & \textbf{82.07}\\
&Counterfactual GAN based-training \cite{ramaswamy2021fair}  &0.344  &0.230  &0.34 &  79.21 \\
&ASAC (with FGSM) &\textbf{0.34} & \textbf{0.29} & 0.342 & 82.01\\
&ASAC (with PGD) &0.383 & 0.243 &\textbf{ 0.34} & 81.99\\

\bottomrule
\end{tabular*}

\end{table*}

\begin{table}[t]
\caption{Performance across accuracy and fairness metrics on the UTK dataset with age being the protected attribute and gender being the target classifier.}
\label{tab:age}
%\vspace*{-0.15in}
\begin{center}

\begin{tabular}{lrrrr}
\toprule
Method & \hspace{3mm} DEO & \hspace{3mm} DEOp & \hspace{3mm} DDP & \hspace{3mm} ACC \\
\midrule

Base & 0.216	&0.180	& \textbf{0.155}	&65.81\\
ASAC &\textbf{0.195}&	\textbf{0.175}&	0.184	&\textbf{71.08}
 \\
\bottomrule
\end{tabular}

\end{center}

\end{table}
\begin{table}[t]
\caption{Comparison of accuracy and fairness metrics between base classifier and fine-tuning using the proposed approach across different backbone architectures.}
\label{tab:backbone_results}

\begin{center}

\begin{tabular}{llrrrr}
\toprule
Method & Model & DEO & DEOp &DDP& ACC\\
\midrule
% \multirow{}{}{} & \multirow{}{}{\centering Wavy  Hair} & \multirow{}{}{} \\
\midrule

Base  & ResNet-18 &0.088 &0.066 &0.150 &84.29 \\
Base  & ResNet-50 &0.077 &0.064 &0.148 &86.60\\
Base  & DenseNet-121 &0.090  &0.075  &0.146 &80.36 \\
Base  & DenseNet-169 &0.092 &0.075 &0.139 &80.95\\
ASAC & ResNet-18 &0.050 &0.043 &0.150 &\textbf{91.91}\\
ASAC & ResNet-50 &\textbf{0.047} &\textbf{0.042} &\textbf{0.137} &90.78\\
ASAC & DenseNet-121 &0.052 &0.046 &0.153 &91.32\\
ASAC & DenseNet-169 &0.070 &0.051 &0.158 &91.08\\

\bottomrule
\end{tabular}

\end{center}

\end{table}

\subsubsection{Comparing the proposed method across different target attributes on different datasets} As shown in Table \ref{tab:celeba_results}, we quantitatively evaluate the performance and fairness of our approach on the CelebA \cite{liu2015faceattributes} dataset, focusing on target attributes such as Smile, Big Nose, and Wavy Hair, with Gender as the protected attribute. Using ResNet-18 \cite{he2016deep} as the backbone, we compare models trained with and without our proposed approach and evaluate against the original baseline model and two different methods: an adversarial training approach \cite{zhang2020towards} and a debiasing approach by Ramaswamy et al. \cite{ramaswamy2021fair} that generates counterfactual images using GANs. Our approach, utilizing FGSM \cite{goodfellow2014explaining} and PGD \cite{madry2017towards} adversarial methods (with $\epsilon=\{0,0.01,0.001\}$), outperforms these previous methods in terms of both accuracy and fairness metrics. The GAN-based counterfactual approach \cite{ramaswamy2021fair} reduces the overall accuracy of the system post training, however, our method not only improves accuracy for all three target attributes, but also improves fairness metrics (for all target attributes except big nose). 

%We conduct the same analysis on the LFW dataset \cite{LFWTech}, targeting the attributes Smiling, Bags Under Eyes, and Wavy Hair, considering the influence of Gender as the protected attribute. As shown in Table \ref{tab:lfw_results}, our findings on our proposed approach on the LFW dataset \cite{LFWTech} are consistent with those on the CelebA dataset \cite{liu2015faceattributes}, indicating an increase in overall accuracy and improvements across most fairness metrics. %\dhruv{I've rewritten 5.1. I think we should end the paragraph here. Thoughts?}, particularly for Big Nose \lc{this is not what I'm reading from the table}. \pushkar{Made changes here}.The results demonstrate our approach's superiority across different architectures, showing significant improvements in both accuracy and fairness metrics.\pushkar {Above paragraph is rewritten to incorporate comments on different architectures and baselines}
\begin{figure*}

  \centering  
   \includegraphics[trim=100 0 100 0, clip,width=\textwidth]{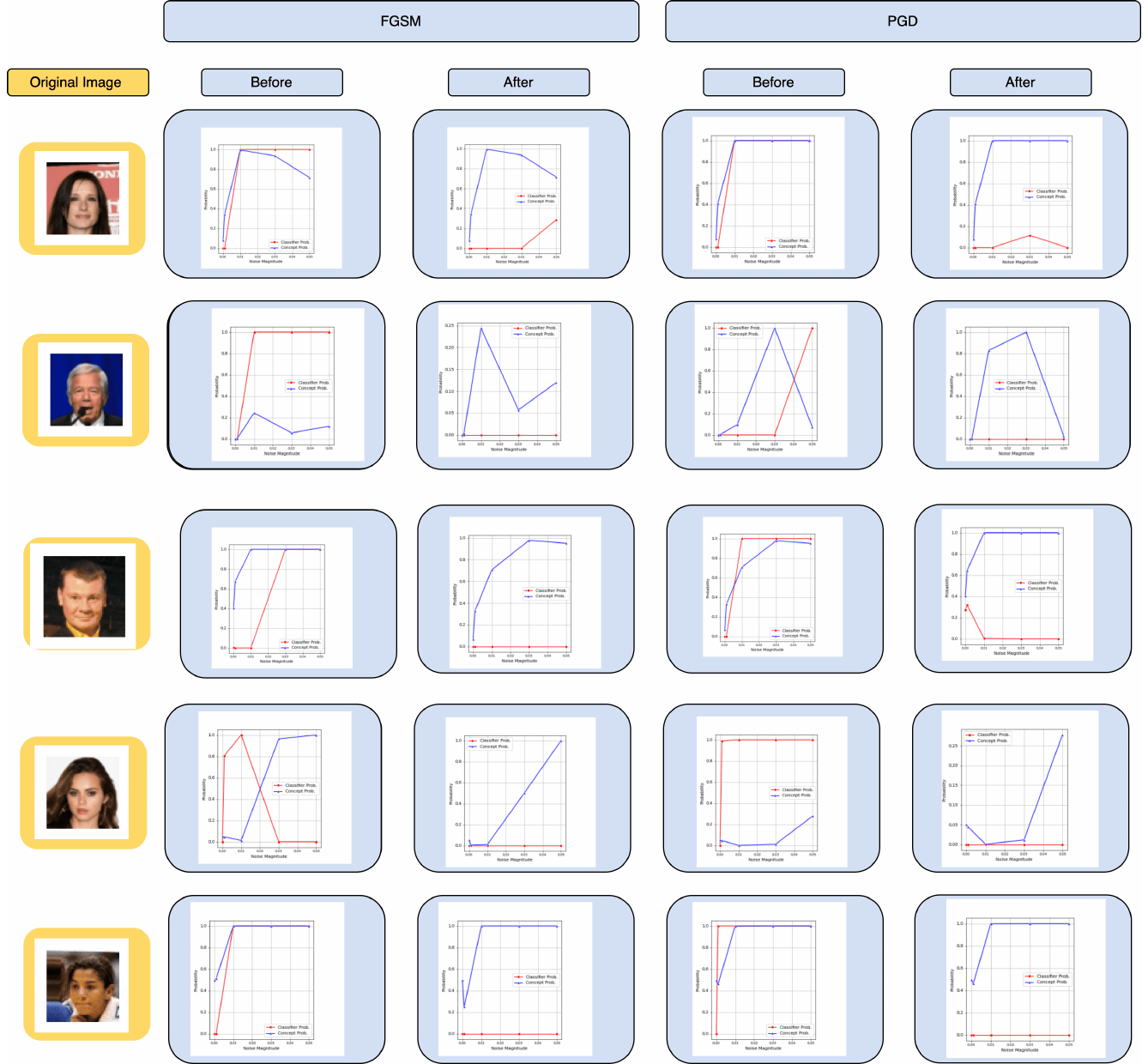}
   \caption{Qualitative results showing that our trained model becomes robust to ASACs after training. The blue curve represents the sensitivity to adversarial noise before training, and the red curve represents the sensitivity after training. Post-training, the images showed increased robustness to adversarial noise, with predictions remaining unchanged even at higher noise levels.}
\label{fig:rob_postive}
\end{figure*}
Most of our experiments focus on Gender as the protected attribute due to its availability in relevant datasets and clear binary classes. However, we also test our approach on a different protected attribute class using a different dataset. Specifically, we use the UTK dataset \cite{zhifei2017cvpr} to train a gender classifier with Age as the protected attribute, categorizing individuals as "older" (40+) or "younger" ($<$ 40), given the dataset's roughly equal split around this age. Using a ResNet-18 architecture, our binary age classifier demonstrates that the proposed approach improves both fairness and accuracy metrics, consistent with previous settings. 
\vspace*{-0.1in}
\subsubsection{Comparing the proposed approach across different backbones}Table \ref{tab:backbone_results} presents a quantitative comparison of our approach using different backbones (ResNet-18 \cite{he2016deep}, ResNet-50 \cite{he2016deep}, DenseNet-121 \cite{huang2017densely}, and DenseNet-169 \cite{huang2017densely}) on the CelebA dataset \cite{liu2015faceattributes}. The overall size of the models vary from 8M to 24M parameters. Similar to our evaluation on the CelebA and UTK datasets, we use the FGSM attack with $\epsilon=\{0,0.01,0.001\}$ to construct ASAC mini-batches across each batch sampled. Our method enhances both accuracy and fairness, demonstrating its scalability and generalization across model sizes.
%Our method obtains significant improvements in accuracy while also improving the fairness of the model, showing that our approach generalizes across architectures and scale of the model.
\subsubsection{Evaluation on a wider set of features}
\label{sup:ram_eval}
We also perform evaluations on a wider range of target labels and aggregate the results, similar to the evaluation protocol in Ramaswamy et al \cite{ramaswamy2021fair}. We show the results before and after averaging fairness and performance metrics in Table~\ref{tab:ASAC_mean_results} and Table~\ref{tab:ASAC_individual_results} respectively. Our setting was similar to the setting explained in Table \ref{tab:cl_order} of their paper. We considered 4 attributes: Young, Pointy Nose, Attractive, and Arched Eyebrows. We use Gender as the protected attribute. Te experiments were conducted on the CelebA dataset. The mean metrics aggregated across these attributes are given below.
\begin{table*}[h]
    \centering
   \caption{Following  Ramaswamy et al. \cite{ramaswamy2021fair}, we show the mean fairness metrics and accuracy before and after fine-tuning, aggregated over target attributes - Young, Pointy Nose, Attractive and Arched Eyebrows with the protected attribute as Gender. }
    \begin{tabular}{lcccc}
        \toprule
        Method & DEO & DEOp & DDP & ACC \\
        \midrule
        Base & 0.29$\pm$0.04 & 0.28$\pm$0.07 & 0.30$\pm$0.04 & 75.70$\pm$0.05 \\
        Proposed Approach & \textbf{0.28$\pm$0.01} & \textbf{0.23$\pm$0.01} & \textbf{0.26$\pm$0.01} & \textbf{77.20$\pm$0.03} \\
        \bottomrule
    \end{tabular}
   
    \label{tab:ASAC_mean_results}
\end{table*}
\begin{table*}[h]
    \centering
     \caption{Fairness and accuracy results for individual attributes before and after fine-tuning}
    \begin{tabular}{lcccc}
        \toprule
        Attribute & DEO & DEOp & DDP & ACC \\
        \midrule
        Young & 0.103 & 0.213 & 0.233 & 82.47 \\
        Young (ASAC training) & \textbf{0.101} & \textbf{0.213} & \textbf{0.247} & \textbf{85.12} \\
        \midrule
        Pointy Nose & 0.289 & 0.218 & 0.187 & 69.85 \\
        Pointy Nose (ASAC training) & \textbf{0.282} & \textbf{0.206} & \textbf{0.180} & \textbf{72.97} \\
        \midrule
        Attractive & 0.285 & 0.268 & 0.408 & \textbf{74.65} \\
        Attractive (ASAC training) & \textbf{0.281} & \textbf{0.221} & \textbf{0.338} & 73.10 \\
        \midrule
        Arched Eyebrows & 0.482 & 0.363 & 0.357 & 75.90 \\
        Arched Eyebrows(ASAC training) & \textbf{0.438} & \textbf{0.301} & \textbf{0.281} & \textbf{77.84} \\
        \bottomrule
    \end{tabular}
   
    \label{tab:ASAC_individual_results}
\end{table*}
\subsection{Qualitative Results}
\label{sub:qual_results}

%\subsubsection{Robustness evaluation}
\label{subsub:robust_eval}

%\begin{figure}[tb]
  %\centering
   %\includegraphics[width=\linewidth]{fig/ablation_pre_and_post_training.png}
   %\caption{A comparison of the confidence of the smile classifier in its prediction, in response to ASACs constructed at various noise magnitudes. The response is measured before (in blue) and after (in red) the proposed fine-tuning procedure.}% We observe more consistency in the model's ability to maintain its prediction, indicating resilience to ASAC-induced decision flips.}
   %\label{fig:effect_of_noise}
%\end{figure}

\subsubsection{Robustness evaluation}
\label{subsub:robustness}
 To assess the impact of adversarial training using ASACs, we examine how adversarial images designed to attack a gender classifier affect the decision of the smile classifier, pre and post-training with our proposed approach. Given an image, we add varying magnitudes of adversarial noise and provide the corrupted images to the smile classifier. We observe that before using our proposed method, the adversarial noise intended to mislead the gender classifier also misleads the smile classifier, indicating a possible correlation between gender and smile attributes in these examples. Such an outcome suggests that manipulating the protected attribute gender can influence predictions on smile. We perform an identical analysis on the models after our proposed fine-tuning approach. We observe that after fine-tuning, the target classifier improves its discriminative ability while the protected attribute classifier used to construct ASACs continues to be deceived. This indicates that the fine-tuning procedure helps to decouple spurious correlations between the target and protected attribute, enabling the target classifier to make more accurate predictions independent of the protected attribute. Examples of these results have been shown in Figure \ref{fig:rob_postive}. For each image we look at how the target classifer (smile) changes at different levels of noise. The blue curve shows the sensitivity to adversarial noise before training, while the red curve shows the sensitivity after training. After training, the images were found to be more robust to adversarial noise, with predictions remaining stable even at higher noise magnitudes.

\subsection{Ablation Studies}
\label{sub:ablations}
\subsubsection{\textbf{Evaluating the impact of Curriculum Learning}}
\label{sub:curriculum_ablations}

%In our ablation study, we use a ResNet-18 backbone with the target attribute as Smile and the protected attribute as Gender. We utilize FGSM with $\epsilon=\{0,0.01,0.001\}$ to generate the ASACs for each batch sampled from the CelebA dataset.

\begin{table}[t]
\caption{Comparing different curriculum learning strategies. Here is CL stands for curriculum learning and $\uparrow$ denotes that examples are shown with increasing order of difficulty scores while $\downarrow$ denotes that examples are shown in decreasing order of difficulty scores.}
\label{tab:cl_order}
%\vspace*{-0.15in}
\begin{center}

\begin{tabular}{lrrrr}
\toprule
Method & \hspace{3mm} DEO & \hspace{3mm} DEOp & \hspace{3mm} DDP & \hspace{3mm} ACC \\
\midrule
Baseline & 0.088 & 0.066 & 0.15 & 84.29 \\ 
Randomized  & 0.055	&0.044	&\textbf{0.14}	&91.23\\
CL in $\uparrow$ & \textbf{0.050}	&\textbf{0.043}	& 0.15	&91.79\\
CL in $\downarrow$   &0.058&	0.048&	0.17	&\textbf{92.08}
 \\
\bottomrule
\end{tabular}

\end{center}
\vskip -0.1in
\vspace*{-0.15in}
\end{table}

In the initial set of ablation analyses, we explore different training strategies. We implement a curriculum learning approach, wherein the model is trained in ascending order of difficulty. We contrast this strategy with an alternative approach where the model encounters the most challenging examples at the outset, followed by progressively easier ones. Table \ref {tab:cl_order} presents a comparison of these strategies across various performance metrics. An intriguing pattern emerges: while exposing the model to difficult examples towards the end of training yields superior performance on fairness metrics, the reverse strategy—starting with challenging examples—results in the highest accuracy attained by the system thus far. This observation sheds light on the impact of curriculum learning on the model. One strategy enhances the accuracy of the model, while the other demonstrates greater improvement in fairness metrics. This highlights the crucial role of curriculum design in shaping the trade-off between accuracy and fairness in machine learning models.

\subsubsection{Comparison across different noise magnitudes}
% \label{subsub:noise_ablation}

In our second ablation study (Table \ref{tab:ASAC_noise_mags_alpha_ablation2}), we investigate the effect of varying noise magnitudes on our smile classifier model's performance. Using FGSM as our chosen adversarial image generation method, we test different magnitudes of $\epsilon$ before implementing curriculum learning. Our results show that $\epsilon= {0.03}$ and $\epsilon= {0.05}$ yield optimal performance. We then incorporate examples generated at these magnitudes, along with original images, into our curriculum learning setup. We perform these experiments on a smile classifier with gender as the protected attribute that is trained on the CelebA dataset.  This experimentation offers valuable insights into selecting noise magnitudes that enhance the model's robustness and effectiveness in adversarial scenarios. 

\begin{table}[t]
\caption{Ablation on the effect of different values of noise magnitudes ($\epsilon$) on the performance and fairness metrics of the target classifier.}
\label{tab:ASAC_noise_mags_alpha_ablation2}
%\vspace*{-0.15in}
\begin{center}

\begin{tabular}{lcrrrr}
\toprule
 \hspace{3mm} Setting & \hspace{3mm} DEO & \hspace{3mm} DEOp & \hspace{3mm} DDP& \hspace{3mm} ACC \\
\midrule
% $\epsilon$ & \{0.001,0.03,0.05\}&0.065	&0.047	&0.16	&91.00\\ % is this last one really .00 (which I added to "91")?
 \{0.001, 0.01\}&\textbf{0.050} &\textbf{0.043}	&0.15	&\textbf{91.91}\\
 % & \{0.001\}&0.054	&0.043	&\textbf{0.14}	&91.62\\
 % & \{0.05\}&0.055	&0.044	&0.14	&91.71\\
 %& \{0.001,0.03,0.05\}&0.057 &	0.048	&0.16 &	91.73\\
  \{0.001,0.03\}&0.055	&0.043	&\textbf{0.14}	&91.42\\
 \{0.001,0.05\}&0.057	&0.047 &0.14	&91.38\\
\{0.001,0.03,0.05\}&0.065	&0.047	&0.16	&91.00\\ % is this last one really .00 (which I added to "91")?
  \{0.001\}&0.054	&0.043	&0.14	&91.62\\
  \{0.05\}&0.055	&0.044	&0.14	&91.71\\
\midrule
%\midrule
%$\alpha$ & 0.3 &0.06	&0.04	&0.16& 92.04  \\
 %& 0.5  &0.058 &0.050 &	0.16&	91.84 \\
 %& 0.7 &0.057&0.047	&0.14&91.90 \\
%\bottomrule
\end{tabular}

\end{center}

\end{table}

\subsubsection{Experiments over multiple seeds}
To ensure that our results are not due to overfitting we conduct experiments on a small scale over five runs on the ResNet18 architecture with different seeds. As shown below, across five trials, our method averaged improved accuracy as well as across our fairness metrics. The results are presented in Table \ref{sup:error_bars}.

\begin{table*}[h]
    \centering
 \caption{Comparison of classifiers before and after fine-tuning averaged over five different seeds. These experiments were performed on Smile classifier trained on CelebA dataset with a ResNet-18 backbone.}
    \begin{tabular}{lcccc}
        \toprule
        Method & DEO & DEOp & DDP & ACC \\
        \midrule
        Base & 0.14 & 0.12 & \textbf{0.13} & 84.31 \\
        Proposed Approach & \textbf{0.12$\pm$0.04} & \textbf{0.10$\pm$0.03} & 0.17$\pm$0.08 & \textbf{90.50$\pm$0.09} \\
        \bottomrule
    \end{tabular}
   
    \label{sup:error_bars}
\end{table*}

\section{Conclusion}
We propose a novel method for generating attribute-specific adversarial counterfactuals to improve fairness and performance of computer vision models. These counterfactuals are constructed via adversarial perturbations along protected attributes and are used to fine-tune existing classifiers. Unlike traditional generative counterfactuals, our approach preserves the semantic content of the original image, reducing the risk of introducing spurious correlations or stereotypes. We further introduce a curriculum learning-based fine-tuning regime, which assigns adversarial examples of increasing difficulty to progressively train the model.

Our experiments across multiple datasets demonstrate that fine-tuning with these adversarial counterfactuals significantly improves fairness metrics without sacrificing accuracy—in fact, in many cases, overall accuracy is enhanced. Qualitative analysis also suggests that the model is better able to disentangle predictions from protected attributes. This positions our framework as a dual-purpose tool for both performance improvement and fairness enhancement. Looking forward, we aim to extend our approach to handle multiple protected attributes and explore how adversarial counterfactuals can be leveraged for interpretable model diagnostics.

This work directly addresses ethical concerns surrounding biased decision-making in vision systems by mitigating discrimination in already trained models. However, the approach comes with important limitations. First, it assumes access to model weights, making it inapplicable to black-box systems. Second, it requires explicit knowledge of protected attributes, which may not always be available or appropriate to collect. While our method offers a promising step toward post hoc fairness correction, its deployment in real-world systems must be coupled with careful consideration of privacy, consent, and downstream social implications.

\subsection{Limitations}
While our work introduces an in-situ method for generating counterfactuals—thereby preventing biases from external or third-party models from influencing the system—a key limitation remains: it does not address the biases that may already exist within the model itself. These internal biases can significantly affect the fairness and reliability of the system. Additionally, as with CAVLI, our approach does not automate the selection of relevant concepts for bias mitigation; this process remains manual and dependent on human input. Furthermore, the datasets used to fine-tune attribute classifiers may themselves carry underlying biases, which can propagate into the final model behavior. These limitations highlight important directions for future work, aimed at automating concept selection, auditing internal model bias, and ensuring fairness in the training data.
\def\modelname{TIBET\xspace}
\chapter{TIBET: Using counterfactuals for identifying and evaluating biases in Text-to-Image generative models}

\epigraph{“The eye sees only what the mind is prepared to comprehend.”}{\textit{Robertson Davies}}
\begin{figure*}[ht]
  \centering
   \includegraphics[width=0.9\linewidth]{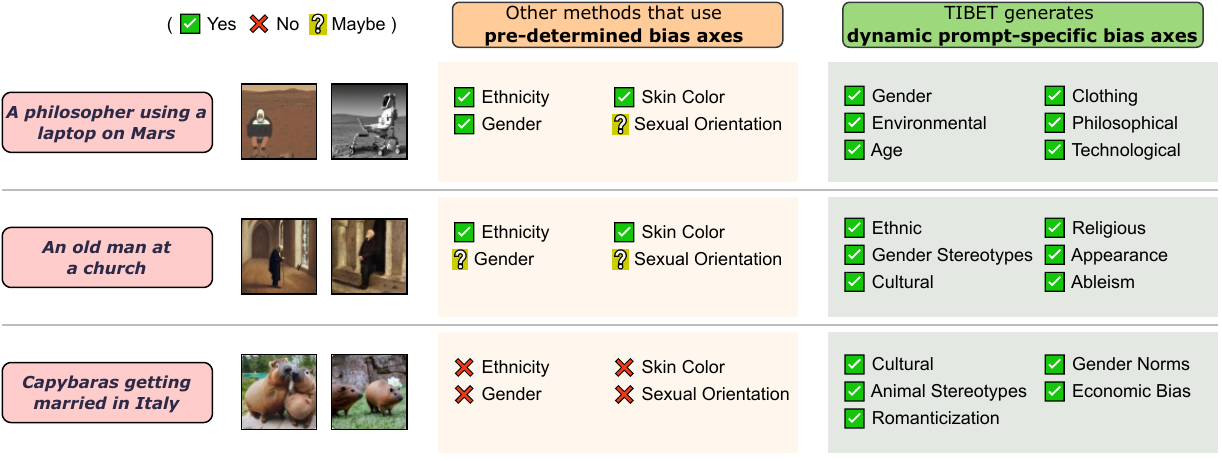}
   \caption{ In generative modeling, the dimensions along which bias manifests are highly dependent on the input prompt. For example, the relevant axes of bias for “a photo of a philosopher” may involve gender, age, or cultural presentation, while a prompt like “capybaras getting married in Italy” may invoke romanticization, geographic, or animal stereotypes. Unlike prior methods that evaluate models using a fixed set of demographic axes (e.g., gender or race), our approach dynamically surfaces context-relevant bias dimensions per prompt—enabling more nuanced bias analysis in text-to-image generation systems.}
   \label{fig:Tibet_header}
\end{figure*}
\clearpage

\section{Introduction}
\label{sec:intro}

In prior chapters, we explored counterfactual reasoning as a powerful methodology for explaining model decisions (via CAVLI) and for mitigating bias in vision models through carefully constructed counterfactuals (via ASACs). These chapters demonstrated how changing one concept—while holding others fixed—can reveal the dependencies and assumptions baked into model behavior.

We now extend this methodology to generative multimodal systems, specifically text-to-image (TTI) models. Unlike classification systems, where a label must be predicted, TTI models generate complex outputs conditioned on natural language prompts making them particularly susceptible to subtle, prompt-sensitive biases. A small change in prompt wording can drastically alter generated content (e.g., “an old man at a church” and “an Asian old man at a church”). This chapter introduces a framework for dynamically identifying and evaluating such biases in TTI systems through prompt-level counterfactuals.

Generative text-to-image (TTI) models have emerged as a prominent research area in computer vision over the past few years. These models are capable of producing high-quality images based on natural language descriptions and have found applications in various fields, including online content creation and image editing. However, despite their promise, TTI models have demonstrated various kinds of biases in the images they generate, as shown in prior research \cite{ghosh2023person,esposito2023mitigating,bianchi2023easily,friedrich2023fair,cho2023dall,wang2023t2iat}. Therefore, the identification and mitigation of biases is crucial in order to fully harness the capabilities of these models.

Existing approaches for measuring biases in TTI models typically employ a predefined set of bias axes (gender, age, and skin color) along which biases are assessed, aggregating over a fixed domain of prompts (e.g., occupation prompts \cite{luccioni2023stable}). This line of work is useful in measuring the relative bias of TTI models. However, biases evaluated in one prompt domain may vary from another, and may even vary from prompt to prompt.
In such cases, averaging across prompts may even mask certain biases. For instance, Figure \ref{fig:Tibet_header} illustrates three distinct input prompts, each associated with a different axis of bias. Here, measuring a predefined bias (e.g., gender) across all prompts is less meaningful (and may underestimate or mask bias due to prompt irrelevance). Ultimately, having the ability to asses biases for individual prompts is as important as doing so in aggregate over a domain of prompts, and the former is lacking from most existing approaches \cite{ghosh2023person,esposito2023mitigating,bianchi2023easily,cho2023dall,wang2023t2iat}.

Additionally, images for a user-provided prompt, or set of prompts, can exhibit different types of biases. These biases may be societally harmful in nature (\textit{societal biases}), or simply be a result of common co-occurrences in the real world or in data that the TTI model was trained on (\textit{incidental correlations}). For example, a computer programmer is often depicted as male (\textit{societal bias}) wearing glasses (\textit{incidental correlation}). While societal biases are most important and are generally analyzed in related works \cite{luccioni2023stable, cho2023dall}, the existence of incidental correlations \cite{bhatt2024mitigating} may also lead to reduced diversity in the generated images, and therefore must also be identified. Henceforth, for simplicity, we use the term ``\textit{bias}'' to represent both societal biases and incidental correlations. Finally, a good measurement method should not only be capable of quantifying biases, but also of providing interpretable insights.

In this chapter, we introduce a novel framework called \modelname\ (\textbf{T}ext to \textbf{I}mage \textbf{B}ias \textbf{E}valuation \textbf{T}ool) for examining, quantifying, and explaining a wide range of biases in images generated by TTI models. Our approach is designed to be compatible with any TTI model, and versatile across any user-provided prompts. In contrast to prior work that rely on a predefined set of biases, we dynamically identify potential biases relevant to the given prompt by leveraging an LLM like GPT-3.5. Next, we generate counterfactual prompts for the identified bias axes, and images sets for the input prompt and all counterfactual prompts using the TTI model we want to evaluate. Finally, we compare the images from the initial prompt and the counterfactual prompts, using a new metric, the Concept Association Score ($\textrm{CAS}$), and further quantify biases using Mean Absolute Deviation ($\textrm{MAD}$). Our model has the ability to provide post-hoc explanations to gain qualitative insights about biases in images generated by TTI models. Furthermore, we can aggregate our metrics over a domain of prompts with the same biases and counterfactuals, in line with previous work. 

Our experiments demonstrate that TIBET not only excels in scenarios where previous approaches \cite{wang2023t2iat,cho2023dall} have been employed, such as detecting gender stereotypes in occupational prompts, but it can also be effectively combined with bias mitigation techniques like ITI-GEN \cite{zhang2023iti}. ITI-GEN is a bias mitigation technique that focuses on editing prompts in text-to-image generation to reduce the presence of stereotypes. It works by automatically generating counterfactual prompts along underrepresented or marginalized attributes and rebalancing the output distributions of generative models. By doing so, it helps mitigate biases by ensuring that generated images better reflect diversity across attributes like gender, race, and age. This combination offers a more comprehensive and automated approach to bias mitigation in TTI models. Moreover, we conduct user studies to validate our approach with human judgement. 

\vspace{0.1in}
 Our contributions can be summarized as follows. First, we propose an automated approach for identifying and measuring biases in images generated by TTI models, accommodating the dynamic nature of biases across different input prompts. Unlike prior works \cite{wang2023t2iat,cho2023dall,luccioni2023stable}, our framework evaluates images on a diverse set of bias axes encapsulating both societal and incidental biases. Second, we propose  novel quantitative metrics, $\textrm{CAS}$ and $\textrm{MAD}$, that can be used to quantify these biases and also offer post-hoc explanations along different dimensions of biases. 

\section{Related Works}

\noindent \textbf{Measuring biases in TTI models.} Much research has been conducted on evaluating and mitigating common social biases in image-only models \cite{buolamwini2018gender,seyyed2021underdiagnosis,hendricks2018women,meister2023gender,wang2022revise,liu2019fair,joshi2022fair,wang2020towards,wang2023overwriting} and text-only models \cite{bolukbasi2016man,hutchinson2020social,shah2020predictive,garrido2021survey,ahn2021mitigating}. However, recent research is extending these studies to include multimodal models and datasets, exploring various aspects of language and vision. These investigations encompass biases found in embeddings \cite{hamidieh2023identifying}, text-to-image \cite{cho2023dall,bianchi2023easily,seshadri2023bias,ghosh2023person,zhang2023iti,wang2023t2iat,esposito2023mitigating}, retrieval \cite{wang2022assessing}, image captioning \cite{hendricks2018women,zhao2021scaling}, and visual question-answering models \cite{park2020fair,aggarwal2023fairness,hirota2022gender}.

Nonetheless, limited attention has been given to understanding biases in text-to-image (TTI) models. Existing approaches such as T2IAT \cite{wang2023t2iat}, DALL-Eval \cite{cho2023dall}, and other works \cite{ghosh2023person,esposito2023mitigating,bianchi2023easily,friedrich2023fair} for evaluating and mitigating biases in TTI models differ from our work in several key ways. They mainly focus on predefined bias axes like gender \cite{wang2023t2iat, cho2023dall,ghosh2023person,esposito2023mitigating,bianchi2023easily}, skin tone \cite{wang2023t2iat, cho2023dall,ghosh2023person,esposito2023mitigating,bianchi2023easily}, culture \cite{esposito2023mitigating,wang2023t2iat}, or location \cite{esposito2023mitigating}, whereas our approach is dynamic, allowing for more flexible bias measurement. Additionally, many of these existing methods \cite{wang2023t2iat,esposito2023mitigating} require specific prompt structures, whereas our approach can assess bias for any input prompt. Moreover, our approach goes a step further by offering post-hoc concept-level explanations. This helps users analyze the presence or absence of different semantic concepts in the images, enhancing their understanding of these biases, and providing insight into our metrics.Current research on bias identification is compactly summarized in Table \ref{tab:relatedworks}.
\begin{table}[tb]
  \centering
  \caption{{\bf Summary of select prior work.} Three relevant characteristics are considered for each method.}
  \label{tab:relatedworks}
  \resizebox{0.8\columnwidth}{!}{
  \begin{tabular}{@{}lccc@{}}
    \toprule
     Related Work & Bias-Axes & Counterfactual  & Concept-level Explainability \\
    
    \midrule
    T2IAT \cite{wang2023t2iat}  & Predefined & \checkmark & - \\
    DALL-Eval \cite{cho2023dall}  & Predefined & - & \checkmark \\
    Stable Bias \cite{luccioni2023stable} & Predefined & \checkmark & \checkmark\\
    Esposito et al. \cite{esposito2023mitigating}  & Predefined & \checkmark & - \\
    
    \bottomrule
    \textbf{Ours}  & \textbf{Dynamic} & \textbf{\checkmark} & \textbf{\checkmark} \\
    \bottomrule
  \end{tabular}}
\end{table}
\section{Method}
\label{sec:method}

\begin{figure*}[tb]
  \centering
   \includegraphics[width=0.9\linewidth]{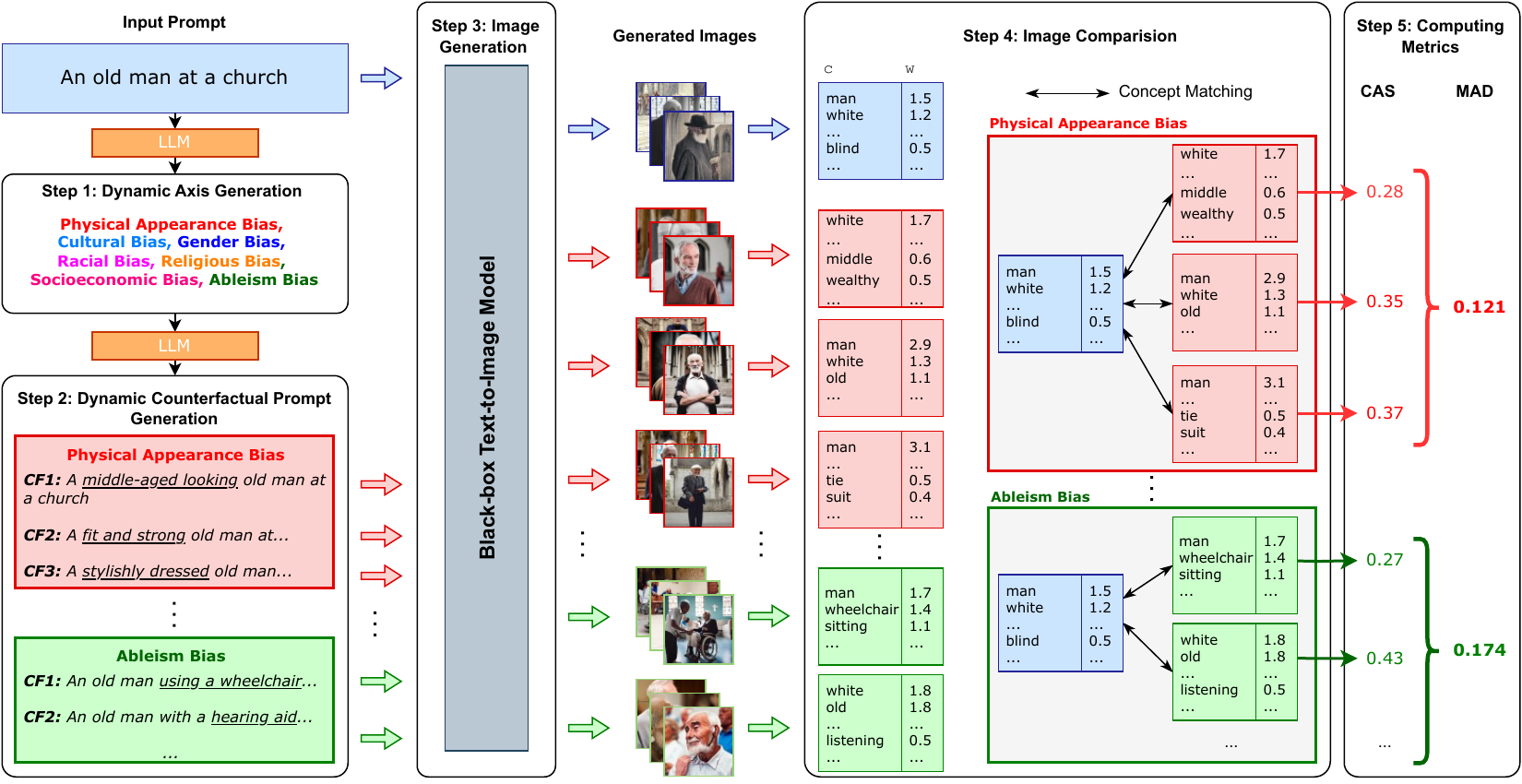}
   \caption{\textbf{\modelname}. Given an input prompt, we query an LLM (GPT-3) to identify axes of biases (Step 1), and generate counterfactual prompts for each axis of bias (Step 2). Here, we show a sample of three counterfactual prompts for the physical appearance bias, and two for the ableism bias. Next, we use a black-box TTI model (Stable Diffusion \cite{rombach2022high}) to generate images for the initial prompt as well as each counterfactual for all axes of bias (Step 3). In this example, we leverage VQA-based concept extraction to obtain a list of concepts and their frequencies for each set of images, and compare the concepts of the initial set with concepts of each counterfactual to obtain $\textrm{CAS}$ scores (Step 4). Finally, we compute $\textrm{MAD}$, a measure of how strong the bias is in the images generated by the initial prompt (Step 5).}
   \label{fig:TIBET_main}
   \vspace{-0.2cm}
\end{figure*}

Given an input prompt $P$, we first dynamically generate bias axes relevant to $P$, and then generate counterfactuals along each bias axes (Steps 1-2 in Figure \ref{fig:TIBET_main}). This process is detailed in Section \ref{sub:dynamiccfg}. We then generate images using a black box TTI model for the input prompt, and each of the counterfactual prompts across all bias axes (Step 3 in Figure \ref{fig:TIBET_main}; Section \ref{sub:t2imodel}). Finally, we compare image sets of the input prompt with image sets of each counterfactual prompt, through the use of VQA-based concept decomposition and CLIP, and a novel metric, the Concept Association Score ($\textrm{CAS}$) (Step 4-5 in Figure \ref{fig:TIBET_main}; Section  \ref{sub:imagecompare}). Furthermore, we detail quantitative ($\textrm{MAD}$) and qualitative metrics for bias evaluation and post-hoc explainability in Section \ref{sub:metrics}.

\subsection{Dynamic Bias Axes and Counterfactuals}
\label{sub:dynamiccfg}
We propose a prompt-dependent, dynamic bias-axes and counterfactual generation scheme that, given an input prompt $P$, generates relevant bias axes, followed by the generation of counterfactual prompts along those axes. Counterfactuals for an input prompt are generated in two steps, using chain-of-thought reasoning in LLMs \cite{wei2022chain, zhang2023automatic}. Firstly, the input prompt is used to dynamically create a list of bias axes representing dimensions of biases that are potentially present in the model (Step 1 in Figure. \ref{fig:TIBET_main}).
The creation of these bias axes is facilitated by Large Language Models (GPT-3 \cite{brown2020language}), leveraging their ability to comprehend complex relationships. These axes then serve as the foundation for generating counterfactual prompts within their respective dimensions (Step 2 in Figure \ref{fig:TIBET_main}). 

\subsubsection{Counterfactual prompt generation using Chain of Thought Prompting}
\label{sup:dynmaicccfg}

We use GPT-3 (specifically, \texttt{gpt-3.5-turbo}) for bias axis and counterfactual generation, through a series of well-defined queries,
\begin{enumerate}
     \item  For the image generation prompt, \texttt{\small <initial prompt>}, what are some of the axes where the prompt may lead to biases in the image?
    \item  Generate many counterfactuals for each axis. Create counterfactuals for all diverse alternatives for an axis. Each counterfactual should look exactly like the original prompt, with only one concept changed at a time.
    \item  Convert these to a $\textrm{JSON}$ dictionary where the axes are the keys and the counterfactuals are list for each key. Only return json.
\end{enumerate}

\noindent where \texttt{\small <initial prompt>} is replaced with the user-provided initial prompt to the model. All GPT-3 generations were done using the OpenAI API, in October and November 2023.

\subsection{Text-to-Image Generation}
\label{sub:t2imodel}

The initial prompt and the counterfactual prompts are fed into a black-box Text-to-Image (TTI) model (that is to be evaluated), which generates a set of images for the input prompt, $I^P$, and counterfactual prompts, $I^{P_{cf}}$ (Step 3 in Figure. \ref{fig:TIBET_main}). Our approach works for any black-box TTI model, and we experiment with Stable Diffusion 1.5 and 2.1. For each input prompt and counterfactual prompt, we generate 48 images.

\subsection{Image Comparison}
\label{sub:imagecompare}

The primary motivation for employing counterfactuals is to discern the proximity or differences between images generated for a given prompt $P$ and those produced for prompts altered along an axis of bias. This comparison enables us to gauge whether images generated for a specific prompt exhibits bias towards a particular counterfactual. Hence, we propose an image comparison module to compare two sets of images. This module can utilize any existing framework or model for comparing image sets. In our study, we investigate two distinct methods inspired by previous works \cite{wang2023t2iat, luccioni2023stable, cho2023dall}. Expanding upon these approaches, we introduce a novel metric termed Concept Association Score ($\textrm{CAS}$) to quantify the similarity between image sets.

\subsubsection{Method 1: VQA-based concept extraction}

\begin{figure}[t]
  \centering
   \includegraphics[width=0.5\linewidth]{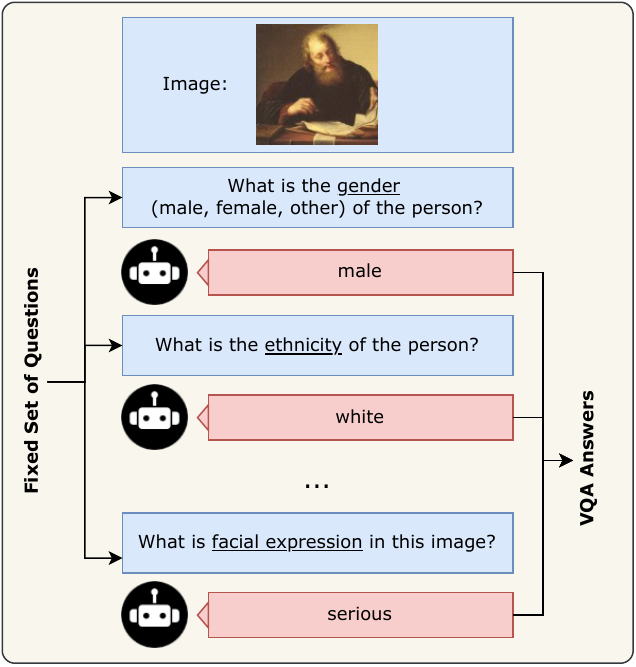}
   \caption{\textbf{VQA-based Image Comparison.} The VQA process in the Image Comparison Module. For each image in the initial set and the counterfactual set, we ask a list of questions to the VQA model, MiniGPT-v2, to extract concepts. The underlined words indicate the axis of bias that is associated with that question.}
   \label{fig:vqa}
\end{figure}

In this approach, we use MiniGPT-v2 \cite{chen2023minigptv2}, a recent vision-language model with competitive performance in various VL tasks, in a question-answer format to extract information from generated images. An example of this process is illustrated in Figure \ref{fig:vqa}. Given an initial prompt $P$, we generate a set of questions that are aligned with the axes of bias $B$ that may be present in the images. For commonly occurring axes of bias, such as gender, age, or ethnicity, we hand-design VQA questions that can be used to query each image. For example, if ``gender bias'' is an axis of bias for a prompt, then we add ``\texttt{\small What is the gender of the person?}'' to the set of VQA questions. For other axes of biases, we implement a template question ``\texttt{\small What is \{bias-name\} in the image?}'' where \texttt{\small \{bias-name\}} is simply replaced with the type of bias (see Table  \ref{tab:vqa-questions}). The questions asked for a prompt $P$ and its counterfactuals $P_{cf}$ remain the same.

We use the VQA setting of MiniGPT-v2 to obtain answers to axis-specific questions, as well as a caption for the image. For frequently occurring axes of bias, we write well-defined questions. These questions are in Table \ref{tab:vqa-questions}.

\begin{table*}
  \centering
  \caption{For commonly occurring axes of bias, we pre-define VQA questions.}
    \label{tab:vqa-questions}
  \begin{tabular}{@{}ll@{}}
    \toprule
     \textbf{Axis of Bias} & \textbf{VQA Question} \\
     \midrule
        age & What age group (young, middle, old) is the person in the image? \\
        ethnicity & What is the ethnicity of the person in the image? \\
        race & What is the ethnicity of the person in the image? \\
        racial & What is the ethnicity of the person in the image? \\
        socioeconomic & What is the socioeconomic status of the person in the image? \\
        gender & What is the gender (male, female, other) of the person in the image? \\
        nationality & What is the nationality of the person in the image? \\
        style & What is the style of the image? \\
        setting & What is the setting of the image? \\
        color & What color is the image? \\
        emotion & What is the emotion of the person in the image? \\
        occupation & What is the occupation of the person in the image? \\
        culture & What is the culture depicted in the image? \\
        fashion & What is the person wearing? \\
        clothing & What is the person wearing? \\
        appearance & Describe the appearance in the image. \\
        background & Describe the background of the image. \\
        \bottomrule
    \end{tabular}
\end{table*}

All VQA answers for all set of images of $P$ and $P_{cf}$ are combined and pre-processed, to obtain a list of entities that describe the set of images. We measure the occurrence of each entity by calculating its frequency over the answers and captions.
The final list of entities, and their frequencies, are considered as the concepts set $C$ that are extracted for a set of images generated by one prompt. Ultimately, we have $C_{\textrm{init}} = \{(c^i_1,w^i_1)\ldots\}$ for the initial prompt, and $C_{\textrm{cf}} = \{(c^{cf}_1,w^{cf}_1)\ldots\}$ for a counterfactual prompt, where $c$ is a concept described in natural language, and $w$ is the frequency of that concept in the VQA answers across the given set of images.

$\textrm{CAS}$ measures the similarity between generated images for the initial prompt and each counterfactual prompt in terms of relevant concepts. $\textrm{CAS}_{\textrm{VQA}}$ uses a concept-level matching algorithm to compare concepts generated for the two sets, as defined in Algorithm \ref{alg:cas}. This concept-level matching algorithm merges synonym words in $C_{\textrm{init}}$ and $C_{\textrm{cf}}$, and reduces the concept lists to two histograms of word frequencies for the initial and counterfactual concept sets. Now, we define $\textrm{CAS}$ as the histogram Intersection-over-Union between the frequency ($\mathcal{W}$) of the two sets of concepts:
\begin{align}
    \mathcal{W}^\cap &=  \textrm{min}(w^i, w^{\textrm{cf}}); \forall_{w^i,w^{cf}\in C_{init},C_{cf}} \\
    \mathcal{W}^\cup &= \textrm{max}(w^i, w^{\textrm{cf}}); \forall_{w^i,w^{cf}\in C_{\textrm{init}},C_{cf}} \\
    \textrm{CAS} &= \frac{\sum_i \mathcal{W}^\cap_i} {\sum_j \mathcal{W}^\cup_j  }
\end{align}

\noindent where $w^i$ and $w^{cf}$ are the concept frequencies for the same concept in the initial and counterfactual concept sets.

\begin{algorithm}
\caption{Concept Association Score}
\label{alg:cas}
\begin{algorithmic}[1]
\State{ In: $C_{\textrm{init}} = \{(c^i_1,w^i_1)\ldots\}$ and $C_{cf} = \{(c^{cf}_1,w^{cf}_1)\ldots\}$}

\State{\textbf{Step 1: Merge Synonym Concepts}}
\State{Build concept vocabulary, $\{c^i_1, \ldots, c^{cf}_1, \ldots\}$}
\State{Obtain synonyms for each concept in the vocabulary using WordNet}
\For{all concepts in $C_{\textrm{init}}$}
    \If{$c^i_j$ is synonym of $c^i_k$}
        \State{merge $(c^i_k,w^i_k)$ into $(c^i_j,w^i_j)$ to get $(c^i_j,w^i_j+w^i_k)$}
        \State{remove $(c^i_k,w^i_k)$}
    \EndIf
\EndFor
\State{ Repeat loop above for $C_{cf}$}

\State{\textbf{Step 2: Add missing concepts}}
\State{For any concept that is present in $C_{init}$ but not in $C_{cf}$, add the concept into $C_cf$ with a frequency of 0, and vice versa.}

\State{\textbf{Step 3: Compare Histograms}}
\State{Re-order $C_{init}$ and $C_{cf}$ to the same order, as in the vocabulary, so that corresponding concept frequencies can be compared.}
\State{ $\textrm{CAS} = \textrm{HistIoU}(w^i_*, w^{cf}_*)$ where $w^i_*$ and $w^{cf}_*$ are the frequencies in $C_{\textrm{init}}$ and $C_{cf}$ respectively.}
\end{algorithmic}
\end{algorithm}

\subsubsection{Method 2: Vision-Language Embedding models} In this approach, we directly embed all images using CLIP \cite{radford2021learning}, and compare each image in the initial prompt set, to every image in the counterfactual prompt set using the cosine similarity metric. We then compute $\textrm{CAS}^{\textrm{CLIP}}$ as the mean of the cosine similarity scores, as follows:

\begin{align}
    CAS^{CLIP} &= mean \left([cosine(CLIP(I^i), CLIP(I^{cf}))]_{\forall{I^i,I^{cf}\in I^P, I^{P_{cf}}}}\right).
\end{align}

In both methods, $\textrm{CAS}$ values range between $[0,1]$, where $0$ indicates no association and  $1$ indicates complete matching of the two concept sets. Unlike $\textrm{CAS}$, which is derived from VQA concepts, $\textrm{CAS}^{\textrm{CLIP}}$ scores are limited in terms of post-hoc explainability as they only rely on image embeddings.

\subsection{Metrics for Bias Evaluation} %\vspace{-0.2cm}
\label{sub:metrics}

$\textrm{CAS}$ scores measure the similarity between the input prompt and each counterfactual for a given axis of bias.  If we have $K$ counterfactuals generated for a bias axis $b$, then we obtain a distribution of $K$ $CAS$ scores as follows:
\begin{align}
    \textrm{CAS}_{K}^\textrm{\textrm{b}} &= [\textrm{CAS}_{1}; \dots; \textrm{CAS}_{k}]^\textrm{b}
\end{align}

If this distribution of CAS scores is uniform, i.e., each counterfactual image set is equally similar to the initial set, it indicates that there high diversity in the initial set and low bias along that axis. Conversely, if this distribution is skewed towards one counterfactual, it indicates that the initial set is heavily biased towards that set. Therefore, a measure of variability in a distribution of $\textrm{CAS}$ scores can allow us to quantify the amount of bias along an axis, and compare the degree of bias along one axis against another. To that end, employ a statistical measure, Mean Absolute Deviation (MAD), for bias evaluation. 

Moreover, we propose two qualitative metrics, \textit{Top-K concepts}, and \textit{Axis-aligned Top-K concepts}, attempt to provide post-hoc explanations about commonly occurring concepts in images generated for a prompt, when VQA-based concept extraction is used.

\subsubsection{Quantitative Metric: Mean Absolute Deviation (MAD)}

$\textrm{MAD}$ is used to measure the degree of bias with respect to a bias-axis. $\textrm{MAD}$ is given as

\begin{align}
    \textrm{MAD} &= \frac{1}{K}\sum^K_{i=1}{|\textrm{CAS}_{i} - \overline{\textrm{CAS}_K}|}
\end{align}
\noindent where $K$ is the number of counterfactuals for a bias axis $b \in B$, $CAS_k$ represents the weight represented for corresponding to the $k^{th}$ counterfactual, and $\overline{\textrm{CAS}_K}$ is the mean of the scores. We normalize $\textrm{MAD}$ against the $\textrm{MAD}$ score of the most skewed distribution with length K (where all scores are 0, except a single 1) in order to be able to compare bias axes even when K varies between them. Additional details regarding this are in  \ref{sub:MADappendix}

\begin{figure*}[tb]
  \centering
   \includegraphics[trim=0 0 275 0, clip,width=0.9\linewidth]{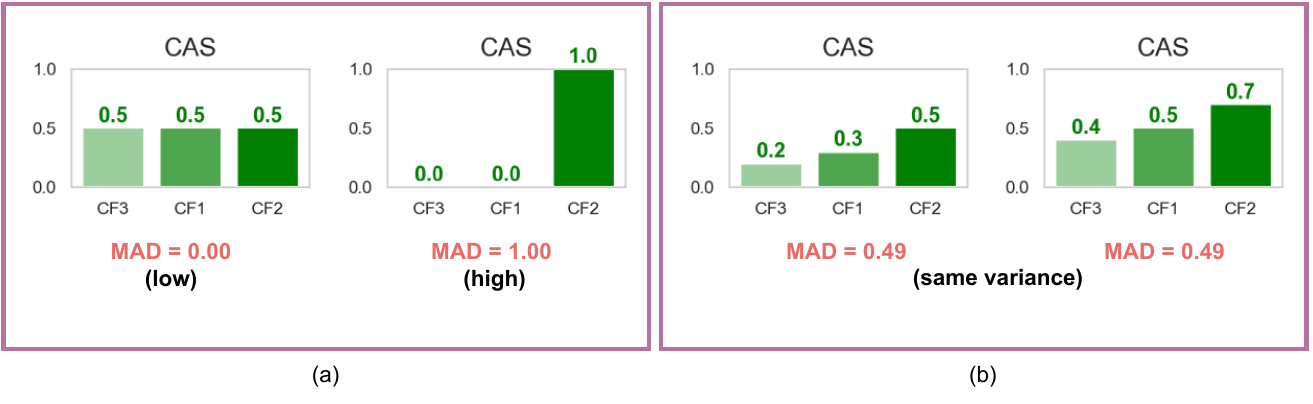}
   \vspace{-0.2cm}
   \caption{\textbf{Metrics:} (a) $\textrm{MAD}$ is low when the $CAS$ scores are uniform across all counterfactuals, and high when the $\textrm{CAS}$ scores are skewed. (b) $\textrm{MAD}$ is only dependent on variability in $\textrm{CAS}$, not on amount of $\textrm{CAS}$}
   \label{fig:MAD_metrics}
\end{figure*}

Once normalized, $\textrm{MAD}$ scores are between [0,1], where a low score suggests that the images generated for the initial prompt exhibits relatively low bias along a specific axis, and a high score indicates a strong association between the initial prompt and one counterfactual, indicating a higher likelihood of bias in the images. We illustrate the behaviour of $\textrm{MAD}$ in Figure \ref{fig:MAD_metrics}.

\subsubsection{Qualitative Measures}
\label{sec:qualmeasures}

In addition to our quantitative metrics, we also provide qualitative concept-based explanations (for the VQA-based method) to reason about the measured change in concepts between the initial set and the counterfactual set using two simple qualitative metrics defined below. Our qualitative metrics are: 
\begin{itemize}
    \item \textbf{Top-K concepts.} For a given input prompt, the Top-K concepts show the most commonly occurring concepts in images generated for a prompt.
    \item \textbf{Axis-aligned Top-K concepts.} The Axis-aligned Top-K concepts show the most frequently occurring concepts in a given image for a specific bias axis. To calculate this measure we extract concepts from questions specific to a bias axis and sort them in order of frequency over the image set.
\end{itemize}

\section{Dataset}
\label{sec:dataset}

\noindent\textbf{Predefined prompts for gender stereotypes in occupations.} 
In order to evaluate our method against existing methods like T2IAT \cite{wang2023t2iat}, DALL-Eval \cite{cho2023dall}, and Stable Bias \cite{luccioni2023stable} that studied gender stereotypes in occupational images generated by TTI models, we use pre-defined prompts for 11 occupations, including ``computer programmer'', ``elementary school teacher'', ``architect'' and others. These prompts follow the format \texttt{\small ``A photo of a <occupation>''}, with \texttt{\small <occupation>} representing one of the 11 occupations. We also create male and female counterfactuals mirroring the ones used by T2IAT, and generate 48 images for each set using Stable Diffusion 1.5 and Stable Diffusion 2.1. 

\noindent\textbf{Varied Text Prompts for Evaluation.}
As our method is capable of using any input prompt, we create a set of 100 prompts to comprehensively assess our method's performance in bias evaluation, including:
(1) Creative Prompts: This subset includes diverse and imaginative prompts meticulously written to evaluate our method thoroughly. Some examples are ``\texttt{\small astronauts cooking dinner on the moon}'' and ``\texttt{\small a boy at a museum}''. 
(2) Prompts from DiffusionDB: We also sample prompts from DiffusionDB \cite{wangDiffusionDBLargescalePrompt2022}, a database of 2.3M distinct human-generated TTI prompts across two sets, \texttt{2M} and \texttt{Large}. These prompts undergo pre-processing where we extract the most descriptive substrings.% The entire list of 100 prompts, pre-processing code, and the biases identified by GPT-3 for each prompt is presented in Table \ref{tab:fulldataset}

\clearpage
\begin{figure}[t]
  \centering
   \includegraphics[width=0.70\linewidth]{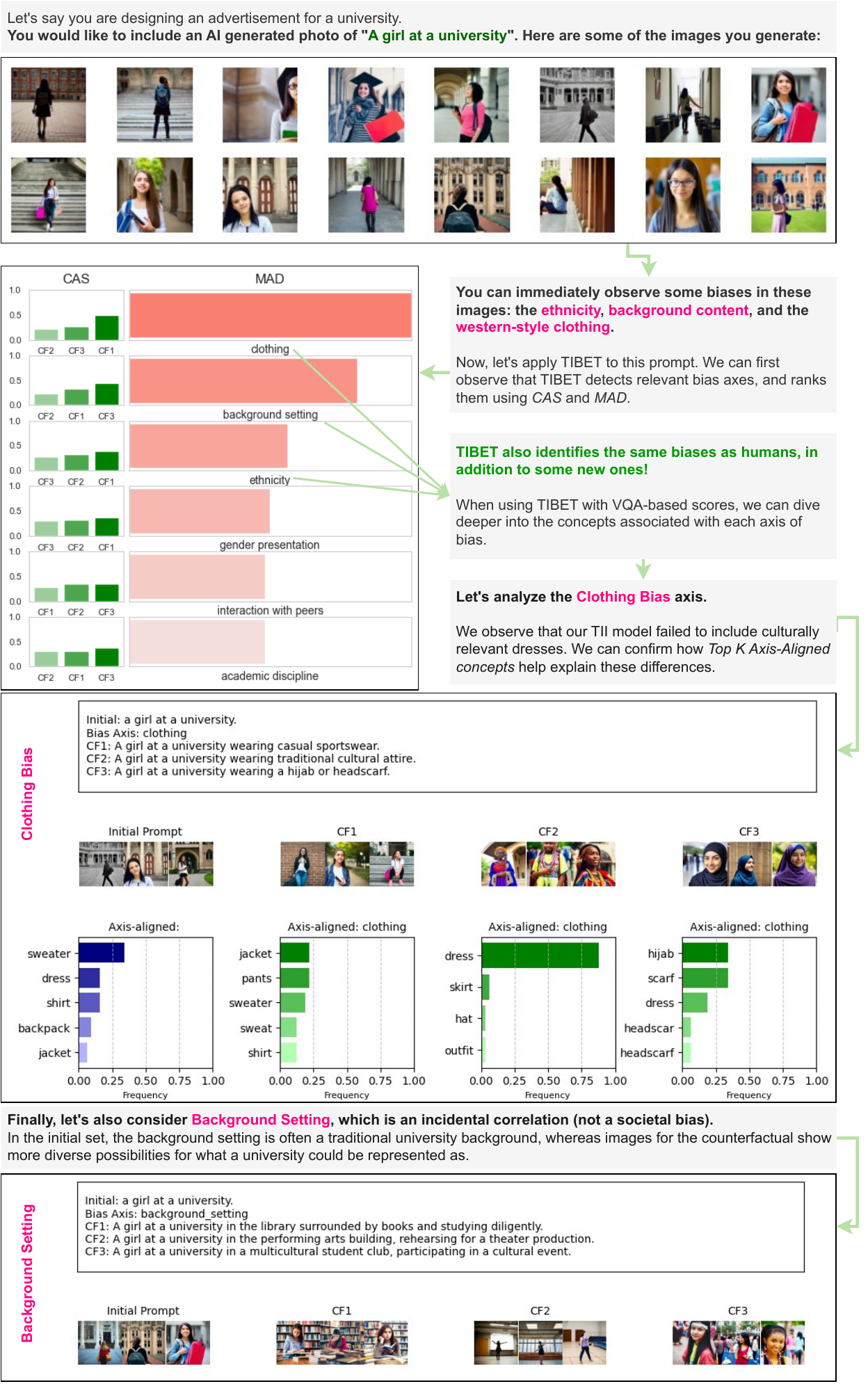}
   \caption{\textbf{Usefulness of TIBET}. In this example setting, we show how TIBET can be useful to a user concerned about biases in the images generated by a TTI model. We show how TIBET can analyse biases along human-observable axes of bias, with post-hoc explainablity.}
   \label{fig:fullpageexample}
\end{figure}
\clearpage
\section{Experiments}
\label{sec:experiments}
%\vspace{-0.2cm}

The section can be broadly divided into three parts. Firstly, in Section \ref{sec:qualitative} and Section \ref{sec:genderstereotypes}, we utilize our approach to examine biases in various prompts. Secondly, in Section \ref{sec:senstivity}, we analyze how biases in VQA models impact our approach. Thirdly, in Section \ref{sec:userstudies}, we present human evaluations aimed at assessing the alignment between human judgments and our approach.% Beyond these, additional experiments such as a user study to analyze VQA models (Appendix 6), comparisons of metrics (Appendix 2.3-2.4), and more qualitative examples to demonstrate TIBET's explainable capabilities (Appendix 5.1) are incorporated in supplementary material.

\begin{figure*}[tb]
  \centering
  \includegraphics[width=1.0\linewidth]{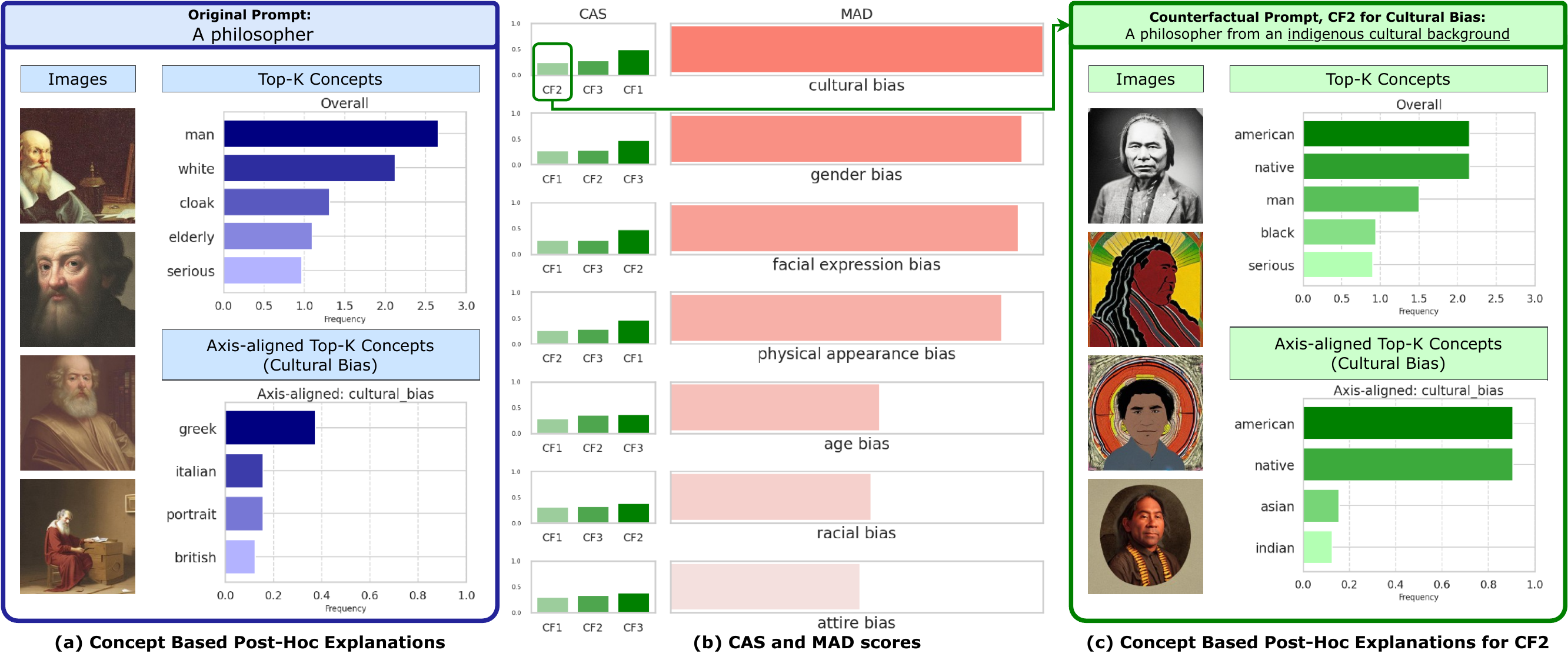}
   \caption{{\bf Analysis enabled by TIBET.} Our approach calculates $\textrm{CAS}$ and $\textrm{MAD}$ scores to measure association with counterfactual prompts and bias degree in generated images. Qualitative metrics like Top-K Concepts and Axis-Aligned Top-K Concepts offer post-hoc model explanations. Additionally, our approach enables comparisons with counterfactual explanations.}
   \label{fig:variance}
   %\vspace{-0.1in}
\end{figure*}

\subsection{Qualitative Results}
\label{sec:qualitative}

We show an example of analysing biases in images generated by the prompt ``\textit{a philosopher}'' using \modelname\ with Stable Diffusion 2.1, in Figure \ref{fig:variance}. In (a), we show images generated for the initial prompt, as well as Top-K concepts and Axis-Aligned Top-K concepts for Cultural Bias. These concepts are ordered by their frequency. In (b), we show $\textrm{CAS}$ (in green) and $\textrm{MAD}$ (in orange) scores of all counterfactuals across all axes. The $\textrm{MAD}$ score tells us which bias may be stronger in the initial set of images. This plot provides a birds-eye view of the biases that are most prominent in the initial set (here, we notice that cultural, gender and facial expression biases are most prominent, as they have higher $\textrm{MAD}$ scores), and study the $\textrm{CAS}$ scores of each counterfactual for every bias axis. Finally, in (c), we show an example of one counterfactual that has a low $\textrm{CAS}$ score, and show that the images and concept frequencies can be compared to those in (a). By observing concepts in the Top-K concepts list, we can validate what the metrics tell us. Here, ``man'' and ``serious'' are the only two concepts that remain in the top five, indicating the large difference between the two image sets, explaining low $\textrm{CAS}$ score. Furthermore, comparing Axis-aligned Top-K Concepts helps us understand the significant differences in cultural depictions by the TTI model for the initial prompt, compared to the counterfactual, where the initial prompt has mostly ``Greek'' philosophers, whereas the counterfactual has ``Native American'' philosophers. 

Another example is shown in Figure \ref{fig:fullpageexample}, where we illustrate how TIBET helps users identify and understand biases in images generated by a Text-to-Image model. For instance, given the prompt ``\textit{A girl at a university},'' our metric highlights potential biases across different axes. The user can then explore these axes in detail-for example, examining the ``\textit{clothing}'' dimension and observing that the model consistently depicts girls in sweaters or dresses, thereby revealing a stereotypical bias embedded in the prompt.

% Overall, \modelname\ allows us to gain deep prompt-specific insights, allowing users to not only quantify biases, but also validate the metrics with concept-level explanations. We provide additional examples, with $\textrm{CAS}$ and $\textrm{CAS}^{CLIP}$, in Appendix 5.

\subsection{Measuring Gender Stereotypes in Occupations}
\label{sec:genderstereotypes}
%\vspace{-0.2cm}
Previous research \cite{bianchi2023easily, wang2023t2iat, cho2023dall} has brought attention to the issue of gender bias in generated images for profession-related text prompts. Building upon these findings, we embarked on a similar investigation to explore gender stereotypes in images generated by TTI models when provided with occupational prompts (as detailed in Section \ref{sec:dataset}). In our study, we assess the disparity between the $\textrm{CAS}$ values for male and female counterfactuals. This assessment allows us to determine whether the images for that profession lean male or female, shedding light on potential gender-related biases in the generated images.

Our analysis of the generated images (in Figure \ref{fig:gender-mitigation} (a) and (b)) indicates that when using Stable Diffusion 1.5, images generated for  ``elementary school teachers'' and ``librarians'' are female dominated, whereas ``announcer,'' ``chef,'' and most other occupations are male dominated. Stable Diffusion 2.1 seems to reduce bias among a few of these professions, notably ``accountant'' and ``pharmacist.'' The overall trends observed in our bias metrics align with those found in previous studies like T2IAT and DALL-Eval. There are, however, discrepancies in the strengths of trends because our metric assesses image sets based on concept-level information, which distinguishes it from previous approaches.

Furthermore, we can use TIBET across a set of prompts with the same bias axes, to compute the an aggregate measure of bias. We observe an average $\textrm{CAS}$ score of 0.56 for the male counterfactual and  0.44 for the female counterfactual across all 11 occupations, for images from Stable Diffusion 1.5. Whereas with Stable Diffusion 2.1, we get 0.52 and 0.48 for male and female counterfactuals, respectively, indicating lower gender bias in the newer model.

\begin{figure*}[tb]
  \centering
   \includegraphics[trim=630 0 0 0, clip,width=0.6\linewidth]{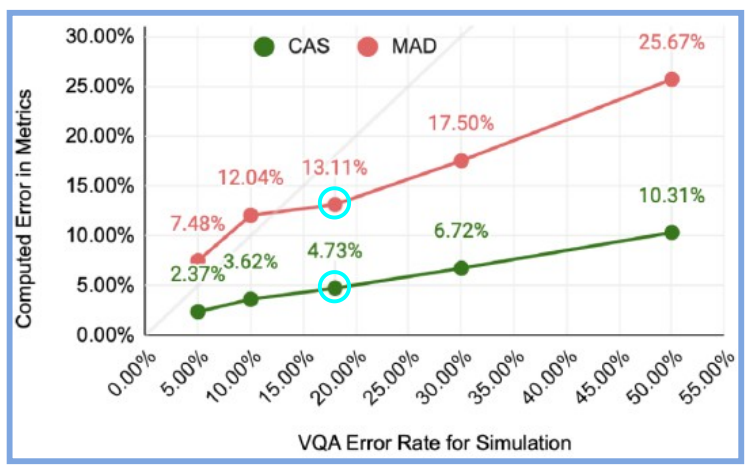}
   \vspace{-0.2cm}
   \caption{\textbf{Sensitivity Analysis} on $\textrm{CAS}$ and $\textrm{MAD}$ for errors in VQA. Per User Study 3 , we estimate an $18\%$ error rate in VQA, leading to $4.73\%$ and $13.11\%$ error in $\textrm{CAS}$ and $\textrm{MAD}$ respectively.}
   \label{fig:metrics}
\end{figure*}

\subsection{Sensitivity of Metrics to Errors in the VQA Model}
\label{sec:senstivity}

Our goal with TIBET is to provide an accurate analysis of potential biases in the images generated by a TTI model for a given input prompt, inexpensively and efficiently. We are required to use models such as MiniGPT-v2 to conduct automatic analysis without expensive human annotations. These VL models carry their own biases, and these biases may be propagated to our metrics. Therefore, it is essential to conduct a sensitivity analysis of $\textrm{CAS}$ and $\textrm{MAD}$ scores. For VQA, we do this by simulating errors in answers. We assume IID errors at the image level, and average our rate of change of $\textrm{CAS}$ and $\textrm{MAD}$ across 10 simulation runs for 30 random prompts in our dataset, across all bias axes.

In Figure \ref{fig:metrics}, we show the results of our sensitivity analysis. We recognize that $\textrm{CAS}$ and $\textrm{MAD}$ do propagate error in VQA into our scores, but do so at a rate lower than the original rate of error. As we use a large set of images in each set that we compare, the top concepts from VQA remain less affected by the per-VQA errors. For an error rate of $18\%$ (established in User Study 3) in VQA, we observe that $\textrm{CAS}$ only changes by $4.73\%$ and $\textrm{MAD}$ by $13.11\%$.

\subsection{Human Evaluation}
\label{sec:userstudies}

\textbf{User Study 1: Evaluating Dynamically Generated Bias Axes.}
\label{sec:userstudy1}
We conduct a user study to evaluate the concurrence of axes of bias chosen by human participants and those generated by LLMs across 100 input prompts. A high level of agreement serves as an indicator of the effectiveness of LLMs in generating bias axes that are both contextually relevant and aligned with human perspectives. Participants are tasked to identify if, for a given prompt, an axis of bias (e.g., gender) may potentially cause societal or incidental biases in the generated images. For each prompt, we present 10 axes of bias, including the ones that the LLM generated, and the rest from a random sample of biases. Each question is answered by five MTurk workers. Further details about the setup of this study are provided in Appendix \ref{sup:userstudy1}. 

We perform two experiments, measuring precision and recall across all axes (overall) and specifically for commonly occurring societal biases (societal). 
The results in Table \ref{tab:userstudy1} reveal a high precision of 0.90 in both experiments, showing human agreement with the LLM on generated biases. In the overall case, a recall of 0.54 suggests that LLMs capture only a subset of human-indicated biases. However, in the societal case, a recall of 0.87 demonstrates GPT-3's strong ability to identify harmful societal biases in prompts.
    
\begin{table}[tb]
    \parbox{.48\columnwidth}{
  \caption{\textbf{User Study 1: Can GPT-3 detect relevant biases?} The high precision in both experiments indicate that Humans and GPT-3 agree on the biases that GPT-3 selected. The high recall in the societal case indicates that GPT-3 is better at capturing societal biases, compared to other types of biases.
  }
  \label{tab:userstudy1}
  \centering
  \resizebox{0.48\columnwidth}{!}{
  \begin{tabular}{@{}lcc@{}}
    \toprule
    Experiment & Precision & Recall \\
    \midrule
    Human-vs-GPT (Overall) & 0.90 & 0.54 \\
    Human-vs-GPT (Societal) & 0.90 & 0.87 \\
    \bottomrule
  \end{tabular}}}
  \hfill
  \parbox{.48\columnwidth}{
  \caption{\textbf{User Study 2: Do humans see the same biases as our model?}. We use prompts with multiple societal biases (`gender', `age', ...), and compute accuracy and ranking correlation.}
    \label{tab:combined_comparison}
    \centering
    \resizebox{0.48\columnwidth}{!}{
    \begin{tabular}{@{}lcccc@{}}
        \toprule
        & \multicolumn{2}{c}{Accuracy} & Ranking \\
        \cmidrule(lr){2-3} \cmidrule(lr){4-4}
        Metric/Baseline & Top-1 & Top-2 & Correlation \\
        \midrule
        \multicolumn{4}{c}{Prompts with Societal Biases} \\
        \midrule
        Bipartite Matching & 41\% & 76\% & -0.08 \\
        CLIP ($\textrm{CAS}^{CLIP}$) & 50\% & 58\% & +0.07\\
        VQA ($\textrm{CAS}$) & \textbf{75\%} & \textbf{83\%} & \textbf{+0.51}\\
        \bottomrule
    \end{tabular}}}
\end{table}

\vspace{-0.4cm}
\subsubsection{User Study 2: Validating $\textrm{CAS}$ and $\textrm{MAD}$ metrics}
\label{sec:userstudy2}

This user study examines the alignment between human bias ratings and $\textrm{MAD}$ scores for a set of images generated from the initial prompt. Due to the subjectivity of identifying biases, we focus solely on a predefined set of societal biases (refer to Appendix \ref{sup:userstudy2}). We present participants with 10 randomly sampled images for each input prompt from our dataset of 100 prompts that may contain two or more societal biases. They rate bias presence on a 1–5 scale. With 10 participants per question, we assess the Spearman correlation between the median value of human bias ratings and our $\textrm{MAD}$ scores. We also calculate Top-1 and Top-2 accuracy for prompts with three or more societal biases.

Our results in Table \ref{tab:combined_comparison} indicate that there is a positive correlation of $+0.51$ between human ratings and our $\textrm{MAD}$ score, indicating that our model is aligned with humans in ranking societal biases. We also compare both our image comparison methods against Bipartite Matching of concepts with cosine similarity of CLIP text embeddings. The poor ranking correlation and lower Top-K accuracy of CLIP also demonstrates the benefits of using $\textrm{CAS}$ over $\textrm{CAS}^{CLIP}$. This is in line with recent works \cite{cho2023dall, singh2023divide}. We also observe a similar dip when we use CLIP scores in bipartite matching, likely due to incorrect matches.

\subsubsection{User Study 3: Evaluating MiniGPT-v2}
\label{sup:userstudy3}

We conduct a third user study to evaluate the quality of the MiniGPT-v2 model for detecting concepts. While each set of images can have several different concepts, we are primarily interested in the Top-K Overall concepts across all images, as those are most influential in calculating $CAS$ scores. Accordingly, we set up an MTurk task, where we ask 3 participants to select all concepts, from a list of 10 concepts, that are relevant to the given set of 10 randomly sampled images. We create 80 such tasks, where we first sample an initial prompt or any counterfactual prompt from our dataset, then sample 10 images for the prompt, and obtain the top 5 concepts that are present in the set of images (from $C_{\textrm{init}}$ or $C_{cf}$). We then also add 5 other random concepts that are \textit{not} predicted by MiniGPT-v2 to make the list of 10 concepts for that set. Each HIT is \$0.10, and all workers are from the US.  A HIT (Human Intelligence Task) refers to a small unit of work assigned to human annotators, typically through a crowdsourcing platform like Amazon Mechanical Turk.

We calculate the accuracy of our model based on the concepts that humans select. Of all the concepts that a human said yes to, our model selected 82.8\% of those concepts (higher is better). This indicates that MiniGPT-v2 is fairly accurate at detecting useful concepts from the images, assuming we observe over a large set of images (48 images in our case). Moreover, for all the concepts that were randomly selected for this task (that our model did not produce for a set of images), humans only select 7.4\% of concepts (lower is better).

For our sensitivity analysis, we roughly estimate a VQA error of 18\%, assuming that our model missed about 17–18\% of the concepts that humans had also said yes to. Because this is only a rough estimate, we report numbers for higher and lower error rates in our sensitivty analysis graph in Figure \ref{fig:metrics}.

%Our user studies show promising results towards the identification of biases using \modelname. We share additional details, user study results, and challenges regarding conducting bias-related user studies in Appendix 6.

\subsection{Comparing CLIP and VQA for Image-Based Bias Measurement}
\label{sec:clip_vs_vqa}

A key component of our framework involves comparing generated images across counterfactual prompts to measure semantic differences that may reflect bias. While prior works have often used CLIP embeddings to compare image sets holistically, we investigate whether incorporating more localized, attribute-specific cues through Visual Question Answering (VQA) provides better alignment with human judgment.

\begin{figure}[t]
  \centering
   \includegraphics[width=0.8\linewidth]{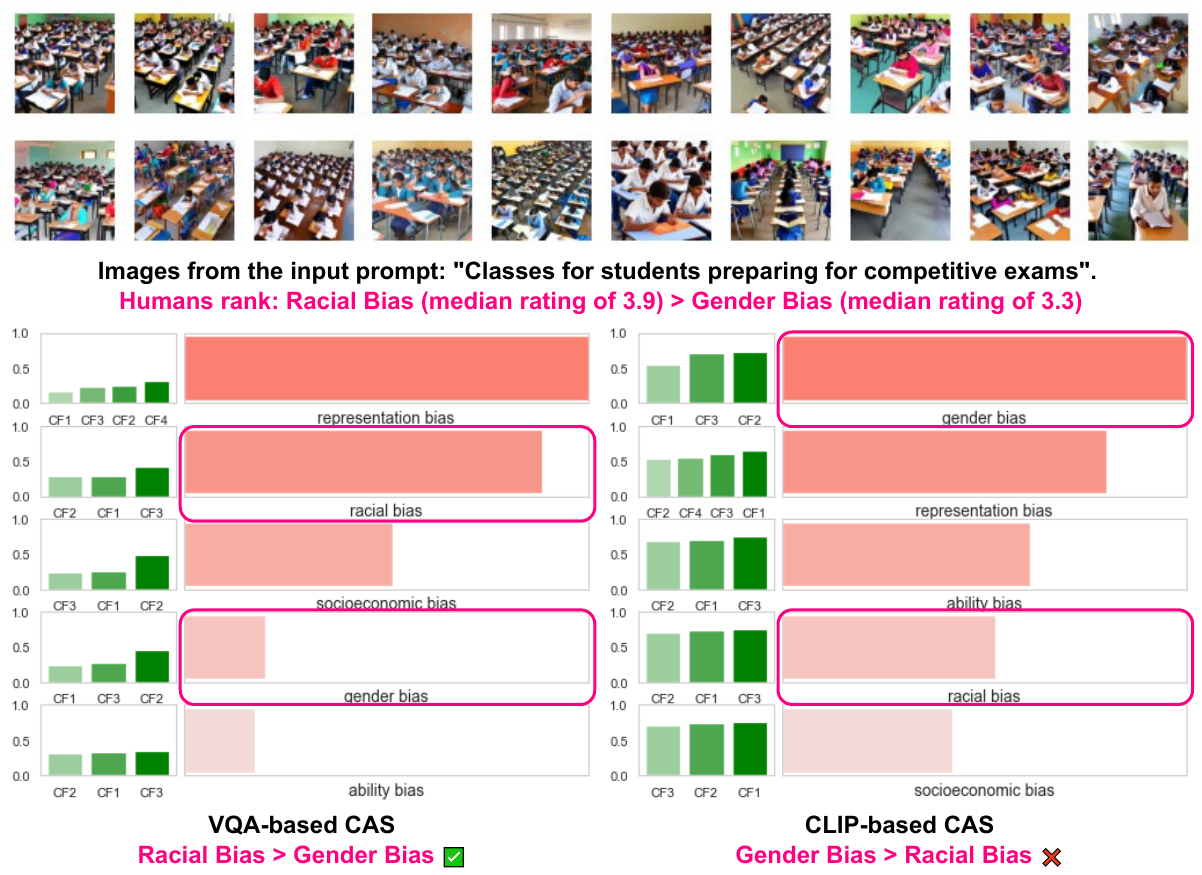}
   \caption{\textbf{Comparing our VQA and CLIP methods for Image Comparison}. In this example, we see that humans rank racial bias to be more significant compared to gender bias, which is also observable in the images. We compare our VQA-based method to our CLIP-based method, and observe that the VQA-based method better aligns with human ranking. This is because, in most cases, biases are attributed to specific characteristics or parts of an image (which VQA helps us obtain), and not the semantic information of the image as a whole (which CLIP embeddings provide). This is in line with what we observe in User Study 2.}
   \label{fig:vqavsclip}
\end{figure}

To evaluate this, we compare the results of our CLIP-based and VQA-based image comparison methods on a set of prompts exhibiting known biases. In particular, we examine cases where both racial and gender biases may be present. As shown in Figure~\ref{fig:vqavsclip}, human annotators consistently rank racial bias as more prominent than gender bias in the given image sets. Our VQA-based method better captures this distinction, whereas CLIP-based similarity scores often blur such attribute-specific differences due to their global, embedding-level nature.

This observation aligns with our findings from User Study 2, where participants noted that perceived biases in image generations often stemmed from distinct visual cues—such as skin tone, clothing, or hairstyle—rather than overall scene semantics. Since VQA enables targeted queries on specific visual elements (e.g., "What is the skin color of the person?"), it allows for more fine-grained bias assessments that mirror human reasoning. This underscores the value of incorporating VQA-based methods for bias detection in TTI models.

\section{Applications}

Having the capability to detect biases and provide concept-level explainability for any input prompt enables several downstream use-cases. Two downstream applications of \modelname\ are described below.

\subsection{Mitigating Biases in TTI models}
\label{sec:itigen}
\begin{figure*}[tb]
  \centering
   \includegraphics[width=\linewidth]{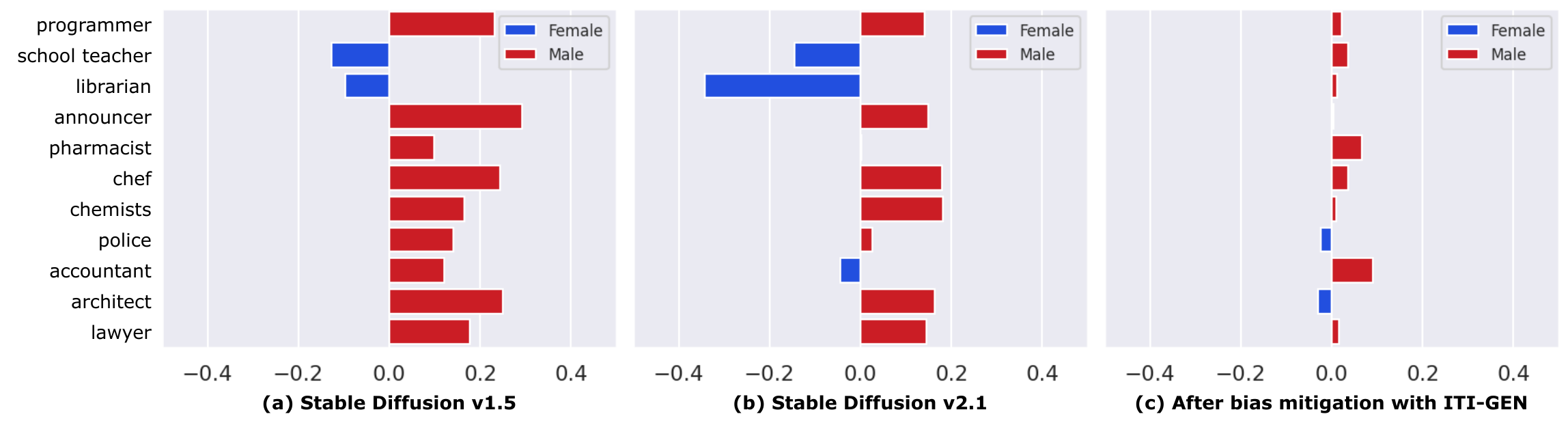}
   \vspace{-0.2cm}
   \caption{\textbf{Bias identification and mitigation}. We compute difference in $\textrm{CAS}$ scores for male and female counterfactuals for 11 occupation prompts. (a) and (b) show male and female leaning professions using Stable Diffusion 1.5 and 2.1 respectively. (c) shows how the difference in $\textrm{CAS}$ scores after using ITI-GEN to mitigate gender bias.}
   \label{fig:gender-mitigation}
\end{figure*}

\begin{figure*}[t]
    \includegraphics[width=1.0\linewidth]{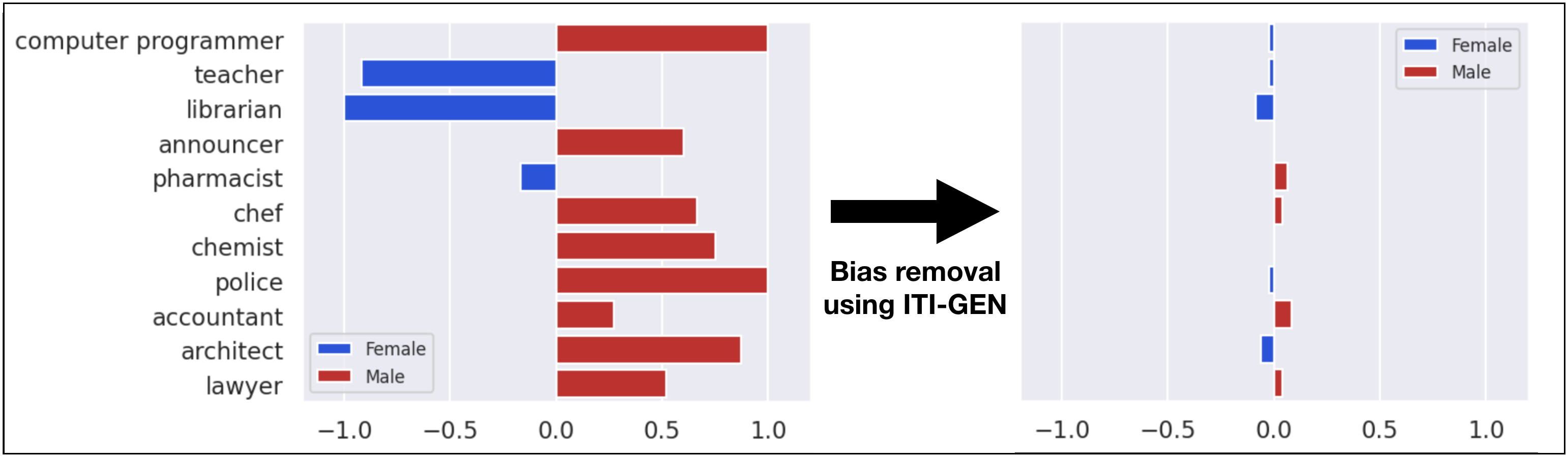}
    \centering
    \caption{\textbf{Bias Identification and Mitigation using TIBET and ITI-GEN - Ground Truth}. Here, we show ground truth gender differences in the initial set of images before bias mitigation, and after bias mitigation. The reduction in gender bias is in line with what we observe using $\textrm{CAS}$ scores.}
    \label{fig:gender-groundtruth}
\end{figure*}

While approaches like ITI-GEN \cite{zhang2023iti} have shown to decrease bias along a given axis, they are incapable of identifying what bias axes and what images to train on. Our approach can complement these approaches by automatically identifying relevant bias axes and producing counterfactual images for these axes. Further, our proposed metrics also measure the degree of bias changes along these axes. We conduct one such experiment where we use our method along with ITI-GEN to mitigate gender biases in occupational prompts. For each occupational prompt, we already generate male and female counterfactual images along the gender axis. These counterfactual images are used as input reference image sets in ITI-GEN. Post-training ITI-GEN, we generate 48 new images for each occupational prompt. As illustrated in Figure \ref{fig:gender-mitigation}(c), the difference in $\textrm{CAS}$ values for the majority of occupations exhibits a notable decrease, underscoring the successful mitigation of bias. Our proposed metric for bias identification effectively captures the reduction of bias achieved by a state-of-the-art method like ITI-GEN, reinforcing the credibility of our metrics. 

In Figure \ref{fig:gender-groundtruth}, we show the percentage difference (male\%-female\%) based on ground truth gender classification for the 48 images we generate for the initial prompt, and the 48 images generated after doing bias mitigation using ITI-GEN for the same prompt. This gender classification is conducted by a human participant who manually went through all 48 images before and after bias mitigation, and classified each image as ``male", ``female", or ``other", where ``other" usually implies that the image does not have a person in it, or only a part of the body is seen (e.g., hands) that is insufficient to classify gender. We can observe that the decrease in gender bias is consistent with what we observed based on our $\textrm{CAS}$ scores in Figure \ref{fig:gender-mitigation}(c).

\section{Discussion and Conclusion}
\label{sec:discussion}

We propose \modelname, an approach to automatically detect and evaluate biases present in images generated by TTI models in an explainable manner. Our approach has the potential to address previously unexplored issues related to bias in TTI models, including reasoning about intersectionality of different bias axes and comprehensive and automated bias mitigation. Our hope is that \modelname\ can serve as the foundation for future research in the these directions. While our approach holds significant promise, it is essential to acknowledge the limitations of our model. Despite designing the tool with the ultimate aim of reducing biases, we have identified several flaws within our model. This section is dedicated to a thorough exploration of these limitations.

%\noindent\textbf{Limitations and Ethical Considerations.}
%Although there are many benefits to our method, we acknowledge that incorrect bias detection can be harmful. In our work, we use LLMs (GPT-3) and VLMs (MiniGPT-v2, CLIP) that may have their own limitations and biases. Our sensitivity analysis improves the transparency of our pipeline, and can measure the effect of these biases. Ultimately, our approach is modular and not dependent on any specific versions of these models. We expect that as fairer and more capable LLMs and VLLMs emerge, they will replace the current models used in our method. Finally, we note that we conduct user studies in accordance with ethics guidelines.

\textit{It is important to note that, while these limitations exist, we view this work as an initial step towards conducting comprehensive bias evaluations for any prompt for TTI models.}

\subsection{Biases in Language Models (LLMs)}
Our approach rests on the assumption that Large Language Models (LLMs) excel at detecting biases in prompts for Text-to-Image (TTI) models. While our user studies validate that humans agree with potential bias axes, there is always the possibility that some of the generated axes may not be meaningful, or may be a result of hallucination. Even though solutions like human intervention and Automated External Sources (AES) filtering can mitigate these issues, the approach cannot be foolproof. Further research and development are necessary. While LLMs may have their own issues, they are the fastest and most capable way to identify biases in any prompt, and the task that TIBET does would take large amounts of time and money to conduct manually.

\subsection{Interpretation of Bias Axes}
Another interesting challenge is that LLMs may generate a completely valid yet orthogonal set of bias axes compared to humans. While LLMs offer diversity in generating bias axes, we advocate for human intervention to validate these axes of bias. We also make clear that the interpretation of results is ultimately up to humans.

\subsection{Biases in Vision Language Models (VLLMs)}
Utilizing Vision-Language Models (VLLMs) for image comparision introduces an additional dimension of bias. We have observed that current models do not perform perfectly, and require significant improvement in their image concept understanding capabilities. Additionally, VLLM models need more comprehensive training or fine-tuning on concept detection tasks to generate more relevant concepts. Automation of predefined questions for VLLMs is also a crucial step for a more comprehensive approach. Ultimately, we have tried to design our metrics to be robust to small errors in VQA, and our sensitivity analysis shows a weaker-than-linear correlation between changes caused by VQA errors on the values of our $\textrm{CAS}$ and $\textrm{MAD}$ scores. 

\subsection{Challenges in Metric Evaluation}
We propose the use of the Concept Association Score ($\textrm{CAS}$) and Mean Absolute Deviation ($\textrm{MAD}$) as metrics and suggest user studies to measure them. However, we lacked a comprehensive dataset with ground truth values to evaluate our metrics. In the future, we aim to conduct a more extensive evaluation on a larger, well-labeled dataset. Finally, our metrics rely on diverse counterfactuals, and make the assumption that the TTI model being evaluated is sensitive to the changes (made by the LLM) from the initial input prompt to the counterfactual prompts. TIBET may fail in the rare case when TTI models fail to incorporate the changes made in counterfactual prompts and the image generation may not always be faithful to the counterfactual prompts \cite{hu2023tifa}. We empirically observe in our qualitative examples that this is not an issue with the Stable Diffusion models we use in our work.

\subsection{Challenges with Bias-related user studies}
\label{sup:uschallenges}

Conducting bias-related user studies is challenging in several ways, including:
\begin{enumerate}
    \item As all individuals observe bias differently, teaching participants what bias is, what each type of bias means, and how it is relevant to our task is a challenge. Participants may misunderstand, or ignore the definitions of bias we provide, and rely on their subjective understanding of biases when answering questions. We provide extensive training and qualification tests to reduce subjectivity as much as possible.
    \item Participants may consider societal biases more important (and therefore more likely, as with User Study 2) compared to incidental biases, as incidental biases are not frequently talked about in society and may not be inherently harmful in any way.
\end{enumerate}

\newcommand{\modelnameintersect}{BiasConnect\xspace}
\newcommand{\modelnamegraph}{BiasGraph\xspace}
\newcommand{\isscore}{Intersectional Sensitivity\space}
\chapter{Using counterfactuals to understand Intersectional Bias Dynamics in Text-to-Image Models}
\epigraph{“There is no such thing as a single-issue struggle because we do not live single-issue lives”}{\textit{Audre Lorde}}
\begin{figure}[ht]
    \centering
    \includegraphics[width=0.8\textwidth]{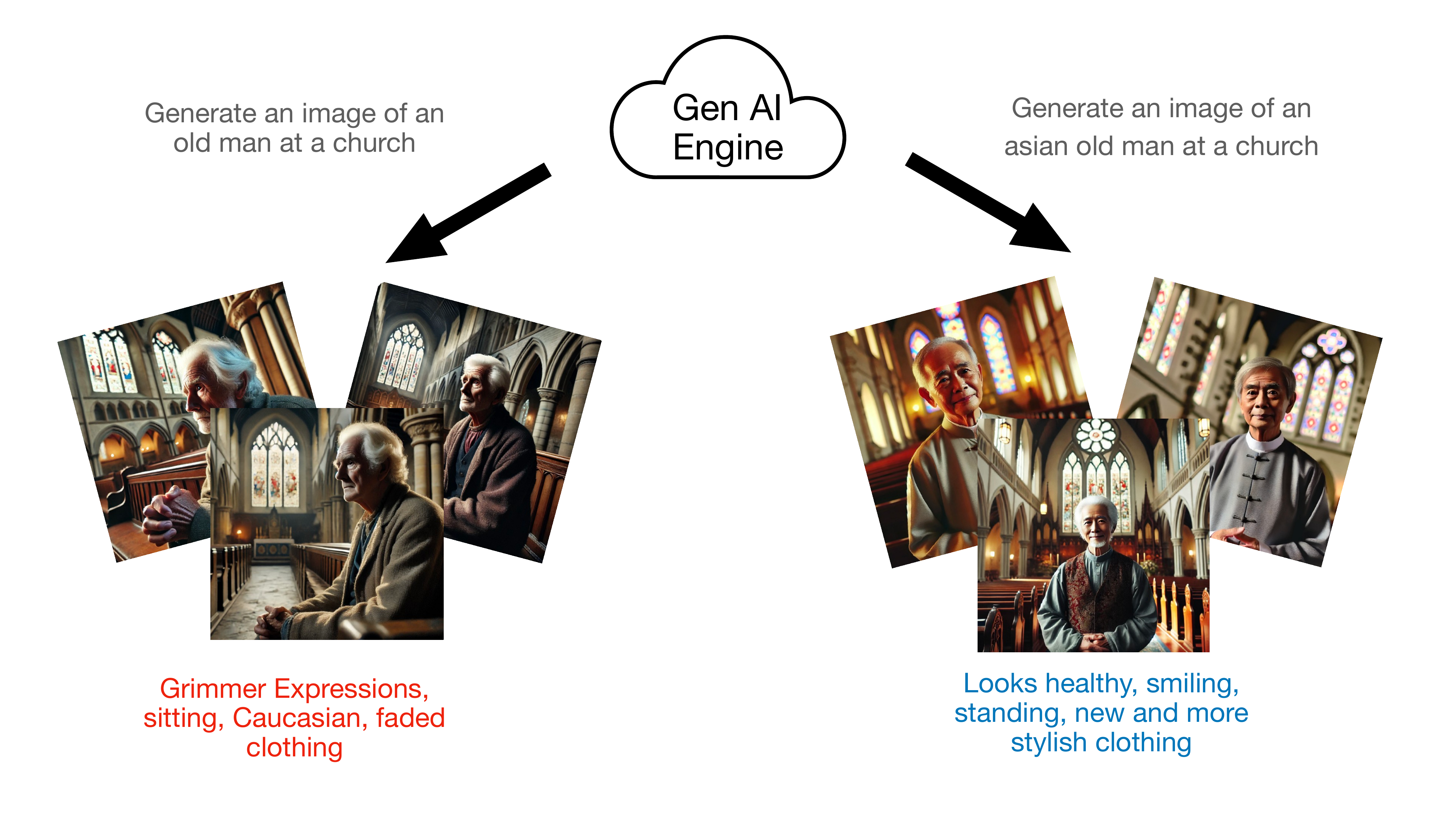}
    \caption{This figure demonstrates how a minimal prompt change—from “\textit{An old man at a church}” to “\textit{An Asian old man at a church}”—can lead to significant differences in generated outputs. While the former yields grim, seated Caucasian men, the latter produces smiling, upright Asian men in brighter scenes. This shift illustrates the intersectional nature of generative model behavior, where the addition of a single identity attribute (ethnicity) modifies not just appearance but also emotion, posture, and context—highlighting how social dimensions interact rather than operate independently.}
    \label{fig:intro_teaser}
\end{figure}

\section{Introduction}
In the previous chapter, we introduced TIBET, a diagnostic framework that applies counterfactual reasoning to dynamically evaluate bias in text-to-image (TTI) generative models. By leveraging context-specific prompt variations, TIBET enables fine-grained auditing of bias along individual social dimensions such as gender, age, or ethnicity. However, this framework raises a deeper question central to the fairness of generative systems: what happens to other bias dimensions when we intervene on one? In other words, if we mitigate bias along a particular axis (e.g., gender), do other forms of bias—such as age, race, or clothing—remain unchanged, improve, or worsen?

Consider a simple example: the prompt “an old man at a church.” The generated images consistently depict seated, somber white men in shabby clothing. When the prompt is modified to “an Asian old man at a church,” the outputs change significantly: the subjects appear happier, more vibrant, better dressed, and are standing. This qualitative shift suggests that altering one axis of identity (ethnicity) influences multiple other attributes, including emotion, body posture, and perceived social status. Such phenomena reveal that biases in TTI models are not merely independent, but often interconnected.

Answering these questions is crucial for understanding the inner workings of TTI models and for designing mitigation strategies that respect the intersectional nature of social identity. To this end, this chapter focuses on how counterfactual reasoning can be extended beyond axis-isolated evaluation to quantify intersectionality in generative systems.

We begin by revisiting the concept of intersectionality and its relevance to AI fairness. We then briefly review how TIBET can be adapted to qualitatively surface intersectional effects through prompt-based variations. However, we also highlight a core limitation of TIBET: it lacks a quantitative mechanism for modeling or measuring how different bias axes influence one another.

To address this gap, we introduce two complementary tools: BiasConnect and BiasGraph. Built upon a novel metric called Intersectional Sensitivity (IS), these tools enable us to quantify and structure the relationships between bias axes. BiasConnect uses counterfactual interventions to populate an Intersectionality Matrix, revealing whether intervening on one dimension improves or harms another. BiasGraph builds on this by transforming the matrix into a causal graph, where directional dependencies between bias axes can be analyzed at scale. Together, these tools offer a principled, interpretable framework for diagnosing intersecting and entangled biases in TTI models—laying the groundwork for future mitigation strategies.

\section{Understanding Intersectionality and Its Role in Bias Diagnosis for Generative Models}
\label{sec:inter_definition}

Intersectionality is a theoretical framework developed to understand how multiple systems of oppression—such as racism, sexism, classism, and ageism—interact to shape the lived experiences of individuals with multiple marginalized identities. First introduced by legal scholar Kimberlé Crenshaw~\cite{crenshaw1989demarginalizing,crenshaw1991mapping}, intersectionality emphasizes that the discrimination faced by, for example, a Black woman, cannot be fully understood by analyzing race and gender in isolation. Instead, these identities intersect to produce qualitatively distinct experiences of marginalization, which are more than just the additive effects of each category. This concept has since become foundational in feminist theory, critical race studies, and increasingly in algorithmic fairness and bias analysis in machine learning.

Within this framework, two major models of interaction have emerged in scholarly discourse \cite{curry2018killing}:

\begin{itemize}
    \item The additive model, which assumes that disadvantage accumulates linearly across marginalized dimensions (e.g., racism + sexism = double discrimination).
    \item The interactive model, which suggests that dimensions of identity interact \textit{synergistically}, producing novel or amplified effects that cannot be explained by additive logic alone.
\end{itemize}

While intersectionality originated in legal and sociological theory, it has increasingly informed the field of algorithmic fairness. Machine learning researchers have adapted its core principles to evaluate fairness in AI systems by measuring performance or representation disparities across intersectional subgroups—e.g., Black women vs.\ white men—in domains such as classification~\cite{diana2021minimax,foulds2020intersectional}, natural language processing~\cite{lalor2022benchmarking,guo2021detecting}, computer vision~\cite{wang2020towards,steed2021image}, and large language models~\cite{kirk_bias_2021,ma2023intersectional}. However, most of these approaches align with the additive paradigm, treating subgroup disparities as separable and largely static.

This chapter departs from that assumption. We adopt the interactive model of intersectionality, using counterfactual interventions to examine how changing one dimension of identity—such as gender—can alter model behavior along another, like age or emotion. This shift is especially critical in  Text-to-Image (TTI) generative models.
\subsection {Generative models and the need for understanding intersectional biases}
TTI models such as \textsc{DALL-E}~\cite{ramesh2021zero}, \textsc{Imagen}~\cite{saharia2022photorealistic}, and \textsc{Stable Diffusion}~\cite{rombach2022high} have become powerful tools for synthesizing visual content from textual prompts. Yet these systems often reproduce harmful social stereotypes embedded in their training data~\cite{wang_towards_2022}, manifesting biases across dimensions such as gender, race, age, clothing, and setting. Although a growing body of work has sought to evaluate or mitigate such biases~\cite{wang2023t2iat,cho2023dall,ghosh2023person,esposito2023mitigating,bianchi2023easily}, most approaches treat each dimension in isolation. For example, gender and race are often audited separately, and mitigation efforts typically focus on a single axis at a time.

This assumption of axis-independence fails to capture a key behavior of generative models: biases often interact. For instance, attempting to mitigate gender bias in a prompt like ``a doctor treating patients'' may inadvertently reduce diversity in age or ethnicity, leading to outputs that are gender-balanced but skewed toward young or racially homogenous representations. These shifts suggest that intervening on one dimension can distort others, resulting in unanticipated trade-offs.

Such entangled effects highlight the need for intersectional bias assessment in TTI models. Without tools that account for the interactive nature of identity, we risk designing evaluations or mitigations that improve fairness along one axis while silently worsening it along others. To address this, we explore in this chapter how counterfactual reasoning can be extended to diagnose inter-axis dependencies—that is, how interventions on one attribute (e.g., race) affect distributions over other attributes (e.g., emotion, clothing, posture).

Through this lens, we introduce two diagnostic tools—\modelnameintersect and \modelnamegraph—built on a novel metric called \isscore. These tools move beyond axis-isolated evaluation, enabling structured, interpretable, and quantitative analyses of how social biases in generative models interact.

\section{Using TIBET to Identify Prompt-Level Intersectional Bias}
\label{sec:inter_TIBET_recap}

\modelname is a diagnostic framework, introduced in Chapter 5, for measuring social biases in text-to-image (TTI) generative models using a counterfactual reasoning approach. Unlike prior work that evaluates bias along static, pre-defined axes, \modelname dynamically identifies the bias dimensions most relevant to a given input prompt and uses them to guide a systematic evaluation.

Given an input prompt $P$, \modelname first detects and constructs a set of relevant social axes (e.g., gender, age, race, emotion) based on the semantics of $P$. It then generates a series of counterfactual prompts by varying one axis at a time while holding the rest of the prompt fixed. For each of these prompts—including the original—TIBET uses a black-box TTI model (e.g., Stable Diffusion) to synthesize a corresponding set of images.

To evaluate the effect of each counterfactual change, TIBET compares the images generated from the original prompt with those generated from each counterfactual. This comparison is done using a combination of vision-language tools: a VQA pipeline to extract interpretable concepts (e.g., clothing, emotion, background), and CLIP embeddings to capture high-level semantic differences. These comparisons are quantified using the Concept Association Score ($\textrm{CAS}$), which measures the directional association between concept distributions across different prompts.

\modelname supports both quantitative metrics—such as the Mean Attribute Difference ($\textrm{MAD}$)—and qualitative visualization tools to provide interpretable explanations of bias. The framework enables users to identify which social attributes are disproportionately associated with a given prompt and how these associations shift under controlled counterfactual interventions.

\begin{figure*}[tb]
  \centering
   \includegraphics[width=0.95\linewidth]{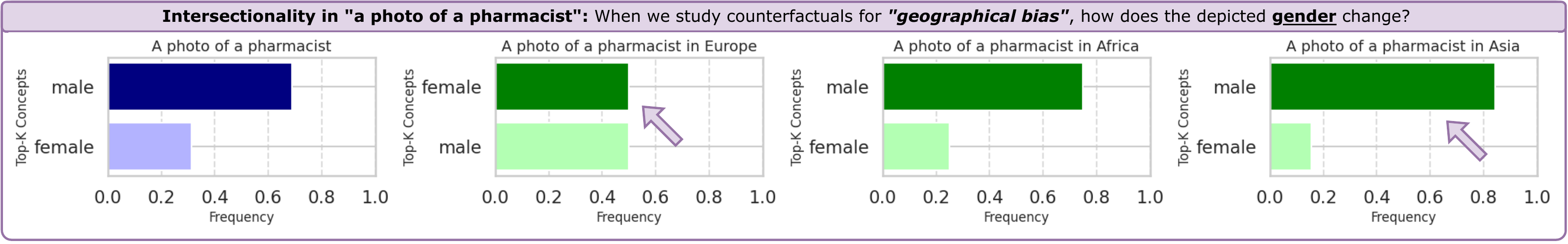}
   %\vspace{-0.2cm}
   \caption{\textbf{Exploring Intersectionality of Biases:} Analysing the Top-K concepts shows that \textit{pharmacists in Europe} and \textit{Asia} are depicted with different gender distributions.}
   \label{fig:intersectionality}
   \vspace{-0.2cm}
\end{figure*}

 Using \modelname, we can study intersectionality by observing counterfactuals along one bias axis, and comparing changes in concepts along another bias axis. This can uncover the interconnectedness between bias axes, showing that modifying one bias may unintentionally amplify biases in other dimensions.

An illustrative example of these interactions is shown in Figure~\ref{fig:intersectionality}. The Axis-Aligned Top-K Concepts corresponding to secondary bias dimensions reveal noteworthy insights into the behavior of the TTI model. For instance, the male-to-female ratio is observed to be higher in images generated for prompts like “pharmacist in Asia” and “pharmacist in Africa”, whereas the ratio is lower for “pharmacist in Europe”—relative to a neutral prompt baseline. These patterns indicate that gender representations are not fixed but vary significantly depending on geographic modifiers in the prompt.

\begin{figure*}[t]
    \includegraphics[width=1.0\linewidth]{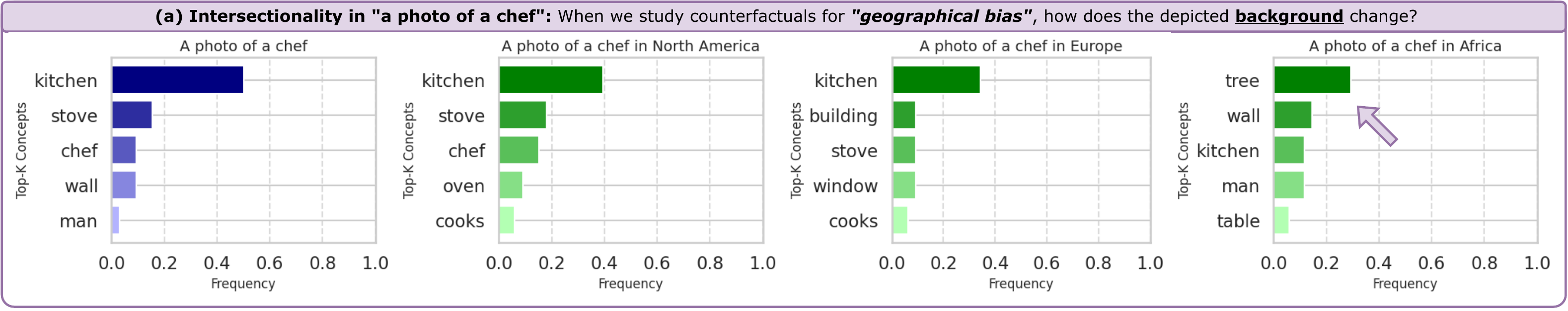}
    \centering
    \caption{\textbf{Intersectional Observations using TIBET}. We observe that images generated for a chef in Africa may be depicted outdoors (tree) unlike chefs in other regions of the world.}
    \label{fig:intersectionalityappx}
\end{figure*}

A similar trend is visible in Figure~\ref{fig:intersectionalityappx}, where regional context influences background composition. Specifically, prompts like “a chef in Africa” are more likely to be rendered in outdoor settings—e.g., with trees or natural elements—unlike depictions of chefs in other regions, which tend to be shown indoors. These observations highlight how secondary attributes such as location can conditionally shape other visual dimensions, reinforcing the need for intersectional bias evaluation.

\subsection{Moving beyond TIBET for deeper intersectional analysis}
\label{sec:inter_TIBET_LIMITATIONS}
While TIBET provides a powerful and flexible framework for dynamically diagnosing social bias in text-to-image models, it has important limitations when it comes to evaluating and reasoning about interactions between multiple bias dimensions.

First, TIBET is fundamentally a visual diagnostic tool. It allows users to observe how varying one social attribute (e.g., race or gender) affects the visual content generated by a model, and to qualitatively detect cases where changes in one axis appear to influence others. For example, as discussed earlier, modifying ethnicity in a prompt may unintentionally alter associated attributes such as emotion or clothing. However, these interactions are primarily surfaced through visual comparisons or concept ranking, rather than being explicitly measured or formalized.

Second, TIBET lacks a quantitative mechanism to assess whether and how biases across different dimensions are causally entangled. That is, while TIBET can reveal that two attributes change together, it does not offer a structured way to determine the direction, magnitude, or consistency of these interactions across prompts or datasets. There is no formal metric to evaluate whether mitigating one bias (e.g., gender) improves, worsens, or leaves unchanged the fairness of another (e.g., age).

Third, TIBET does not provide a scalable or generalizable structure for modeling bias dependencies across multiple axes. Each evaluation is prompt-specific and axis-specific, making it difficult to reason globally about the underlying entanglement structure of the model's latent space.

Finally, TIBET is limited to qualititative evaluation and offers no guidance for how to use observed interactions in downstream tasks such as bias mitigation.

To address the challenges outlined above and better understand intersectionality in generative models, we introduce two frameworks in this chapter: \modelnameintersect and \modelnamegraph. These methods aim to: (1) quantify how different bias dimensions interact, using a new metric called \textit{Intersectional Sensitivity}, and (2) represent these interactions as graphs that support model auditing and guide mitigation strategies. While \modelname visualizes interactions between bias axes, these approaches take a step further by directly quantifying these interactions, enabling a more precise analysis of model behavior and providing concrete tools for understanding intersectional biases.

\section{\modelnameintersect: Quantifying Bias Relationships}
To understand how biases in text-to-image (TTI) models influence one another, we propose \modelnameintersect, a framework that evaluates societal biases in TTI models while accounting for intersectional relationships, rather than treating bias dimensions independently. \modelnameintersect  identifies and quantifies how interventions on one bias axis affect others, revealing cases where mitigating one bias may inadvertently worsen another. Unlike prior methods, our approach supports prompt-level intersectional analysis and enables aggregate evaluations across a set of prompts. This allows us to examine how model architecture, training data, or objectives contribute to bias entanglement. Given an input prompt, \modelnameintersect  uses counterfactual interventions to measure how changes along one bias axis influence others in the generated outputs. These interactions are quantified using a metric called the \textit{Intersectional Sensitivity Score} (\isscore), which captures the direction and magnitude of such effects. This section describes the \modelnameintersect\ framework in detail and discusses the utility of this approach for intersectional bias analysis in TTI models.

\begin{figure}[t]
  \centering
   \includegraphics[width=0.8\linewidth]{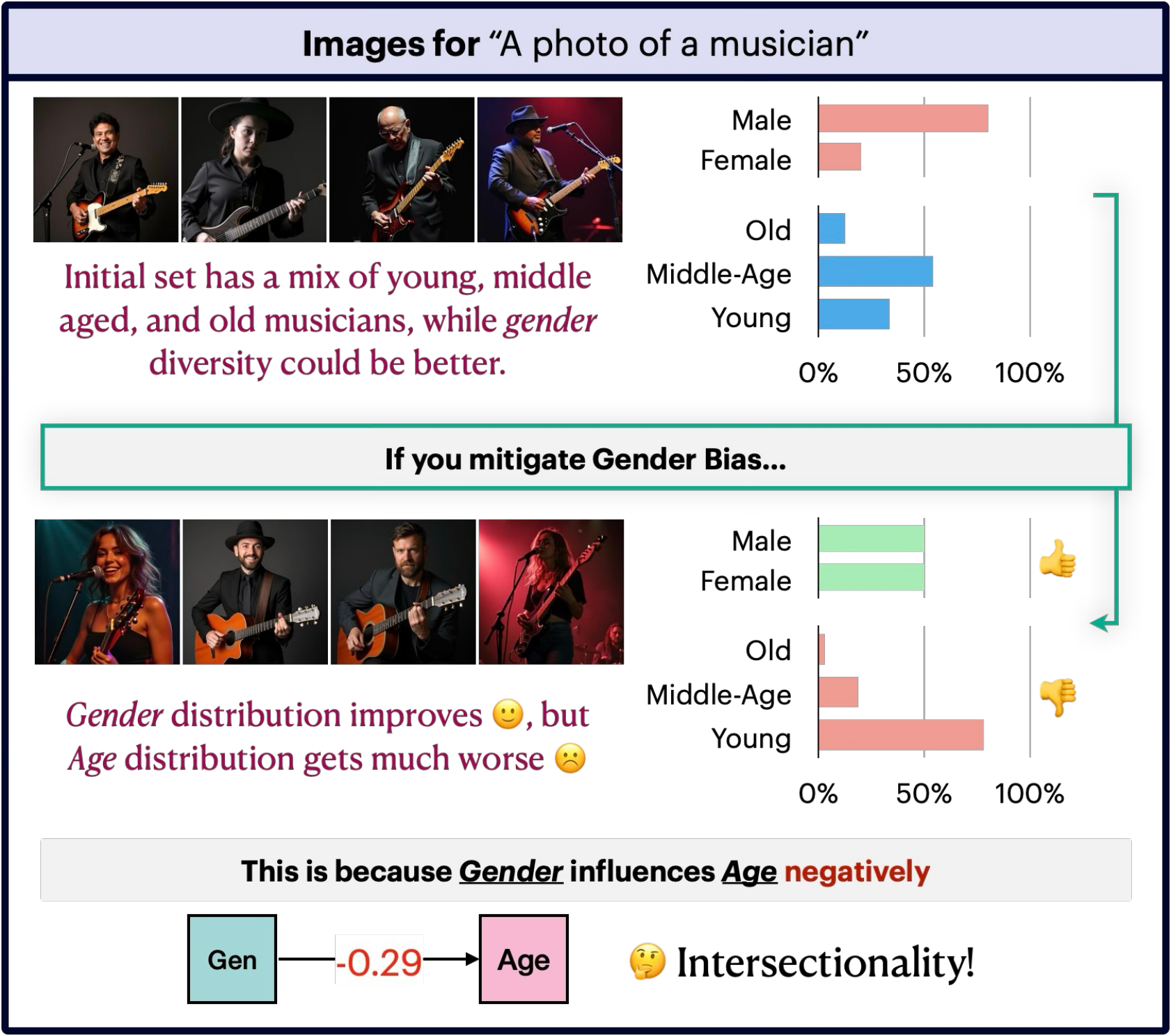}
   \caption{An example for which \modelnameintersect estimates a negative impact of bias mitigation along one axis on another axis. For this query, increasing the gender diversity (Gen) skews age distribution (Age) for images of musicians generated by Flux-dev.}
   \label{fig:Bias_connect_header}
\end{figure}

Figure~\ref{fig:Bias_connect_header} illustrates a representative example where \modelnameintersect detects a negative interaction between two bias axes. In this case, we examine a prompt related to musicians generated by the Flux-dev \cite{flux2024} model. The figure shows that increasing gender diversity leads to a skewed age distribution—specifically, a reduction in the representation of older individuals. This reflects a common failure mode in TTI models, where mitigating bias along one axis (e.g., gender) unintentionally exacerbates bias along another (e.g., age). Such cases underscore the need for intersectional analysis tools like \modelnameintersect, which make these trade-offs visible and quantifiable.

\subsection{Approach}
\label{sec:inter_Approach}

\begin{figure*}
  \centering
   \includegraphics[width=\linewidth]{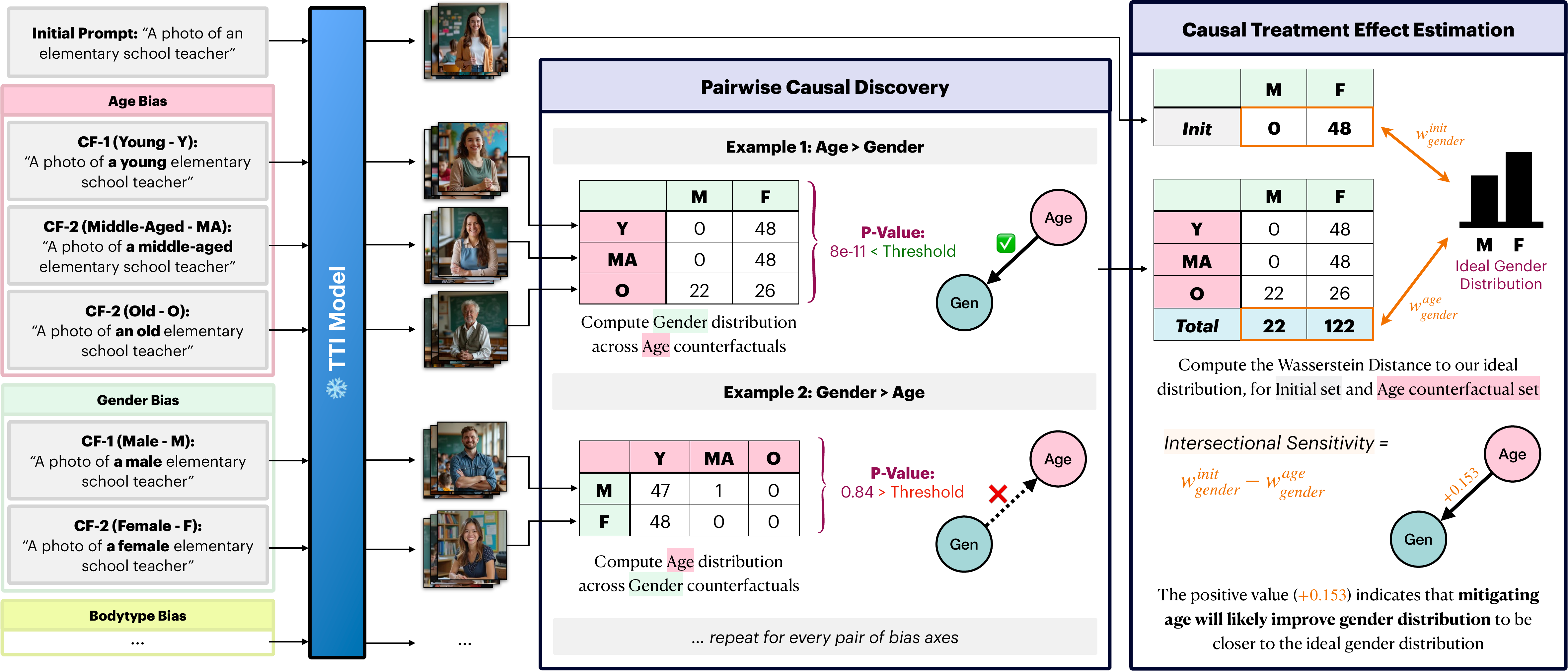}
   \caption{\textbf{An overview of \modelnameintersect}. We use a counterfactual-based approach to measure pairwise causality between bias axes. For dependent axes, we measure the causal effect, estimating how bias mitigation on one axis impacts another.}
   \label{fig:approach}
   % \vspace{-0.2cm}
\end{figure*}
The objective of \modelnameintersect\ is to identify and quantify the intersectional effects of intervening on one bias axis (\(B_x\)) to mitigate that bias, on any other bias axis (\(B_y\)). \modelnameintersect works by systematically altering input prompts and analyzing the resulting distributions of generated images (see Figure. \ref{fig:approach}). To achieve this, we leverage counterfactual prompts by modifying specific attributes (e.g., male and female) along a bias axis (e.g., gender) and examine how these interventions impact other bias dimensions (e.g., age and ethnicity). If modifying one bias axis through counterfactual intervention causes significant shifts in the distribution of attributes along another bias axis, it indicates an intersectional dependency between these axes. We first construct prompt counterfactuals and generate images using a TTI model (Section \ref{sec:Approach:Imggen}). Subsequently, to identify bias-related attributes in the generated images, we use a Visual Question Answering (VQA) model (Section \ref{sec:Approach:VQA}). Finally, to quantify the intersectional effects, and to identify whether these effects are positive or negative, we compute the  \isscore\ (Section \ref{sec:Approach:CausalEffect}). 

\subsubsection{Counterfactual Prompts \& Image Generation}
\label{sec:Approach:Imggen}

Given an input prompt \( P \) and bias axes \( B = \{B_1, B_2, \dots, B_n\} \), we generate counterfactual prompts 
%(\( B_i \) has counterfactuals 
\(\{\textrm{CF}_i^1, \dots, \textrm{CF}_i^j\} \) for each bias $B_i \in B$. The original prompt \( P \) and its counterfactuals are then used to generate images with the TTI model to measure intersectional effects.

\subsubsection {VQA-based Attribute Extraction}
\label{sec:Approach:VQA}
 
To facilitate the process of extracting bias-related attributes from the generated images, we use VQA. This is inspired by previous approaches on bias evaluation, like \modelname and OpenBias \cite{d2024openbias}, where a VQA-based method was used to extract concepts from generated images. Following TIBET, we use MiniGPT-v2 \cite{chen2023minigptv2} in a question-answer format to extract attributes from generated images.

For the societal biases we analyze, we have a list of predefined questions  similar to \modelname corresponding to each bias axis in $B$, and each question has a choice of attributes to choose from. For example, for the gender bias axis, we ask the question ``\texttt{\small [vqa] What is the gender (male, female) of the person?}''. Note that every question is multiple choice (in this example, \texttt{\small male} and \texttt{\small female} are the two attributes for gender). For datasets where counterfactuals are dynamically generated (e.g. TIBET dataset), an LLM-generated set of questions is used instead. The questions asked for all images of prompt \( P \) and its counterfactuals \( \textrm{CF}_i^j \) remain the same. With the completion of this process, we have attributes for all images, where each image has one attribute for each bias axis in $B$.

\subsubsection{Computing Intersectional Sensitivity}
\label{sec:Approach:CausalEffect}
Our objective is to understand how the impact of interventions on \( B_x \) affects \( B_y \) in a positive or negative direction concerning an ideal distribution. To address this, we propose a metric that quantifies the impact of bias mitigation on dependent biases with respect to an ideal distribution.
\noindent{\bf Defining an Ideal Distribution.} We first define a desired (ideal) distribution \(D^*\), which represents the unbiased state we want bias axes to achieve. This can be a real-world distribution of a particular bias axis, a uniform distribution (which we use in our experiments), or anything that suits the demographic of a given sub-population. 

\noindent{\bf Measuring Initial Bias Deviation.} Given the images of initial prompt \(P\), we compute the empirical distribution of attributes associated with bias axis \(B_y\), denoted as \( D_{B_y}^{\text{init}} \). We then compute the Wasserstein distance between this empirical distribution and the ideal distribution:
\begin{equation}
w_{B_y}^{\text{init}} = W_1(D_{B_y}^{\text{init}}, D^*)
\end{equation}
\noindent where \( W_1(\cdot, \cdot) \) represents the Wasserstein-1 distance. The Wasserstein-1 distance (also known as the Earth Mover's Distance) between two probability distributions \( D_1 \) and \( D_2 \) is defined as:
\begin{equation}
W_1(D_1, D_2) = \inf_{\gamma \in \Pi(D_1, D_2)} \mathbb{E}_{(x,y) \sim \gamma} [|x - y|]
\end{equation}
\noindent where \( \Pi(D_1, D_2) \) is the set of all joint distributions \( \gamma(x, y) \) whose marginals are \( D_1 \) and \( D_2 \), and \( |x - y| \) represents the transportation cost between points in the two distributions.

We use $\overline{w}^{\textrm{init}}_{B_y}$ to measure the amount of bias in the image set, where $\overline{w}_{B_y}$ is computed by normalizing $w_{B_y}$ based on the number of counterfactuals in $B_y$. $\overline{w}^{\textrm{init}}_B \in [0,1]$ where 1 indicates that the distribution is completely biased and 0 indicates no bias.

\noindent{\textbf{Intervening on $B_x$.}}
Next, say we intervene on \(B_x\) to simulate the mitigation of bias $B_x$. This intervention ensures that all counterfactuals of \(B_x\) are equally represented in the generated images. For example, if \(B_x\) is gender bias, we enforce equal proportions of male and female individuals in the dataset. This intervention is in line with most bias mitigation methods proposed for TTI models, like ITI-GEN \cite{zhang2023iti}. Using our counterfactuals along \(B_x\), we sum the distributions on \(B_y\) across all counterfactuals of \(B_x\). This sum across the counterfactuals of \( B_x \) yields a new empirical distribution of \( B_y \), denoted \( D_{B_y}^{B_x} \), simulating the effect of mitigating \( B_x \) (See Figure \ref{fig:approach}). We compute its Wasserstein distance from the ideal distribution.

\begin{equation}
w_{B_y}^{B_x} = W_1(D_{B_y}^{B_x}, D^*)
\label{eqn:biasscore}
\end{equation}
\noindent{\textbf{Computing \isscore.}} To quantify the effect of mitigating \(B_x\) on \(B_y\), we define the metric, \isscore, as:
\begin{equation}
\textrm{IS}_{xy} = \overline{w}_{B_y}^{\text{init}} - \overline{w}_{B_y}^{B_x} 
\label{eqn:metric}
\end{equation}
\noindent as Wasserstein distance is sensitive to the number of counterfactuals, and $\textrm{IS}_{xy} \in [-1,1]$. A positive value (\( \textrm{IS}_{xy} > 0 \)) indicates that mitigating \( B_x \) improves \( B_y \), bringing it closer to the ideal distribution, while a negative value (\( \textrm{IS}_{xy} < 0 \)) suggests it worsens \( B_y \), moving it further from the ideal. If \( \textrm{IS}_{xy} = 0 \), mitigating \( B_x \) has no effect on \( B_y \). This approach enables us to assess whether addressing one bias (e.g., gender) improves or worsens another (e.g., ethnicity) in generative models, providing a systematic way to evaluate trade-offs and unintended consequences in bias mitigation strategies.

\subsubsection{Visualization}

To visualize $\textrm{IS}$ scores comprehensively, we use a \textit{Bias Intersectionality Matrix} $\mathbf{S}$, where each entry $\textrm{IS}_{ij}$ quantifies the effect of intervening on row $B_i$ on column $B_j$ for mitigation. This matrix captures directional dependencies and enables a structured analysis of intersectional bias effects.

\subsection{Experiments}
\label{sec:biasconnectExperiments}

\subsubsection{Experiment Setup}
\label{sec:ModelsDatasets}
We conduct experiments on two prompt datasets, across six TTI models:

\textbf{Occupation Prompts}:  To facilitate a structured evaluation, we develop a dataset with 26 occupational prompts, along eight distinct bias dimensions: gender, age, ethnicity, environment, disability, emotion, body type, and clothing. We generate 48 images for all initial counterfactual prompts using five TTI models: Stable Diffusion 1.4, Stable Diffusion 3.5, Flux \cite{flux2024}, Playground v2.5 \cite{li2024playground} and Kandinsky 2.2 \cite{kandinsky22,razzhigaev2023kandinsky}. We consider 26 occupations: \textit{ computer programmer, elementary school teacher, librarian, announcer, pharmacist, chef, chemist, police, accountant, architect, lawyer, philosopher, scientist, doctor, nurse, engineer, musician, journalist, athlete, social worker, sales person, politician, farmer, mechanic, firefighter, gardener}. The different prompts and axis used for each prompt have been presented in Table \ref{tab:fulldataset}.
\begin{table}
  \centering
  \caption{The prompts for the occupation dataset constructed to evaluate BiasConnect}
  \resizebox{0.40\columnwidth}{!}{
  
  \begin{tabular}{@{}ll@{}}
    \toprule
     \textbf{Axis} & \textbf{Counterfactuals} \\
     \midrule
\multirow{2}{*}{gender bias}  &  A photo of a male \textit{occupation} \\
 &  A photo of a female \textit{occupation} \\
\midrule
\multirow{3}{*}{age bias}  &  A photo of an old \textit{occupation} \\
  &  A photo of a middle-aged \textit{occupation} \\
 &  A photo of a young \textit{occupation} \\
\midrule
\multirow{6}{*}{ethnicity bias} &  A photo of a white \textit{occupation} \\
 &  A photo of a african american \textit{occupation} \\
  &  A photo of an asian \textit{occupation} \\
 &  A photo of a south asian \textit{occupation} \\
  &  A photo of a middle eastern \textit{occupation} \\
  &  A photo of a hispanic \textit{occupation} \\
\midrule
\multirow{3}{*}{bodytype bias} &  A photo of a obese \textit{occupation} \\
  &  A photo of a slim \textit{occupation} \\
  &  A photo of a muscular \textit{occupation} \\
\midrule
\multirow{2}{*}{environment bias}  &  A photo of a \textit{occupation} working indoors \\
 &  A photo of a \textit{occupation} working outdoors \\
\midrule
\multirow{2}{*}{clothing bias}  &  A photo of a \textit{occupation} in formal attire \\
 &  A photo of a \textit{occupation} in informal attire \\
\midrule
\multirow{4}{*}{emotion bias}  &  A photo of a \textit{occupation} who is happy \\
 &  A photo of a sad \textit{occupation} who is sad \\
  &  A photo of a \textit{occupation} who is serious \\
 &  A photo of a \textit{occupation} who is tired \\
\midrule
\multirow{4}{*}{disability bias}  &  A photo of a \textit{occupation} who is fit \\
 &  A photo of a blind \textit{occupation} \\
  &  A photo of a \textit{occupation} with a hearing aid \\
  &  A photo of a \textit{occupation} on a wheelchair \\
    \bottomrule
    \end{tabular}}
    
    \label{tab:fulldataset}
\end{table}

. 

\textbf{TIBET dataset}: 
We use the same dataset introduced in the TIBET framework, which consists of 100 creative prompts paired with unique LLM-generated bias axes and corresponding counterfactuals. This dataset enables evaluation across a diverse range of social and aesthetic biases. For each prompt and its counterfactuals, the dataset includes 48 images generated using Stable Diffusion 2.1, providing a rich basis for systematic bias analysis.

\subsubsection{Studying prompt-level intersectionality}
\modelnameintersect supports prompt-level analysis of intersectional biases (Figure. \ref{fig:examples}), helping users identify key bias axes and effective mitigation strategies. For example, in Figure. \ref{fig:examples}(a), Stable Diffusion 3.5 shows a causal link between clothing and emotion—informal attire leads to happier depictions of librarians (\isscore = 0.31), suggesting clothing changes can diversify emotional portrayal. In contrast, Figure. \ref{fig:examples}(c) shows ethnicity negatively affecting gender diversity, with South Asian athletes mostly depicted as male (\isscore = -0.40), indicating that addressing ethnicity alone may worsen gender bias. 
\begin{figure*}[ht]
  \centering
   \includegraphics[width=\linewidth]{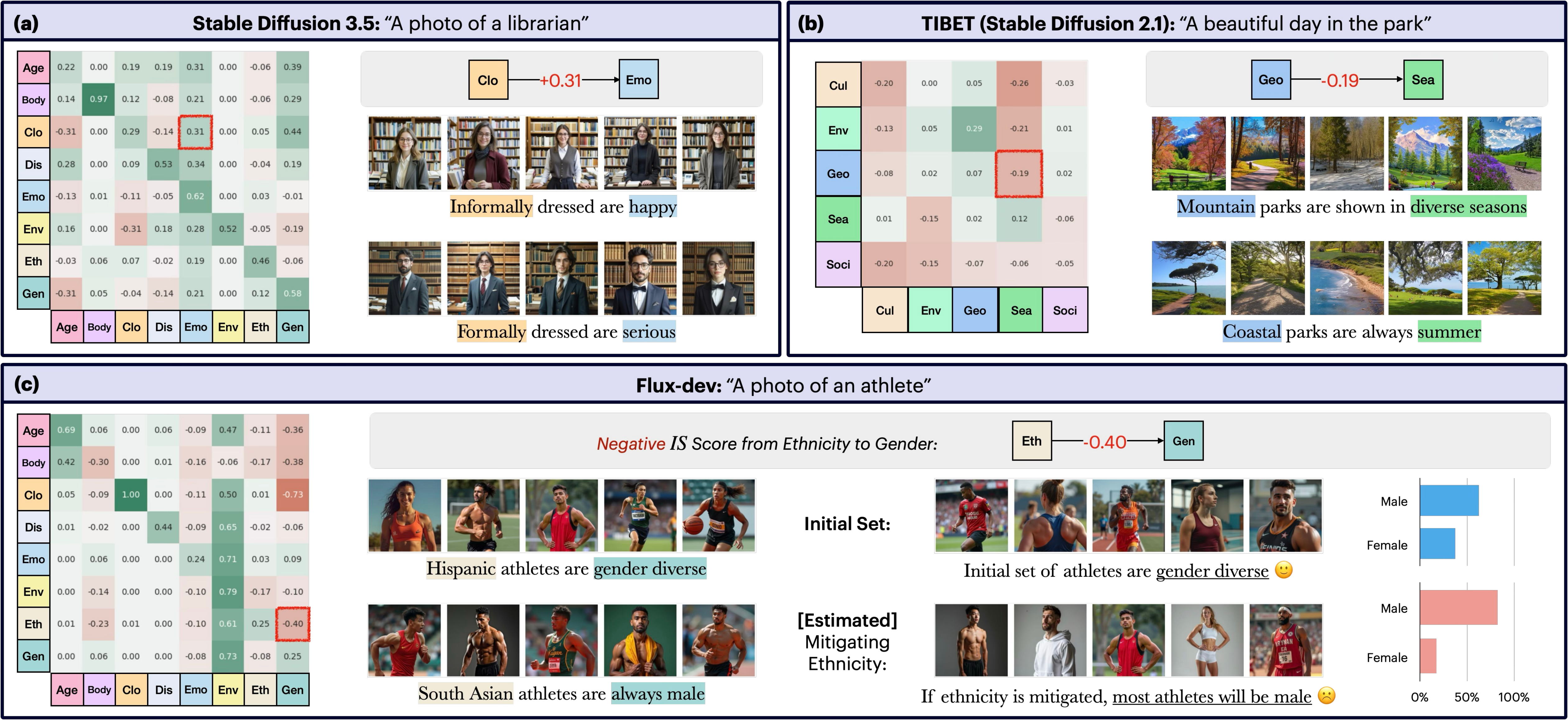}
   \caption{Analyzing bias intersectionality matrices from \modelnameintersect. (a) Shows how mitigating clothing bias also mitigates emotion bias. (b) Explores interactions between non-traditional bias axes in the TIBET dataset. (c) Reveals that generating ethnically diverse athletes reduces gender diversity. \modelname can allow the user the user to understand whether interventions along one dimension impact other dimensions positively or negatively.   }
   \label{fig:examples}
   % \vspace{-0.2cm}
\end{figure*}

\subsubsection{Validating \isscore}
\label{sec:inter_mitigation}
To evaluate the effectiveness of our framework, we examine how well our \textit{Intersectional Sensitivity Score} (\isscore) predicts the downstream effects of counterfactual-based bias mitigation. Specifically, our goal is to estimate how intervening on one bias dimension ($B_x$) impacts another ($B_y$), without actually performing the full mitigation step. This predictive capability is crucial, as real-time mitigation across large prompt sets and multiple models is often computationally expensive.

To validate our estimates, we conduct an experiment using two mitigation strategies: ITI-GEN on Stable Diffusion 1.4 and hardprompt on Stable Diffusion 3.5. For each of the 26 occupations in our dataset, we consider all directed bias relationships $B_x \to B_y$ and independently apply mitigation along $B_x$. We then regenerate image sets for each mitigated prompt and extract updated VQA-based attribute distributions for $B_y$. Using these updated distributions, we compute a new post-mitigation \isscore, which we define as:

\begin{equation}
w_{B_y}^{B_x} = W_1(D_{B_y}^{\text{mit}(B_x)}, D^*)
\end{equation}
\begin{equation}
\textrm{IS}_{xy}^{\text{mit}(x)} = \overline{w}{B_y}^{\text{init}} - \overline{w}{B_y}^{B_x}
\end{equation}

Here, $D_{B_y}^{\text{mit}(B_x)}$ represents the distribution of attribute values along axis $B_y$ after mitigating $B_x$, and $D^*$ denotes the ideal target distribution. 

We then compute the Pearson correlation between the initial \isscore\ values (before mitigation) and the post-mitigation values $\textrm{IS}_{xy}^{\text{mit}(x)}$ across all $B_x \to B_y$ pairs. This correlation quantifies how accurately our original estimates capture the real-world effect of mitigation.

Our results show a strong average correlation of +0.65 across occupations using ITI-GEN, with particularly high correlations for specific prompts such as musician (+0.91), accountant (+0.81), and lawyer (+0.82). For hardprompt, the correlation is even stronger—averaging +0.95—which is expected since the mitigation prompts are closely aligned with the counterfactuals used in our estimation process.

These high correlations demonstrate that our framework effectively predicts the impact of counterfactual-based interventions on secondary bias dimensions without requiring actual mitigation. This capability allows practitioners to anticipate fairness trade-offs and make informed decisions about which bias axes to mitigate, how those interventions may affect others, and whether the expected benefits are worth the cost. Our approach provides a practical and interpretable means of assessing intersectional bias entanglement in TTI models, offering a valuable pre-mitigation diagnostic tool.

\begin{figure}
  \centering
   \includegraphics[width=0.8\linewidth]{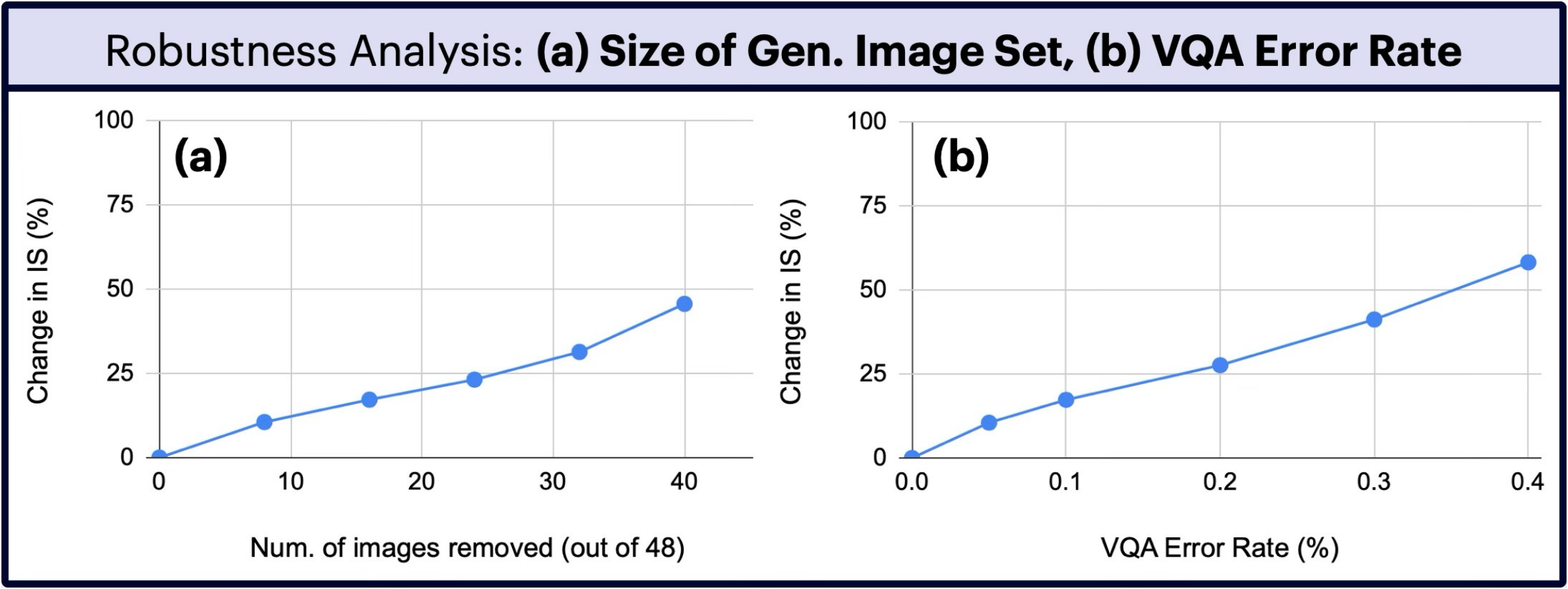}
   \caption{\textbf{Sensitivity analysis on \modelnameintersect}. We evaluate the robustness of our approach by analyzing the impact of VQA errors and the effect of the number of images on \isscore.}
   \label{fig:sensitivity}
   % \vspace{-0.2cm}
\end{figure}

\subsubsection{Robustness of \modelnameintersect}
\label{sec:robustness}

We analyze the robustness of our method by evaluating the impact of  number of images (Figure. \ref{fig:sensitivity}(a))  and VQA error rate (Figure. \ref{fig:sensitivity}(b)) on \isscore\ values. Our method uses 48 images per prompt to study bias distributions. Removing 8 images (16.6$\%$) results in a 10.5$\%$ drop, an removing 32 images (66.6$\%$) yields a 31.3$\%$ drop. This sub-linear impact suggests that TTI models often generate similar bias distributions (e.g., always depicting nurses as females), preserving overall trends despite fewer images. Therefore, our approach is robust to moderate reductions in image count, but very small sets of images will significantly affect \isscore\ values. To test the robustness over VQA errors, we randomly change the VQA answers to a different answer (simulating an incorrect answer), from 5\% to 40\% of the time. We observe that even with low error rates of 5\% and 10\%, \isscore\ values change by 10\% and 17.3\% respectively. Here, the impact is compounded twice, because an error can skew the distribution away from one counterfactual towards another, and that a 5\% error causes 13,478 answers out of a total of 269,568 answers to be changed, which is substantial. Nonetheless, we note that this impact remains linear. As VQA models improve, achieving low error rates for robustness becomes practical.

\section {\modelnamegraph: From Matrices to Causal Graphs}

While the \isscore\ provides a useful measure of how one bias dimension influences another, the resulting intersectional matrices do not convey the underlying causal structure between bias axes. In other words, while these matrices quantify the strength of association or interaction, they do not reveal whether one axis is causally dependent on another. To address this, we extend our framework beyond matrix-based analysis and propose a graph-based approach that captures directional dependencies among bias dimensions.

As before, we begin by generating prompt-level counterfactuals and synthesizing corresponding images using a TTI model (Section~\ref{sec:Approach:Imggen}). We then apply a visual question answering (VQA) pipeline to extract interpretable bias-related attributes from the generated images (Section~\ref{sec:Approach:VQA}). Since the \isscore\ can be interpreted analogously to a causal treatment effect—measuring how an intervention on one axis affects another—we use it to guide a causal discovery process. Specifically, we perform pairwise conditional independence testing between bias axes to identify statistically significant dependencies (Section~\ref{sec:Approach:CausalDiscovery}).

This graph-based extension offers several advantages. First, it moves beyond symmetric correlation to uncover asymmetric, directional relationships, helping us identify which bias axes act as ``sources" (i.e., interventions here propagate effects) and which act as ``sinks" (i.e., axes affected by others). Second, we visualize this information using a directed graph structure, rather than a flat matrix, improving interpretability in high-dimensional settings. Third, we apply this approach to measure intersectional dependence only among strongly connected bias axes, ensuring that the structure reflects meaningful entanglement rather than noise.

Finally, we highlight several potential uses of this graph-based analysis. It allows for comparative audits across multiple TTI models, helps identify optimal mitigation paths by targeting upstream bias sources, and reveals how bias interactions encoded in training data or real-world prompts may be altered—or even amplified—in generative model outputs.

\section{Pairwise Causal Discovery}
\label{sec:Approach:CausalDiscovery}
Given an initial set of bias axes \( B \), we define an intersectional relationship between a pair of biases \((B_x, B_y)\) as \( B_x \to B_y \), indicating that a counterfactual intervention on \( B_x \) to mitigate its bias also affects \( B_y \). As a first step, we intervene across all \( n \times n \) bias relationships. Using attributes extracted by the VQA, we can count the attributes for a bias axis $B_y$ over any set of images. We construct a contingency table where rows represent the intervened bias axis \( B_x \) , and columns capture the distribution on the target axis \( B_y \) (e.g., age with old, middle-aged, and young categories). The values in the contingency tables are the counts of attributes of $B_y$ over the counterfactual image sets of $B_x$.

Next, we refine these relationships by extracting only statistically significant ones. This ensures that only strong dependencies between different bias pairs are retained. We apply conditional independence testing using the Chi-square (\(\chi^2\)) test, pruning bias pairs with respect to \( B_x \) if their p-value exceeds a predefined threshold (p-value$>0.0001$). Bias pairs with a p-value below this threshold are considered strongly dependent, indicating that intervening on \( B_x \) results in a significant change in the other bias axis. This process is applied iteratively for all bias axes. This step is referred to as Pairwise Causal Discovery, and it returns a set of bias pair relationships where mitigating along one bias axis has led to a strong change in another bias dimension. \isscore is used as weights for these edges.

\subsection{Visualization}

Following the process above, we have a set of pairwise causal relationships for all significant intersectional bias pairs $B_x \to B_y$. Furthermore, each pair $B_x \to B_y$ has an \isscore\ score to quantify the intersectional effects. There are many ways to represent these pairwise relationships, including building an $n \times n$ matrix, or a graph with $n$ nodes and directed edges that represent the relationships between these nodes. 

A usermay want to understand all important intersectional effects together. To that end, we adopt a graph representation for our output. This graph is referred to as a \modelnamegraph in the rest of the paper. Figures \ref{fig:examples} and \ref{fig:main} show examples of such graphs. To interpret this graph, first pick a focal node where the intervention takes place. All outgoing edges from this node indicate intersectional relationships that are statistically significant. The weights of the edges show the \isscore\ and can be interpreted as the impact of intervention on the bias axis for the focal node.
\section{Applications}
\label{sec:inter_Applications}

%In this section, we explore various applications of our tool in analyzing bias interactions across different Text-to-Image (TTI) models. Additionally, we discuss its potential use cases and highlight the importance of gaining a deeper understanding of different bias interactions.

\subsection{Applying \modelnamegraph\ to analyze TTI models}
\label{sec:globalanalysis}

\begin{figure*}
  \centering
   \includegraphics[width=\linewidth]{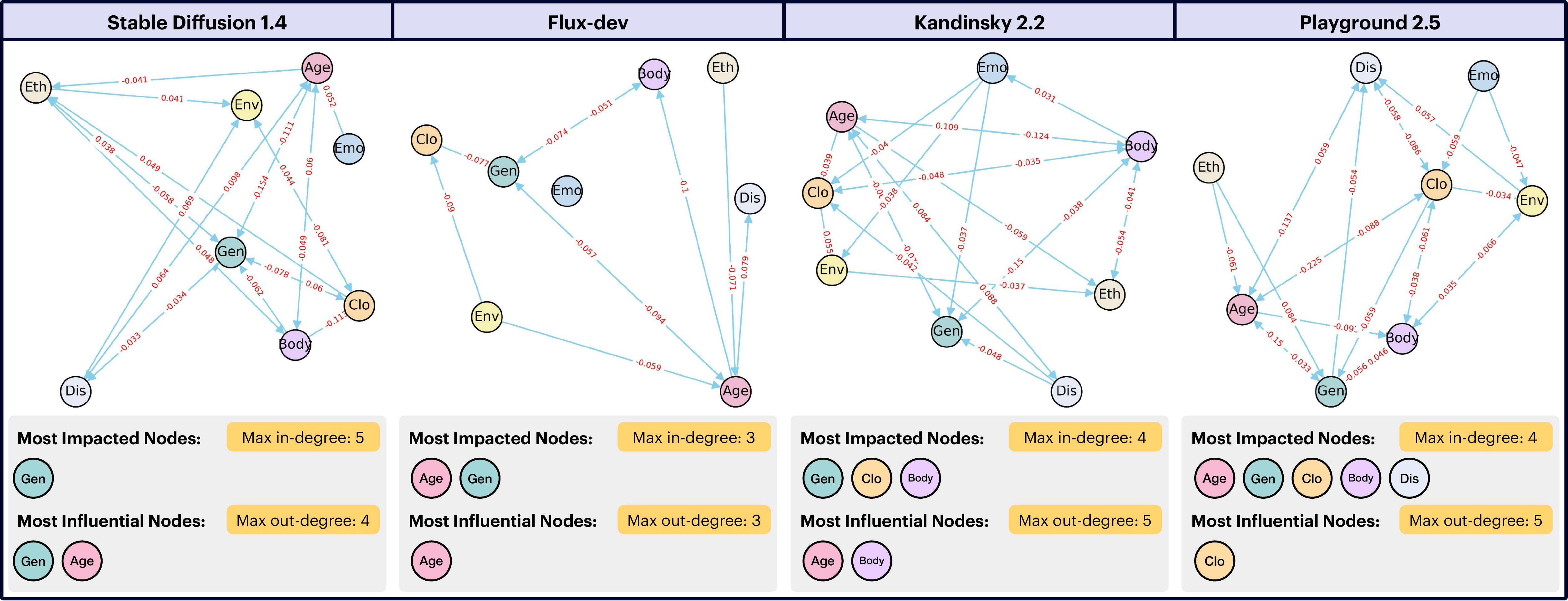}
   
   \caption{We compare aggregated causal graphs for four models: Stable Diffusion 1.4, Flux-dev, Kandinsky 2.2, and Playground 2.5. These graphs combine pairwise causal relationships across all bias axes, accumulated from occupation prompts in our dataset.}
   \label{fig:main}
   % \vspace{-0.2cm}
\end{figure*}

\subsubsection{Global Aggregations}
\label{sup:aggregation}

In order to do a comparative analysis of intersectionality across models over a dataset of prompts, we perform an aggregation step. For the 26 occupation prompts, we first start by using counterfactuals and VQA to identify attributes over all bias axes in $B$. Now, in the Causal Discovery step, we build contingency tables that aggregate attributes over all $CF$ prompts across all the occupations. For example, when considering the intersectional relationship $Gender \to Age$, we consider all images for \texttt{male} occupation and \texttt{female} occupation for all occupations for the rows of the contingency matrix, and count over the $Age$ attributes \texttt{young, middle-aged, old} to find the overall global distribution. This gives us the global contingency table for any bias pair. We follow the steps in Sec. \ref{sec:Approach:CausalDiscovery} to obtain this list of bias intersectionality relationships that are significant. Next, in order to compute \isscore, we use the same contingency table and sum across its columns to get $D_{B_y}^{\textrm{global}(B_x)}$. For the initial distribution, we accumulate attributes across all initial prompt images for all occupations, to give us $D_{B_y}^{\textrm{global}(init)}$. We can now compute \isscore\ as:

\begin{equation}
w_{B_y}^{\textrm{global}(\text{init})} = W_1(D_{B_y}^{\textrm{global}(\text{init})}, D^*)
\end{equation}
\begin{equation}
w_{B_y}^{\textrm{global}(B_x)} = W_1(D_{B_y}^{\textrm{global}(B_x)}, D^*)
\end{equation}
\begin{equation}
\textrm{IS}_{\textrm{global}(xy)} = w_{B_y}^{\textrm{global}(\text{init})} - w_{B_y}^{\textrm{global}(B_x)} 
\end{equation}

Given the large number of images (as we aggregate over multiple sets), we choose to use a p-value threshold of $0.00005$, and we further discard edges in the pairwise causal graph where the $-0.03 > \textrm{IS}_{\textrm{global}(xy)} > 0.03$.

\subsubsection{Uncovering optimal Bias mitigation strategies}

\noindent\textbf{Identifying high-impact biases.} Some biases act as primary sources, influencing multiple others, while some function as effects, shaped by upstream factors. A node’s impact is measured by its outgoing edges (\textbf{MaxImp}, Table \ref{tab:mitigationcorr}), while its susceptibility to influence is quantified by incoming edges (\textbf{MaxInf}). This helps in model selection based on specific bias priorities.

As an example, let's analyze how this information can help in selecting appropriate models using the global graphs in Figure \ref{fig:main}. If a user prioritizes robustness to age-related bias when selecting a model, Kandinsky 2.2 would be the best choice, as its Age node is the least influenced by other biases in the global analysis. This means that modifying other attributes (e.g., gender or clothing) has minimal unintended effects on age representation, ensuring more stable and independent age depictions across generated images.

Similarly, if the goal is to generate occupation-related images while minimizing unintended bias propagation across other attributes, Playground 2.5 is the optimal choice. In this model, variations in body type have the least impact on other biases, meaning changes in body shape do not disproportionately affect other attributes like gender, ethnicity, or perceived professionalism. This makes Playground 2.5 preferable in scenarios where maintaining fairness across multiple dimensions while altering body type is critical. By analyzing bias influence and susceptibility, users can make informed choices based on fairness priorities, whether aiming for stability in a bias axis or minimizing unintended shifts in related attributes.
\begin{table}[t]
  \small
  \centering
   \caption{\textbf{Correlation Between Estimates and Post-Mitigation Evaluation on ITI-GEN.} The high correlation validates our mitigation estimates. For each prompt, we report one of the most influenced node (MaxInf) and the node with the greatest impact on others (MaxImp).
 }
  \begin{tabular}{lcccc}
    \toprule
    \textbf{Prompt} & \textbf{Edges} & \textbf{Corr.} & \textbf{MaxInf} & \textbf{MaxImp}\\
    \midrule
    Pharmacist  &  12  &  +0.399 & Gender & Age\\
    Scientist  &  9  &  +0.600 & Clothing & Ethnicity\\
    Doctor  &  9  &  +0.638 & Age & Disability \\
    Librarian  &  14  &  +0.805 & Emotion & Age \\
    Nurse  &  5  & +0.997  & Age & Disability\\
    Chef  &  8  &  +0.757 & Bodytype & Ethnicity\\
    Politician  &  10  &  +0.782 & Emotion & Disability \\
    \midrule
    \textbf{Overall} & - & \textbf{+0.696} & Gender & Age\\
    \bottomrule
  \end{tabular}
  %\vspace{-0.2cm}

\label{tab:mitigationcorr}
  %\vspace{-0.2cm}
\end{table}
\begin{figure*}
  \centering
   \includegraphics[width=0.6\linewidth]{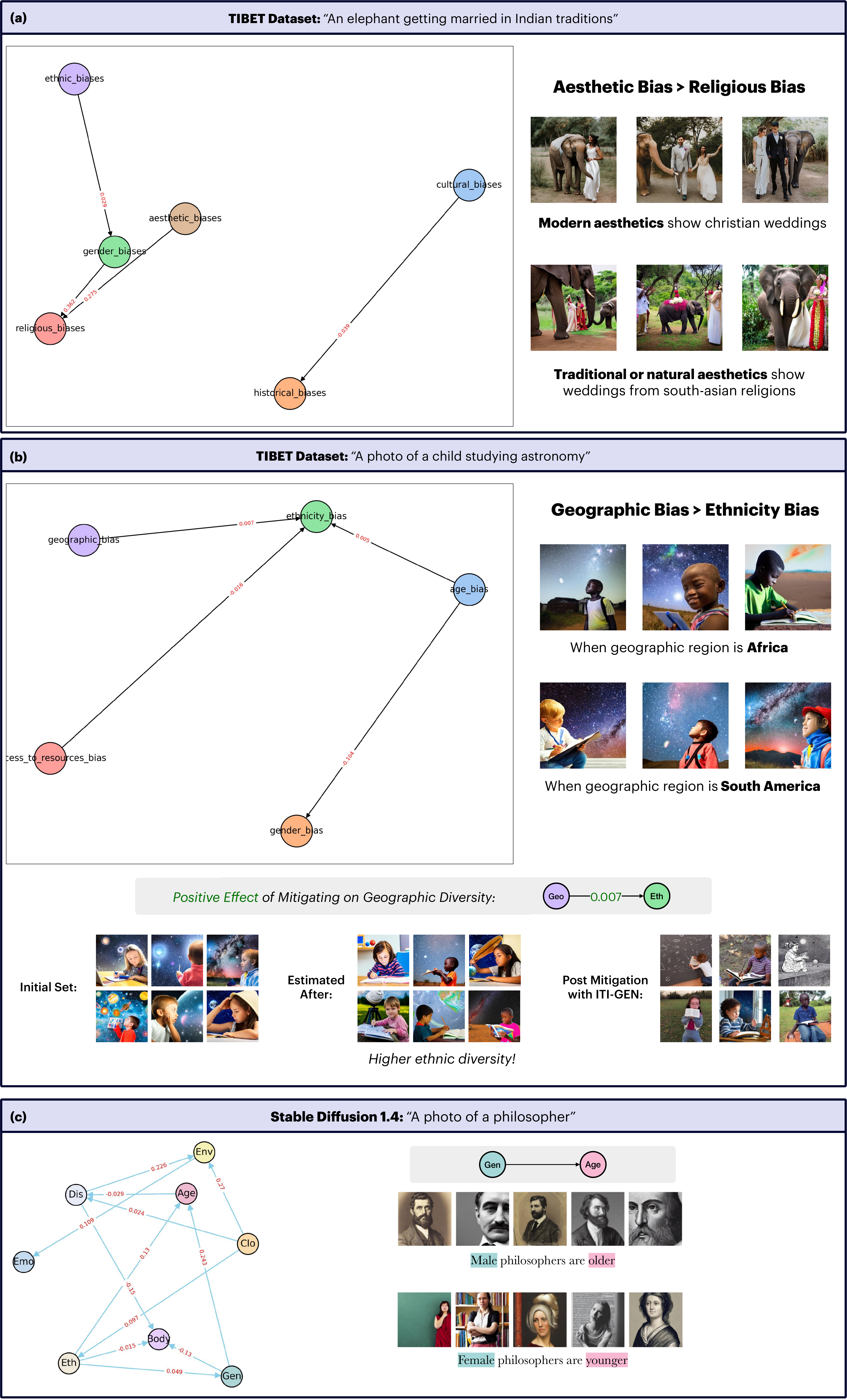}
   \caption{Additional examples on TIBET dataset (a-b) and Occupation prompt (c) on prompt-level analysis provided by \modelnamegraph. }
   \label{fig:moreegs}
   % \vspace{-0.2cm}
\end{figure*}

\begin{figure}
  \centering
   \includegraphics[width=0.8\linewidth]{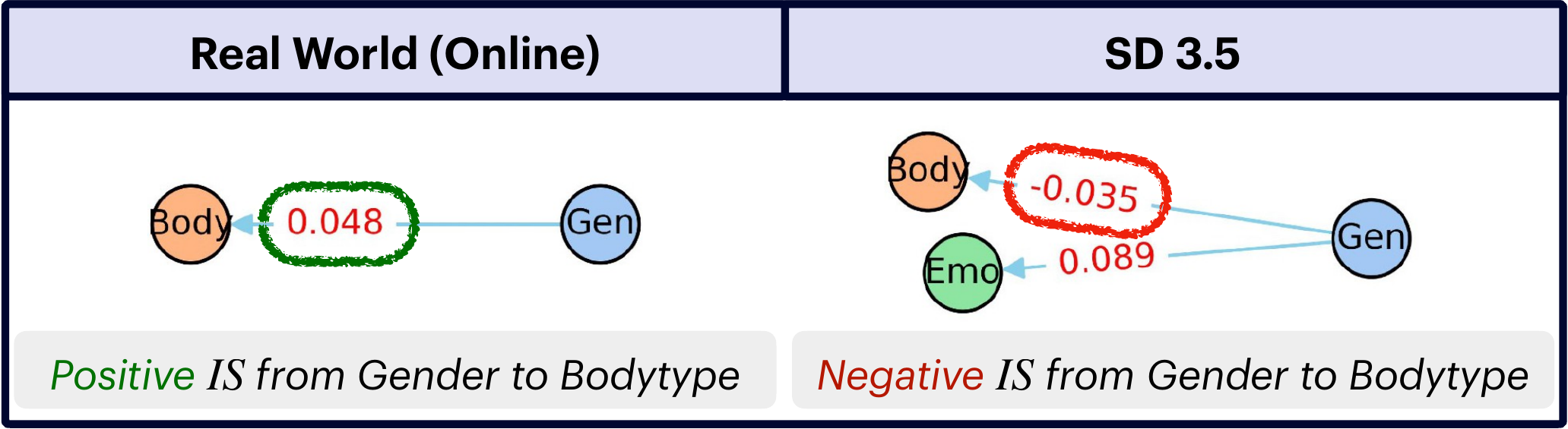}
    \caption{Comparison of real-world and Stable Diffusion 3.5 pairwise causal relationships for gender in computer programmer images. In the real world, gender diversification increases body type diversity, whereas in Stable Diffusion 1.4, it has a negative impact. }
   \label{fig:realworld}
   % \vspace{-0.2cm}
\end{figure}

\begin{figure*}[ht]
  \centering
   \includegraphics[width=\linewidth]{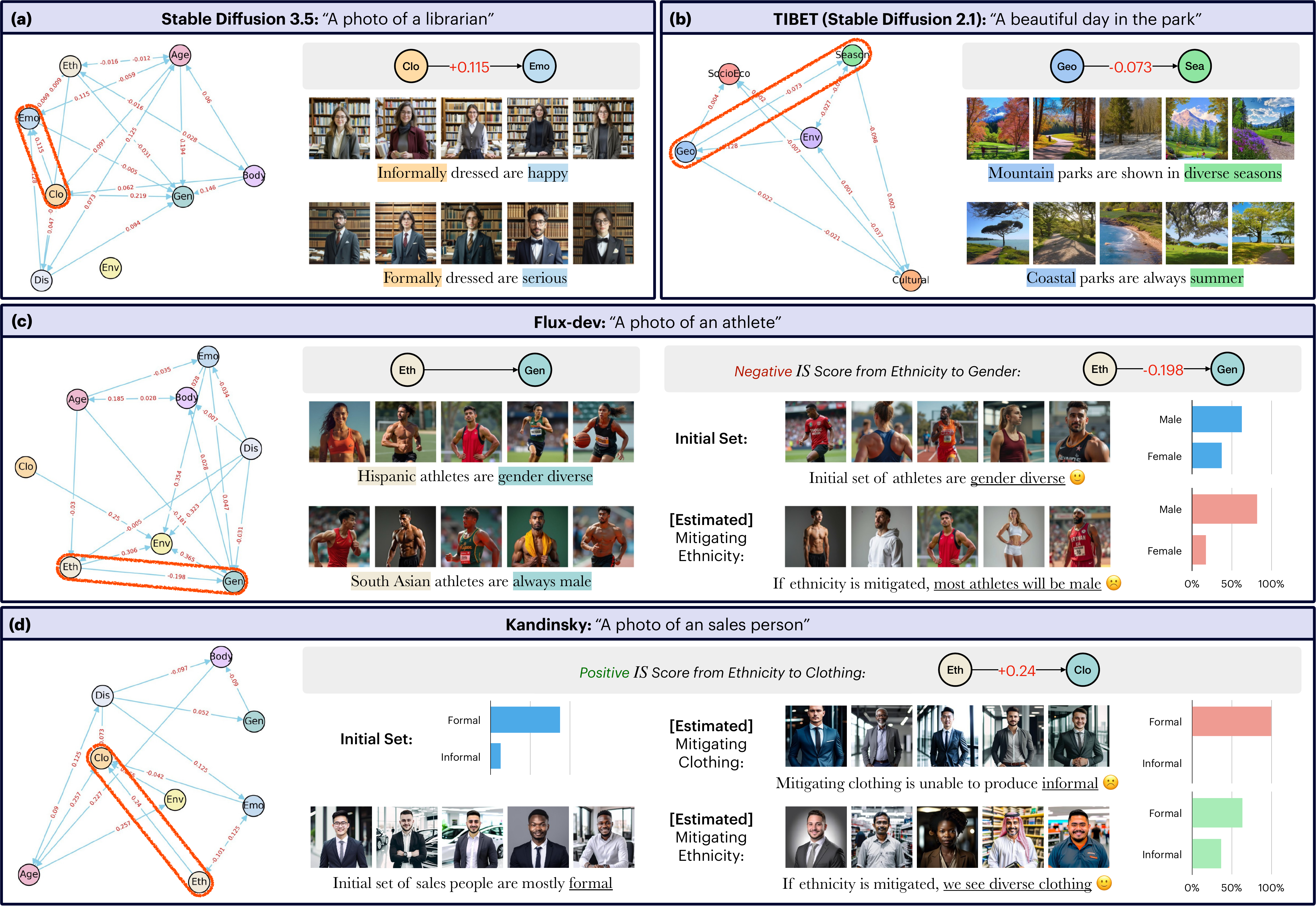}
   \caption{The figure illustrates bias interpretations from Bias Connects, combining all pairwise graphs into one. (a) Shows how mitigating clothing bias also mitigates emotion bias. (b) Explores interactions between non-traditional bias axes in the TIBET dataset. (c) Reveals that generating ethnically diverse athletes reduces gender diversity. (d) Demonstrates that diversifying salesperson clothing is best achieved by increasing ethnic diversity rather than directly specifying clothing variation. }
   \label{fig:examples}
   % \vspace{-0.2cm}
\end{figure*}
\subsection{Studying Real-World Biases}
\label{sec:real_world_biases}

\modelnamegraph\ can also be used to compare the distribution of images generated by TTI models against real-world data, offering insights into how generative models may distort or amplify real-world bias structures. Instead of assuming a uniform target distribution for computing bias sensitivity, we extend our framework by using empirical distributions derived from real-world image datasets as the reference. This allows us to evaluate whether a TTI model's internal bias dynamics align with those observed in the real world.

Given a prompt $P$ (e.g., \textit{``A computer programmer''}), let $B = [B_1, B_2, ..., B_n]$ denote the set of relevant bias axes (e.g., gender, age, race). For each axis $B_y$, we define:
\begin{itemize}
    \item $D_{B_y}^{\text{real}}$: the real-world distribution of $B_y$, obtained from a curated image dataset or known statistics.
    \item $D_{B_y}^{\text{TTI}}$: the distribution of $B_y$ attributes in the TTI-generated images.
\end{itemize}

We first measure the initial bias deviation by computing the Wasserstein-1 distance between these two distributions:
\begin{equation}
    w_{B_y}^{\text{init}} = W_1(D_{B_y}^{\text{TTI}}, D_{B_y}^{\text{real}})
\end{equation}

Next, to assess how intervening on another bias axis $B_x$ affects $B_y$, we compute the post-intervention distribution $D_{B_y}^{B_x}$ and its distance from the real-world reference:
\begin{equation}
    w_{B_y}^{B_x} = W_1(D_{B_y}^{B_x}, D_{B_y}^{\text{real}})
\end{equation}

The change in alignment with the real-world distribution is captured by the \isscore\ score:
\begin{equation}
    \textrm{IS}_{xy} = w_{B_y}^{\text{init}} - w_{B_y}^{B_x}
\end{equation}

A positive $\textrm{IS}_{xy}$ indicates that intervening on $B_x$ brings $B_y$ \textit{closer} to its real-world distribution, while a negative value indicates divergence. To measure the overall degree of intersectional distortion or alignment, we define an aggregate metric:
\begin{equation}
    \mathcal{I} = \sum_{x \neq y} |\textrm{IS}_{xy}|
\end{equation}
where a higher $\mathcal{I}$ signifies greater intersectional bias amplification and entanglement, while a lower value suggests closer alignment with real-world distributions.

To illustrate this analysis, we curated a small dataset comprising 48 real-world images of computer programmers, as well as 48 male and 48 female programmer images sourced from the internet. We then compared the intersectional dependencies in this real-world dataset with those in images generated by Stable Diffusion 3.5 for the same prompt. As shown in Figure~\ref{fig:realworld}, in real-world data, gender diversification positively influences body type diversity. However, in Stable Diffusion 3.5, gender not only impacts body type negatively—reducing its diversity—but also influences emotion, revealing a divergence from real-world relationships. Such discrepancies highlight how TTI models may amplify or alter real-world bias interactions, reinforcing the importance of intersectional auditing with tools like \modelnamegraph.

\section{Conclusion}
\label{sec:Conclusion}
Our study proposes a tool to investigate intersectional biases in TTI models. While prior research has explored bias detection and mitigation in generative models, to the best of our knowledge, no previous work has focused on understanding how biases influence one another. We believe our work makes a significant contribution by enabling a more nuanced analysis of bias interactions.
Beyond academic research, \modelnameintersect\ and \modelnamegraph\ have practical applications, including comparing biases dependencies learned across different models, establishing empirical guarantees for mitigation, and determining optimal mitigation approaches that account for intersectionality. We hope that this tool will facilitate more informed decision-making for AI practitioners, policymakers, and developers, ultimately leading to more equitable and transparent generative models.

While  \modelnameintersect and  \modelnamegraph\ provides a valuable framework, it represents only an initial step toward a more comprehensive causal approach to understanding intersectionality. Our current setup does not allow us to reason about indirect causal effects, or develop an optimal bias mitigation strategy that utilizes our tool to mitigate multiple biases simultaneously. Addressing these challenges presents an important avenue for future research. We also need to develop mitigation strategies that consider these interdependencies.

\noindent{\textbf{Ethical Considerations.}}
We acknowledge that the presence of biases in generative AI models can lead to real-world harms, reinforcing stereotypes and disproportionately affecting marginalized groups. Our tool is intended to provide researchers and practitioners with a means to better understand and mitigate these biases, rather than to justify or amplify them. Additionally, we recognize that bias analysis can be sensitive to the choice of datasets, evaluation methods, and experimental assumptions, and we encourage future work to refine and expand upon our approach.

\newcommand{\mitname}{InterMit\xspace}
\newcommand{\hardprompt}{{\textsl{PM}}\space}
\chapter{\mitname: A Modular Algorithm for Intersectional Bias Mitigation in TTI Models}
\epigraph{“A system cannot fail those it was never meant to protect.”}{\textit{W.E.B. Du Bois}}
%\epigraph{
	%\footnotesize{
		%.%”

%}}{\textit{W.E.B. Du Bois}}

%\vspace*{-0.8\baselineskip}

\begin{figure}[h]
  \centering
  \includegraphics[trim=140 30 140 30, clip,width=0.5\textwidth]{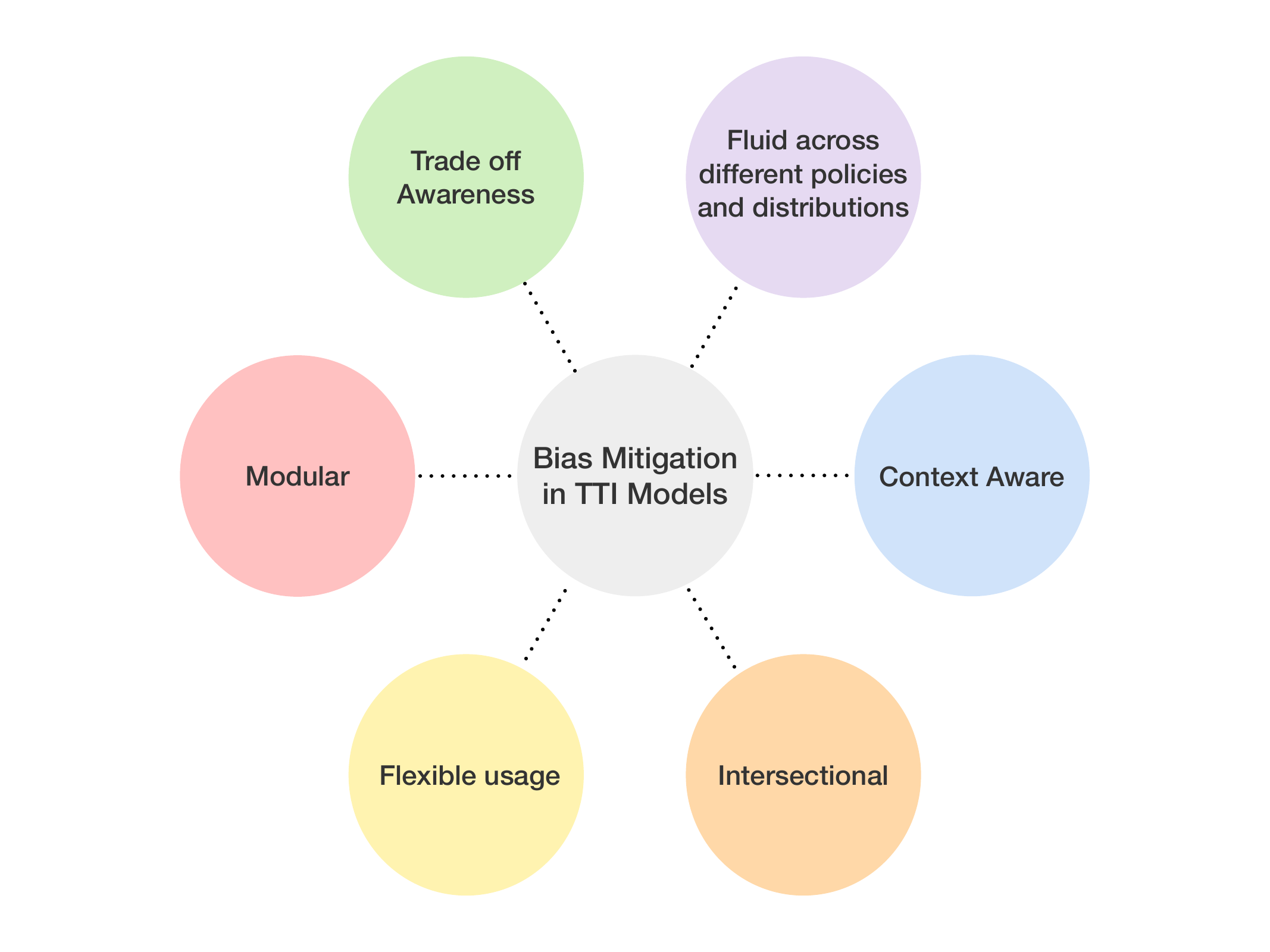}
  \caption{Key challenges in mitigating bias in TTI models. Effective methods must handle how bias changes with different prompts, capture how multiple biases interact, and let users choose which biases to address. They should support different fairness goals across cultures and help users understand the trade-offs involved in reducing specific types of bias.}
  \label{fig:Intersectionality_teaser}
\end{figure}

\clearpage

\section{Introduction}
In the previous chapters, we introduced a suite of diagnostic tools, \modelname, \modelnameintersect, and \modelnamegraph, to systematically evaluate bias in Text-to-Image (TTI) generative models. \modelname enables dynamic, prompt-sensitive audits by using counterfactual perturbations to measure how attributes like gender, race, or age influence model outputs. \modelnameintersect and  \modelnamegraph build on this by revealing how interventions on one attribute (e.g., gender) affect others (e.g., age or ethnicity). Together, these tools offer a powerful framework for uncovering where and how bias manifests in TTI systems—highlighting not just individual failures, but structural patterns of representational harm. 

Yet while these diagnostic frameworks shed light upon the structure of the problem, they offer no guidance on how to act upon it. Should models always try to fix gender, race, and age bias at the same time? Or does it depend on the situation? For example, a healthcare tool that creates patient materials might focus more on racial diversity, while a fashion app might care more about showing different body types. Different users also have different goals: one person might want to remove gender stereotypes, while another might want to keep cultural details related to race. These choices are not just about style—they show deeper values. Also, fixing one type of bias but ignoring others can cause new problems. For example, if we improve gender balance in images of leaders but only show young women, we might end up reinforcing age bias or narrow ideas about gender roles, even while trying to be fair.

 First, these models are inherently context-sensitive: a term like ``man'' or ``doctor'' cannot be mitigated in isolation, as its meaning and visual portrayal depend on surrounding concepts and scene structure (e.g., ``a confident man leading a boardroom meeting'' versus ``an old man praying alone in a church''). Fairness in TTI is difficult to measure directly—the outputs are not discrete predictions but high-dimensional images, where bias must be inferred through proxy classifiers or semantic cues. These models are highly underconstrained and susceptible to non-linear prompt perturbations: small textual edits can yield disproportionate and unpredictable visual changes, making mitigation strategies brittle and difficult to scale.

%Existing approaches to bias mitigation in generative models suffer from several limitations. Many require retraining or fine-tuning, which is computationally prohibitive for large-scale diffusion systems and often inaccessible to end users. Others rely on prompt rewriting or latent editing (e.g., via GANs), but these interventions often introduce unintended visual artifacts or reinforce the very stereotypes they aim to counter—for example, associating femininity with makeup, or masculinity with aggressive facial features. More fundamentally, most techniques focus on a \textit{single bias axis at a time}, failing to address the reality that gender, race, age, and other attributes interact in complex and compounding ways.

A central, often unspoken dilemma in mitigating bias in TTI models is the question of what the ideal distribution of outputs should be. Should generative systems aim for demographic parity—ensuring equal representation of different identity groups across all prompts? Or should outputs reflect real-world statistics, such as actual gender ratios in specific professions or locations? But real-world statistics themselves often encode histories of exclusion and marginalization, making them a poor baseline for fairness. In other cases, fairness might require actively counterbalancing these historical biases—deliberately overrepresenting marginalized groups to correct for entrenched underrepresentation. Further complicating this question is the issue of context: a prompt like ``a nurse in Tokyo'' might demand a very different demographic distribution than ``a global tech CEO.'' Thus, fairness in generative models is not reducible to a single distributional ideal. Most existing mitigation strategies either hard-code a static notion of fairness (often implicitly) or provide no mechanism for users to express their own. 

The question of which axes to mitigate further complicates the mitigation landscape. Should models always address gender, race, and age simultaneously? Or are there domains where one axis is more ethically or contextually salient than others? For example, a healthcare system generating patient education materials might prioritize racial inclusivity, while a fashion recommendation tool might focus on body-type diversity. Even among individual users, fairness goals may vary sharply: one might seek to remove only gender stereotypes, while another may wish to preserve race-specific cultural cues. These preferences are not merely stylisticthey reflect deeper beliefs about what is fair or important. Also, fixing bias in one area while ignoring others can have unexpected and harmful effects. For example, improving gender balance in leadership roles but only showing younger women could still reinforce age bias or stereotypes about sexuality, even though it might appear fair at first glance.

These challenges are further amplified when considering intersecting identities, where biases do not operate independently but interact in compounding and often unpredictable ways. A mitigation strategy that improves gender balance in the aggregate may still fail to generate equitable representations of older Black women or nonbinary South Asian professionals. In some cases, mitigating one bias dimension may actively worsen another—e.g., interventions that diversify gender portrayals may inadvertently homogenize race or age in the process. This raises a critical question: when such trade-offs are unavoidable, what should the mitigation system optimize for? Should it minimize the average bias, focus on the most underrepresented intersections, or follow explicit user directives? The answer depends heavily on context and on whose notion of fairness is prioritized. InterMit is designed to make these trade-offs visible and tunable, offering users diagnostic transparency and actionable control.

Behind the technical challenges of building fair systems is a deeper question: who gets to decide what fairness means in a generative model? Is it the people who build the model, the companies that run the platform, the users, or the communities shown (or left out) in the results? Today, most systems are built with hidden ideas about fairness—often based on Western values—without ways for people to question or change them. But fairness isn’t a fixed rule. It depends on culture, history, and politics. So, any real attempt to reduce bias must ask: Who sets the rules, who decides what trade-offs are okay, and who is responsible when things go wrong? If systems aren’t open, user-focused, and built with community input, even the best models can end up repeating the same harms they were supposed to fix.

\begin{figure}[h]
  \centering
  \includegraphics[trim=140 30 140 30, clip,width=\textwidth]{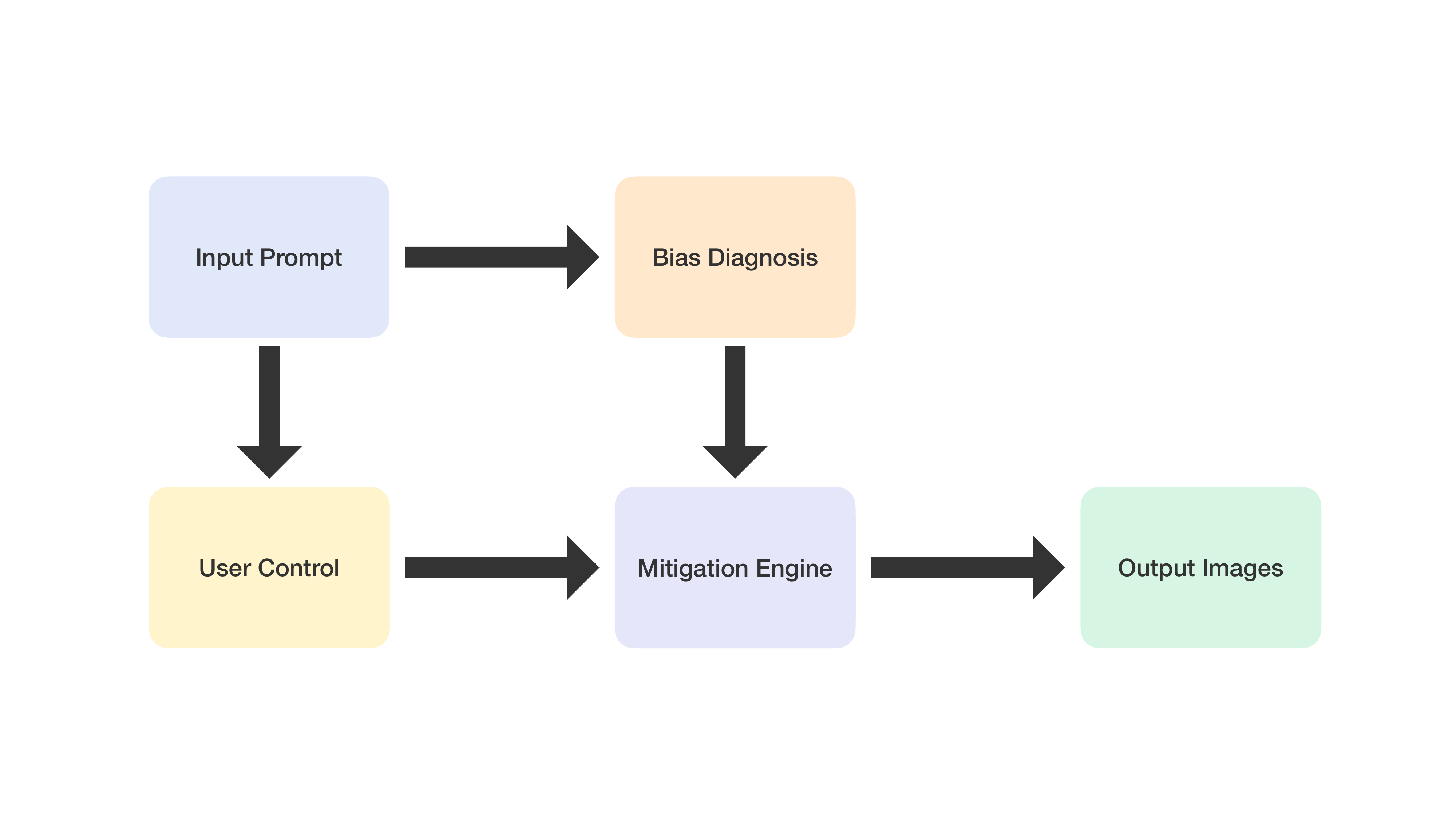}

\caption{Overview of the \mitname\ pipeline. Prompts are first analyzed using BiasConnect and the \isscore metric. Users then define their fairness preferences by specifying a priority vector and an ideal target distribution for each bias axis they wish to mitigate. \mitname uses this input to iteratively adjust the model’s outputs, accounting for intersectional dependencies and mitigating biases along the specified directions. Throughout the process, users are informed about the trade-offs associated with each mitigation step (Section \ref{sec:multi_step_mitigation}).}
\label{fig:intermit_pipeline}
\end{figure}

\section{What Should a Bias Mitigation Strategy for TTI Models Look Like?}

Given the challenges of building fair text-to-image (TTI) models, we outline key requirements that any effective bias mitigation strategy should meet.

\begin{enumerate}
    \item \textbf{Context-Aware and Prompt-Sensitive} Bias in TTI models is not a fixed property of the model, but a dynamic function of input prompts. The same identity attribute (e.g., gender) may be rendered in radically different ways depending on scene elements, occupations, or aesthetics. Therefore, mitigation must occur at the level of prompt semantics, not token-level substitutions. An effective strategy must dynamically analyze how prompts interact with demographic concepts and ensure that fairness is maintained across a range of plausible phrasings, genres, and prompt templates.

    \item \textbf{Robust to Bias Intersections} Biases in text-to-image models often don’t appear alone. A prompt that seems to reflect gender bias may also involve race, age, or profession in ways that are deeply connected. To truly reduce bias, we can’t just look at one category at a time. We need methods that examine how different biases interact and influence each other. This means going beyond simple checks that look at only one type of bias. Instead, we need tools like causal graphs, which show how one bias might affect another, and sensitivity matrices, which help identify which combinations of features lead to biased results. 

    \item \textbf{Prioritizes user needs} Fairness is not a universal constant—it varies by context, audience, and intention. A TTI model deployed in education may require different fairness constraints than one used in entertainment or advertising. As such, any mitigation strategy must be configurable. Users should be able to encode their ethical goals and fairness priorities through an explicit, interpretable interface—such as a priority vector that governs which axes of bias are most important, and how trade-offs should be managed when fairness improvements along one dimension risk degrading another.

    \item \textbf{Modular} Most deployed TTI models are large, closed-source, and difficult to retrain. Thus, any practical bias mitigation strategy must be \textit{training-free}, operating as a lightweight post-processing or pre-generation module. Such a strategy should be modular, capable of interfacing with a variety of TTI architectures (e.g., diffusion models, transformers) and diagnostic tools (e.g., prompt auditors, concept classifiers). This modularity also enables compatibility with evolving fairness frameworks and domain-specific applications.

    \item \textbf{Transparent and Trade-off Aware} Bias mitigation in generative systems inherently involves trade-offs—between performance and fairness, between different identity groups, and between competing stakeholder values. An ideal mitigation strategy must make these trade-offs auditable and visible. Users should be informed about the potential consequences of mitigation actions, including which intersections were improved or worsened, and what assumptions about the ideal distribution were made in the process. Without such transparency, fairness interventions risk being opaque, brittle, or misaligned with user goals.

    \item \textbf{Adaptive to different definitions of Fairness} Finally, a mitigation strategy must acknowledge that fairness is culturally situated and socially contested. What counts as appropriate representation varies across communities, and no technical fix can substitute for participatory engagement with those affected by generative outputs. Therefore, an ideal system must support workflows where fairness preferences can be set, contested, and updated—whether by end users, institutions, or stakeholder communities.

\end{enumerate}

\medskip

\section{Proposed Solution}

Taking into account the challenges outlined above, we propose \mitname a modular and training-free framework that addresses intersectional biases during the mitigation process. \mitname\ is designed to meet all the key requirements discussed in the previous section. First, it is context-aware and prompt-sensitive mitigation operates at the prompt level using counterfactual perturbations and causal sensitivity analysis, allowing it to adapt to the nuanced semantics of diverse generation tasks. Second, it is intersectionally robust, leveraging causal graphs from \modelnameintersect to diagnose entangled biases and prioritize interventions that avoid collateral harm across social dimensions. Third, \mitname\ is user-aligned and configurable through its use of a priority vector, which enables users to select specific bias axes to focus on.  Fourth, it is fully training-free and modular, requiring no access to the TTI model’s internals and can be integrated on top of any bias mitigation approach.  Fifth, it is trade-off aware and transparent, quantifying and visualizing how mitigation along one axis influences others—thus alerting users when fairness improvements come at the cost of unintended harms. 
Sixth, it is flexible to different definitions of fairness and allows users to choose ideal distribution for each dimesnion of bias making  and explicitly encode trade-offs according to their own fairness goals. 
 In our evaluation, \mitname\ outperforms existing methods by mitigating biases more effectively, producing higher-quality images, and requiring fewer mitigation steps. Moreover, unlike other methods, it can handle a larger number ($>3$) of bias axes and alerts users when mitigation along one axis adversely affects others. 

\section{Methodology}

This section introduces \mitname, an iterative strategy for mitigating intersectional biases. We begin by outlining the general input framework presented to the user (Section~\ref{sec:user_interface}). Next, we briefly describe how our mitigation algorithm builds on ideas from previous chapters, specifically leveraging \modelnameintersect and the \isscore metric as its foundation (Section \ref{sec:model_ineractions}. Finally, we explain the full mitigation process, which is detailed in Algorithm~\ref{alg:intersectional_mitigation} (Section \ref{sec:multi_step_mitigation})

\subsection{User Interface}
\label{sec:user_interface}
Given an input prompt $P$ and a Text-to-Image (TTI) model $M$, our approach allows users to control how bias is mitigated based on their own fairness goals. The process begins by selecting a subset of relevant bias dimensions $B^* \subseteq B$, where $B = \{b_1, b_2, \dots\}$ is a larger set of possible bias axes. This full set of dimensions can either be predefined or generated dynamically using a method like \modelname. For each selected bias dimension $b_i \in B^*$, the user specifies an ideal distribution $D^{*}_i$. This distribution represents how the user believes the model should behave if the bias were fully mitigated along that axis. In addition, the user assigns a priority score to each selected bias dimension through a priority vector $\mathbf{p}$, where each value indicates the relative importance of that dimension. The vector is normalized such that $|\mathbf{p}|_1 = 1$, ensuring the weights are interpretable and comparable.

\paragraph{\textbf{Example} :}Suppose the input prompt is:
$P$: ``A CEO giving a presentation''. 
\begin{itemize}
    \item The system offers a set of possible bias dimensions: $B = \{\text{gender}, \text{race}, \text{age}\}$. 
    \item The user selects $B^* = \{\text{gender}, \text{age}\}$ as the relevant biases to address.
    \item For gender, the ideal distribution is defined as $D^*_{\text{gender}} = [50\% \text{ male},\ 50\% \text{ female}]$.
    \item For age, the ideal distribution is defined as $D^*_{\text{age}} = [30\% \text{ young},\ 40\% \text{ middle-aged},\ 30\% \text{ older adults}]$.
    \item The user then defines a priority vector $\mathbf{p} = [0.7,\ 0.3]$, indicating that gender bias is more important to mitigate than age bias in this context.
    
\end{itemize}

This setup allows the user to specify which biases to address, define what fairness should look like along each axis, and determine acceptable trade-offs based on their specific needs.

\subsection{Understanding Bias Interactions using \modelnameintersect}
\label{sec:model_ineractions}
Once the user has defined their bias priorities, we use \modelnameintersect\ to analyze how different bias dimensions interact. This step allows the system to assess whether mitigating one type of bias (e.g., gender) will impact other types (e.g., age or race), either positively or negatively. It helps uncover dependencies that can inform more effective mitigation strategies. Notably, if the user selects only a single bias axis, this step is skipped, as intersectional effects are not relevant in that case.

\modelnameintersect\ is designed to identify and quantify how mitigating one bias axis, denoted \( B_x \), affects another axis \( B_y \). This enables the evaluation of intersectional dependencies between biases. The approach works by systematically modifying prompts along specific bias dimensions and analyzing how these interventions change the distributions of other attributes in the generated images.

\paragraph{Counterfactual Prompts \& Image Generation.}
\label{sec:Approach:Imggen}

Given an input prompt \( P \) and a set of bias axes \( B = \{B_1, B_2, \dots, B_n\} \), we generate counterfactual prompts \( \{\textrm{CF}_i^1, \dots, \textrm{CF}_i^j\} \) for each \( B_i \in B \). These prompts are created by altering specific attributes (e.g., changing gender from male to female) and used with a TTI model to generate corresponding images. These counterfactual generations simulate interventions on a specific axis \( B_x \).

\paragraph{VQA-based Attribute Extraction.}
\label{sec:Approach:VQA}

To extract semantic attributes relevant to bias analysis, we use a Visual Question Answering (VQA) model—MiniGPT-v2 \cite{chen2023minigptv2}—to answer predefined, multiple-choice questions for each generated image. For example, for the gender axis, we ask:  
\texttt{[vqa] What is the gender (male, female) of the person?}

\paragraph{Quantifying Intersectional Sensitivity.}
\label{sec:Approach:CausalEffect}

To measure how intervening along one bias affects another, we had defined a metric called Intersectional Sensitivity in the previous chapter, which captures the directional effect of mitigating \( B_x \) on \( B_y \). This involves the following steps:

\begin{enumerate}
    \item  \textbf{Measuring Initial Bias.}  
For the original prompt \( P \), we compute the empirical distribution of attributes for axis \( B_y \), denoted \( D_{B_y}^{\text{init}} \), and measure its deviation from the ideal using the Wasserstein-1 distance:
\begin{equation}
w_{B_y}^{\text{init}} = W_1(D_{B_y}^{\text{init}}, D^*)
\end{equation}

We normalize this value to obtain

\begin{equation}
\overline{w}^{\text{init}}_{B_y} \in [0, 1],
\end{equation}

where 0 indicates no bias and 1 indicates maximal deviation from the ideal.

    \item \textbf{ Intervening on \( B_x \).}  
We simulate a mitigation of \( B_x \) by ensuring equal representation of its counterfactual values during image generation. This results in a new attribute distribution for \( B_y \), denoted \( D_{B_y}^{B_x} \), and we compute its Wasserstein distance to the ideal:

\begin{equation}
w_{B_y}^{B_x} = W_1(D_{B_y}^{B_x}, D^*)
\end{equation}

\end{enumerate}

We then compute the intersectional sensitivity score as:

\begin{equation}
\textrm{IS}_{xy} = \overline{w}_{B_y}^{\text{init}} - \overline{w}_{B_y}^{B_x}
\end{equation}

A positive score (\( \textrm{IS}_{xy} > 0 \)) indicates that mitigating \( B_x \) improves fairness in \( B_y \), while a negative score (\( \textrm{IS}_{xy} < 0 \)) suggests it worsens \( B_y \). A score of zero means there is no effect. This directional metric helps identify both positive outcomes and unintended consequences of single-axis interventions.

\paragraph{Visualization.}

To summarize and interpret intersectional effects, we build a bias intersectionality matrix \( \mathbf{S} \), where each entry \( IS_{ij} \) represents the effect of mitigating bias axis \( B_i \) (row) on bias axis \( B_j \) (column). This matrix supports structured analysis of intersectional dependencies and can guide better prioritization and sequencing of bias mitigation strategies.

\subsection{Multi Step Mitigation}
\label{sec:multi_step_mitigation}

Given the aforementioned information, we first calculate a bias score for initial model $M^{(0)}$ by taking the dot product of the priority vector $\mathbf{p}$ and the initial measures of biases $B^*$ ($\overline{w}^{\textrm{init}}_{B*}$) computed using Eq. \ref{eqn:biasscore}. This is denoted by $\tau =\langle \overline{w}^{\textrm{init}}_{B*},p \rangle$ and measures the overall bias of the model on $B^*$ at any timestep. We proceed to mitigation if $\tau$ is greater than a threshold $\epsilon$. 

To choose which bias axis to mitigate on, we extract the submatrix $\mathbf{S}' \in \mathbb{R}^{n \times |B^*|}$ consisting of the relevant columns from $\mathbf{S}$ obtained using \modelnameintersect. For each row $\mathbf{s}'_i$ of $\mathbf{S}'$, we compute a similarity score $\gamma_i = \langle \mathbf{s}'_i, \mathbf{p} \rangle$, which quantifies the alignment between the $i$-th intersectional bias and the desired direction of mitigation. The bias axis $i^* = \arg\max_i \gamma_i$ with the highest alignment score is selected for targeted mitigation in the current iteration. The model is then updated to reduce bias along the direction corresponding to $i^*$, using a mitigation method, giving $M^{(1)}$. After mitigation, we generate a new set of images, recompute $\tau$, and continue the mitigation process if $\tau > \epsilon$. 

\begin{algorithm}
\caption{\mitname: Intersectional Mitigation}
\label{alg:intersectional_mitigation}
\begin{algorithmic}[1]
\Require Relevant bias axes $B^* \subseteq B$, priority vector $\mathbf{p} \in \mathbb{R}^{|B^*|}$ with $\|\mathbf{p}\|_1 = 1$, sensitivity matrix $\mathbf{S} \in \mathbb{R}^{n \times {|B^*|}}$, bias threshold $\epsilon$, TTI model $M$
\Ensure Final mitigated model $M^{(t)}$ with $\tau < \epsilon$
\State Initialize model $M^{(0)}$, set iteration counter $t \gets 0$
\Repeat
    %\State Select relevant axes $B^* \subseteq B$
    \State Extract submatrix $\mathbf{S}' \in \mathbb{R}^{n \times |B^*|}$ from $\mathbf{S}$
    \State Extract priority vector $\mathbf{p} \in \mathbb{R}^{|B^*|}$
    \For{$i = 1$ to $n$}
        \State Compute similarity score $\gamma_i \gets \langle \mathbf{s}'_i, \mathbf{p} \rangle$
    \EndFor
    \State Identify target axis: $i^* \gets \arg\max_i \gamma_i$
    \State Mitigate axis $i^*$ to update model: $M^{(t+1)}$
    \State Compute  bias score $\tau^{(t+1)} =\langle \overline{w}^{\textrm{init}}_{B*},p \rangle $
    \State $t \gets t + 1$
\Until{$\boldsymbol{\tau}^{(t)} < \epsilon$}
\State \Return $M^{(t)}$
\end{algorithmic}
\end{algorithm}

%%%%%%%%%
\section{Experiments}
\subsection{Analyzing \mitname\ for Mitigation}
\label{sec:compareitigen}

To evaluate the effectiveness of our intersectional bias mitigation strategy, we compare \mitname\ against ITI-GEN \cite{zhang2023iti}, a widely used method designed specifically for Stable Diffusion 1.4 (SD1.4). ITI-GEN supports up to three simultaneous bias mitigations but requires retraining and lacks scalability. In contrast, our method combines a simple, modular prompting technique that uses hardprompts with a powerful intersectional optimization algorithm \mitname.

\noindent{\textbf{Prompt Modification (\hardprompt})} Our method uses sequential prompt modification to incorporate bias counterfactuals. Consider mitigating `environment' and `clothing' bias for the prompt "a nurse":
\label{sup:hardprompt}
\begin{itemize}
    \item First, we replace the base prompt with ``a nurse working indoors" and ``a nurse working outdoors", ensuring a 50/50 split.
    \item Next, we extend to clothing: formal/informal variants of both environment prompts. This leads to 4 total prompts, evenly sampled.
\end{itemize}
This modular setup allows compound mitigation, efficient image generation, and compatibility with hardware optimizations.

\noindent\textbf{Mitigation.} \mitname\ can use any sequential mitigation method, but we consider a simple training-free mitigation method using only prompt modifications. At each mitigation step, we modify the initial prompt to introduce counterfactual concepts associated with the mitigated bias axis. Over multiple steps, we create collections of counterfactual prompts that include all permutations of all mitigated axes (see \ref{sup:hardprompt}). We emperically set $\epsilon = 0.35$ for all our experiments. To compare our method to a traditional mitigation approach, we select ITI-GEN \cite{zhang2023iti}, as it uses a similar FairToken-based permutation approach.

\noindent\textbf{Experimental Setup.} We evaluate our mitigation framework using two different TTI models: Stable Diffusion 1.4 (SD 1.4) and Stable Diffusion 3.5 (SD 3.5). For SD1.4, we select random subsets of occupation-based biases and assign equal weights. For SD 3.5, we evaluate on 15 occupation prompts with user-defined priority vectors. We quantify mitigation using (1) \textbf{MitAmt}: the average bias score $\tau^T$ after mitigation; and (2) \textbf{MitSteps}: the ratio of mitigated bias axes to the total biases in $\mathbf{p}$. Visual quality is measured using CLIP-IQA \cite{wang2023exploring} and a VQA prompt: ``\texttt{[vqa] Is there a person in the image?}''.

\noindent\textbf{Main Results.} As shown in Table~\ref{tab:mitigationres}, \mitname-\hardprompt\ achieves a lower post-mitigation bias (0.33 vs.\ 0.52 for ITI-GEN) while using only 75.6\% of the mitigation steps. Moreover, it preserves visual quality better (0.82 vs.\ 0.73), with more realistic and natural images. For SD 3.5, \mitname\ achieves a bias score of 0.27 with only 76\% of the mitigation steps, again without retraining. These results demonstrate that \mitname\ provides a training-free, efficient, and high-quality solution for intersectional bias mitigation.

\begin{table*}[t]
\small
  \centering
  \caption{\textbf{Comparing our Mitigation Algorithm to ITI-GEN}. We mitigate a randomly chosen subset of 2–5 biases for prompts in the occupation set, and compute {\color[HTML]{F56B00} visual quality} metrics and {\color[HTML]{00009B} mitigation outcomes}. We find that our algorithm uses 22\% fewer mitigation steps, while still yielding higher mitigation amount and quality. $^*$Indicates we use a different prompt set and priority on SD 3.5, so these should not be compared to SD1.4 results.
 }
  \begin{tabular}{l|ccccc|cc}
  \toprule 
\textbf{Method} & {\color[HTML]{F56B00} \textbf{quality}} $\uparrow$ & {\color[HTML]{F56B00} \textbf{real}} $\uparrow$ & {\color[HTML]{F56B00} \textbf{natural}} $\uparrow$ & {\color[HTML]{F56B00} \textbf{colorfulness}} $\uparrow$ & {\color[HTML]{F56B00} \textbf{IsP?}} $\uparrow$ & {\color[HTML]{00009B} \textbf{MitAmt}} $\downarrow$ & {\color[HTML]{00009B} \textbf{MitSteps}} $\downarrow$ \\
\midrule
ITI-GEN (SD1.4)  & 0.73    & 0.92   & 0.37    & 0.45     & 92.8\%    &    0.52   & 100\%  \\
\mitname-\hardprompt\ (SD1.4) & \textbf{0.82}   & \underline{0.98}  & \underline{0.58} & \underline{0.66} & \underline{99\%}  &  \textbf{0.33} & \textbf{75.6\%}   \\
\midrule
\mitname-\hardprompt\ (SD 3.5)  & 0.78    & \textbf{0.99}     & \textbf{0.92}  & \textbf{0.74}   & \textbf{100\%}    & 0.27$^*$   & 76\%$^*$   \\                  
\bottomrule
\end{tabular}
  %\vspace{-0.2cm}
  
  \label{tab:mitigationres}
  %\vspace{-0.2cm}
\end{table*}

\subsection{Uncovering optimal bias mitigation strategies} 
\begin{figure*}
  \centering
  \includegraphics[width=\linewidth]{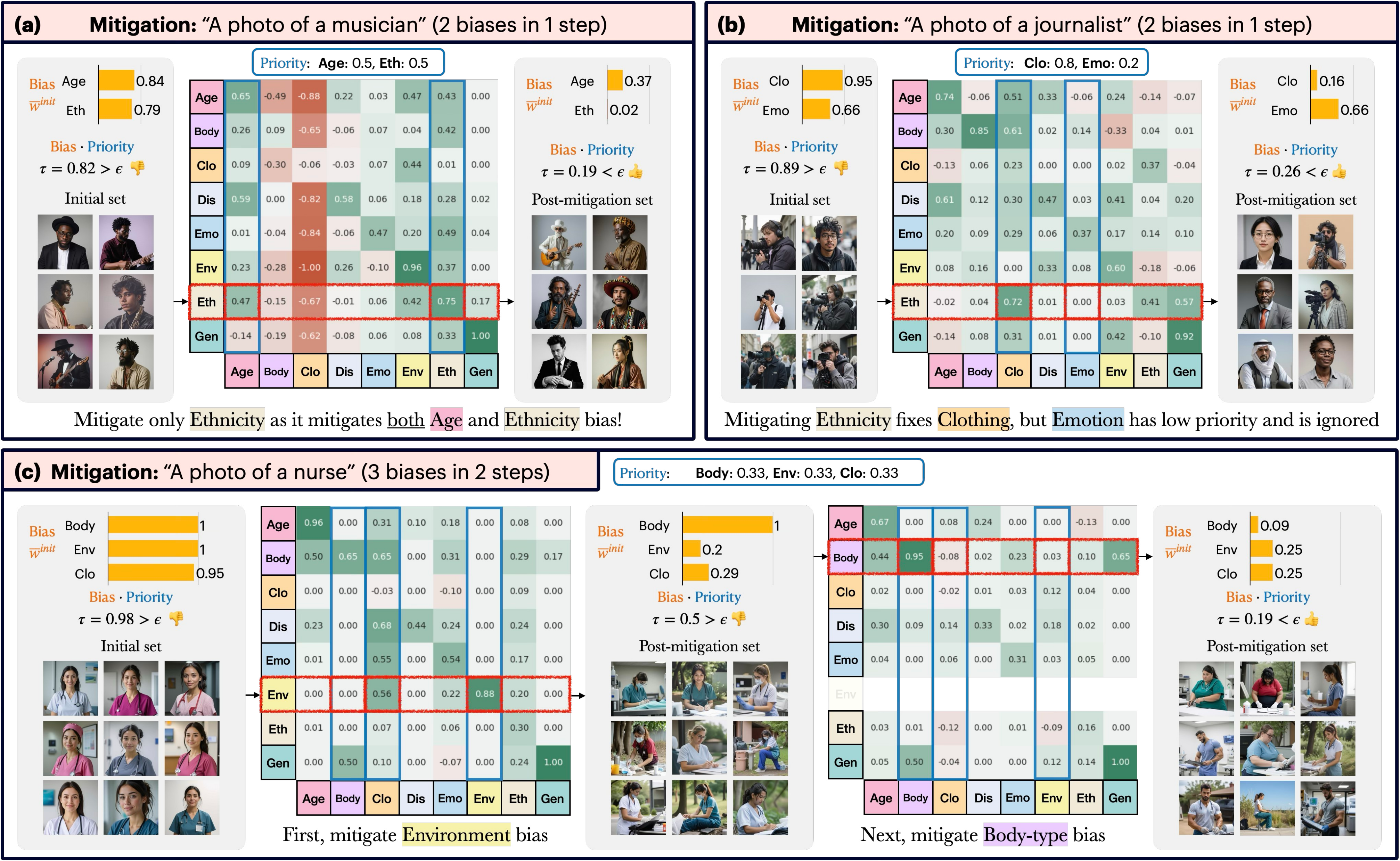}
  \caption{Examples of \mitname\ mitigation. Blue columns are mitigated. (a) and (c) show efficient multi-bias mitigation. (b) highlights user-guided prioritization. Cross-axis effects like Ethnicity $\rightarrow$ Emotion are observable where mitigating ethnicity improves diversity in clothing but based on the user priority ignores emotion.}
  \label{fig:applications}
\end{figure*}

\mitname\ is flexible and supports any set of user-specified bias axes. As shown in Figure \ref{fig:applications}(a) \& (c), it often achieves effective mitigation in fewer steps than the user-defined threshold. By leveraging inter-axis relations, it identifies optimal strategies. In Figure \ref{fig:applications}(a), when age and ethnicity are equally prioritized, mitigating ethnicity alone can reduce both due to demographic overlap, and a single intervention meets the threshold. In Figure \ref{fig:applications}(c) in two mitigation steps, the bias profile progressively aligns with the priority vector (dot product $\tau$ drops: 0.98 → 0.50 → 0.19). Notably, mitigating environment also reduces clothing bias due to strong intersectionality, showing how our method leverages inter-axis relationships for efficient mitigation. Moreover, if \mitname\ fails to reach the desired bias threshold or if mitigating one axis negatively impacts another, it can alert the user to these trade-offs, enabling informed decision-making.
\section {Discussion}

\mitname\ is a modular approach for mitigating biases in TTI models. In this section, we highlight key features of \mitname\ that enable it to meet the requirements of a dynamic bias mitigation framework for TTI models, as outlined in the earlier sections of this chapter.

\noindent\textbf{Extension to Other Approaches.}
\label{sec:other_approaches}
We propose a general framework for mitigating intersectional biases in TTI models. As shown in Alg. \ref{alg:intersectional_mitigation}, our method can be layered on top of any sequential bias mitigation strategy. At each step, a single bias axis is mitigated, and the intersectionality matrix $\mathbf{S}$ is recomputed to capture updated dependencies between bias dimensions. This enables an iterative process where mitigation actions are informed not only by the user’s priorities but also by how interventions on one bias axis may impact others. While we demonstrate our approach using a simple method like \hardprompt, it can be readily extended to more sophisticated strategies. For example, weight-editing approaches such as UCE \cite{gandikota2024unified} could leverage the embeddings of bias-relevant keywords to adjust attention weights dynamically. These adjustments can be made in accordance with the user’s specified ideal distributions, allowing the system to fine-tune the weight of each keyword depending on whether a particular attribute needs to be amplified or suppressed to achieve a desired distributional outcome. Similarly, any LoRA -based method can incorporate the intersectionality matrix 
$\mathbf{S}$ directly into its low-rank decomposition setup by aligning the decomposition adjustments with the directional dependencies captured by 
$\mathbf{S}$. This integration allows the model to modulate its representations in a targeted manner, mitigating intersectional biases more holistically without requiring retraining from scratch. In all these cases, our framework serves as a modular layer that complements existing mitigation techniques by explicitly accounting for the complex interactions between multiple bias dimensions.

\noindent\textbf{Role of Priority Vectors and Visual Examples.} Incorporating user-defined priorities enables flexible and targeted bias mitigation, allowing the system to align with diverse user goals and fairness considerations. By specifying a priority vector over different bias dimensions, users can explicitly control which biases should be addressed more aggressively and which can be deprioritized depending on the context of the application. For example, as shown in Figure. ~\ref{fig:applications}(c), the user assigns equal weights to body type, environment, and clothing, which prompts the model to distribute the mitigation effort evenly across these three dimensions. This ensures that no particular bias dominates the output, resulting in balanced representation across all specified attributes.

In contrast, Figure. ~\ref{fig:applications}(b) illustrates a different scenario where the user places a higher priority on mitigating clothing diversity while assigning lower weight to emotion-related biases. In this case, the model focuses its efforts primarily on reducing biases related to clothing while leaving minor or less impactful changes in the emotional expressions of the generated subjects. This targeted approach allows users to steer mitigation strategies toward dimensions they deem most critical, reflecting practical needs or ethical concerns relevant to the deployment context.

This flexibility is crucial in real-world applications, where fairness requirements can vary significantly across domains. For instance, in fashion or marketing industries, users may prioritize attributes like body type and clothing diversity, whereas in healthcare or education, factors like age, gender, or race might take precedence. By allowing fine-grained control over mitigation efforts, our approach supports a broad spectrum of fairness objectives and adapts to the values and expectations of different stakeholders.

\begin{figure*}
  \centering
   \includegraphics[width=0.8\linewidth]{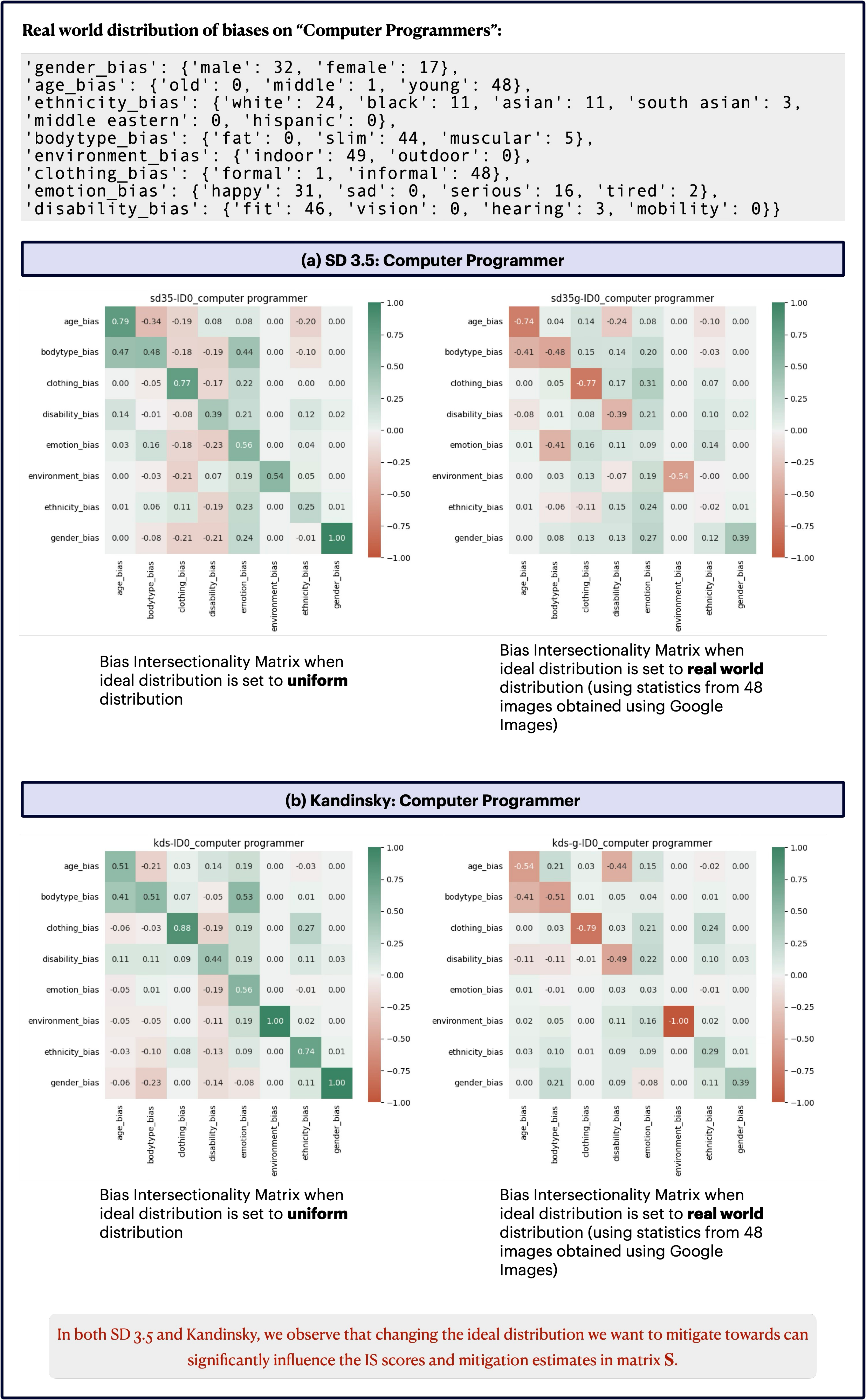}
   \caption{Modifying $D^*$ (ideal distribution) in \mitname\ can have a significant effect on the \isscore\ values. Intermit accoounts for this change during the mitigation process and allows user to specify the ideal distribution they want to have.}
   \label{fig:distribution_change}
   % \vspace{-0.2cm}
\end{figure*}

\noindent\textbf{Generalization to Custom Target Distributions.} \mitname allows users to define context-specific target distributions $D^*$ for each bias dimension. In one of our experiments, we collect 48 real-world images of computer programmers to construct a distribution $D^{*}$ based on observed data. Modifying the ideal distribution $D^{*}$ within \mitname has a significant impact on the resulting \isscore values and, consequently, on the overall mitigation strategy. \mitname\ explicitly accounts for such changes by allowing users to specify the desired target distribution they wish to achieve. In Figure. ~\ref{fig:distribution_change}, we demonstrate this effect by comparing two different choices of $D^*$ for the occupation of computer programmers: one based on the real-world distribution $D^{*}$ and the other using a uniform distribution. The resulting confusion matrices show substantial differences between these two settings. Notably, some relationships—such as the correlation between body type and age—reverse direction depending on the choice of $D^*$. As a result, the mitigation strategy must adapt to align with the user-defined fairness objectives. This example underscores the importance of carefully selecting the ideal distribution, as it directly shapes which biases are prioritized and how intersectional dependencies are handled during mitigation.
\label{sec:targetdist}

\section{Conclusion}

This chapter presented \mitname, a method for reducing intersectional bias in text-to-image models. It is designed to meet the main needs of a good bias mitigation strategy. Below, we summarize how it addresses each of these needs:

\begin{itemize}
    \item \textbf{Context-aware and prompt-sensitive:} \mitname operates at the prompt level. It adjusts mitigation based on how prompts are written and what context they imply. The outcome may vary across different prompt sets.

    \item \textbf{Causally informed:} It uses sensitivity scores from \modelnameintersect\ and \isscore\ to understand how one bias axis affects another. This is based on counterfactual estimates, where \isscore\ acts like a treatment effect showing the impact of changes on other axes.

    \item \textbf{Intersectionally robust:} \mitname supports multiple bias axes through a priority vector. Users select which biases to focus on and assign weights to reflect trade-offs.

    \item \textbf{Supports user-defined distributions:} Users can define a target distribution for each bias axis. These can be uniform, real-world-based, or context-specific, removing the need to assume a fixed “ideal” distribution.

    \item \textbf{Modular and training-free:} \mitname does not require model retraining or internal access. It can be used with different TTI models as a separate component.

    \item \textbf{Trade-off aware and transparent:} It reports how mitigation on one axis affects others. This helps users understand and manage cross-axis effects.

    \item \textbf{Aligned with community and contextual norms:} Users set their own fairness goals. This allows adaptation to different use cases and cultural settings.

    \item \textbf{Empirical results:} \mitname reduced bias more than ITI-GEN, needed fewer steps, and preserved image quality even when mitigating multiple axes.
\end{itemize}
\subsection{Ethical Consideration}

While \mitname offers flexibility in how users define and control fairness goals, this same flexibility can also be misused. Users can arbitrarily modify the priority vector or define biased target distributions to push outputs toward harmful or exclusionary representations. The system does not enforce any default notion of fairness, and so users must take care to ensure that mitigation choices align with broader ethical standards and do not reinforce new forms of bias.

Additionally, in designing this system, we simplify complex social attributes such as gender and ethnicity into discrete, categorical variables. This design choice enables quantification, measurement, and intervention but deviates from how these concepts are experienced in real life—where identity is often fluid, non-binary, and shaped by culture and context. We acknowledge this limitation as a necessary trade-off to make the problem tractable and actionable within current generative AI systems.

We encourage future work to explore how these representations can be made more nuanced, participatory, and reflective of lived experiences, while maintaining technical feasibility.

\chapter{Conclusion}
\epigraph{
``In the process of conducting experiments, testing hypotheses, and pushing the boundaries of research during your PhD, you acquire a meta-skill — a skill that is hard to put into words but makes you wise.''
}{\textit{Reflection during my PhD journey}}
This thesis presents my efforts to develop algorithms for evaluating and improving computer vision models. While counterfactual reasoning serves as a unifying theme, other ideas recur throughout the chapters. For instance, Chapters 3 and 5 focus on explainability, while Chapters 4, 6, and 7 emphasize fairness. Chapters 3, 5, and 6 propose methods for evaluating AI models, whereas Chapters 4 and 7 extend this by introducing techniques to improve model fairness. Additionally, Chapters 3 and 4 deal with classification models, while Chapters 5 through 7 focus on generative modeling.

\section{Key Contributions}
A PhD thesis is, at its core, an attempt to push the boundaries of existing research. While the long-term impact of any such work can only be judged over time, the primary motivation behind this thesis has been to contribute meaningful advances to the growing body of research on computer vision safety and explainability. This thesis set out to challenge prevailing assumptions about fairness auditing in machine learning and to advocate for more context-sensitive, concept-aligned, and causally grounded methods—particularly within vision and generative models.

Across multiple chapters, we operationalized counterfactual reasoning at varying levels of abstraction—concept attribution (CAVLI), adversarial counterfactual training (ASACs), dynamic bias evaluation (TIBET), causal bias diagnosis (BiasConnect), and intersectional mitigation (InterMit). Each of these contributions embodies the same core principle: isolate the influence of meaningful interventions while holding irrelevant factors constant.

Below, we summarize the key contributions of this thesis:
\begin{itemize}
\item \textbf{Dynamic, Context-Dependent Bias Evaluation and Mitigation for Generative Models}
A central theme of this thesis is the push for dynamic rather than static evaluations of bias. Prior work in bias detection and mitigation often focuses on predefined dimensions such as gender or ethnicity, assuming these axes of bias are fixed and independent. However, in practice—especially in generative models like Text-to-Image (TTI) systems—the influence of these attributes is highly context-dependent. The same identity label can have vastly different semantic and visual implications depending on surrounding prompt cues. 

To address this, the thesis introduces TIBET, a framework for dynamic, prompt-sensitive bias evaluation in TTI models. We also proposed InterMit, a context-aware bias mitigation algorithm that adapts its interventions based on observed interactions between bias axes and user-defined fairness priorities. These methods offer a more realistic and flexible evaluation and intervention pipeline that reflects the fluid semantics of generative tasks. This thesis moves in a different direction than previous approaches by emphasising on the need of dynamic bias evaluation and mitigation platforms. 
\item  \textbf{A Rigorous Framework for Studying Intersectional Biases in Generative Models.}
Intersectionality is a widely recognized concept in social sciences and phliosophy\cite{curry2018killing,diana2023correcting}, few technical frameworks exist for quantifying and diagnosing how multiple biases interact within machine learning models—especially generative ones. In Chapter 6,  this gap  is addressed through BiasConnect, a causal, counterfactual-based diagnostic tool that measures the ripple effect of mitigating one bias axis (e.g., gender) on others (e.g., age or race). Unlike observational fairness metrics, our method uncovers deeper, causal relationships between protected attributes—offering a principled way to understand bias entanglement. Further \mitname was proposed not only to mitigate intersectional biases simulataneously but also alert the user about the trade-off about these biases. As per my knowledge this is the first concrete work to tackle intersectionality of biases in TTI models in such a comprehensive manner.

\item \textbf{A Flexible, Modular Bias Mitigation Algorithm for TTI Models.}
Chapter 7 argues that bias mitigation in TTI models is uniquely challenging due to the open-ended nature of generation, the lack of explicit labels, and the entanglement of social attributes in latent representations. We proposed InterMit, a novel mitigation algorithm that is modular, training-free, and sensitive to user-defined fairness trade-offs. InterMit leverages causal sensitivity scores to guide intervention, ensuring that debiasing in one dimension does not inadvertently worsen outcomes along others. This approach represents a step toward practically deployable, intersectionally aware mitigation tools for generative systems. One significant contribution of this thesis is outlaying the challenges associated with bias mitigation in TTI models and presenting a tool that takes a step in that direction. 

\item \textbf{Using Adversarial Examples as Tools for Fairness.}
In Chapter 4 adversarial examples are presented not as threats to model robustness, but as tools for fairness enhancement. Through ASACs (Attribute-Specific Adversarial Counterfactuals), we demonstrate that adversarial perturbations targeting protected attributes can be repurposed to fine-tune biased classifiers in a curriculum-based manner. This method not only reduces fairness gaps but also improves overall model accuracy—offering an ethical and effective alternative to GAN-based counterfactuals, which often reinforce stereotypes.
\item \textbf{Concept-Level Interpretability for Individual Model Decisions.} In Chapter 3, CAVLI, a novel hybrid method combining TCAV and LIME is introduced that produces localized, concept-based explanations. CAVLI quantifies the degree to which a classifier’s decision on an individual image depends on human-defined concepts like ``grassland" or ``stripes." By analyzing the spatial overlap between regions important for concept representation and those driving the model’s decision, CAVLI provides a Concept Dependency Score (CDS). This enables practitioners to go beyond pixel saliency and ask: Is the model predicting 'cow' because of the animal, or because of the grassy background?

Our results show that such fine-grained, interpretable diagnostics can uncover spurious correlations and dataset biases—thereby guiding more trustworthy model design. In sensitive applications like medical imaging or facial recognition, such explanations can help ensure decisions are grounded in appropriate semantic features.

\end{itemize}

\section{Limitations}

\subsection{Counterfactual Approaches}
\label{subsec:cf_limitations}

Counterfactual methods help us understand how models behave by asking “what if” questions, but they come with important limitations. First, they rely on assumptions about what changes are small and realistic. In social settings, this is hard because traits like race, gender, and class are often linked in ways that make it unrealistic to change one without affecting others.

Many methods also assume that features can be changed one at a time or that effects add up in simple ways. These assumptions often break down, especially in complex models like text-to-image systems where features interact in messy and unpredictable ways.

There are also practical limits. It’s not possible to test every possible counterfactual, so we use a small sample to keep things fast. But this means we might miss some important examples. There is often a higher computational costs that is associated with generating a large set of counterfactual images. 

Finally, there is no single way to judge whether a counterfactual is good. We use model scores and concept measures, but these do not always reflect what people find fair, realistic, or helpful. Without human feedback or ground truth, it is difficult to determine when a counterfactual truly makes sense.

These limitations do not mean that counterfactuals are not useful, but they remind us to be careful — especially when applying them in areas that directly impact people’s lives.

\subsection{Dynamic Bias Mitigation}
\label{subsec:dynamic_limitations}

Dynamic bias mitigation methods adjust outputs based on the prompt, model state, or user-defined fairness goals. While this flexibility is useful, it can cause problems. These systems often behave differently for similar inputs, which makes it hard to reproduce results or audit fairness over time. Since they make decisions on the fly, they also need more computation, which slows things down and makes large-scale deployment harder. It's also tough to judge what the “right” mitigation is, since fairness means different things to different people and contexts.

These methods can also run into deeper problems. They may fix one kind of bias but unintentionally make others worse, especially in intersectional cases. When users control fairness settings, the system may overfit to their short-term goals and miss long-term harms. It's also harder to explain how or why a certain output was generated, which can reduce trust. Finally, letting users or institutions decide what to fix raises questions about whose idea of fairness the system follows. These trade-offs show why dynamic approaches must be used carefully, especially in areas that affect people’s lives.

\subsection{Adversarial Images for Fairness}
\label{subsec:adv_fairness}

Adversarial images can help us test and improve fairness in computer vision models. These are images that are slightly changed to see if the model gives a different output in ways that might show bias. For example, if a model changes its decision when skin tone or clothing is adjusted slightly, that might be a sign of unfairness. Researchers also use adversarial training to make models less sensitive to certain features by adding these images during training and making the model learn not to rely on them.

However, this approach has limits. Adversarial images may look strange or unrealistic, which can confuse both the model and the human trying to interpret the result. It’s also hard to decide what counts as a fair change—sometimes a small tweak might flip the output even though it's not meaningful. Most adversarial fairness tests focus on one attribute at a time, and don’t capture more complex combinations like race and gender together. Finally, if done carelessly, this method might make the model ignore useful signals or cause new types of bias to appear.

\section{Ethical Considerations}
\label{sec:ethics}

This thesis touches on questions that sit at the boundary between technical modeling and social meaning. In building systems to detect and reduce bias, we work with sensitive social categories like race, gender, and ethnicity. These categories are shaped by history, culture, and power, and there is no single way to define or formalize them. As computer scientists, we may not always engage deeply with the full range of views on these topics, but we believe it is important to reflect on the assumptions behind our approach.

Our method takes a clear stance: making patterns of bias visible is a step toward accountability. While our use of quantitative scores and fairness weights requires abstraction, we see this not as erasing social meaning, but as making it traceable. By letting users set fairness goals and inspect how outputs change, our method can also be prone to misuse if given in the wrong hands. We believe that making these decisions transparent—even when imperfect—is better than hiding them. This helps reduce the risk of covert manipulation and supports open discussion about what fairness should look like in different settings.

We do not claim that our way of modeling intersectionality is the only valid one. We work within a broader view that accepts both additive and interactive patterns of disadvantage. The categories we use are not fixed; they can overlap, shift, and take on different meanings in different contexts. Rather than fix their meaning, we treat them as flexible tools to help surface how a model responds to real-world complexity.

There are also concerns about using visual cues to infer features like ethnicity. We understand that these features are socially constructed and may not always have clear boundaries. Still, in the absence of agreement on how to define them, we choose to stay neutral on the deeper debates. What matters for our work is not the ultimate nature of these categories, but how models respond to them in practice. By focusing on how models treat identity-related prompts and outputs, we provide a way to study patterns of harm—even if we cannot settle what these identities ``really" are.

Finally, we acknowledge that no system can prevent misuse. But we can reduce the risk by making our assumptions clear, our settings open, and our results reproducible. We see transparency not only as a design choice, but as an ethical safeguard. In building tools that diagnose bias and allow users to control fairness goals, we aim to support open, responsible, and adaptive use—without claiming to offer a final answer to the deep social and philosophical questions at play.

\section{Future Directions}
\label{sec:future}

There are several promising directions for extending this work. One key challenge is to better understand how intersectional identities interact in generative systems. Our current method models these identities through compositional prompts and measurable dimensions of bias, but deeper work is needed to capture how social meanings shift across contexts. Future research could explore richer prompt structures, broader cultural inputs, and participatory methods that reflect how affected groups view fairness and representation.

The work on CAVLI and ASAC can be extended to automated concept discovery approaches, removing the need to specify predefined concepts such as “grassland” or “gender.” Instead, the system could automatically identify which concepts or prompts are most relevant and either mitigate biases along these dimensions (as in ASAC) or quantify the model’s dependence on them (as in CAVLI). Additionally, the approach presented in Chapter 4 is currently designed to mitigate one attribute at a time. Extending it to support simultaneous mitigation of multiple attributes would make it more effective and better suited for real-world scenarios where multiple biases often coexist.

Our work in Chapter 6 introduces \modelnamegraph\ and \modelnameintersect\ as tools to diagnose the intersectionality of biases. However, these tools should be seen as initial steps toward this goal. Further work is needed to develop a fully causal framework for understanding intersectionality—one capable of evaluating higher-order dependencies and constructing comprehensive causal graphs. Similarly, the mitigation strategies presented in Chapter 7 could be extended to handle more than two bias dimensions simultaneously, leveraging higher-order methods that consider multiple axes of bias together rather than in isolation.

Another current limitation of the approach discussed in Chapter 7 is efficiency. At each step of our bias mitigation algorithm, we compute a full intersectionality matrix, denoted as \texttt{S}, which slows down the process. This points to the need for faster and more scalable approaches. Future research could also explore alternatives to the sensitivity matrix \texttt{S} and investigate better ways to incorporate such representations into mitigation frameworks. An important direction is to develop more robust, adaptive mitigation strategies. While our current approach allows users to set fairness goals dynamically, future work could incorporate feedback loops where users, auditors, or policy makers co-define mitigation objectives over time. This would help the system respond to evolving norms while maintaining transparency and accountability.

We also see potential in building connections between formal modeling and qualitative perspectives. Rather than treating abstraction and lived experience as opposites, future methods could combine structured diagnosis with narrative or case-based feedback. This would allow researchers to test how well formal scores align with how people experience bias in practice.

The approaches discussed in this thesis can also be extend to domains other than images. While scaling these tools to work with more complex generative pipelines—such as multi-turn conversations, video synthesis, or multi-agent systems—will require both technical and ethical advances. These systems will likely raise new fairness concerns that go beyond individual prompts, involving interaction histories, social roles, or emergent behaviors. Understanding and addressing bias in these settings will require interdisciplinary collaboration and continued reflection on the limits of current methods.

\newpage

\bibliographystyle{unsrtnat}
\bibliography{main}
\clearpage
%\renewcommand{\thechapter}{A}
%\renewcommand{\thesection}{\thechapter. \arabic{section}}
%\setcounter{chapter}{1}
%\setcounter{section}{0}

%\setcounter{chapter}{1}
%\setcounter{section}{0}
%\chapter*{Appendix}

%\input{chapters/Appendix/TIBET}

\appendix
% \renewcommand{\thesection}{\Alph{section}}
% \setcounter{section}{0}
% \renewcommand{\thefigure}{A\arabic{figure}}
%\chapter{APPENDIX}
%\setcounter{chapter}{1}
%\setcounter{section}{0}

\chapter{Definitions of Biases}
\label{sup:biasdefinitions}

Our aim is to quantify and establish a framework for analyzing biases in generative Text-to-Image (TTI) models. While these biases can take diverse forms, it's helpful to categorize them into two distinct groups:

\vspace{0.05in}
\noindent
\textbf{Societal Biases}: These biases encompass the biases that are of societal concern. They are characterized by the presence of unfair or harmful associations between attributes within the generated images \cite{whittaker2018ai}. These biases can stem from various sources, including the training data, and they have the potential to perpetuate and reinforce societal inequalities.

\vspace{0.05in}
\noindent
\textbf{Incidental Correlations}: This includes non-harmful correlations in the generated dataset, stemming from statistical training data correlations, incidental endogeneity, or spurious connections introduced by the TTI model \cite{bhatt2023mitigating,fan2014challenges}. While not directly harmful, they can impact image generation diversity.

As mentioned in Section \ref{sec:intro}, for simplicity, we use the word `bias' to refer to either societal biases or incidental correlations, as the LLMs we use can detect both kinds of bias.

\section{CAS scores}
\label{sup:conceptdecompostion_alg}

The Concept Association Score ($CAS$) tells us how similar the set of images for the initial prompt are to the set of images of a counterfactual prompt. 

In the case of VQA, for each set of images:
\begin{enumerate}
    \item First, we use VQA with MiniGPT-v2 to obtain answers for each image. All answers are combined into a single string.
    \item This string undergoes processing to remove punctuation and stop words.
    \item We use the FreqDict function in NLTK (\url{https://www.nltk.org/api/nltk.probability.FreqDist.html}) to obtain a list of words and their corresponding word frequencies. We normalize this frequency by the number of images in the set (in our setting, 48). This represents a set of concepts, $C = \{(c_1,w_1)...\}$.
\end{enumerate}
Once we complete this process for the initial set (to obtain $C_{init}$) and for a counterfactual set (to obtain $C_{cf}$), we apply Algorithm \ref{alg:cas} to obtain the $CAS$ score.

\section{MAD scores}
\label{sub:MADappendix}

We leverage the MAD metric to measure the amount of bias by computing the variability in CAS scores. We normalize MAD to make it comparable across bias axes with different number of counterfactuals. Specifically, we normalize such that our MAD score for the "most skewed" list of CAS scores for any length K is equal to 1. Therefore, we first create a vector of length K such that all numbers are 0, except one 1. The MAD score for this vector is the maximum MAD score we can obtain for a CAS score list of length K. For simplicity, let us call this $MAD_{K}$, normalized as follows:
\begin{align}
    MAD_{normalized} &= \sqrt{\frac{MAD}{MAD_{K}}}.
\end{align}

For simplicity, we use $MAD$ to refer to $MAD_{normalized}$ in our results and figures.

\noindent\textbf{Alternatives to $MAD$.} We considered different alternatives in place of MAD; however, we selected our (normalized) MAD as it achieved low error amplification on the sensitivity analysis compared to other metrics. At \textbf{18\% VQA error rate}, we observed a $MAD$ change of \underline{\textbf{13.11\%}} (lower is better). We attempted to use two other metrics: (1) Wasserstein distance between our scores, $CAS_{K}^b$ from Eq. 5, to the uniform distribution of $CAS$ scores of length $K$ (which indicates no bias). With the same normalization strategy, this metric changes by \underline{15.15\%}. (2) Standard deviation in $CAS_{K}^b$ scores, which changes by \underline{28.65\%}. As $MAD$ is less sensitive to mistakes by the VLLM VQA model, we select it over these alternative metrics.

\chapter{User Studies}
\label{sup:userstudies}

\section{User Study 1}
\label{sup:userstudy1}

\subsection{Setup}
We conduct the user study using the Amazon Mechanical Turk platform. Prior to commencing the study, we provide workers with comprehensive information regarding various aspects of our research. This includes explanations about TTI models, their purposes, the types of inputs and outputs they handle, and an elucidation of biases.

Specifically, we categorize biases into two distinct types, which we previously referred to as "societal" and "incidental" biases. We offer a detailed definition of these biases and provide practical examples within the training materials to help users grasp these concepts.

In the course of the study, each user is tasked with evaluating input prompts for the presence of two types of bias:
\begin{enumerate}
    \item Societal Bias Related to "Axes": \texttt{Is there evidence of "Axes-related" societal bias in the prompt?}
    \item Incidental Bias Related to "Axes": \texttt{Is there evidence of "Axes-related" incidental bias in the prompt?}
\end{enumerate}

\noindent where "Axes" is replaced with the actual bias axis name obtained using the LLM. 

\begin{figure*}[t]
  \centering
   \includegraphics[width=1.0\linewidth]{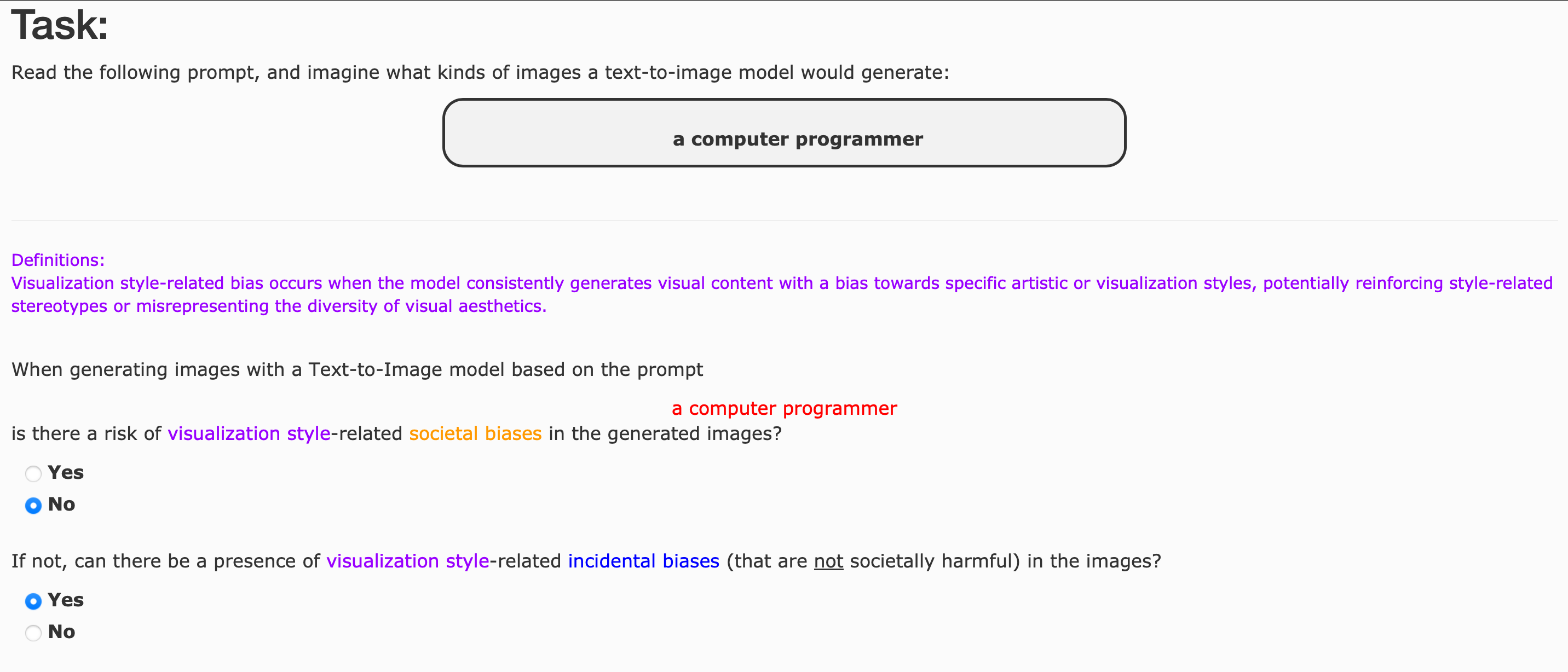}
   \caption{\textbf{User Study 1}. This is the task that a Amazon Mechanical Turk participant sees.}
   \label{fig:us1}
\end{figure*}

This approach ensures that users are equipped to identify and assess these specific biases in the prompts they encounter during the study. Figure \ref{fig:us1} shows an example of the user study task that a participant can see.

Each user participating in our study is presented with five biases related to the displayed prompt in every HIT. In our study, we only display the prompts to the user, omitting the accompanying images. This is because GPT-3 is also coming up with bias axes just by looking at the prompt. Additionally, we include attention-check questions for all users to ensure they are actively engaged in the study. Users are compensated at a rate of \$0.15 for each task they complete.

\subsection{Training Details and Qualification Test}

The participants in our user study are from Canada, the United States, and the United Kingdom. We administer a qualification test that assesses their comprehension of the training materials and includes four straightforward questions resembling those in the primary study. Users who achieve a score above 90\% on the qualification exam are eligible to participate in the main user study.

% \subsection{Results}
% A histogram of biases
% \\Biases that have been detected the most

\section{User Study 2}
\label{sup:userstudy2}

\subsection{Setup}
Our approach to User Study 2 closely mirrors that of User Study 1. We administer the study through the Amazon Mechanical Turk platform. Before participants begin the study, we ensure they have access to detailed information about several key aspects of our research. This information encompasses thorough explanations about TTI models, their intended applications, the range of inputs and outputs these models handle, and a clear explanation of the concept of biases.

\begin{figure*}[t]
  \centering
   \includegraphics[width=1.0\linewidth]{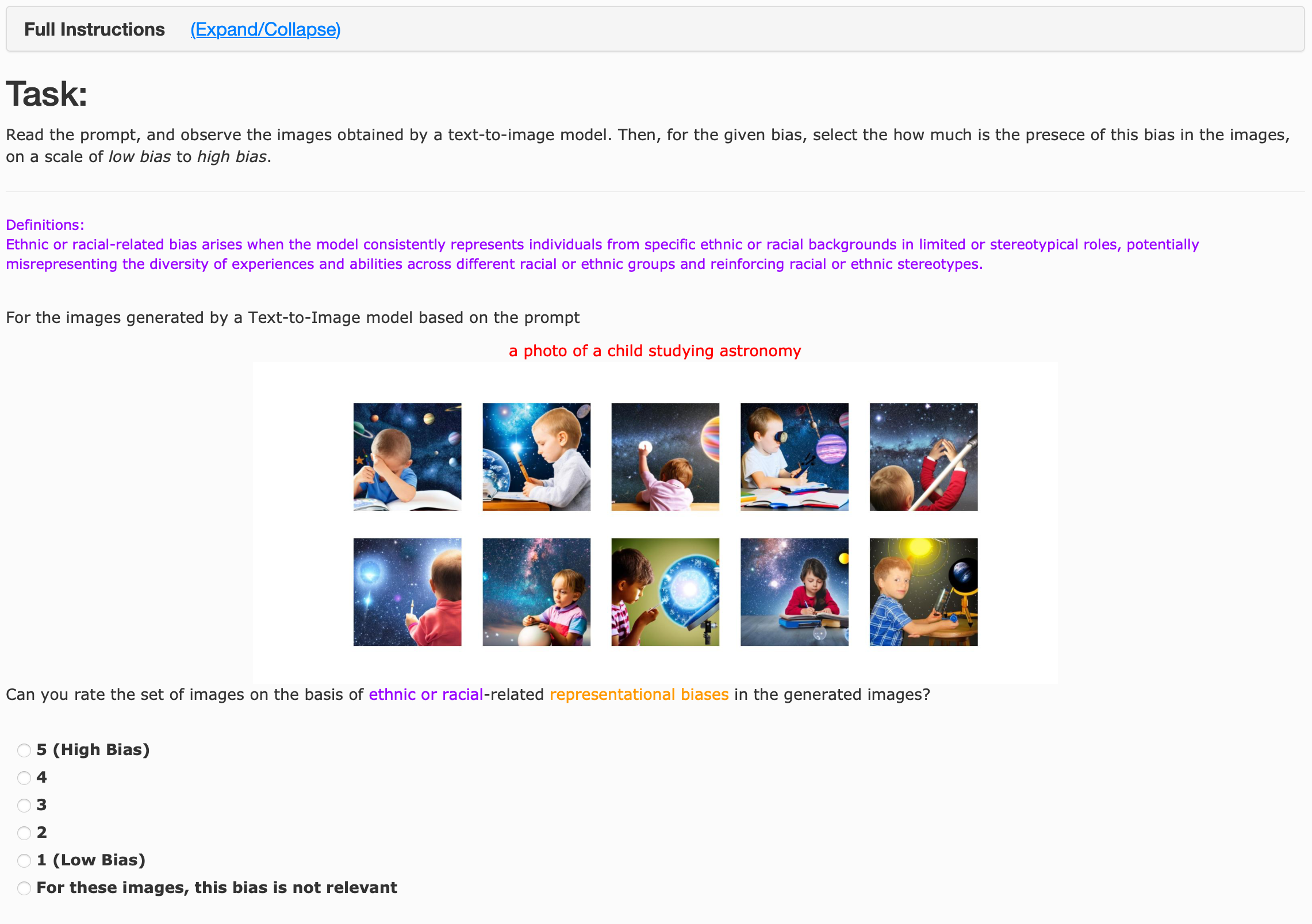}
   \caption{\textbf{User Study 2}. This is the task that a Amazon Mechanical Turk participant sees.}
   \label{fig:us2}
\end{figure*}

In this study, we present users with 10 randomly sampled images for a prompt and request them to rate the presence or absence of bias on a 5-point Likert scale (as depicted in Figure \ref{fig:us2}). To compute our correlations, we only consider prompts with two or more societal biases, specifically, bias axes names containing `gender', `age', `race', `racial', `geographic', `ethnicity', `cultural', as we find that humans are not very reliable at observing non-societal biases.

\subsection{Training Details and Qualification Test}

In alignment with our approach in User Study 1, our study participants are drawn from Canada, the United States, and the United Kingdom. We employ a qualification test designed to gauge their understanding of the training materials, consisting of four straightforward questions akin to those used in the primary study. Participants who attain a score exceeding 90\% on the qualification exam qualify for participation in the main user study.

\end{document}